\documentclass[a4paper,fleqn]{cas-dc}
\usepackage[authoryear]{natbib}
\usepackage{hyperref}
\usepackage{graphicx}
\usepackage{subfigure}
\usepackage{longtable}
\usepackage{epstopdf}
\usepackage{color}
\usepackage[noend]{algpseudocode}
\usepackage{algorithmicx,algorithm}
\usepackage{CJKutf8}
\usepackage[export]{adjustbox}
\usepackage{tabularx}
\usepackage{booktabs}
\usepackage{multirow}
\usepackage{makecell}
\usepackage{etoolbox}
\usepackage{bm}
\AtBeginEnvironment{tabular}{\rmfamily}
\usepackage{xcolor}
\usepackage{ulem}

\newcolumntype{\cm}[1]{>{\centering\arraybackslash}m{#1}}
\def\tsc#1{\csdef{#1}{\textsc{\lowercase{#1}}\xspace}}
\tsc{WGM}
\tsc{QE}

\begin{document}
    \let\WriteBookmarks\relax
	\def\floatpagepagefraction{1}
	\def\textpagefraction{.001}
	
	\shorttitle{Dogfight Search}    
	\shortauthors{Yujing Sun et al.}  
	
	\title [mode = title]{Dogfight Search: A Swarm-Based Optimization Algorithm for Complex Engineering Optimization and Mountainous Terrain Path Planning}
    
	\author[1]{Yujing Sun}
    \ead{2211060216@stu.lntu.edu.cn}
    \credit{Conceptualization, Methodology, Writing - Original Draft, Software, Validation}

    \author[1]{Jie Cai}
    \ead{2311010201@stu.lntu.edu.cn}
    \credit{Validation, Formal analysis, Investigation, Visualization}
    
    \author[2]{Xingguo Xu}[orcid=0009-0008-8286-7367]
    \cormark[1]
    \ead{xuxingguo@mail.dlut.edu.cn}
    \credit{Writing - Review \& Editing, Supervision}

    \author[3]{Yuansheng Gao}[orcid=0000-0003-3278-8835]
    \cormark[1]
    \ead{y.gao@zju.edu.cn}
    \credit{Writing - Review \& Editing, Supervision, Project administration}
    
    \author[2]{Lei Zhang}
    \ead{bhgleizhang@mail.dlut.edu.cn}
    \credit{Investigation}

    \author[4]{Kaichen Ouyang}
    \ead{oykc@mail.ustc.edu.cn}
    \credit{Writing - Review \& Editing}

    \author[2]{Zhanyu Liu}
    \ead{ramirez@mail.dlut.edu.cn}
    \credit{Visualization}

    \affiliation[1]{organization={College of Science, Liaoning Technical University},
            city={Fuxin},
            postcode={123000}, 
            country={China}}

    \affiliation[2]{organization={School of Mathematical Sciences, Dalian University of Technology},
            city={Dalian},
            postcode={116024}, 
            country={China}}
            
    \affiliation[3]{organization={College of Computer Science and Technology, Zhejiang University},
            city={Hangzhou},
            postcode={310027}, 
            country={China}}
            
    \affiliation[4]{organization={Department of Physics, University of Science and Technology of China},
            city={Hefei},
            postcode={230026}, 
            country={China}}
            
    \cortext[1]{Corresponding author}

	
	\begin{abstract}
		Dogfight is a tactical behavior of cooperation between fighters. Inspired by this, this paper proposes a novel metaphor-free metaheuristic algorithm called Dogfight Search (DoS). Unlike traditional algorithms, DoS draws algorithmic framework from the inspiration, but its search mechanism is constructed based on the displacement integration equations in kinematics. Through experimental validation on CEC2017 and CEC2022 benchmark test functions, 10 real-world constrained optimization problems and mountainous terrain path planning tasks, DoS significantly outperforms 7 advanced competitors in overall performance and ranks first in the Friedman ranking. Furthermore, this paper compares the performance of DoS with 3 SOTA algorithms on the CEC2017 and CEC2022 benchmark test functions. The results show that DoS continues to maintain its lead, demonstrating strong competitiveness. The source code of DoS is available at https://ww2.mathworks.cn/matlabcentral/fileexchange/183519-dogfight-search.
	\end{abstract}
	
	
	\maketitle
    
    \section{Introduction}
        
        Optimization problems are widely present in various real-world domains, including engineering design \citep{hao2025composite}, healthcare \citep{osaba2019discrete}, logistics management \citep{chu2021physarum} and financial portfolio optimization \citep{dye2012finite}. The quality of the obtained solutions often directly affects system performance, resource allocation efficiency, and economic benefits. Therefore, designing efficient and robust optimization algorithms to obtain the global optimal solution has become a core research direction in the field of intelligent computing.

        Over the years, researchers have proposed a series of traditional optimization algorithms, such as Newton’s method \citep{su2011two} and gradient descent \citep{hosseinali2024accelerated}, based on the linear characteristics of objective functions. These methods exhibit high computational efficiency and convergence accuracy in solving linear or convex optimization problems. However, their performance is often limited when dealing with complex real-world problems such as non-convexity or non-differentiability \citep{keivanian2022novel, deb2002fast, myers2016application, gong2010bbo}. Moreover, traditional algorithms heavily rely on gradient information during iteration. As problem dimensionality increases, the complexity of gradient computation also rises sharply, further undermining their efficiency in solving high-dimensional engineering problems.

        In contrast to the limitations of traditional methods, metaheuristic algorithms have shown greater adaptability and potential in complex optimization tasks due to their simple implementation, low parameter dependency, and strong global search capabilities \citep{dhiman2019seagull}. For example, Gholizadeh et al. proposed the Newton Metaheuristic Algorithm (NMA) \citep{gholizadeh2020new} for the seismic design of steel frame structures with discrete performance criteria, effectively overcoming the limitations of traditional methods in nonlinear and discrete design spaces. Houssein et al. developed a hybrid metaheuristic algorithm, BES-GO \citep{houssein2025recent}, which demonstrated superior convergence performance and practical applicability in structural design benchmarks. In addition, Lu et al. introduced the Multi-Strategy Beetle Swarm Optimization (MBSO) \citep{lu2025research} algorithm to optimize air volume in mine ventilation networks, significantly improving the energy efficiency of the system. These studies highlight the broad applicability and significant development potential of metaheuristics in solving nonlinear, high-dimensional, and constrained engineering optimization problems. Representative contributions are summarized in Table~\ref{tab1}.

        \begin{table*}[htbp!]
        \centering
        \caption{Contributions of metaheuristic algorithms to real-world engineering problems.}
        \label{tab1}
        \begin{tabular}{\cm{0.25\textwidth}\cm{0.65\textwidth}}
            \toprule
            Algorithm & Engineering Contribution \\
            \midrule
            Dual-Pheromone Crossover Ant Colony Optimization (DPX-ACO) \citep{wang2025towards} & The proposed method significantly reduces the overall production costs in large-scale cotton blending optimization. Comprehensive experimental analyses further demonstrate its effectiveness, with several benchmark scenarios achieving cost reductions exceeding 30\%. \\
            Improved Discrete Particle Swarm Optimization (IDPSO) \citep{wu2025improved} & Provided superior scheduling solutions for automated guided vehicle manufacturing processes by considering dynamic changes, operator skills, and equipment maintenance, leading to a notable reduction in task completion time. \\
            Artificial Rabbits Optimization (ARO) \citep{kuo2025artificial} & Optimized control parameters of biped robots, enabling fast adaptation to external disturbances and maintaining balance in complex environments such as disaster response scenarios. \\
            Improved Ant Colony Optimization (IACO) \citep{wangying2025scheduling} & Applied to helicopter route planning for forest fire prevention, reducing total flight distance by 4.1\% and improving emergency response efficiency by 7.3\% in dynamic environments. \\
            \bottomrule
        \end{tabular}
    \end{table*}
    
        The core idea of metaheuristic algorithms is to build nature-inspired mathematical models and perform global search by simulating biological behaviors, physical phenomena, or natural laws. According to the No Free Lunch Theorem \citep{wolpert2002no}, no single algorithm can solve all optimization problems optimally, which highlights the importance of structural innovation tailored to specific problem characteristics.
        
        The performance of metaheuristic algorithms largely depends on the balance between exploration and exploitation \citep{arani2013improved}. However, most existing algorithms overly rely on the current best solution as a guiding mechanism. For instance, Particle Swarm Optimization (PSO), proposed by Kennedy et al. based on bird foraging behavior \citep{kennedy1995particle}; Grey Wolf Optimizer (GWO), introduced by Mirjalili et al. based on the cooperative hunting strategy of grey wolves \citep{mirjalili2014grey}; and Black Winged Kite Algorithm (BKA), proposed by Wang et al. based on the aerial hunting behavior of black-winged kites \citep{wang2024black}. While these algorithms perform well in specific applications, experimental results show that excessive dependence on the current best solution tends to reduce population diversity, leading to premature convergence \citep{gao2025freedom}. This brings forth the first research question: How to design more reasonable guidance mechanisms to maintain population diversity and effectively avoid local optima in complex search spaces?

        On the other hand, many metaheuristic algorithms adopt only a single search strategy, making it difficult to dynamically balance exploration and exploitation throughout the entire optimization process. For example, both PSO and the PID-based Search Algorithm (PSA) proposed by Gao \citep{gao2023pid} suffer from this limitation. Although JADE, proposed by Zhang et al. \citep{zhang2009jade}, introduces dynamic parameter adaptation to enhance search diversity, experimental results still show that JADE’s population diversity remains limited. This leads to the second research question: How to achieve a more efficient balance between exploration and exploitation through multi-strategy collaboration and adaptive regulation mechanisms?

        In recent years, an increasing number of researchers have drawn attention to the phenomenon of over-metaphorization in metaheuristic algorithm design \citep{sorensen2015metaheuristics}. These algorithms tend to rely heavily on natural metaphors while overlooking the underlying mathematical principles. Although such algorithms have achieved success in some engineering applications, their abstract nature and lack of generalizability limit both extensibility and innovation potential \citep{camacho2023exposing}. Therefore, designing metaphor-free metaheuristic algorithms with strong performance and practical value has become a new research trend.

        Several studies have made progress in this direction. For example, Logistic Gauss Circle Optimizer, proposed by Wang et al. \citep{wang2025logistic}, though inspired by three types of chaotic maps, essentially regulates exploration and exploitation through the cooperative use of Logistic, Gauss, and Circle chaos mechanisms. Similarly, the Delta Plus algorithm developed by Gao et al. completely abandons metaphor-driven design and instead constructs an optimization framework based purely on the general search principles of metaheuristics, offering a new perspective for the development of metaphor-free algorithms.

        Based on the above motivations, we propose a metaphor-free metaheuristic algorithm inspired by the dogfight behavior of fighter jets, called Dogfight Search (DoS). Different from traditional algorithms such as Genetic Algorithm (GA) \citep{holland1992genetic} and PSO, DoS introduces a dual formation structure and distinguishes the solutions within the formation as "leader aircraft" and "wing aircraft", where wing aircraft possess the ability to dynamically select search strategies to achieve adaptive adjustment between exploration and exploitation. In the search process, DoS incorporates the displacement integration equations from kinematics \citep{zohar2011mobile}, coordinating five search strategies through the adjustment of direction vectors, flight speed, and flight time, thereby enhancing the overall algorithm performance. Although the concept of DoS originates from the dogfight behavior of fighter jets, its search mechanism is detached from metaphors and it achieves results superior to SOTA algorithms in CEC2017 and CEC2022 benchmark test functions. The main contributions of this paper are:

        \begin{enumerate}[(1)]
            \item A metaphor-free metaheuristic algorithm, named DoS, is proposed;
            \item In CEC2017 (30, 50, 100 dimensions) and CEC2022 (10, 20 dimensions) benchmark test functions, DoS demonstrated significantly superior performance compared to seven advanced competitors; Furthermore, DoS achieved the top Friedman ranking when compared against three SOTA algorithms: LSHADE \citep{tanabe2014improving}, LSHADE-SPACMA \citep{mohamed2017lshade} and AL-SHADE \citep{li2022novel}, demonstrating strong competitiveness;
            \item DoS ranks first in overall performance on real-world engineering optimization problems. It remains statistically superior in problems with 91 inequality constraints, confirming its engineering applicability;
            \item In mountainous terrain path planning tasks with and without no-fly zones, DoS also demonstrates significantly superior performance compared to competitors.
        \end{enumerate}

        The rest of this paper is organized as follows: Section 2 introduces related research on metaheuristic algorithms; Section 3 describes the design inspiration and mathematical model of DoS; Section 4 evaluates DoS numerically on multidimensional CEC2017 and CEC2022 benchmark test functions; Section 5 verifies the performance of DoS on solving constrained optimization problems through various real-world tasks; Section 6 further validates the advantages of DoS in solving mountainous terrain path planning problems; Section 7 discusses the advantages and limitations of DoS; Section 8 summarizes the work and points out potential future research directions based on DoS.

    \section{Related studies}

\begin{figure*}[htbp!]
            \centering
            \includegraphics[width=0.6\linewidth]{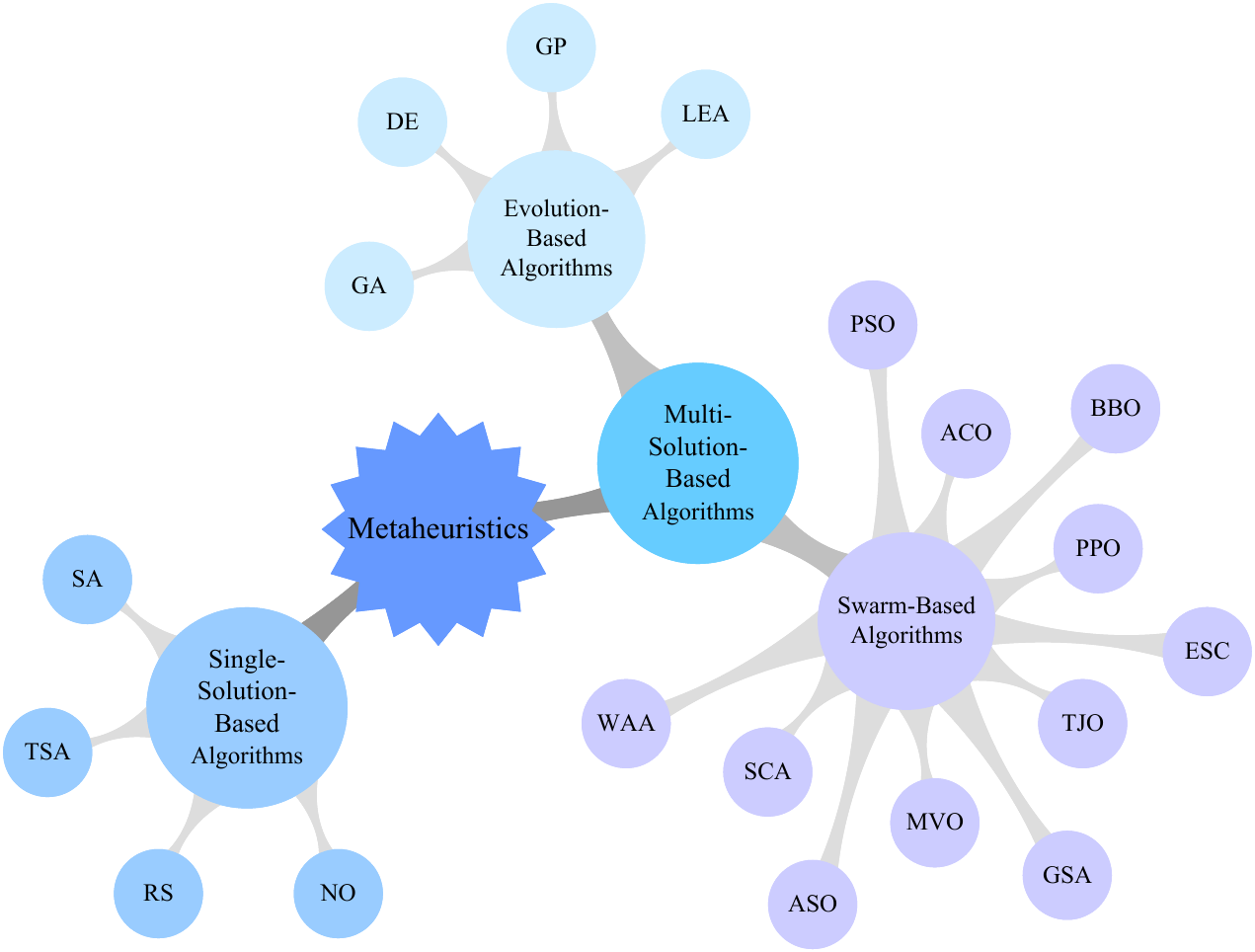}
            \caption{The classification of metaheuristic algorithms.}
            \label{fig1}
        \end{figure*}

        The classification of metaheuristic algorithms is mainly reflected in two aspects. As shown in Fig.~\ref{fig1}, based on the number of solutions, algorithms can be categorized into two major types: single-solution-based algorithms and multi-solution-based algorithms.

        The classification of single-solution-based algorithms lies in their search process relying on a single-solution. A representative example is the Simulated Annealing (SA) proposed by Kirkpatrick et al. in 1983 \citep{kirkpatrick1983optimization}, which was inspired by the energy changes in the cooling process of solids. Due to the limited global information obtained from a single-solution, SA introduces a probability mechanism for accepting worse solutions to escape local optima and enhance exploration ability. Other representative algorithms include the Tabu Search Algorithm (TSA) \citep{prajapati2020tabu}, Random Search (RS) \citep{zabinsky2009random} and Non-monopolize Search (NO) \citep{abualigah2024non}. Although this type of algorithm has advantages in convergence speed, its low information utilization efficiency often makes it difficult to deal with complex high dimensional problems, which in turn has promoted the development of multi-solution-based algorithms.

        Multi-solution-based algorithms utilize information from multiple solutions to guide the search, thereby enhancing global search ability. Although the design inspirations of such methods are diverse, including animal behaviors, natural phenomena, physical laws and mathematical functions, their classification depends on the search mechanisms. Multi-solution-based algorithms can be further categorized into evolution-based algorithms and swarm-based algorithms \citep{ezugwu2021metaheuristics}.
        
        Evolution-based algorithms are based on Darwin's theory of evolution and stress genetic crossover and mutation operations on decision variables throughout the optimization process. A typical representative is the Genetic Algorithm (GA) proposed by Forrest et al., which continuously optimizes candidate solutions through mechanisms such as selection, crossover and mutation. Other evolutionary approaches include Differential Evolution (DE) \citep{storn1997differential}, Genetic Programming (GP) \citep{koza1994genetic} and Love Evolution Algorithm (LEA) \citep{gao2024love}.

        Swarm-based algorithms form the largest subclass, drawing inspiration from biological behaviors, physical processes, and mathematical models. Among them, typical algorithms inspired by biological activities include PSO and Ant Colony Optimization (ACO) \citep{dorigo1991ant}. The latter was proposed by Dorigo in 1991 and simulates the pheromone mechanism of ants for optimization. A representative of physics inspired algorithms is the Gravitational Search Algorithm (GSA) \citep{rashedi2009gsa}, which simulates gravitational interactions between solutions to achieve global guidance. Mathematics inspired algorithms include the Sine Cosine Algorithm (SCA) \citep{mirjalili2016sca}, which leverages the periodicity of trigonometric functions to enhance search diversity. In addition to the above typical methods, swarm-based algorithms also include: Beaver Behavior Optimizer (BBO) \citep{ouyang2025beaver}, Philoponella Prominens Optimizer (PPO) \citep{gao2025escape}, Escape Algorithm (ESC) \citep{ouyang2024escape}, Traffic Jam Optimizer (TJO) \citep{wang2025traffic}, Multi-Verse Optimizer (MVO) \citep{mirjalili2016multi}, Atom Search Optimization (ASO) \citep{zhao2019atom} and Weighted Average Algorithm (WAA) \citep{cheng2024weighted}.

        Although these algorithms perform well in various problems, most swarm-based algorithms share two common shortcomings. First, the guidance information overly depends on the population’s best solution and lacks diversified guidance, making the search prone to local optima. Second, the number of search strategies is usually no more than three, resulting in monotonous search behavior. These limitations reduce their search efficiency in complex feasible regions.

        To address the above issues, the proposed DoS introduces multiple guiding solutions in its design to enhance diversity. In addition, DoS assigns wing aircrafts a dynamic search strategy selection mechanism to effectively balance exploration and exploitation.
        
    \section{Methodology}
        
        This section will introduce the design inspiration, mathematical model, pseudocode and theoretical analysis of the computational complexity of DoS.

        \subsection{Inspiration}

            Dogfighting \citep{ashraf2021dogfight} is a highly dynamic air combat behavior, which is typically divided into two phases: attack and evasion. In the attack phase, fighter jets carry out strikes by chasing and locking onto enemy aircraft and launching missiles. In the evasion phase, they rely on irregular and chaotic flight trajectories to escape enemy lock-on and missile tracking \citep{greenwood1992differential}. Two fighter jet formations strive to achieve the best combat readiness through adjustments to their positions and combat strategies, thereby securing victory. This process bears a striking resemblance to the optimization solutions employed in metaheuristic algorithms. This process is highly similar to the optimization solving process in metaheuristic algorithms. Therefore, inspired by this behavior, we designed five types of search strategies for DoS: free flight, maneuver lock-on, missile attack, maneuver evasion and flare evasion. To better explain the algorithm design concept, we abstract the inspirations as follows:

            \begin{enumerate}[(1)]
                \item Solution: the position of a fighter jet. We divide the fighter group into two formations, which are adversarial to each other;
                \item Feasible domain: the airspace in which the fighter jets conduct their search;
                \item Function value: the combat state of the fighter jet. For minimization problems, the better the combat state, the smaller the function value;
                \item Global optimal solution: the location in the airspace with the best combat state.
            \end{enumerate}

            \begin{figure*}[htbp!]
                \centering
                \begin{tabular}{\cm{0.48\textwidth}\cm{0.48\textwidth}}
                    \multicolumn{2}{c}{\includegraphics[width=0.96\linewidth]{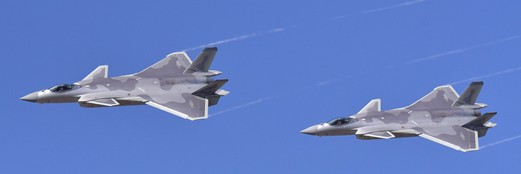}} \\
                    \multicolumn{2}{c}{(a) Free flight strategy} \\
                    \includegraphics[width=0.95\linewidth]{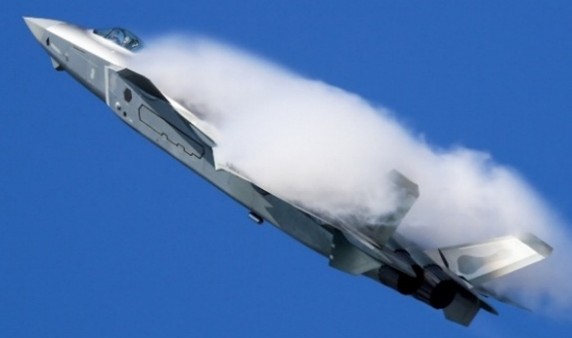} & \includegraphics[width=0.95\linewidth]{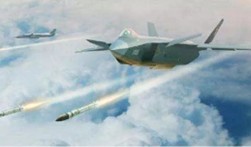} \\
                    (b) Maneuver lock-on strategy & (c) Missile attack strategy \\
                    \includegraphics[width=0.95\linewidth]{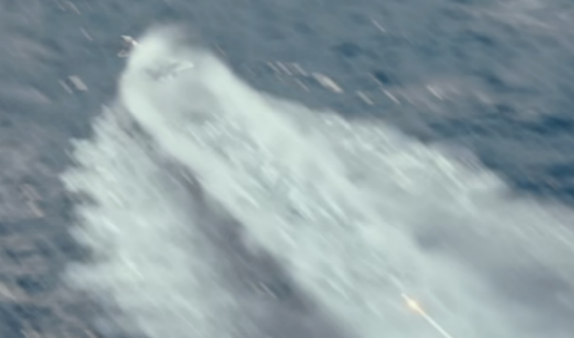} & \includegraphics[width=0.95\linewidth]{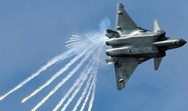} \\
                    (d) Maneuver evasion strategy & (e) Flare evasion strategy
                \end{tabular}
                \caption{The inspiration diagram of DoS search strategies.}
                \label{fig2}
            \end{figure*}

            Fig.~\ref{fig2}(a) shows the inspiration source of the free flight strategy in DoS \citep{olsder1974role}. In this strategy, the leader aircraft and wing aircraft randomly search the airspace based on commands from the command center, while the wing aircraft continuously cooperates with the leader aircraft, enabling efficient exploration of the airspace. Fig.~\ref{fig2}(b) and (c) illustrate the maneuver lock-on and missile attack strategies, respectively. Both strategies focus on enemy aircraft as core targets and exhibit strong convergence and guidance, promoting local exploitation and improving solution accuracy. Fig.~\ref{fig2}(d) and (e) illustrate the maneuver evasion and flare evasion strategies. In this phase, fighter jets rapidly disengage from the current airspace and move toward new flight regions to avoid threats. From the perspective of search behavior, this process shows clear global exploration capability. The synergy of these five strategies enables the solutions to continuously move toward better regions, thereby improving overall search efficiency.

            It is important to emphasize that although DoS is inspired by dogfighting behavior, its search strategies are mathematically modeled based on the displacement integration equations in kinematics. Specifically, we conduct diversified modeling of key elements such as the direction vector, flight speed and flight duration in DoS, thereby achieving metaphor-free search operations.

        \subsection{Mathematical model of the search strategy}
            
\begin{figure*}[htbp!]
                \centering
                \includegraphics[width=0.7\linewidth]{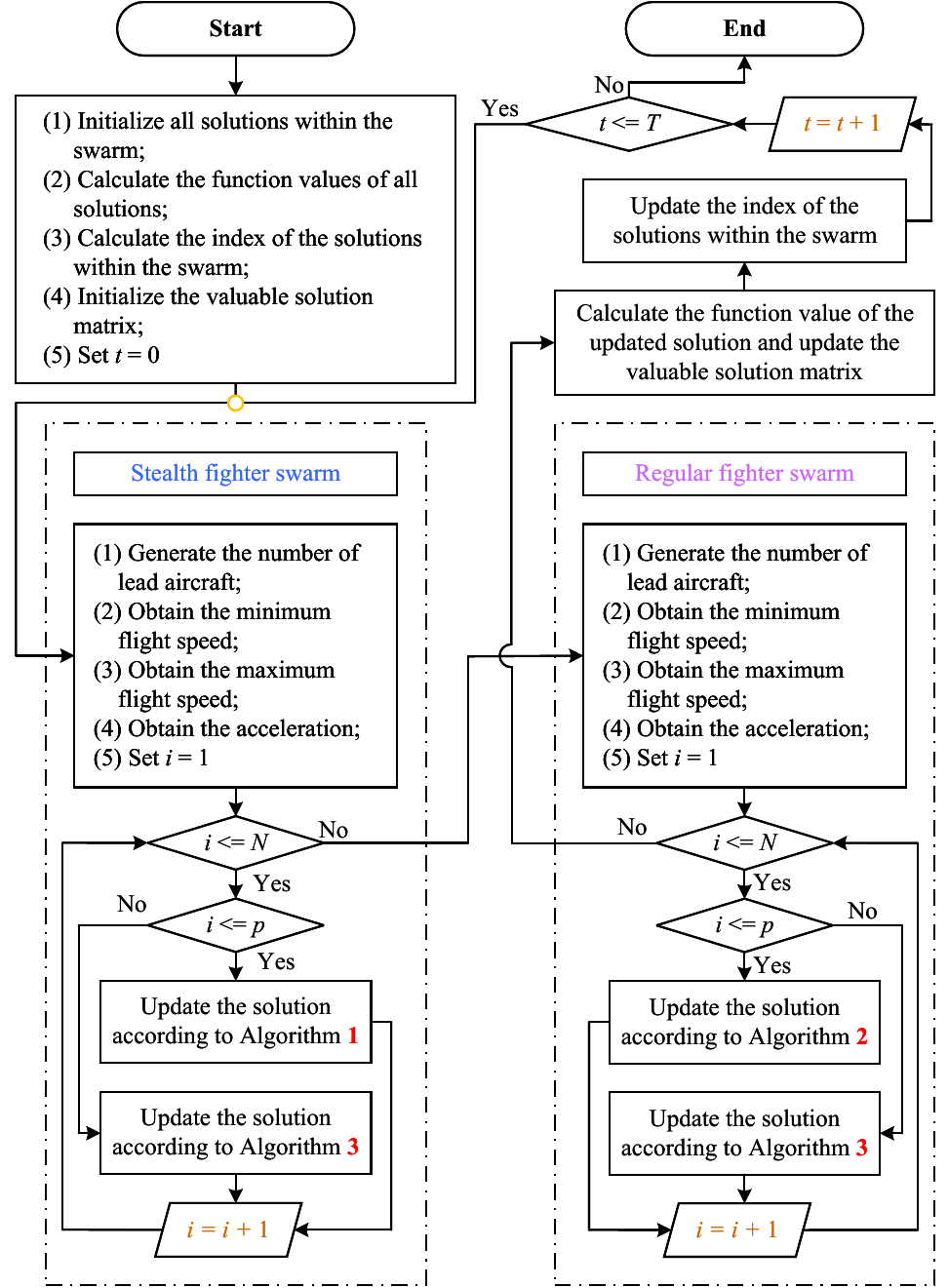}
                \caption{The iterative flowchart of DoS.}
                \label{fig3}
            \end{figure*}

            The workflow of DoS is shown in Fig.~\ref{fig3}, which sequentially includes initialization (see Section 3.2.1), the calculation of the number of leader aircrafts and the influencing factors of search strategies (see Sections 3.2.1 and 3.2.2), the search strategies of the two formations (see Sections 3.2.2 to 3.2.4) and the dynamic selection mechanism of wing aircraft (see Section 3.3).

            To facilitate understanding of the dual formation design in DoS, Section 3.2.1 first introduces the construction of the dual formation and the method for calculating the number of leader aircrafts in each iteration; Section 3.2.1 also explains the initialization strategy tailored to the dual formation design of DoS. Considering that the five types of search strategies follow the same fundamental principles, Section 3.2.2 provides a unified explanation of the common principles behind these strategies, including the mathematical meanings of key factors such as direction vector, flight speed and flight duration. Subsequently, Sections 3.2.3 to 3.2.5 respectively present the specific settings of the relevant parameters in the five search strategies.

            \subsubsection{Initialization of Dual Formations}

                To meet the requirements of the dual formation structure in DoS, we assume that the total number of formation solutions ($N$) is an even number. Among them, $N/2$ solutions are defined as the stealth fighter formation ($\bm{X}$) and the remaining $N/2$ solutions are defined as the regular fighter formation ($\bm{Y}$).

                To avoid convergence to local optima caused by relying on a single guiding solution, leader aircrafts serve as the guiding solutions and wing aircrafts adopt search strategies through a dynamic selection mechanism. Furthermore, to broaden the exploration range of the formation, the number of leader aircrafts is randomly selected within a defined interval, as expressed in Equation~\ref{eq1}.
                \begin{equation}
                    p=\text { round }\left\{\left[k_{1}+\left(\frac{1}{2}-k_{1}\right)-r_{1}\right]-\frac{N}{2}\right\}
                    \label{eq1}
                \end{equation}

                \noindent
                where $p$ denotes the number of leader aircrafts, $k_{1}$ is a tunable hyperparameter in DoS and $r_{1}$ is a random number uniformly distributed in the interval $[0, 1]$. It can be seen that the number of leader aircrafts is constrained within the range $[k_{1}N, 0.25N]$, which not only expands the exploration radius of the formation but also reserves a sufficient proportion of wing aircrafts to better achieve a balance between exploration and exploitation.

                Considering that DoS adopts a dual formation structure and integrates multiple search strategies, more dispersed initial formations yield more effective searches. Therefore, this paper proposes an initialization strategy different from traditional methods, whose form is shown in Equation~\ref{eq2}.
                \begin{equation}
                    \begin{aligned}
                        \bm{X}^{ij} &= \bm{L}^{j} + \left( \frac{\bm{U}^{j} + \bm{L}^{j}}{2} - \bm{L}^{j} \right) r_{2}, \\
                        \bm{Y}^{ij} &= \frac{\bm{U}^{j} + \bm{L}^{j}}{2} + \left( \bm{U}^{j} - \frac{\bm{U}^{j} + \bm{L}^{j}}{2} \right) r_{2}, \\
                        i &= 1,2,\ldots,N/2, \quad j = 1,2,\ldots,D
                    \end{aligned}
                    \label{eq2}
                \end{equation}

                \noindent
                where $\bm{X}$ represents the stealth fighter formation, $\bm{Y}$ represents the regular fighter formation, $\bm{U}$ and L are the upper and lower bound vectors of the decision variables, $D$ is the problem dimension and $r_{2}$ is a random number uniformly distributed within $[0, 1]$. This initialization strategy divides the feasible domain into two sub intervals, upper and lower based on the hyperplane formed by the midpoints of each dimension \citep{winder1966partitions}, which are then used as the initialization ranges for the stealth fighter formation and the regular fighter formation, respectively.

                By placing the two formations in spatially distinct regions based on this domain partitioning, the strategy enhances the spread of solutions in the early search stage. This spatial separation also supports the subsequent attack and evasion strategies, thereby improving the exploration efficiency of DoS.
                
                \subsubsection{Common Principles of Search Strategies}

                    Although DoS uses dogfighting behavior as an external expression, the mathematical models of the 5 search strategies are based on the kinematics and displacement integration equations shown in Equation~\ref{eq3}.
                    \begin{equation}
                        X_{new}^{i} = X^{i} + \int_{0}^{\Delta \xi} \bm{u} \cdot v \, d\xi, \quad i = 1,2,\ldots,N/2
                        \label{eq3}
                    \end{equation}
                
                    \noindent
                    where $\bm{u}$ denotes the direction vector, v denotes the flight speed and   represents the flight duration. Since DoS sorts the solutions within each formation in ascending order based on their function values at the end of each iteration, solutions with better function values are ranked earlier. The top p solutions are defined as leader aircrafts and the remaining solutions are defined as wing aircrafts. In Sections 3.2.2 to 3.2.4, based on the distinction between leader aircrafts and wing aircrafts, the direction vector u is further divided into $\bm{u_{leader}}$ and $\bm{u_{wing}}$.

                    As shown in Equation~\ref{eq3}, regardless of which search strategy is adopted, the update of a solution depends on three key factors: u, v and  . This paper leverages differentiated definitions of these three factors to implement the effects of the five search strategies.

                    \textbf{Decomposition of the direction vector:}

                    In most metaheuristic algorithms, a solution’s direction is typically determined by intraformation information alone. As a result, it is often difficult to escape local optima when solving complex optimization problems. To address this, DoS decomposes $u$ into two components: pilot guidance ($\bm{u^{\mathrm{Pilot}}}$) and command guidance ($\bm{u^{\mathrm{Head}}}$). Specifically:
                
                    \begin{enumerate}[(1)]
                        \item $\bm{u^{\mathrm{Pilot}}}$: the direction vector determined by intra-formation pilot information;
                        \item $\bm{u^{\mathrm{Head}}}$: the direction vector determined by informative and high-quality solutions.
                    \end{enumerate}

                    Therefore, uHead helps solutions escape from local optima in a timely manner and explore new potential regions. Since uPilot must be designed to meet the exploration and exploitation needs of different strategies, its definition is elaborated in Sections 3.2.2 to 3.2.4. The mathematical form of uHead is shown in Equation~\ref{eq4}.
                    \begin{equation}
                        \begin{aligned}
                            &R_1 = \mathrm{round}(NR \cdot r_3), \quad R_2 = \mathrm{round}(NP \cdot r_4), \\
                            &\bm{u^{\mathrm{Head}}} = \bm{X}^{R_{1}} - \bm{Z}^{R_{2}}
                        \end{aligned}
                        \label{eq4}
                    \end{equation}

                    where $R_{1}$ denotes the index of a randomly selected high-quality solution, $NR$ denotes the number of high quality solutions (fixed at $0.1N$), $Z$ is the storage matrix of valuable solutions, $R_{2}$ denotes the index of a randomly selected valuable solution, $NP$ denotes the number of stored valuable solutions (with a maximum of $2.5N$) and $r_{3}$ and $r_{4}$ are random numbers uniformly distributed in the range $[0, 1]$.

                    \textbf{Minimum speed, maximum speed and acceleration:}

                    In DoS, the motion state of solution updates is modeled as either uniform or uniformly accelerated motion. Therefore, $v$ is determined by three influencing factors: minimum speed ($v_{min}$), maximum speed ($v_{max}$) and acceleration ($a$). As shown in Equation~\ref{eq3}, $v$ is closely related to the step size of solution updates. If speed can be appropriately quantified to guide the update of solutions toward positions with lower function values, the search efficiency can be significantly improved. To achieve this, DoS records promising solutions (i.e., those with decreasing function values) at the end of each iteration and uses the velocity and function value information of these promising solutions to update the initial velocity for the next iteration. Details are shown in Equation~\ref{eq5}.
                    \begin{equation}
                        V_{new} = \frac{\sum_{i\in \Omega} \bm{W}_i \cdot \bm{V}_i^2}{\sum_{i\in \Omega} \bm{W}_i \cdot \bm{V}_i}, \quad \bm{W}_i = \frac{d\bm{f}_i}{\left| \sum_{i\in \Omega} d\bm{f}_j \right|}
                        \label{eq5}
                    \end{equation}

                    \noindent
                    where $V_{new}$ denotes the initial velocity for the next iteration, $\Omega$ represents the set of indices of promising solutions, $\bm{V}_{i}$ and $d\bm{f}_{i}$ denote the flight velocity and function value difference of the $i$-th promising solution, respectively. To enhance solution diversity, DoS adds real-valued random perturbations to the initial velocity, as shown in Equation~\ref{eq6}.
                    \begin{equation}
                        V = V_{new} + \tan\left[ \frac{\pi}{2} (r_5 - 0.5) \right] / 10
                        \label{eq6}
                    \end{equation}

                    \noindent
                    where $r_{5}$ is a random number uniformly distributed in the range $[0, 1]$ and the velocity $V$ obtained from Equation~\ref{eq6} can be further used to define the minimum speed, maximum speed and acceleration of a solution in the next iteration, as shown in Equation~\ref{eq7}.
                    \begin{equation}
                        \begin{aligned}
                            &v_{\min} = \min\{k_5 V, V\}, \quad v_{\max} = \max\{k_5 V, V\}, \\
                            &a = (v_{\max} - v_{\min}) / 1.2
                        \end{aligned}
                        \label{eq7}
                    \end{equation}

                    \noindent
                    where $k_{5}$ is a tunable hyperparameter in DoS and belongs to the interval $[0, 1]$.

                    \textbf{Flight duration}:

                    In Equation~\ref{eq3}, $\Delta \xi$ is also a key factor shaping the solution’s update radius. Since $v$ has already achieved sufficient diversification of the movement radius, this paper $\Delta \xi$ designs as a control factor for diversified search regions, allowing solutions to be updated within a certain range. The specific formulation is shown in Equation~\ref{eq8}.
                    \begin{equation}
                        \Delta \xi = 0.8 + 0.4 r_6
                        \label{eq8}
                    \end{equation}

                    \noindent
                    where $r_{6}$ is a random number uniformly distributed in the range $[0, 1]$ and thus is a uniformly distributed random value within the interval $[0.8, 1.2]$. This setting allows solutions to perform fine grained exploitation within a small range while also expanding the exploration radius over a larger range, thereby improving overall search efficiency.

            \subsubsection{Free Flight Strategy}

                In this subsection and the subsequent descriptions of the mathematical models for search strategies, we focus on the design of the direction vector and flight speed. It is important to note that, to improve the search efficiency of DoS in the early iterations, the two formations differ only in the strategy selection mechanism during the early stage (see Section 3.3), while their mathematical expressions remain identical. Therefore, this paper uses the stealth fighter formation as the unified example for explanation.

                In the early stage of iteration, due to the lack of prior knowledge about the solution space of the optimization problem, it is necessary to conduct global exploration of the feasible domain. As information about the solution space accumulates, a gradual balance between exploration and exploitation should be achieved. Based on this, a “free flight” strategy is designed for DoS. Under this strategy, the direction vectors of leader aircrafts and wing aircrafts are defined as follows, as shown in Equation~\ref{eq9}.
                \begin{equation}
                    \begin{aligned}
                        &\bm{u_{\mathrm{leader}}^{\mathrm{Pilot}}} = \bm{X_{rand}} - \bm{X}^{i}, \quad \bm{u_{\mathrm{wing}}^{\mathrm{Pilot}}} = \bm{X}^{H} - \bm{X}^{i}, \\
                        &\bm{X_{rand}} = \bm{L} + (\bm{U} - \bm{L}) r_7
                    \end{aligned}
                    \label{eq9}
                \end{equation}

                \noindent
                where leader denotes the leader aircraft, wing denotes the wing aircraft, $H$ indicates the index of the leader aircraft followed by the wing aircraft and $r_{7}$ is a random number uniformly distributed in the range $[0, 1]$. Under this mechanism, leader aircrafts conduct random exploration within the feasible domain, while wing aircrafts perform local exploitation along the flight path of leader aircrafts, thereby significantly improving the discovery efficiency of high-quality solutions during the early stages of iteration.

\begin{figure*}[htbp!]
                    \centering
                    \includegraphics[width=0.75\linewidth]{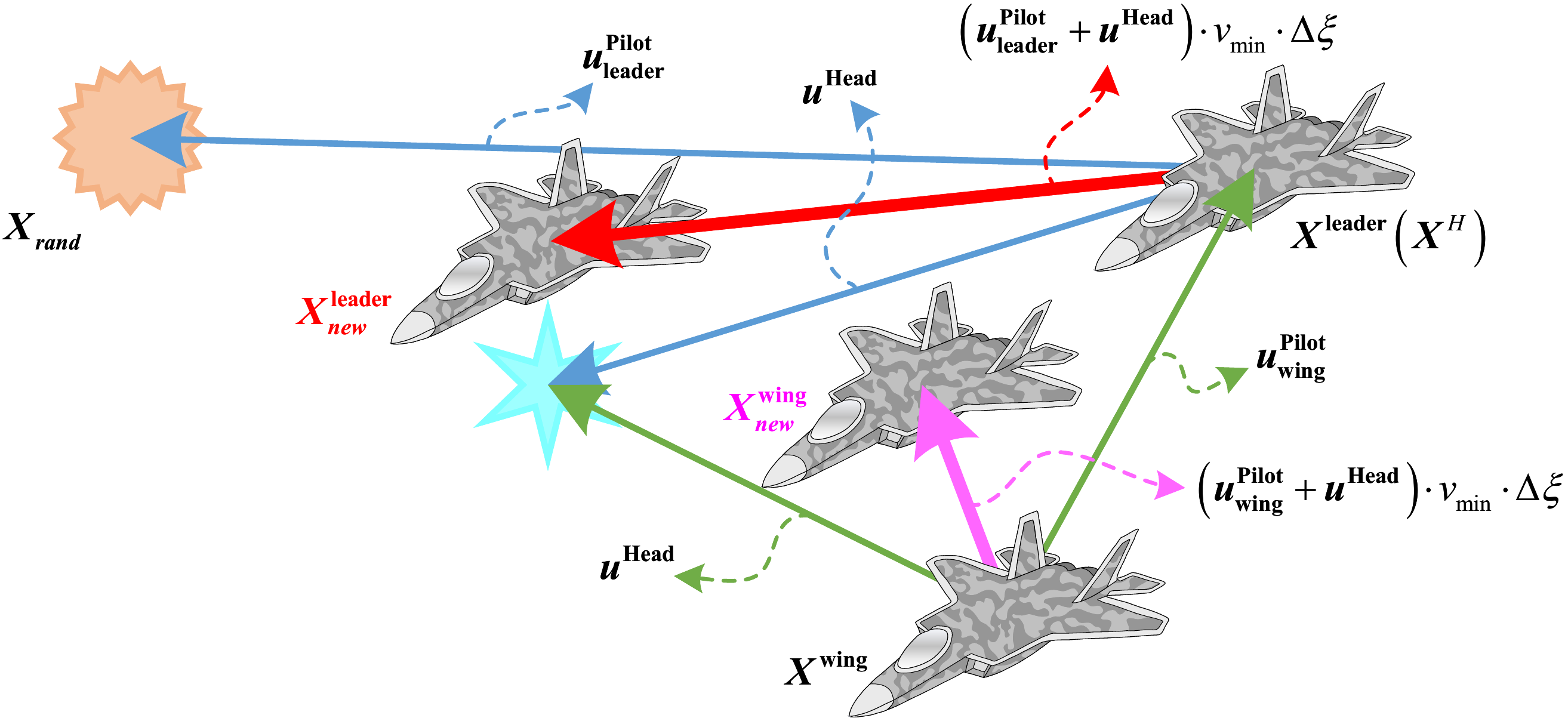}
                    \caption{Visualization of the free flight strategy.}
                    \label{fig4}
                \end{figure*}

                Since an excessively large step size may cause the update process to miss some valuable solutions, the motion state under this strategy is modeled as uniform motion with a speed of $v_{min}$. By incorporating the flight duration defined in Equation~\ref{eq8} and substituting it into Equation~\ref{eq3}, the mathematical model of the free flight strategy is obtained, as shown in Equation~\ref{eq10}.
                \begin{equation}
                    \begin{cases}
                        \bm{X_{new}}^{i} = \bm{X}^{i} + \left( \bm{u_{\mathrm{leader}}^{\mathrm{Pilot}}} + \bm{u^{\mathrm{Head}}} \right) v_{\min} \cdot \Delta \xi, & i \le p \\
                        \bm{X_{new}}^{i} = \bm{X}^{i} + \left( \bm{u_{\mathrm{wing}}^{\mathrm{Pilot}}} + \bm{u^{\mathrm{Head}}} \right) v_{\min} \cdot \Delta \xi, & i > p
                    \end{cases}
                    \label{eq10}
                \end{equation}

                where $\bm{X_{new}}^{i}$ denotes the updated position of the $i$-th leader aircraft or wing aircraft. To intuitively demonstrate the behavioral characteristics of this strategy, it is visualized in a two dimensional plane, as shown in Fig~\ref{fig4}.

            \subsubsection{Offensive Strategy}

                Although the free flight strategy can achieve a dynamic balance between exploration and exploitation in the early stages of iteration, relying on a single search strategy throughout the entire iteration process will inevitably limit the diversity of search behavior. To address this, two offensive strategies are designed for DoS to enhance its local exploitation ability around valuable solutions. These strategies are inspired by the processes of pursuing and locking onto enemy aircraft, as well as missile attacks during dogfights and are thus named the maneuver lock-on and missile attack strategies. The mathematical models of these strategies are described below.

                \textbf{Maneuver lock-on strategy:}

                Most metaheuristic algorithms rely on a single population, which makes them prone to falling into local optima, resulting in many iterations without meaningful improvement. Sudholt \citep{sudholt2019benefits} pointed out that increasing the number of populations can improve search diversity. Based on this, DoS leverages the advantages of its dual formation structure, allowing both offensive strategies to obtain exploitation information from the opposing formation. The computation function for the pilot guided direction vector in the maneuver lock-on strategy is shown in Equation~\ref{eq11}.
                \begin{equation}
                    \begin{aligned}
                        &\bm{u_{1}^{\mathrm{Pilot}}} = \bm{Y}^{h} - \bm{X}^{i}, \quad \bm{u_{2}^{\mathrm{Pilot}}} = \bm{Y_{pre}}^{h} - \bm{X}^{i}, \\
                        &\bm{u^{\mathrm{Pilot}}} = r_{8} \cdot \bm{u_{1}^{\mathrm{Pilot}}} + (1 - r_{8}) \bm{u_{2}^{\mathrm{Pilot}}}
                    \end{aligned}
                    \label{eq11}
                \end{equation}

                \noindent
                where $\bm{Y}$ denotes the coordinate matrix of the regular fighter formation, $h$ is the index of the selected regular fighter, $\bm{Y_{pre}}^{h}$ represents the predicted position and $r_{8}$ is a random number uniformly distributed in the range $[0, 1]$. According to the equation, the $\bm{u^{\mathrm{Pilot}}}$ under the maneuver lock-on strategy is no longer distinguished between leader aircraft and wing aircraft. Instead, it is determined by two sub vectors:

                \begin{enumerate}[(1)]
                    \item $\bm{u_{1}^{\mathrm{Pilot}}}$: determined based on the solution information from the opposing formation;
                    \item $\bm{u_{2}^{\mathrm{Pilot}}}$: determined based on both the opposing formation and the best solution information, with the definition of $\bm{Y_{pre}}^{h}$ given in Equation~\ref{eq12}.
                \end{enumerate}
                \begin{equation}
                    \bm{Y_{pre}}^{h} = \bm{Y}^{h} + \left( \bm{Y^{best}} - \bm{Y}^{h} \right) r_{9}, \quad h \leq p
                    \label{eq12}
                \end{equation}

                \noindent
                where $r_{9}$ is a random number uniformly distributed in the interval $[0, 1]$ and both $\bm{Y}^{h}$ and $\bm{Y^{best}}$ are leader aircraft within the regular fighter formation. The linear combination of these two yields $\bm{Y_{pre}}^{h}$, which possesses high exploitation value. Since $\bm{Y_{pre}}^{h}$ does not belong to the opposing formation, this design ensures exploitation while also effectively preventing the two formations from converging prematurely.

                To enrich the distribution diversity of solutions after executing the maneuver lock-on strategy, the motion state is modeled as uniformly accelerated motion. The corresponding function describing velocity as a function of time is given in Equation~\ref{eq13}.
                \begin{equation}
                    v_{\xi} = v_{\min} + a \cdot \xi
                    \label{eq13}
                \end{equation}

                By combining Equations~\ref{eq11}, ~\ref{eq13}, ~\ref{eq8} and~\ref{eq3}, the complete mathematical formulation of the maneuver lock-on strategy can be derived, as shown in Equation~\ref{eq14}.
                \begin{equation}
                    \begin{aligned}
                        \bm{X_{new}}^{i} &= \bm{X}^{i} + \int_{0}^{\Delta \xi} \left( \bm{u^{\mathrm{Pilot}}} + \bm{u^{\mathrm{Head}}} \right) v_{\xi} \cdot d\xi \\
                        &= \bm{X}^{i} + \left( \bm{u^{\mathrm{Pilot}}} + \bm{u^{\mathrm{Head}}} \right) \left( v_{\min} + \frac{1}{2} a \cdot \Delta \xi^{2} \right), \\
                        i &= 1,2,\ldots,\frac{N}{2}
                    \end{aligned}
                    \label{eq14}
                \end{equation}

\begin{figure*}[htbp!]
                    \centering
                    \includegraphics[width=0.75\linewidth]{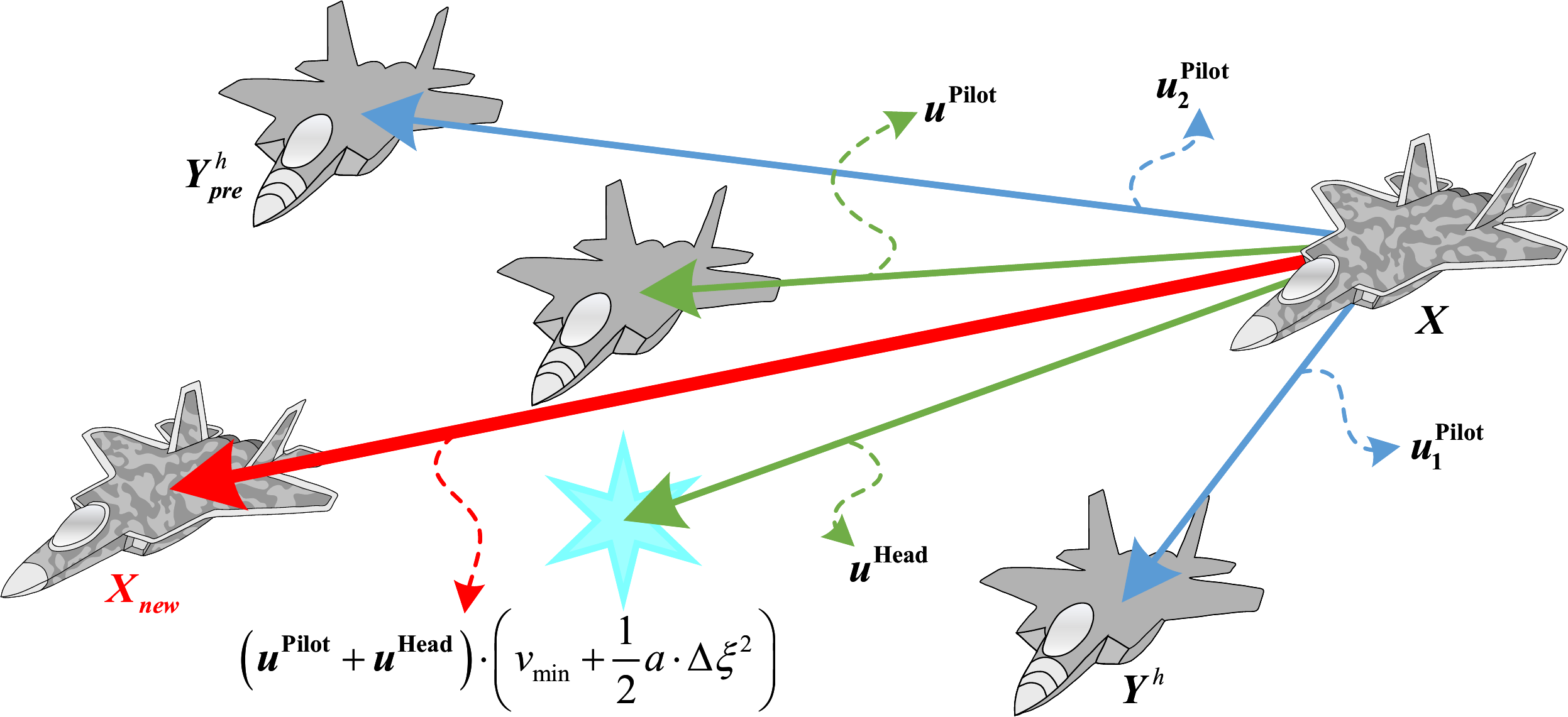}
                    \caption{Visualization of the maneuver locking strategy.}
                    \label{fig5}
                \end{figure*}

                To intuitively demonstrate the search behavior of this strategy, the solution update process is visualized in a 2 dimensional space, as shown in Fig.~\ref{fig5}.

                \textbf{Missile attack strategy:}

                Although the maneuver lock-on strategy can prevent the two formations from converging prematurely, the algorithm’s reliance on exploitation does not completely eliminate potential risks. When the solution approaches the region containing the global optimal solution, if DoS relies solely on the maneuver lock-on strategy for exploitation, it may be misled by attractive local optima and fail to further explore the global optimal solution region. To address this limitation, a missile attack strategy is proposed in this paper. This strategy relies exclusively on information from the opposing formation. The definition of $\bm{u^{\mathrm{Pilot}}}$ under this strategy is provided in Equation~\ref{eq15}.
                \begin{equation}
                    \bm{u^{\mathrm{Pilot}}} = \bm{Y}^{h} - \bm{X}^{i}, \quad h \leq p
                    \label{eq15}
                \end{equation}

                Although this setting may cause the two formations to converge prematurely to the same solution region, the execution frequency of this strategy has been appropriately reduced within the iterative process of DoS. As long as the majority of solutions still adopt the maneuver lock-on strategy, this issue can be effectively avoided.

                Since the missile attack strategy requires solutions to move away from their original regions and gradually approach the global optimal, the motion state under this strategy is modeled as uniformly decelerated motion. The corresponding velocity function is given in Equation~\ref{eq16}.
                \begin{equation}
                    v_{\xi} = v_{\max} - a \cdot \xi
                    \label{eq16}
                \end{equation}

                By combining Equations~\ref{eq15}, ~\ref{eq16}, ~\ref{eq8} and~\ref{eq3}, the complete mathematical formulation of the missile attack strategy can be derived, as shown in Equation~\ref{eq17}.
                \begin{equation}
                    \begin{aligned}
                        \bm{X_{new}}^{i} &= \bm{X}^{i} + \int_{0}^{\Delta \xi} \left( \bm{u^{\mathrm{Pilot}}} + \bm{u^{\mathrm{Head}}} \right) v_{\xi} \cdot d\xi \\
                        &= \bm{X}^{i} + \left( \bm{u^{\mathrm{Pilot}}} + \bm{u^{\mathrm{Head}}} \right) \left( v_{\max} - \frac{1}{2} a \cdot \Delta \xi^{2} \right), \\
                        i &= 1,2,\ldots,\frac{N}{2}
                    \end{aligned}
                    \label{eq17}
                \end{equation}

                In summary, the exploitation mechanism of DoS integrates both the maneuver lock-on and missile attack strategies: the former utilizes high quality information beyond the opposing formation to achieve efficient exploitation and avoid convergence consistency, while the latter ensures deep exploration of the global optimal solution. Their synergy allows DoS to maintain exploitation diversity while improve global performance.
                
            \subsubsection{Evasive Strategy}

                The offensive strategies proposed in Section 3.2.4 have already satisfied the exploitation requirements of DoS; however, exploration is also a critical factor throughout different stages of iteration. If exploration is insufficient in the early stage, the efficiency of gathering information about the solution space will be low; if it is insufficient in the middle and later stages, the algorithm is prone to getting trapped in local optima and may fail to converge to the global optimal solution. To address the exploration needs in the early stage, the free flight strategy has been designed. For the middle and later stages, in order to further enhance the exploration capability of DoS, two evasion strategies are proposed, namely the maneuver evasion and flare evasion strategies. These strategies are inspired by the evasive maneuvers and flare release behaviors of fighter jets during dogfights.

                \textbf{Maneuver evasion strategy:}

                In the middle and later stages of iteration, the main role of exploration strategies is to help solutions escape local optima. Therefore, $\bm{u^{\mathrm{Pilot}}}$ should be decoupled from the current exploitation region and reoriented within the feasible domain. However, if completely random search, as in the free flight strategy, is adopted again, the optimization accuracy may be severely degraded. To avoid this, the mathematical model of $\bm{u^{\mathrm{Pilot}}}$ in this paper is based on the set of stored valuable solutions, as defined in Equation~\ref{eq18}.
                \begin{equation}
                    \bm{u^{\mathrm{Pilot}}} = \sum_{j=1}^{n} \frac{\bm{Z}^{j}}{n} - \bm{X}^{i}, \quad n = \mathrm{round}(NR \cdot r_{4})
                    \label{eq18}
                \end{equation}

                \noindent
                where $\bm{u^{\mathrm{Pilot}}}$ is determined by the centroid of half of the solutions in the valuable solution matrix. This design enables the strategy to depart from the original solution region while effectively suppressing ineffective exploration. To ensure that solutions have a sufficient movement radius and to enrich the diversity of movement ranges, the maneuver evasion strategy is modeled as uniformly accelerated motion. By combining Equations~\ref{eq18}, ~\ref{eq13}, ~\ref{eq8} and~\ref{eq3}, the complete mathematical formulation is derived, as shown in Equation~\ref{eq19}.
                \begin{equation}
                    \begin{aligned}
                        \bm{X_{new}}^{i} &= \bm{X}^{i} + \int_{0}^{\Delta \xi} \left( \bm{u^{\mathrm{Pilot}}} + \bm{u^{\mathrm{Head}}} \right) v_{\xi} \cdot d\xi \\
                        &= \bm{X}^{i} + \left( \bm{u^{\mathrm{Pilot}}} + \bm{u^{\mathrm{Head}}} \right) \left( v_{\min} + \frac{1}{2} a \cdot \Delta \xi^{2} \right), \\
                        i &= 1,2,\ldots,\frac{N}{2}
                    \end{aligned}
                    \label{eq19}
                \end{equation}

\begin{figure*}[htbp!]
                    \centering
                    \includegraphics[width=0.75\linewidth]{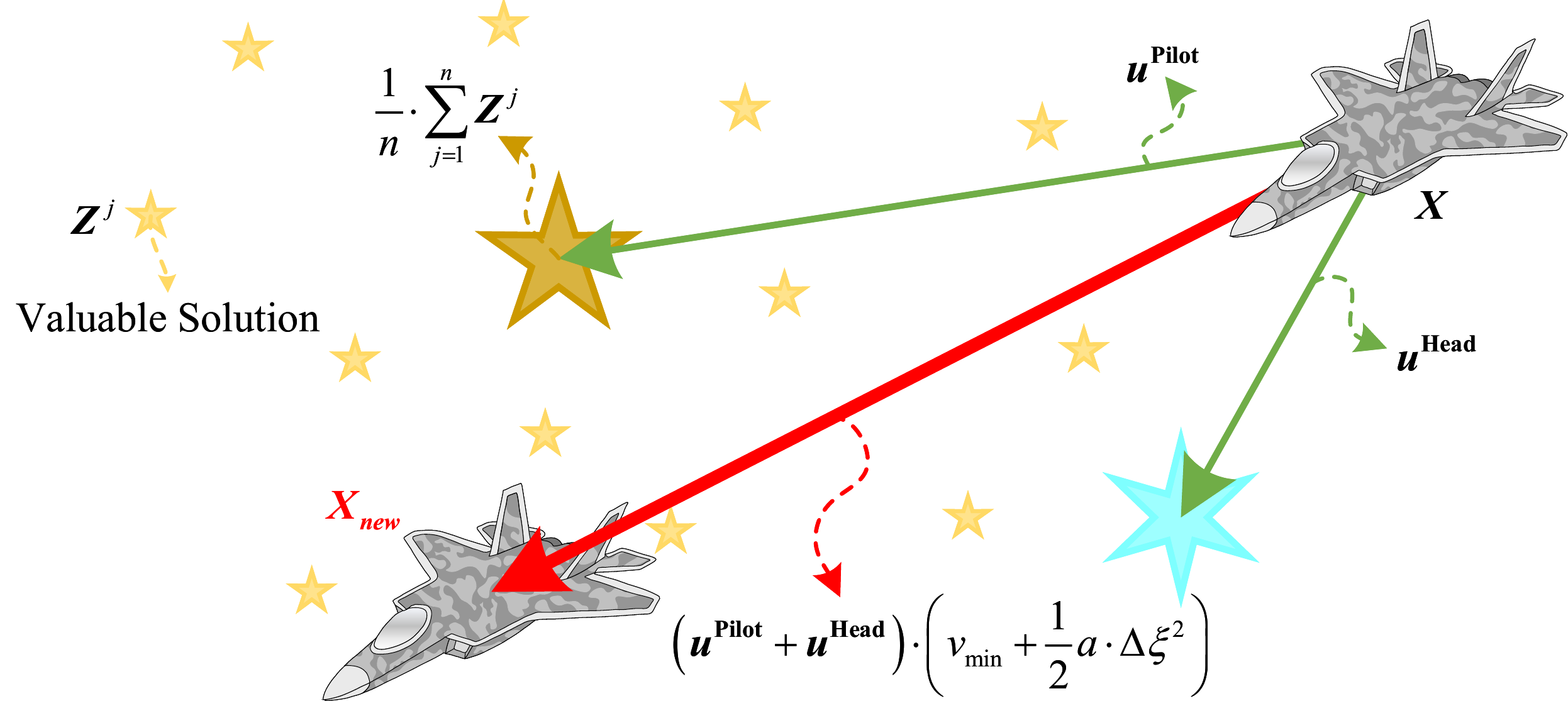}
                    \caption{Visualization of the maneuver evasion strategy.}
                    \label{fig6}
                \end{figure*}

                To intuitively demonstrate the search trajectory of the maneuver evasion strategy, a visualization is conducted in a 2-dimensional plane, as shown in Fig.~\ref{fig6}.

                \textbf{Flare evasion strategy:}

                The role of maneuvering evasion strategies is to guide solutions away from local optima and migrate toward other valuable solutions, while the flare evasion strategy is designed from a different perspective. As is well known, the randomness in metaheuristic algorithms usually stems from random numbers embedded in strategies or parameters. However, most algorithms tend to concentrate random searches near the center of the feasible domain, resulting in weaker exploration near the boundaries. When the global optimum is near the boundary, this drawback can markedly degrade performance.
                \begin{equation}
                    \begin{aligned}
                        \bm{u^{\mathrm{Pilot}}} &= \bm{X_{rand}^{boundary}} - \bm{X}^{i}, \\
                        \bm{X_{rand}^{boundary}} &= \bm{X_{rand}} \cdot I_{1} + I_{2} \left[ \bm{L} \cdot I_{2} + \bm{U} (1 - I_{2}) \right]
                    \end{aligned}
                    \label{eq20}
                \end{equation}

                To tackle this, this paper proposes the flare evasion strategy, which is specifically designed to strengthen the exploration capability in boundary regions. The definition of $\bm{u^{\mathrm{Pilot}}}$ for this strategy is given in Equations~\ref{eq20} and~\ref{eq21}.
                \begin{equation}
                    I_{1} =
                    \begin{cases}
                        0, & r_{10} < 0.5 \\
                        1, & r_{10} \geq 0.5
                    \end{cases}, \quad
                    I_{2} =
                    \begin{cases}
                        0, & r_{11} < 0.5 \\
                        1, & r_{11} \geq 0.5
                    \end{cases}
                    \label{eq21}
                \end{equation}

                \noindent
                where $\bm{X_{rand}^{boundary}}$ is a randomly generated boundary point within the feasible domain. The generation principle of $\bm{X_{rand}}$ is the same as in Equation~\ref{eq9} and the settings of $I_{1}$ and $I_{2}$ are defined in Equation~\ref{eq21}, where $r_{10}$ and $r_{11}$ are random numbers uniformly distributed within the range $[0, 1]$. Therefore, the exploration information of $\bm{u^{\mathrm{Pilot}}}$ originates from randomly selected boundary points. Since boundary search requires a larger movement radius for support, this strategy is modeled as uniform motion with maximum speed. The complete mathematical model is provided in Equation~\ref{eq22}.
                \begin{equation}
                    \begin{aligned}
                        \bm{X_{new}}^{i} &= \bm{X}^{i} + \int_{0}^{\Delta \xi} \left( \bm{u^{\mathrm{Pilot}}} + \bm{u^{\mathrm{Head}}} \right) v_{\max} \cdot d\xi \\
                        &= \bm{X}^{i} + \left( \bm{u^{\mathrm{Pilot}}} + \bm{u^{\mathrm{Head}}} \right) v_{\max} \cdot \Delta \xi, \\
                        i &= 1,2,\ldots,\frac{N}{2}
                    \end{aligned}
                    \label{eq22}
                \end{equation}

                In summary, the exploration capability of DoS is enhanced through two complementary evasion strategies: the maneuver evasion strategy helps solutions effectively escape from local optima and transition to new valuable solutions, while the flare evasion strategy addresses the issue of insufficient exploration in boundary regions. The combination of these two strategies equips DoS with stronger global search ability in the middle and later stages of iteration.

        \subsection{Dynamic Selection Mechanism}

            According to Section 3.2, DoS assigns corresponding search strategies to the early, middle and later stages of the iteration process. However, to achieve a dynamic balance between exploration and exploitation, the dynamic strategy selection mechanism is a key factor. This section introduces the mechanism based on the pseudocode of the iterative process for four types of solutions. For ease of explanation, the search strategies are represented by numeric identifiers: free flight (1), maneuver lock-on (2), missile attack (3), maneuver evasion (4) and flare evasion (5).

            \textbf{Leader aircraft in the stealth fighter formation:}

            As guiding solutions within the formation, leader aircraft adopt a more stochastic approach to strategy selection. According to Sections 3.2.3 to 3.2.5, leader aircraft primarily adopt the free flight strategy during the early stage of iteration, gradually shifting toward offensive and evasion strategies in the middle and later stages. To prevent over reliance on free flight, this paper introduces the following constraint for their execution, as shown in Equation~\ref{eq23}.
            \begin{equation}
                \bm{b_{X}}^{i} (i \leq p) =
                \begin{cases}
                    1, & r_{12} < K \ \ \mathrm{and} \ \ t < k_{3} \cdot T \\
                    \mathrm{Eq.}~\ref{eq25}, & r_{12} \geq K \ \ \mathrm{or} \ \ t \geq k_{3} \cdot T
                \end{cases}
                \label{eq23}
            \end{equation}

            \noindent
            where $\bm{b_{X}}^{i}$ denotes the strategy index selected by the $i$-th leader aircraft in the stealth fighter formation, $k_{3}$ is a tunable hyperparameter in DoS, $t$ and $T$ represent the current and maximum number of iterations, respectively, $K$ is the execution probability of the free flight strategy and $r_{12}$ is a random number uniformly distributed within the range $[0, 1]$. To adapt to the iteration process, $K$ is set as a monotonically decreasing function with respect to the number of iterations, as defined in Equation~\ref{eq24}.
            \begin{equation}
                K = e^{k_{2} - \frac{t}{T}}
                \label{eq24}
            \end{equation}

            If Equation~\ref{eq23} is not satisfied, the solution will select an offensive or evasion strategy based on a random number. The mathematical form is as follows:
            \begin{equation}
                \bm{b_{X}}^{i} (i \leq p) =
                \begin{cases}
                    2, & r_{13} < P^{X} \ \ \mathrm{and} \ \ r_{14} < 0.5 \\[4pt]
                    3, & r_{13} < P^{X} \ \ \mathrm{and} \ \ r_{14} \geq 0.5 \\[4pt]
                    4, & r_{13} \geq P^{X} \ \ \mathrm{and} \ \ r_{14} < 0.8 \\[4pt]
                    5, & r_{13} \geq P^{X} \ \ \mathrm{and} \ \ r_{14} \geq 0.8
                \end{cases}
                \label{eq25}
            \end{equation}

            \noindent
            where $r_{13}$ and $r_{14}$ are random numbers uniformly distributed within the range $[0, 1]$ and $P^{X}$ is the probability coefficient for offensive and evasion strategies, which dynamically changes with the function values of the formation. This coefficient is updated according to the promising-solution selections in each iteration, as defined in Equation~\ref{eq26}.
            \begin{equation}
                P_{new}^{X} = P^{X} + 0.05 (1 - P^{X}) \left( \frac{\sum_{i \in \Omega_{1}} i}{\sum_{j \in \Omega^{1}} j} - \frac{\sum_{i \in \Omega_{2}} i}{\sum_{j \in \Omega^{2}} j} \right) \frac{t}{T}
                \label{eq26}
            \end{equation}
            
            \noindent
            where $P_{new}^{X}$ denotes the updated probability coefficient, $\Omega^{1}$ and $\Omega^{2}$ the sets and represent the indices of solutions that have selected offensive and evasion strategies, respectively. Similarly $\Omega_{1}$ and $\Omega_{2}$ denote the indices of promising solutions that have chosen offensive and evasion strategies, respectively. Since solutions with larger indices tend to have worse function values, they are assigned greater weight in influencing strategy selection probabilities. This allows leader aircraft to reasonably regulate the formation’s need for exploration or exploitation.

            By integrating Equations~\ref{eq23}, ~\ref{eq25} and~\ref{eq26}, the pseudocode for leader aircraft in the stealth fighter formation can be formulated, as shown in Algorithm \ref{alg1}.
            
            \begin{algorithm}[htbp!]
                \caption{Leader Aircraft of Stealth Fighter Formation.}
                \label{alg1}
                \begin{algorithmic}[1]
                    \Require Updated solution index ($i$), search strategy selection factors ($r_{12}, r_{13}, r_{14}$), probability coefficient ($P_X$).
                    \Ensure Updated coordinates ($\bm{X_{t+1}}^i$), selected search strategy ($\bm{b_X}^i$).
                    \If{$r_{12} < K$ \text{and} $t < k_3 T$}
                    \State $\bm{X_{t+1}}^i \gets$ Update $\bm{X_{t}}^i$ using Equation~\ref{eq10};
                    \Else
                    \If{$r_{13} < P_X$}
                    \If{$r_{14} < 0.5$}
                    \State $\bm{X_{t+1}}^i \gets$ Update $\bm{X_{t}}^i$ using Equation~\ref{eq14};
                    \Else
                    \State $\bm{X_{t+1}}^i \gets$ Update $\bm{X_{t}}^i$ using Equation~\ref{eq17};
                    \EndIf
                    \Else
                    \If{$r_{14} < 0.8$}
                    \State $\bm{X_{t+1}}^i \gets$ Update $\bm{X_{t}}^i$ using Equation~\ref{eq19};
                    \Else
                    \State $\bm{X_{t+1}}^i \gets$ Update $\bm{X_{t}}^i$ using Equation~\ref{eq22};
                    \EndIf
                    \EndIf
                    \EndIf
                \end{algorithmic}
            \end{algorithm}

            \textbf{Leader aircraft in the regular fighter formation:}

            According to Section 3.2.3, free flight is crucial for global exploration in the early stages of iteration. To prevent both formations from excessively executing the free flight strategy, which could reduce overall efficiency, this paper sets the leader aircraft in the stealth fighter formation to execute free flight probabilistically, while leader aircraft in the regular fighter formation are restricted to exclusively executing the free flight strategy during the early stage. The corresponding constraint is defined in Equation~\ref{eq27}.
            \begin{equation}
                \bm{b_{X}}^{i} (i \leq p) =
                \begin{cases}
                    1, & t < k_{4} \cdot T \\
                    \mathrm{Eq.}~\ref{eq25}, & t \ge k_{4} \cdot T
                \end{cases}
                \label{eq27}
            \end{equation}

            \noindent
            where $k_{4}$ is a tunable parameter in DoS used to control the demand for random search in the early stages of the algorithm. If the iteration count does not satisfy Equation~\ref{eq27}, the leader aircraft of the regular fighter formation selects the offensive or evasive strategy based on Equations~\ref{eq25} and~\ref{eq26}. This mechanism avoids the accumulation of ineffective exploration in the early stage and ensures the exploration and exploitation balance between the two formations at different stages. The pseudocode for this mechanism is shown in Algorithm~\ref{alg2}.

            \begin{algorithm}[htbp!]
                \caption{Leader Aircraft of Regular Fighter Formation.}
                \label{alg2}
                \begin{algorithmic}[1]
                    \Require Updated solution index ($i$), search strategy selection factors ($r_{13}$ and $r_{14}$), probability coefficients for exploration and exploitation strategies ($P_X$).
                    \Ensure Updated coordinates ($\bm{Y_{t+1}}^i$), selected search strategy ($\bm{b_Y}^i$).
                    \If{$t < k_4 T$}
                    \State $\bm{Y_{t+1}}^i \gets$ update $\bm{Y_{t}}^i$ using Equation~\ref{eq10};
                    \Else
                    \If{$r_{12} < K$ and $t < k_3 T$}
                    \State $\bm{Y_{t+1}}^i \gets$ update $\bm{Y_{t}}^i$ using Equation~\ref{eq10};
                    \Else
                    \If{$r_{13} < P_X$}
                    \If{$r_{14} < 0.5$}
                    \State $\bm{Y_{t+1}}^i \gets$ update $\bm{Y_{t}}^i$ using Equation~\ref{eq14};
                    \Else
                    \State $\bm{Y_{t+1}}^i \gets$ update $\bm{Y_{t}}^i$ using Equation~\ref{eq17};
                    \EndIf
                    \Else
                    \If{$r_{14} < 0.8$}
                    \State $\bm{Y_{t+1}}^i \gets$ update $\bm{Y_{t}}^i$ using Equation~\ref{eq19};
                    \Else
                    \State $Y^{t+1}_i \gets$ update $\bm{Y_{t}}^i$ using Equation~\ref{eq22};
                    \EndIf
                    \EndIf
                    \EndIf
                    \EndIf
                \end{algorithmic}
            \end{algorithm}

            \textbf{Wing aircraft in both formations:}

            To maintain the balance between exploration and exploitation in DoS, this paper draws inspiration from the leader and wingman formation mechanism and designs a dynamic selection mechanism for wing aircraft that relies on the decisions of leader aircraft. Since the dynamic selection mechanisms for wing aircraft are identical in both formations, this subsection uses the stealth fighter formation as an example for explanation. The detailed mathematical model is defined in Equation~\ref{eq28}.
            \begin{equation}
                \bm{b_{X}}^{i} (i > p) =
                \begin{cases}
                    1, & \bm{b_{X}}^{H} = 1 \\
                    2, & \left( \bm{b_{X}}^{H} = 2 \ \ \mathrm{or} \ \ 3 \right) \ \ \mathrm{and} \ \ \left( r_{13} < P^{X} \right) \\
                    4, & \left( \bm{b_{X}}^{H} = 2 \ \ \mathrm{or} \ \ 3 \right) \ \ \mathrm{and} \ \ \left( r_{13} \geq P^{X} \right) \\
                    2, & \left( \bm{b_{X}}^{H} = 4 \ \ \mathrm{or} \ \ 5 \right) \ \ \mathrm{and} \ \ \left( r_{14} < 0.5 \right) \\
                    3, & \left( \bm{b_{X}}^{H} = 4 \ \ \mathrm{or} \ \ 5 \right) \ \ \mathrm{and} \ \ \left( r_{14} \geq 0.5 \right)
                \end{cases}
                \label{eq28}
            \end{equation}

            According to Equation~\ref{eq28}, when the leader aircraft selects the free flight strategy, the wing aircraft also executes free flight to ensure early-stage balance and information sharing. When the leader aircraft adopts an offensive strategy, the wing aircraft can switch between maneuver lock-on and maneuver evasion. If the leader aircraft selects an evasion strategy, the wing aircraft must choose an offensive strategy to enhance exploitation capability. The corresponding pseudocode is presented in Algorithm~\ref{alg3}.

            \begin{algorithm}[htbp!]
                \caption{Wing Aircraft of Stealth Fighter Formation.}
                \label{alg3}
                \begin{algorithmic}[1]
                    \Require Updated solution index ($i$), search strategy selection factors ($r_{13}$ and $r_{14}$), probability coefficients for exploration and exploitation strategies ($P_X$), leader aircraft index ($H$).
                    \Ensure Updated coordinates ($\bm{X_{t+1}}^i$).
                    \If{$\bm{b_X}^H = 1$}
                    \State $\bm{X_{t+1}}^i \gets$ update $\bm{X_{t}}^i$ using Equation~\ref{eq10};
                    \ElsIf{$\bm{b_X}^H = 2$ \textbf{or} $\bm{b_X}^H = 3$}
                    \If{$r_{13} < P^X$}
                    \State $\bm{X_{t+1}}^i \gets$ update $\bm{X_{t}}^i$ using Equation~\ref{eq14};
                    \Else
                    \State $\bm{X_{t+1}}^i \gets$ update $\bm{X_{t}}^i$ using Equation~\ref{eq19};
                    \EndIf
                    \Else
                    \If{$r_{14} < 0.5$}
                    \State $\bm{X_{t+1}}^i \gets$ update $\bm{X_{t}}^i$ using Equation~\ref{eq14};
                    \Else
                    \State $\bm{X_{t+1}}^i \gets$ update $\bm{X_{t}}^i$ using Equation~\ref{eq17};
                    \EndIf
                    \EndIf
                \end{algorithmic}
            \end{algorithm}

            In summary, the complete pseudocode and flowchart of DoS are presented in Algorithm~\ref{alg4} and Fig.~\ref{fig3}, respectively.

            \begin{algorithm}[htbp!]
                \caption{Dogfight Search.}
                \label{alg4}
                \begin{algorithmic}[1]
                    \Require Swarm size ($N$), maximum number of iterations ($T$), decision variable dimensions ($D$), upper and lower bounds of decision variables ($\bm{U}$ and $\bm{L}$) and objective function ($f$).
                    \Ensure Optimal solution ($\mathrm{Best\_Pos}$), optimal objective value ($\mathrm{Best\_Score}$).
                    \State Initialize $\bm{X}$ and $\bm{Y}$ using Equation~\ref{eq2};
                    \State Calculate fitness values $\bm{F}$ for $\bm{X}$ and $\bm{G}$ for $\bm{Y}$;
                    \State Assign indices to swarm solutions based on $\bm{F}$ and $\bm{G}$;
                    \While{$t \le T$}
                    \State Calculate $p$ using Equation~\ref{eq1};
                    \For{$i = 1$ to $N/2$}
                    \If{$i \le p$}
                    \State Update $\bm{X_{t}}^i$ using Algorithm~\ref{alg1};
                    \Else
                    \State Update $\bm{X_{t}}^i$ using Algorithm~\ref{alg3};
                    \EndIf
                    \EndFor
                    \For{$i = 1$ to $N/2$}
                    \If{$i \le p$}
                    \State Update $\bm{Y_{t}}^i$ using Algorithm~\ref{alg2};
                    \Else
                    \State Update $\bm{Y_{t}}^i$ using Algorithm~\ref{alg3};
                    \EndIf
                    \EndFor
                    \State Check boundary constraints of $\bm{X_{t + 1}}$ and $\bm{Y_{t + 1}}$;
                    \For{$i = 1$ to $N/2$}
                    \State Calculate fitness values $F_i$ and $G_i$;
                    \State Update valuable solution matrix;
                    \State Collect potential solution information;
                    \EndFor
                    \State Update probability coefficients $P^X$, $P^Y$, and velocity bounds $v_{\min}, v_{\max}$;
                    \State Update solution indices based on $\bm{F}$ and $\bm{G}$;
                    \State $t \gets t + 1$
                    \EndWhile
                \end{algorithmic}
            \end{algorithm}

        \subsection{Theoretical Complexity Analysis}

            Complexity is one of the key indicators for evaluating  an algorithm \citep{thomas2009introduction}. In computer simulations, complexity is mainly reflected in terms of time complexity and space complexity. This section provides a theoretical analysis of DoS from both perspectives.

            \textbf{Time Complexity Analysis:}

            The time complexity of DoS mainly depends on three factors: the total size of the two formations ($N$), the maximum number of iterations ($T$) and the dimensionality of the decision variables ($D$). In the initialization phase, the algorithm needs to allocate initial positions for the formations, resulting in a time complexity of $O(ND)$. In the iteration phase, although DoS introduces a relatively complex dynamic strategy selection mechanism, each solution executes only one search strategy per iteration, so it does not impose additional computational overhead. The total computational complexity for this phase is $O(NDT)$. Therefore, the overall time complexity of DoS is: $O(ND)+O(NDT) \approx O(NDT)$.

            \textbf{Space Complexity Analysis:}

            The space complexity of DoS mainly comes from the following four components:

            \textbf{Position matrix}: Stores the solution vectors, with complexity $O(ND)$;

            \textbf{Objective value matrix}: Stores the function values of the solutions, with complexity $O(N)$;

            \textbf{Strategy index matrix}: Records the strategy indices selected by solutions, with complexity $O(N)$;

            \textbf{Valuable solution matrix}: Stores information of valuable solutions accumulated during iterations, with a maximum size of $O(2.5ND)$.

            By combining the complexities of the four components, the total space complexity of DoS is: $O(ND) + O(N) + O(N) + O(2.5ND) \approx O(ND)$.

            In summary, DoS maintains the conventional complexity level of metaheuristic algorithms, with a time complexity of $O(NDT)$ and space complexity of $O(ND)$. The diversity of search strategies and the complexity of the dynamic selection mechanism do not theoretically increase the computational overhead.

    \section{Experiments and Discussions}

        This section presents a systematic evaluation of the optimization performance of DoS. The content includes the following aspects: a basic introduction to the benchmark test functions (Section 4.1), performance comparison with advanced competitors (Section 4.2), performance comparison with SOTA algorithms (Section 4.3), analysis of exploration and exploitation capabilities (Section 4.4), convergence analysis (Section 4.5), scalability analysis (Section 4.6), behavior analysis (Section 4.7), and computational time cost analysis (Section 4.8).

        \subsection{Experimental setup}

            To evaluate the performance of DoS on single-objective optimization problems, this study selects the CEC2017 \citep{wu2017problem} and CEC2022 \citep{ahrari2022problem} benchmark test functions for comparative experiments. The comparisons involve 7 advanced competitors and 3 SOTA algorithms. The benchmark test functions and competitors are introduced below.

            CEC2017 contains 29 classical test functions, covering four categories: unimodal, multimodal, hybrid and composition (see Table~\ref{tab2}). Among them, unimodal functions mainly evaluate the exploitation ability of the algorithm; multimodal functions, due to the presence of multiple local minima, rely on the balance between exploration and exploitation; hybrid functions are composed of multiple types of functions, increasing the complexity of the search space; and composition functions place multiple local minima with similar function values in adjacent regions, imposing higher requirements on the algorithm. Therefore, CEC2017 can comprehensively reflect the characteristics of real-world optimization problems and is suitable for verifying algorithmic performance, exploration and exploitation balance and the ability to escape from local optima.

            \begin{table}[htbp!]
                \centering
                \caption{CEC2017 Benchmark Test Functions.}
                \label{tab2}
                \begin{tabular}{\cm{0.01\textwidth}\cm{0.08\textwidth}\cm{0.25\textwidth}\cm{0.03\textwidth}}
                    \toprule
                    No. & Type & Function & $F^{*}$ \\
                    \midrule
                    F1 & \multirow{2}{*}{Unimodal} & Shifted and Rotated Bent Cigar Function & 100 \\
                    F3 &  & Shifted and Rotated Zakharov Function & 300 \\
                    \midrule
                    F4 & \multirow{13}{*}{Multimodal} & Shifted and Rotated Rosenbrock’s Function & 400 \\
                    F5 &  & Shifted and Rotated Rastrigin’s Function & 500 \\
                    F6 &  & Shifted and Rotated Expanded Scaffer’s F6 Function & 600 \\
                    F7 &  & Shifted and Rotated Lunacek Bi-Rastrigin Function & 700 \\
                    F8 &  & Shifted and Rotated Non-Continuous Rastrigin’s Function & 800 \\
                    F9 &  & Shifted and Rotated Levy Function & 900 \\
                    F10 &  & Shifted and Rotated Schwefel’s Function & 1000 \\
                    \midrule
                    F11 & \multirow{10}{*}{Hybrid} & Hybrid Function 1 (N = 3) & 1100 \\
                    F12 &  & Hybrid Function 2 (N = 3) & 1200 \\
                    F13 &  & Hybrid Function 3 (N = 3) & 1300 \\
                    F14 &  & Hybrid Function 4 (N = 4) & 1400 \\
                    F15 &  & Hybrid Function 5 (N = 4) & 1500 \\
                    F16 &  & Hybrid Function 6 (N = 4) & 1600 \\
                    F17 &  & Hybrid Function 6 (N = 5) & 1700 \\
                    F18 &  & Hybrid Function 6 (N = 5) & 1800 \\
                    F19 &  & Hybrid Function 6 (N = 5) & 1900 \\
                    F20 &  & Hybrid Function 6 (N = 6) & 2000 \\
                    \midrule
                    F21 & \multirow{10}{*}{Composition} & Composition Function 1 (N = 3) & 2100 \\
                    F22 &  & Composition Function 2 (N = 3) & 2200 \\
                    F23 &  & Composition Function 3 (N = 4) & 2300 \\
                    F24 &  & Composition Function 4 (N = 4) & 2400 \\
                    F25 &  & Composition Function 5 (N = 5) & 2500 \\
                    F26 &  & Composition Function 6 (N = 5) & 2600 \\
                    F27 &  & Composition Function 7 (N = 6) & 2700 \\
                    F28 &  & Composition Function 8 (N = 6) & 2800 \\
                    F29 &  & Composition Function 9 (N = 3) & 2900 \\
                    F30 &  & Composition Function 10 (N = 3) & 3000 \\
                    \midrule
                    \multicolumn{4}{c}{Search Range: $\left [ -100,100 \right ] ^{D} $} \\
                    \bottomrule
                \end{tabular}
                \vspace{-4pt}
                \begin{flushleft}
                     \footnotesize
                     \rmfamily
                     Type indicates the category of the test function, Function refers to the name of the test function, and $F^{*}$ represents the global optimal solution value of the test function.
                \end{flushleft}
            \end{table}

            CEC2022 is an enhanced extension of CEC2017. Although the number of functions is reduced to 12, their complexity is significantly increased, with more densely distributed local minima. Therefore, it is more representative for evaluating the performance of algorithms in solving complex single objective problems (see Table~\ref{tab3}).

            \begin{table}[htbp!]
                \centering
                \caption{CEC2022 Benchmark Test Functions.}
                \label{tab3}
                \begin{tabular}{\cm{0.01\textwidth}\cm{0.08\textwidth}\cm{0.25\textwidth}\cm{0.03\textwidth}}
                    \toprule
                    No. & Type & Function & $F^{*}$ \\
                    \midrule
                    C1 & Unimodal & Shifted and full Rotated Zakharov Function & 300 \\
                    \midrule
                    C2 & \multirow{8}{*}{Basic} & Shifted and full Rotated Rosenbrock’s Function & 400 \\
                    C3 &  & Shifted and full Rotated Expanded Schaffer’s F6 Function & 600 \\
                    C4 &  & Shifted and full Rotated Non-Continuous Rastrigin’s Function & 800 \\
                    C5 &  & Shifted and full Rotated Levy Function & 900 \\
                    \midrule
                    C6 & \multirow{3}{*}{Hybrid} & Hybrid Function 1 (N = 3) & 1800 \\
                    C7 &  & Hybrid Function 2 (N = 6) & 2000 \\
                    C8 &  & Hybrid Function 3 (N = 5) & 2200 \\
                    \midrule
                    C9 & \multirow{4}{*}{Composition} & Composition Function 1 (N = 5) & 2300 \\
                    C10 &  & Composition Function 2 (N = 4) & 2400 \\
                   C11 &  & Composition Function 3 (N = 5) & 2600 \\
                    C12 &  & Composition Function 4 (N = 6) & 2700 \\
                    \midrule
                    \multicolumn{4}{c}{Search Range: $\left [ -100,100 \right ] ^{D} $} \\
                    \bottomrule
                \end{tabular}
                \vspace{-4pt}
                \begin{flushleft}
                     \footnotesize
                     \rmfamily
                     Type indicates the category of the test function, Function refers to the name of the test function and $F^{*}$ represents the global optimal solution value of the test function.
                \end{flushleft}
            \end{table}

            All experiments in this paper are implemented in MATLAB 2024b under a 64-bit Windows 11 operating system. To ensure fairness and comparability of experiments, all competitors employed the hyperparameter settings from their original papers and strictly adhered to the standard testing conditions recommended by the CEC2017 and CEC2022 conferences. Detailed settings are shown in Table~\ref{tab4}.

            \begin{table}[htbp!]
                \centering
                \caption{Standard Test Conditions.}
                \label{tab4}
                \begin{tabular}{\cm{0.2\textwidth}\cm{0.1\textwidth}\cm{0.15\textwidth}}
                    \toprule
                     & CEC2017 & CEC2022 \\
                    \midrule
                    Selected Dimensions ($D$) & 30, 50, 100 & 10, 20 \\
                    Number of Evaluations ($E$) & $D × 10000$ & 200000 ($D = 10$), 1000000 ($D = 20$) \\
                    Number of Independent Runs & 51 times & 30 times \\
                    \bottomrule
                \end{tabular}
            \end{table}
            
            Recent studies have shown that traditional algorithms such as GA, PSO and GWO have gradually become less competitive compared to newly proposed methods. Based on this, this paper selects seven advanced competitors, AOO \citep{wang2025animated}, PSA, FFA \citep{shayanfar2018farmland}, AVOA \citep{abdollahzadeh2021african}, ETO \citep{luan2024exponential}, SGA \citep{tian2024snow} and WAA, as well as three SOTA algorithms that have demonstrated strong performance in the CEC competitions, LSHADE, LSHADE-SPACMA and AL-SHADE for comparison with DoS. The hyperparameter settings used by DoS in the experiments are listed in Table~\ref{tab5}, while all advanced competitors follow the parameter configurations specified in their original literature.

            \begin{table}[htbp!]
                \centering
                \caption{DoS parameter settings.}
                \label{tab5}
                \begin{tabular}{\cm{0.35\textwidth}\cm{0.04\textwidth}}
                    \toprule
                    Parameters & Value \\
                    \midrule
                    Population size ($N$) & 50 \\
                    Leader to follower ratio ($k_{1}$) & 0.3 \\
                    Probability factor of free flight strategy ($k_{2}$) & -2.5 \\
                    Time ratio of free flight strategy ($k_{3}$) & 0.2 \\
                    Population random search ratio coefficient of regular fighters ($k_{4}$) & 0.05 \\
                    Flight speed ratio factor ($k_{5}$) & 0.5 \\
                    \bottomrule
                \end{tabular}
            \end{table}

        \subsection{Comparison with Advanced Competitors}

            This section provides a systematic analysis of the experimental results of DoS and the advanced competitors mentioned in Section 4.1 on the CEC2017 benchmark test functions (30D, 50D, 100D) and CEC2022 benchmark test functions (10D, 20D). The test results are summarized in Tables~\ref{tab6}, ~\ref{tab7}, ~\ref{tab8}, ~\ref{tab9}, ~\ref{tab10}, ~\ref{tab11}, ~\ref{tab12}, ~\ref{tab13}, ~\ref{tab14}, ~\ref{tab15} and~\ref{tab16}. Since the number of functions in CEC2017 is relatively large, the test results for each dimension are divided into three tables according to function types: unimodal and multimodal, hybrid and composition functions, for better presentation.

            The statistical indicators used include: mean (Mean), Standard Deviation (Std), Kruskal-Wallis test result (Kruskal) \citep{mckight2010kruskal}, Wilcoxon signed-rank test result (WSRT) \citep{wilcoxon1992individual} at the 5\% significance level, Friedman test for average rank (FMR) and Final rank (F-Rank). In the tables, the best performing results or those statistically superior to competitors are marked in bold.

            \begin{table*}[htbp!]
                \centering
                \caption{Experimental results of DoS and advanced competitors on 30D CEC2017 benchmark test functions (Unimodal and Multimodal).}
                \label{tab6}
                \begin{tabular}{\cm{0.02\textwidth}\cm{0.06\textwidth}\cm{0.085\textwidth}\cm{0.085\textwidth}\cm{0.085\textwidth}\cm{0.085\textwidth}\cm{0.085\textwidth}\cm{0.085\textwidth}\cm{0.085\textwidth}\cm{0.085\textwidth}\cm{0.085\textwidth}}
                    \toprule
                    No. & Metrics & DoS & AOO & PSA & FFA & AVOA & ETO & SGA & WAA \\
                    \midrule
                    \multirow{4}{*}{F1} & Mean & \textbf{100} & 3573.6777 & 4778.6883 & 3929809.8 & 3879.4607 & 6.758E+09 & 2853437.8 & 1093664.2 \\
                     & Std & \textbf{1.33E-14} & 4520.0466 & 6066.3438 & 10582599 & 5128.8228 & 3.892E+09 & 6260055.6 & 158463.82 \\
                     & Kruskal & \textbf{26} & 125.80392 & 132.64706 & 276.4902 & 126.92157 & 383 & 265.13725 & 300 \\
                     & WSRT &  & 1.3E-18(+) & 1.3E-18(+) & 1.3E-18(+) & 1.3E-18(+) & 1.3E-18(+) & 1.3E-18(+) & 1.3E-18(+) \\
                    \midrule
                    \multirow{4}{*}{F3} & Mean & \textbf{300} & 300.0103 & 593.21876 & 40345.956 & 308.37 & 42274.951 & 4450.1667 & 308.03883 \\
                     & Std & \textbf{8.51E-14} & 0.0063952 & 1553.372 & 8704.7464 & 22.671363 & 8942.5492 & 1807.7862 & 1.6451758 \\
                     & Kruskal & \textbf{26} & 158.70588 & 101.98039 & 354.15686 & 136.5098 & 360.84314 & 279.35294 & 218.45098 \\
                     & WSRT &  & 1.6E-18(+) & 1.6E-18(+) & 1.6E-18(+) & 1.6E-18(+) & 1.6E-18(+) & 1.6E-18(+) & 1.6E-18(+) \\
                    \midrule
                    \multirow{4}{*}{F4} & Mean & \textbf{410.816} & 492.25512 & 466.4252 & 504.59428 & 482.80955 & 1025.4808 & 513.40804 & 500.48884 \\
                     & Std & \textbf{12.20367} & 16.339536 & 32.834153 & 12.591353 & 22.705696 & 504.39798 & 29.183628 & 19.213575 \\
                     & Kruskal & \textbf{30.80392} & 198.33333 & 122.07843 & 251.03922 & 157.41176 & 383 & 265.47059 & 227.86275 \\
                     & WSRT &  & 6.7E-18(+) & 1.6E-13(+) & 3.3E-18(+) & 1.9E-17(+) & 3.3E-18(+) & 4.4E-18(+) & 3.3E-18(+) \\
                    \midrule
                    \multirow{4}{*}{F5} & Mean & \textbf{542.901} & 586.59329 & 626.82109 & 667.20595 & 696.46241 & 717.85927 & 736.53624 & 813.08839 \\
                     & Std & \textbf{11.40222} & 23.889164 & 33.252734 & 42.26114 & 36.453227 & 26.141061 & 38.094334 & 13.085099 \\
                     & Kruskal & \textbf{27.98039} & 86.235294 & 138.88235 & 193.82353 & 238.19608 & 272.15686 & 296.66667 & 382.05882 \\
                     & WSRT &  & 8.7E-16(+) & 3.9E-18(+) & 3.3E-18(+) & 3.3E-18(+) & 3.3E-18(+) & 3.3E-18(+) & 3.3E-18(+) \\
                    \midrule
                    \multirow{4}{*}{F6} & Mean & \textbf{600.1275} & 610.17691 & 600.27012 & 600.18082 & 635.58839 & 643.58632 & 658.70026 & 667.02719 \\
                     & Std & 0.2042876 & 5.2376552 & 0.7649296 & \textbf{0.177903} & 8.9543386 & 8.0890007 & 7.2989513 & 4.1158917 \\
                     & Kruskal & \textbf{65.88235} & 178.92157 & 69.333333 & 96 & 244.15686 & 273 & 334.11765 & 374.58824 \\
                     & WSRT &  & 3.3E-18(+) & 6.4E-01($\approx$) & 9.0E-04(+) & 3.3E-18(+) & 3.3E-18(+) & 3.3E-18(+) & 3.3E-18(+) \\
                    \midrule
                    \multirow{4}{*}{F7} & Mean & \textbf{778.3365} & 824.48427 & 892.43764 & 851.89009 & 1071.2078 & 1044.6876 & 1183.072 & 1363.2113 \\
                     & Std & \textbf{11.92261} & 27.837361 & 41.590273 & 38.815368 & 75.450567 & 59.264625 & 90.371426 & 30.107359 \\
                     & Kruskal & \textbf{29.15686} & 93.431373 & 163.13725 & 125.29412 & 268.90196 & 254.45098 & 320.43137 & 381.19608 \\
                     & WSRT &  & 4.4E-15(+) & 3.3E-18(+) & 2.3E-17(+) & 3.3E-18(+) & 3.3E-18(+) & 3.3E-18(+) & 3.3E-18(+) \\
                    \midrule
                    \multirow{4}{*}{F8} & Mean & \textbf{840.8458} & 887.16766 & 914.57401 & 981.3438 & 961.87161 & 975.10959 & 979.211 & 1022.3178 \\
                     & Std & \textbf{8.857624} & 17.347282 & 30.699509 & 31.565117 & 33.562664 & 25.363266 & 31.19243 & 14.818309 \\
                     & Kruskal & \textbf{26.07843} & 89.823529 & 135.56863 & 272.88235 & 224.88235 & 254.52941 & 265.4902 & 366.7451 \\
                     & WSRT &  & 4.2E-18(+) & 3.3E-18(+) & 3.3E-18(+) & 3.3E-18(+) & 3.3E-18(+) & 3.3E-18(+) & 3.3E-18(+) \\
                    \midrule
                    \multirow{4}{*}{F9} & Mean & \textbf{906.4729} & 1324.4667 & 3463.5826 & 940.27927 & 5082.4391 & 5877.3694 & 5398.8897 & 6498.7964 \\
                     & Std & \textbf{6.175741} & 513.12248 & 1382.6671 & 23.046509 & 921.53943 & 1702.7937 & 1047.4418 & 630.87715 \\
                     & Kruskal & \textbf{28.21569} & 127.27451 & 201.11765 & 77.647059 & 265.76471 & 304.05882 & 283.94118 & 347.98039 \\
                     & WSRT &  & 3.5E-18(+) & 3.3E-18(+) & 1.8E-15(+) & 3.3E-18(+) & 3.3E-18(+) & 3.3E-18(+) & 3.3E-18(+) \\
                    \midrule
                    \multirow{4}{*}{F10} & Mean & \textbf{2995.011} & 4086.0552 & 4409.5398 & 7968.3505 & 4994.6571 & 6386.2307 & 6004.4391 & 5942.2188 \\
                     & Std & \textbf{258.6889} & 611.91558 & 650.89795 & 288.77962 & 733.4811 & 626.35825 & 756.63324 & 716.13103 \\
                     & Kruskal & \textbf{30.05882} & 103.2549 & 129.54902 & 382.03922 & 176.96078 & 294.27451 & 261.7451 & 258.11765 \\
                     & WSRT &  & 9.2E-15(+) & 1.4E-16(+) & 3.3E-18(+) & 3.3E-18(+) & 3.3E-18(+) & 3.3E-18(+) & 3.3E-18(+) \\
                    \bottomrule
                \end{tabular}
            \end{table*}

            \begin{table*}[htbp!]
                \centering
                \caption{Experimental results of DoS and advanced competitors on 30D CEC2017 benchmark test functions (Hybrid).}
                \label{tab7}
                \begin{tabular}{\cm{0.02\textwidth}\cm{0.06\textwidth}\cm{0.085\textwidth}\cm{0.085\textwidth}\cm{0.085\textwidth}\cm{0.085\textwidth}\cm{0.085\textwidth}\cm{0.085\textwidth}\cm{0.085\textwidth}\cm{0.085\textwidth}\cm{0.085\textwidth}}
                    \toprule
                    No. & Metrics & DoS & AOO & PSA & FFA & AVOA & ETO & SGA & WAA \\
                    \midrule
                    \multirow{4}{*}{F11} & Mean & \textbf{1183.613} & 1219.0968 & 1253.0216 & 1185.991 & 1229.6233 & 2225.483 & 1290.5779 & 1188.8889 \\
                     & Std & 36.046174 & 42.029705 & 47.928244 & \textbf{26.45045} & 48.440969 & 1019.3045 & 52.346656 & 60.84981 \\
                     & Kruskal & 109.21569 & 183.86275 & 244.66667 & 116.52941 & 202.82353 & 381.21569 & 292.52941 & \textbf{105.1569} \\
                     & WSRT &  & 1.8E-05(+) & 2.2E-11(+) & 4.5E-01($\approx$) & 9.3E-07(+) & 3.5E-18(+) & 8.2E-16(+) & 3.7E-01($\approx$) \\
                    \midrule
                    \multirow{4}{*}{F12} & Mean & \textbf{8085.755} & 2137054.1 & 53285.144 & 1571978.5 & 1294660.8 & 185093497 & 9254480.1 & 1990462.5 \\
                     & Std & \textbf{6121.55} & 1627807.7 & 34066.696 & 1114836 & 1098320.4 & 197397007 & 6337323.7 & 1011953 \\
                     & Kruskal & \textbf{26.70588} & 221.94118 & 77.352941 & 201.17647 & 182.4902 & 382.98039 & 313.80392 & 229.54902 \\
                     & WSRT &  & 3.3E-18(+) & 2.7E-17(+) & 3.3E-18(+) & 3.3E-18(+) & 3.3E-18(+) & 3.3E-18(+) & 3.3E-18(+) \\
                    \midrule
                    \multirow{4}{*}{F13} & Mean & \textbf{1705.753} & 101324.66 & 15056.948 & 134904.41 & 46358.217 & 104372665 & 120952.93 & 125780.91 \\
                     & Std & \textbf{497.7931} & 64318.328 & 16046.347 & 281717.73 & 21055.857 & 433967685 & 84997.261 & 66578.534 \\
                     & Kruskal & \textbf{30.62745} & 246.2549 & 95.372549 & 176.70588 & 174.17647 & 382.96078 & 254.17647 & 275.72549 \\
                     & WSRT &  & 3.3E-18(+) & 2.3E-13(+) & 2.0E-17(+) & 3.3E-18(+) & 3.3E-18(+) & 3.3E-18(+) & 3.3E-18(+) \\
                    \midrule
                    \multirow{4}{*}{F14} & Mean & \textbf{1559.072} & 6002.4604 & 13361.966 & 21320.115 & 20496.278 & 186725.27 & 35215.013 & 21934.693 \\
                     & Std & \textbf{69.58593} & 4263.1723 & 9362.4427 & 12927.257 & 15735.06 & 254904.26 & 24687.251 & 16509.427 \\
                     & Kruskal & \textbf{26.13725} & 108.60784 & 184.29412 & 235.41176 & 221.58824 & 351.4902 & 280.17647 & 228.29412 \\
                     & WSRT &  & 5.0E-18(+) & 3.3E-18(+) & 3.3E-18(+) & 3.3E-18(+) & 3.3E-18(+) & 3.3E-18(+) & 3.3E-18(+) \\
                    \midrule
                    \multirow{4}{*}{F15} & Mean & \textbf{1687.1} & 43916.293 & 8934.6881 & 14464.558 & 13178.516 & 1123306.2 & 59591.664 & 35090.024 \\
                     & Std & \textbf{127.3315} & 26040.927 & 10148.567 & 23001.181 & 10220.328 & 4813338 & 50597.464 & 27299.846 \\
                     & Kruskal & \textbf{35.58824} & 273.98039 & 120.11765 & 140.60784 & 163.90196 & 368.80392 & 290.03922 & 242.96078 \\
                     & WSRT &  & 3.3E-18(+) & 1.6E-10(+) & 1.0E-14(+) & 3.3E-18(+) & 3.3E-18(+) & 3.3E-18(+) & 3.3E-18(+) \\
                    \midrule
                    \multirow{4}{*}{F16} & Mean & \textbf{2023.233} & 2283.0587 & 2664.5465 & 3079.7483 & 2822.4906 & 2915.0385 & 3165.9144 & 3339.4033 \\
                     & Std & \textbf{138.6301} & 244.09129 & 268.0909 & 212.1503 & 279.6546 & 304.00652 & 463.13849 & 439.43968 \\
                     & Kruskal & \textbf{37} & 85.45098 & 167.88235 & 286.43137 & 213.35294 & 235.82353 & 287.58824 & 322.47059 \\
                     & WSRT &  & 2.0E-08(+) & 6.7E-17(+) & 3.3E-18(+) & 1.7E-17(+) & 7.5E-18(+) & 4.7E-18(+) & 3.3E-18(+) \\
                    \midrule
                    \multirow{4}{*}{F17} & Mean & \textbf{1798.463} & 1937.46 & 2295.2608 & 1997.9355 & 2358.3617 & 2221.7716 & 2318.1095 & 2896.4661 \\
                    & Std & \textbf{59.48037} & 141.94374 & 202.57959 & 142.11471 & 214.28736 & 188.66926 & 236.98423 & 312.40785 \\
                     & Kruskal & \textbf{37.03922} & 103.45098 & 249.27451 & 131.66667 & 269.23529 & 222.47059 & 253.31373 & 369.54902 \\
                     & WSRT &  & 2.9E-11(+) & 8.9E-18(+) & 1.6E-13(+) & 4.7E-18(+) & 1.1E-17(+) & 8.4E-18(+) & 3.3E-18(+) \\
                    \midrule
                    \multirow{4}{*}{F18} & Mean & \textbf{2087.501} & 163447.15 & 118019.86 & 1071703.4 & 177884.02 & 866851.31 & 740017.84 & 329967.79 \\
                     & Std & \textbf{307.3159} & 116770.49 & 91983.392 & 571179.64 & 153927.97 & 674802.4 & 595229.88 & 268771.98 \\
                     & Kruskal & \textbf{26} & 155.82353 & 126.54902 & 338.72549 & 159.39216 & 308.58824 & 297.84314 & 223.07843 \\
                     & WSRT &  & 3.3E-18(+) & 3.3E-18(+) & 3.3E-18(+) & 3.3E-18(+) & 3.3E-18(+) & 3.3E-18(+) & 3.3E-18(+) \\
                    \midrule
                    \multirow{4}{*}{F19} & Mean & \textbf{1996.864} & 112804.49 & 9359.0025 & 9898.8753 & 10322.404 & 1081434.4 & 79372.589 & 51036.061 \\
                     & Std & \textbf{67.47545} & 103314.23 & 11486.387 & 19160.847 & 11641.369 & 4333856.1 & 82373.244 & 17341.289 \\
                     & Kruskal & \textbf{32.52941} & 288.39216 & 131.27451 & 124.47059 & 151.11765 & 367.45098 & 268.86275 & 271.90196 \\
                     & WSRT &  & 3.3E-18(+) & 6.7E-15(+) & 1.6E-13(+) & 3.3E-18(+) & 3.3E-18(+) & 3.3E-18(+) & 3.3E-18(+) \\
                    \midrule
                    \multirow{4}{*}{F20} & Mean & \textbf{2124.574} & 2330.1334 & 2501.5145 & 2395.3068 & 2560.2715 & 2561.5864 & 2640.4097 & 2946.624 \\
                     & Std & \textbf{61.49456} & 135.36806 & 180.90689 & 131.35557 & 203.7329 & 159.6346 & 194.32465 & 240.49813 \\
                     & Kruskal & \textbf{32.09804} & 123.76471 & 212.19608 & 158.76471 & 235.94118 & 242.01961 & 273.29412 & 357.92157 \\
                     & WSRT &  & 7.7E-14(+) & 1.1E-17(+) & 1.0E-15(+) & 4.7E-18(+) & 3.3E-18(+) & 3.3E-18(+) & 3.3E-18(+) \\
                    \bottomrule
                \end{tabular}
            \end{table*}

            \begin{table*}[htbp!]
                \centering
                \caption{Experimental results of DoS and advanced competitors on 30D CEC2017 benchmark test functions (Composition).}
                \label{tab8}
                \begin{tabular}        {\cm{0.02\textwidth}\cm{0.06\textwidth}\cm{0.085\textwidth}\cm{0.085\textwidth}\cm{0.085\textwidth}\cm{0.085\textwidth}\cm{0.085\textwidth}\cm{0.085\textwidth}\cm{0.085\textwidth}\cm{0.085\textwidth}\cm{0.085\textwidth}}
                    \toprule
                    No. & Metrics & DoS & AOO & PSA & FFA & AVOA & ETO & SGA & WAA \\
                    \midrule
                    \multirow{4}{*}{F21} & Mean & \textbf{2338.601} & 2386.2256 & 2428.9095 & 2457.4887 & 2488.7295 & 2501.4059 & 2531.6737 & 2662.2154 \\
                     & Std & \textbf{8.448033} & 26.486128 & 36.15707 & 43.151154 & 43.680388 & 31.96612 & 51.749341 & 53.280363 \\
                     & Kruskal & \textbf{28.76471} & 88.078431 & 146.98039 & 193.92157 & 239.66667 & 263.54902 & 294.90196 & 380.13725 \\
                     & WSRT &  & 5.6E-16(+) & 3.3E-18(+) & 6.3E-17(+) & 3.3E-18(+) & 3.3E-18(+) & 3.3E-18(+) & 3.3E-18(+) \\
                    \midrule
                    \multirow{4}{*}{F22} & Mean & \textbf{2401.722} & 4562.4384 & 4530.3552 & 2469.8868 & 5455.501 & 7084.6276 & 5545.7902 & 7440.9618 \\
                     & Std & \textbf{505.0165} & 1694.7972 & 1942.0851 & 927.84542 & 2223.7592 & 1818.7774 & 2648.7982 & 1131.3822 \\
                     & Kruskal & \textbf{39.43137} & 172.03922 & 165.84314 & 146.84314 & 216.88235 & 318.56863 & 254.07843 & 322.31373 \\
                     & WSRT &  & 1.3E-14(+) & 3.7E-14(+) & 7.1E-16(+) & 3.9E-15(+) & 2.2E-18(+) & 8.3E-18(+) & 8.4E-19(+) \\
                    \midrule
                    \multirow{4}{*}{F23} & Mean & \textbf{2690.025} & 2748.2516 & 2789.0488 & 2738.5936 & 2911.6209 & 2932.7588 & 3059.6711 & 3485.6388 \\
                     & Std & \textbf{8.240597} & 27.929707 & 40.345631 & 28.711814 & 64.157107 & 53.812124 & 110.0099 & 165.41296 \\
                     & Kruskal & \textbf{28.01961} & 117.86275 & 164.80392 & 103.4902 & 253.15686 & 266.19608 & 319.86275 & 382.60784 \\
                     & WSRT &  & 2.4E-17(+) & 3.3E-18(+) & 1.7E-16(+) & 3.3E-18(+) & 3.3E-18(+) & 3.3E-18(+) & 3.3E-18(+) \\
                    \midrule
                    \multirow{4}{*}{F24} & Mean & \textbf{2859.896} & 2908.1356 & 2961.2945 & 3012.096 & 3097.911 & 3139.7113 & 3219.6786 & 3699.6527 \\
                     & Std & \textbf{9.739557} & 23.439242 & 38.465654 & 18.404057 & 90.272064 & 38.537506 & 111.12692 & 139.87083 \\
                     & Kruskal & \textbf{26.31373} & 82.627451 & 131.2549 & 182.60784 & 245.29412 & 277.37255 & 307.60784 & 382.92157 \\
                     & WSRT &  & 8.4E-18(+) & 3.3E-18(+) & 3.3E-18(+) & 3.3E-18(+) & 3.3E-18(+) & 3.3E-18(+) & 3.3E-18(+) \\
                    \midrule
                    \multirow{4}{*}{F25} & Mean & \textbf{2882.181} & 2893.1476 & 2900.3333 & 2893.0253 & 2900.5516 & 3109.0003 & 2932.5518 & 2890.0572 \\
                     & Std & 9.8631592 & 16.718655 & 19.311325 & 5.208034 & 19.265772 & 114.26641 & 24.838985 & \textbf{3.10024} \\
                     & Kruskal & \textbf{51.62745} & 120.80392 & 192.66667 & 205.43137 & 196.54902 & 382.68627 & 305.29412 & 180.94118 \\
                     & WSRT &  & 1.6E-09(+) & 7.3E-14(+) & 4.2E-14(+) & 1.1E-12(+) & 3.3E-18(+) & 9.4E-17(+) & 3.4E-12(+) \\
                    \midrule
                    \multirow{4}{*}{F26} & Mean & \textbf{3596.484} & 4337.8935 & 5073.1215 & 4202.0084 & 6046.612 & 6237.3857 & 6979.9428 & 9333.7044 \\
                     & Std & \textbf{313.5879} & 609.72323 & 1173.543 & 448.27555 & 1388.0012 & 627.55266 & 1985.8611 & 1898.0906 \\
                     & Kruskal & \textbf{58.60784} & 135.11765 & 181.09804 & 119.47059 & 244.54902 & 258.21569 & 279.29412 & 359.64706 \\
                     & WSRT &  & 7.2E-12(+) & 2.7E-09(+) & 3.5E-12(+) & 4.9E-11(+) & 5.6E-18(+) & 1.6E-11(+) & 7.3E-16(+) \\
                    \midrule
                    \multirow{4}{*}{F27} & Mean & \textbf{3200.006} & 3223.6832 & 3241.4742 & 3211.1006 & 3259.5374 & 3372.2234 & 3377.074 & 3918.8171 \\
                     & Std & \textbf{0.000133} & 14.518918 & 15.621088 & 6.1220123 & 20.14633 & 57.440145 & 129.96702 & 378.51176 \\
                     & Kruskal & \textbf{32} & 128.54902 & 184.05882 & 84.27451 & 221.86275 & 309.68627 & 295.82353 & 379.7451 \\
                     & WSRT &  & 1.6E-14(+) & 3.3E-18(+) & 1.6E-14(+) & 3.3E-18(+) & 3.3E-18(+) & 3.3E-18(+) & 3.3E-18(+) \\
                    \midrule
                    \multirow{4}{*}{F28} & Mean & 3298.2369 & 3188.0497 & 3184.5793 & 3236.2144 & 3181.2 & 3662.6274 & 3264.5353 & \textbf{3174.048} \\
                     & Std & \textbf{3.675618} & 54.330403 & 55.584821 & 18.859915 & 49.388376 & 321.8931 & 22.284333 & 43.92663 \\
                     & Kruskal & 328.84314 & 124.78431 & 120.43137 & 211.94118 & 107.19608 & 383 & 267.21569 & \textbf{92.58824} \\
                     & WSRT &  & 6.3E-17(-) & 3.3E-18(-) & 3.3E-18(-) & 3.3E-18(-) & 3.3E-18(+) & 1.7E-15(-) & 3.3E-18(-) \\
                    \midrule
                    \multirow{4}{*}{F29} & Mean & \textbf{3363.415} & 3660.6339 & 3788.7912 & 3945.0394 & 4027.1821 & 4160.0736 & 4736.6651 & 4790.3267 \\
                     & Std & \textbf{92.8501} & 145.09637 & 195.65842 & 220.42833 & 256.26873 & 264.16975 & 367.91525 & 377.80745 \\
                     & Kruskal & \textbf{28.4902} & 107.45098 & 144.09804 & 192.72549 & 214.39216 & 248.66667 & 346.7451 & 353.43137 \\
                     & WSRT &  & 9.9E-17(+) & 4.0E-17(+) & 8.4E-18(+) & 5.3E-18(+) & 3.5E-18(+) & 3.3E-18(+) & 3.3E-18(+) \\
                    \midrule
                    \multirow{4}{*}{F30} & Mean & \textbf{3617.183} & 524929.73 & 9139.082 & 25318.13 & 36223.583 & 6339751.5 & 1349562.6 & 333842.02 \\
                     & Std & \textbf{1677.274} & 367260.51 & 2875.3313 & 50698.576 & 15216.399 & 4605588.5 & 1402574.4 & 200689.48 \\
                     & Kruskal & \textbf{28.4902} & 275.76471 & 87.254902 & 123.33333 & 172.82353 & 379.90196 & 312.11765 & 256.31373 \\
                     & WSRT &  & 3.3E-18(+) & 2.9E-16(+) & 5.6E-17(+) & 3.3E-18(+) & 3.3E-18(+) & 3.3E-18(+) & 3.3E-18(+) \\
                    \midrule
                    \multicolumn{2}{c}{FMR} & \textbf{1.392833} & 3.4178499 & 3.4016227 & 4.2237999 & 4.4861393 & 6.6396214 & 6.1575389 & 6.280595 \\
                    \multicolumn{2}{c}{F-Rank} & \textbf{1} & 3 & 2 & 4 & 5 & 8 & 6 & 7 \\
                    \bottomrule
                \end{tabular}
                \vspace{-4pt}
                \begin{flushleft}
                     \footnotesize
                     \rmfamily
                     FMR and F-Rank are the statistical results over all CEC2017 benchmark test functions (30D).
                \end{flushleft}
            \end{table*}

            Tables~\ref{tab6}, ~\ref{tab7} and~\ref{tab8} summarize the performance comparison between DoS and advanced competitors on the 30-dimensional CEC2017 benchmark test functions. The results show that DoS achieves the best values in all three indicators, Mean, Std and Kruskal on 25 out of 29 functions, except for F6, F11, F25 and F28, indicating strong stability and overall performance in low dimensional optimization problems.

            In terms of the WSRT indicator, DoS demonstrates statistically significant advantages on all functions except F28. The F28 function contains strong local convergence traps, which cause DoS to be more prone to falling into local optima during the search process, leading to slightly inferior performance compared to some competitors.

            The FMR result from the Friedman test further shows that DoS ranks first among all compared methods, with a 2.11 point lead over the second-best algorithm, PSA, confirming its remarkable optimization capability on low dimensional benchmark tests.

            \begin{table*}[htbp!]
                \centering
                \caption{Experimental results of DoS and advanced competitors on 50D CEC2017 benchmark test functions (Unimodal and Multimodal).}
                \label{tab9}
                \begin{tabular}{\cm{0.02\textwidth}\cm{0.06\textwidth}\cm{0.085\textwidth}\cm{0.085\textwidth}\cm{0.085\textwidth}\cm{0.085\textwidth}\cm{0.085\textwidth}\cm{0.085\textwidth}\cm{0.085\textwidth}\cm{0.085\textwidth}\cm{0.085\textwidth}}
                    \toprule
                    No. & Metrics & DoS & AOO & PSA & FFA & AVOA & ETO & SGA & WAA \\
                    \midrule
                    \multirow{4}{*}{F1} & Mean & \textbf{100} & 7458.7766 & 4205.0167 & 80718.056 & 5228.0557 & 2.015E+10 & 26034150 & 4965931 \\
                     & Std & \textbf{8.27E-10} & 7197.5812 & 6252.8935 & 377927.33 & 4731.6526 & 6.059E+09 & 26262110 & 568339.28 \\
                     & Kruskal & \textbf{26} & 155.92157 & 121.31373 & 195.60784 & 141.15686 & 383 & 329.88235 & 283.11765 \\
                     & WSRT &  & 3.3E-18(+) & 3.3E-18(+) & 3.3E-18(+) & 3.3E-18(+) & 3.3E-18(+) & 3.3E-18(+) & 3.3E-18(+) \\
                    \midrule
                    \multirow{4}{*}{F3} & Mean & \textbf{300} & 300.01759 & 9531.6933 & 155854.74 & 891.77674 & 96977.11 & 23051.704 & 471.10537 \\
                     & Std & \textbf{2.36E-13} & 0.0086403 & 22769.732 & 15777.301 & 827.11499 & 16078.457 & 4098.5324 & 102.90146 \\
                     & Kruskal & \textbf{26} & 78.176471 & 181.33333 & 382.7451 & 192.29412 & 330.70588 & 274.13725 & 170.60784 \\
                     & WSRT &  & 3.0E-18(+) & 3.0E-18(+) & 3.0E-18(+) & 3.0E-18(+) & 3.0E-18(+) & 3.0E-18(+) & 3.0E-18(+) \\
                    \midrule
                    \multirow{4}{*}{F4} & Mean & \textbf{422.0392} & 519.03187 & 494.35771 & 561.1983 & 515.74742 & 3238.7534 & 616.6472 & 557.38907 \\
                     & Std & 35.920017 & 46.454784 & 49.673021 & 35.511733 & 58.226563 & 1243.9116 & 61.231415 & \textbf{33.76841} \\
                     & Kruskal & \textbf{37.33333} & 164.92157 & 129.05882 & 236.84314 & 165.17647 & 383 & 294.35294 & 225.31373 \\
                     & WSRT &  & 1.2E-15(+) & 2.0E-12(+) & 2.1E-17(+) & 8.3E-15(+) & 3.3E-18(+) & 5.9E-18(+) & 4.0E-17(+) \\
                    \midrule
                    \multirow{4}{*}{F5} & Mean & \textbf{616.226} & 692.26334 & 758.30356 & 900.97302 & 841.94139 & 929.5862 & 877.1667 & 912.60771 \\
                     & Std & \textbf{20.86876} & 35.061109 & 51.575974 & 55.761143 & 29.85556 & 37.124135 & 46.495572 & 25.857135 \\
                     & Kruskal & \textbf{28.21569} & 83.137255 & 127.05882 & 297.41176 & 202.33333 & 335.62745 & 254.56863 & 307.64706 \\
                     & WSRT &  & 1.6E-15(+) & 4.2E-18(+) & 3.3E-18(+) & 3.3E-18(+) & 3.3E-18(+) & 3.3E-18(+) & 3.3E-18(+) \\
                    \midrule
                    \multirow{4}{*}{F6} & Mean & \textbf{600.0709} & 621.36133 & 600.8858 & 600.27842 & 637.36958 & 659.05157 & 668.75256 & 671.05589 \\
                     & Std & 0.2564354 & 7.7845019 & 0.9587414 & \textbf{0.166567} & 6.2372134 & 7.840533 & 6.2154284 & 2.0258048 \\
                     & Kruskal & \textbf{32.35294} & 182.58824 & 111.11765 & 87.529412 & 227.21569 & 293.01961 & 342.31373 & 359.86275 \\
                     & WSRT &  & 3.0E-18(+) & 3.2E-15(+) & 1.7E-13(+) & 3.0E-18(+) & 3.0E-18(+) & 3.0E-18(+) & 3.0E-18(+) \\
                    \midrule
                    \multirow{4}{*}{F7} & Mean & \textbf{881.0636} & 973.3422 & 1125.4193 & 1058.854 & 1478.6609 & 1407.5583 & 1647.9132 & 1836.9871 \\
                     & Std & \textbf{24.8489} & 56.212727 & 69.638523 & 73.321046 & 102.85529 & 69.617662 & 118.27748 & 43.187892 \\
                     & Kruskal & \textbf{29.03922} & 85.921569 & 162.92157 & 132.15686 & 274.05882 & 246.39216 & 325.35294 & 380.15686 \\
                     & WSRT &  & 7.0E-15(+) & 3.3E-18(+) & 9.5E-18(+) & 3.3E-18(+) & 3.3E-18(+) & 3.3E-18(+) & 3.3E-18(+) \\
                    \midrule
                    \multirow{4}{*}{F8} & Mean & \textbf{907.918} & 995.45201 & 1057.0124 & 1200.8858 & 1135.5209 & 1241.0952 & 1195.0693 & 1210.3759 \\
                     & Std & \textbf{19.88024} & 45.695405 & 56.869752 & 54.479444 & 37.656796 & 36.671633 & 46.647196 & 34.183085 \\
                     & Kruskal & \textbf{27.60784} & 86.254902 & 129.45098 & 290.31373 & 192.98039 & 342.33333 & 271.23529 & 295.82353 \\
                     & WSRT &  & 2.7E-16(+) & 4.4E-18(+) & 3.3E-18(+) & 3.3E-18(+) & 3.3E-18(+) & 3.3E-18(+) & 3.3E-18(+) \\
                    \midrule
                    \multirow{4}{*}{F9} & Mean & \textbf{1257.971} & 5367.3213 & 12360.52 & 1548.6009 & 12925.665 & 25590.045 & 14749.781 & 20020.578 \\
                     & Std & \textbf{367.3211} & 2039.6856 & 3083.1816 & 375.04432 & 1840.4879 & 5048.4321 & 2004.6052 & 1804.9739 \\
                     & Kruskal & \textbf{39.05882} & 129.27451 & 214.52941 & 64.196078 & 220.13725 & 374.84314 & 258.45098 & 335.5098 \\
                     & WSRT &  & 4.2E-18(+) & 3.3E-18(+) & 2.0E-05(+) & 3.3E-18(+) & 3.3E-18(+) & 3.3E-18(+) & 3.3E-18(+) \\
                    \midrule
                    \multirow{4}{*}{F10} & Mean & \textbf{4782.937} & 6306.9044 & 6871.0237 & 14173.124 & 8326.4449 & 11666.121 & 9463.8427 & 8720.207 \\
                     & Std & 386.18247 & 767.74351 & 725.74363 & \textbf{358.3042} & 906.39592 & 914.61567 & 1408.5348 & 910.50144 \\
                     & Kruskal & \textbf{28.03922} & 94.294118 & 122.07843 & 382.64706 & 204.01961 & 326.47059 & 253.72549 & 224.72549 \\
                     & WSRT &  & 9.6E-16(+) & 3.3E-18(+) & 3.3E-18(+) & 4.2E-18(+) & 3.3E-18(+) & 3.3E-18(+) & 3.3E-18(+) \\
                    \bottomrule
                \end{tabular}
            \end{table*}

            \begin{table*}[htbp!]
                \centering
                \caption{Experimental results of DoS and advanced competitors on 50D CEC2017 benchmark test functions (Hybrid).}
                \label{tab10}
                \begin{tabular}{\cm{0.02\textwidth}\cm{0.06\textwidth}\cm{0.085\textwidth}\cm{0.085\textwidth}\cm{0.085\textwidth}\cm{0.085\textwidth}\cm{0.085\textwidth}\cm{0.085\textwidth}\cm{0.085\textwidth}\cm{0.085\textwidth}\cm{0.085\textwidth}}
                    \toprule
                    No. & Metrics & DoS & AOO & PSA & FFA & AVOA & ETO & SGA & WAA \\
                    \midrule
                    \multirow{4}{*}{F11} & Mean & \textbf{1279.913} & 1314.3262 & 1372.6352 & 1336.712 & 1307.114 & 6080.2766 & 1458.8419 & 1366.6447 \\
                     & Std & 47.390331 & 50.612961 & 80.017588 & 66.267192 & 50.907215 & 2354.975 & 91.302341 & \textbf{32.06185} \\
                     & Kruskal & \textbf{86.37255} & 135.2549 & 215.33333 & 169.23529 & 124.19608 & 383 & 299.05882 & 223.54902 \\
                     & WSRT &  & 2.3E-03(+) & 8.4E-10(+) & 3.2E-05(+) & 1.4E-02(+) & 3.3E-18(+) & 1.2E-16(+) & 6.3E-14(+) \\
                    \midrule
                    \multirow{4}{*}{F12} & Mean & \textbf{6940.079} & 9491633.5 & 531961.24 & 7188695.6 & 4039899.8 & 4.889E+09 & 45078699 & 10512110 \\
                     & Std & \textbf{3988.266} & 4551846.5 & 390543.95 & 3675557.8 & 2179042.5 & 4.201E+09 & 28315049 & 5898223.7 \\
                     & Kruskal & \textbf{26} & 231.78431 & 78.176471 & 202.52941 & 150.17647 & 383 & 327.90196 & 236.43137 \\
                     & WSRT &  & 3.3E-18(+) & 3.3E-18(+) & 3.3E-18(+) & 3.3E-18(+) & 3.3E-18(+) & 3.3E-18(+) & 3.3E-18(+) \\
                    \midrule
                    \multirow{4}{*}{F13} & Mean & \textbf{3664.402} & 153373.79 & 6819.5476 & 87632.24 & 61687.305 & 388392019 & 146683.6 & 288113.36 \\
                     & Std & \textbf{3178.467} & 90205.977 & 7114.717 & 184106.75 & 24297.225 & 1.244E+09 & 87685.879 & 141981.07 \\
                     & Kruskal & \textbf{51.72549} & 250.7451 & 77.196078 & 137.17647 & 179.17647 & 383 & 245.11765 & 311.86275 \\
                     & WSRT &  & 3.3E-18(+) & 9.5E-04(+) & 1.0E-07(+) & 3.5E-18(+) & 3.3E-18(+) & 3.3E-18(+) & 3.3E-18(+) \\
                    \midrule
                    \multirow{4}{*}{F14} & Mean & \textbf{1705.921} & 44096.741 & 29450.302 & 246333.45 & 70966.003 & 931345.3 & 164414.59 & 69433.339 \\
                     & Std & \textbf{115.1329} & 34268.23 & 22095.405 & 169392.84 & 65332.351 & 914056.12 & 105648.79 & 41790.379 \\
                     & Kruskal & \textbf{26} & 151 & 121.29412 & 306.80392 & 188.52941 & 364.84314 & 279.37255 & 198.15686 \\
                     & WSRT &  & 3.3E-18(+) & 3.3E-18(+) & 3.3E-18(+) & 3.3E-18(+) & 3.3E-18(+) & 3.3E-18(+) & 3.3E-18(+) \\
                    \midrule
                    \multirow{4}{*}{F15} & Mean & \textbf{1835.009} & 45955.61 & 8805.2286 & 12452.34 & 27609.218 & 26655984 & 64201.776 & 47242.202 \\
                     & Std & \textbf{170.3198} & 22360.863 & 5349.426 & 12662.092 & 10802.562 & 62129681 & 48340.016 & 33885.092 \\
                     & Kruskal & \textbf{31.23529} & 264.88235 & 103.17647 & 113.84314 & 200.4902 & 383 & 285.76471 & 253.60784 \\
                     & WSRT &  & 3.3E-18(+) & 1.1E-17(+) & 1.8E-12(+) & 3.3E-18(+) & 3.3E-18(+) & 3.3E-18(+) & 3.3E-18(+) \\
                    \midrule
                    \multirow{4}{*}{F16} & Mean & \textbf{2547.262} & 2915.2508 & 3561.9158 & 4683.2115 & 3844.888 & 4034.6808 & 4105.5049 & 4278.5403 \\
                     & Std & 253.96026 & 378.20303 & 415.43788 & \textbf{251.6003} & 425.10843 & 350.97201 & 510.68634 & 519.25189 \\
                     & Kruskal & \textbf{39.60784} & 76.686275 & 164.23529 & 354.88235 & 213.17647 & 245.41176 & 256.62745 & 285.37255 \\
                     & WSRT &  & 2.2E-06(+) & 5.3E-16(+) & 3.3E-18(+) & 5.0E-18(+) & 3.3E-18(+) & 3.9E-18(+) & 3.7E-18(+) \\
                    \midrule
                    \multirow{4}{*}{F17} & Mean & \textbf{2418.729} & 2719.7053 & 3185.1532 & 3568.7941 & 3503.938 & 3271.9265 & 3775.8722 & 3724.1766 \\
                     & Std & \textbf{173.8556} & 273.46237 & 423.69382 & 279.89768 & 425.66708 & 317.99089 & 416.59638 & 272.84868 \\
                     & Kruskal & \textbf{38.78431} & 86.901961 & 179.09804 & 272.4902 & 255.4902 & 194.27451 & 303.72549 & 305.23529 \\
                     & WSRT &  & 3.2E-08(+) & 4.1E-15(+) & 3.7E-18(+) & 4.4E-18(+) & 4.2E-17(+) & 3.3E-18(+) & 3.3E-18(+) \\
                    \midrule
                    \multirow{4}{*}{F18} & Mean & \textbf{2889.954} & 252038.96 & 149527.74 & 5100801.1 & 382040.7 & 7963272.3 & 2078889.8 & 551327.19 \\
                     & Std & \textbf{837.5434} & 136017.17 & 67683.289 & 2225158.8 & 193530.5 & 11746533 & 1518737.6 & 179796.68 \\
                     & Kruskal & \textbf{26} & 134.2549 & 95.607843 & 356.41176 & 176.78431 & 335.37255 & 289.88235 & 221.68627 \\
                     & WSRT &  & 3.3E-18(+) & 3.3E-18(+) & 3.3E-18(+) & 3.3E-18(+) & 3.3E-18(+) & 3.3E-18(+) & 3.3E-18(+) \\
                    \midrule
                    \multirow{4}{*}{F19} & Mean & \textbf{2056.599} & 117421.65 & 17545.893 & 17701.825 & 23796.405 & 6311633.6 & 591776.92 & 152148.13 \\
                     & Std & \textbf{60.36797} & 76454.106 & 10820.538 & 10792.076 & 12846.567 & 7026606.7 & 1079727.3 & 134463.91 \\
                     & Kruskal & \textbf{29.03922} & 254.39216 & 122.29412 & 120.62745 & 146.4902 & 381.84314 & 317.09804 & 264.21569 \\
                     & WSRT &  & 3.3E-18(+) & 3.3E-18(+) & 1.8E-14(+) & 3.3E-18(+) & 3.3E-18(+) & 3.3E-18(+) & 3.3E-18(+) \\
                    \midrule
                    \multirow{4}{*}{F20} & Mean & \textbf{2437.44} & 2839.3975 & 3111.7539 & 3688.2487 & 3346.1449 & 3358.2875 & 3459.6085 & 3839.1618 \\
                     & Std & \textbf{131.8813} & 243.2789 & 320.4854 & 145.48906 & 318.27883 & 232.07294 & 310.20894 & 210.51489 \\
                     & Kruskal & \textbf{30.68627} & 98.294118 & 158.03922 & 317.76471 & 217.11765 & 216.35294 & 248.17647 & 349.56863 \\
                     & WSRT &  & 1.4E-13(+) & 2.3E-17(+) & 3.3E-18(+) & 4.7E-18(+) & 3.5E-18(+) & 4.2E-18(+) & 3.3E-18(+) \\
                    \bottomrule
                \end{tabular}
            \end{table*}

            \begin{table*}[htbp!]
                \centering
                \caption{Experimental results of DoS and advanced competitors on 50D CEC2017 benchmark test functions (Composition).}
                \label{tab11}
                \begin{tabular}{\cm{0.02\textwidth}\cm{0.06\textwidth}\cm{0.085\textwidth}\cm{0.085\textwidth}\cm{0.085\textwidth}\cm{0.085\textwidth}\cm{0.085\textwidth}\cm{0.085\textwidth}\cm{0.085\textwidth}\cm{0.085\textwidth}\cm{0.085\textwidth}}
                    \toprule
                    No. & Metrics & DoS & AOO & PSA & FFA & AVOA & ETO & SGA & WAA \\
                    \midrule
                    \multirow{4}{*}{F21} & Mean & \textbf{2404.251} & 2478.1009 & 2560.141 & 2666.5647 & 2740.9989 & 2766.9726 & 2809.8841 & 3106.3958 \\
                     & Std & \textbf{14.8067} & 40.419561 & 41.070358 & 74.808223 & 83.488362 & 52.063935 & 102.38993 & 112.45899 \\
                     & Kruskal & \textbf{27.13725} & 81.411765 & 131.2549 & 197.96078 & 251.56863 & 274.35294 & 291 & 381.31373 \\
                     & WSRT &  & 9.4E-17(+) & 3.3E-18(+) & 3.3E-18(+) & 3.3E-18(+) & 3.3E-18(+) & 3.3E-18(+) & 3.3E-18(+) \\
                    \midrule
                    \multirow{4}{*}{F22} & Mean & \textbf{6284.981} & 8371.8925 & 8733.8068 & 13007.794 & 10401.36 & 13942.92 & 11350.239 & 10992.011 \\
                     & Std & 892.27191 & \textbf{731.5659} & 1341.8168 & 4826.0349 & 1014.7198 & 1235.6665 & 1299.1237 & 961.13311 \\
                     & Kruskal & \textbf{37.80392} & 111.23529 & 134.96078 & 298.62745 & 214.41176 & 338.96078 & 256.4902 & 243.5098 \\
                     & WSRT &  & 3.5E-18(+) & 1.1E-16(+) & 3.8E-07(+) & 3.3E-18(+) & 3.3E-18(+) & 3.3E-18(+) & 3.3E-18(+) \\
                    \midrule
                    \multirow{4}{*}{F23} & Mean & \textbf{2790.402} & 2929.2471 & 3076.9321 & 2955.7869 & 3357.475 & 3364.0782 & 3669.0811 & 4391.0206 \\
                     & Std & \textbf{21.30589} & 45.640249 & 83.79211 & 87.38988 & 107.30542 & 70.132346 & 174.23964 & 234.55009 \\
                     & Kruskal & \textbf{26.05882} & 101.4902 & 169.31373 & 114.47059 & 257.31373 & 257.70588 & 326.90196 & 382.7451 \\
                     & WSRT &  & 3.5E-18(+) & 3.3E-18(+) & 3.7E-18(+) & 3.3E-18(+) & 3.3E-18(+) & 3.3E-18(+) & 3.3E-18(+) \\
                    \midrule
                    \multirow{4}{*}{F24} & Mean & \textbf{3009.852} & 3093.9456 & 3231.4075 & 3296.9562 & 3591.9587 & 3608.9625 & 3824.4668 & 4377.3548 \\
                     & Std & \textbf{24.03879} & 57.327666 & 107.91972 & 19.957172 & 141.02818 & 76.246627 & 211.80716 & 149.1794 \\
                     & Kruskal & \textbf{28.70588} & 79.784314 & 135.7451 & 168.17647 & 260.96078 & 267.15686 & 313.19608 & 382.27451 \\
                     & WSRT &  & 6.0E-15(+) & 4.2E-18(+) & 3.3E-18(+) & 3.3E-18(+) & 3.3E-18(+) & 3.3E-18(+) & 3.3E-18(+) \\
                    \midrule
                    \multirow{4}{*}{F25} & Mean & \textbf{2968.833} & 3025.8936 & 3047.7998 & 3064.4626 & 3061.4041 & 4638.9591 & 3120.553 & 3073.2565 \\
                     & Std & 33.590393 & 33.362062 & 39.059624 & 24.501291 & 33.164013 & 528.44852 & 28.235566 & \textbf{24.39106} \\
                     & Kruskal & \textbf{39.01961} & 116.33333 & 164.15686 & 198.84314 & 192.2549 & 383 & 318.90196 & 223.4902 \\
                     & WSRT &  & 1.8E-12(+) & 3.5E-13(+) & 1.1E-16(+) & 1.6E-15(+) & 3.3E-18(+) & 3.9E-18(+) & 2.0E-17(+) \\
                    \midrule
                    \multirow{4}{*}{F26} & Mean & \textbf{4028.313} & 5094.9033 & 6989.2165 & 5616.083 & 8868.7991 & 10155.383 & 12019.176 & 12566.885 \\
                     & Std & \textbf{383.5196} & 1459.1209 & 1636.9614 & 647.12254 & 2852.5 & 766.31556 & 1863.6407 & 550.71848 \\
                     & Kruskal & \textbf{54.5098} & 104.68627 & 171.7451 & 121.21569 & 220.03922 & 262.90196 & 338.72549 & 362.17647 \\
                     & WSRT &  & 6.3E-05(+) & 2.1E-12(+) & 3.0E-17(+) & 2.8E-09(+) & 3.3E-18(+) & 6.7E-17(+) & 3.3E-18(+) \\
                    \midrule
                    \multirow{4}{*}{F27} & Mean & \textbf{3200.011} & 3351.9457 & 3522.0545 & 3329.3773 & 3744.9983 & 4161.0991 & 3942.265 & 5902.2563 \\
                     & Std & \textbf{0.00014} & 64.593004 & 133.82913 & 43.778659 & 241.73346 & 188.93897 & 245.07028 & 829.66284 \\
                     & Kruskal & \textbf{26} & 113.7451 & 182.5098 & 101.98039 & 235.82353 & 317 & 276.2549 & 382.68627 \\
                     & WSRT &  & 3.3E-18(+) & 3.3E-18(+) & 3.3E-18(+) & 3.3E-18(+) & 3.3E-18(+) & 3.3E-18(+) & 3.3E-18(+) \\
                    \midrule
                    \multirow{4}{*}{F28} & Mean & 3299.2651 & \textbf{3288.217} & 3301.0225 & 3338.9459 & 3302.6131 & 5045.0724 & 3333.3892 & 3302.2879 \\
                     & Std & \textbf{3.595295} & 21.439867 & 30.503813 & 26.12062 & 20.981256 & 460.34286 & 22.544042 & 20.48227 \\
                     & Kruskal & \textbf{111.3922} & 111.58824 & 130.43137 & 287.29412 & 153 & 383 & 283.23529 & 176.05882 \\
                     & WSRT &  & 9.1E-01($\approx$) & 9.4E-02($\approx$) & 8.7E-16(+) & 3.4E-02(+) & 3.3E-18(+) & 9.6E-16(+) & 9.2E-04(+) \\
                    \midrule
                    \multirow{4}{*}{F29} & Mean & \textbf{3442.742} & 4121.89 & 4299.3474 & 4180.1437 & 4761.6165 & 5815.6851 & 6649.5095 & 5960.7296 \\
                     & Std & \textbf{127.5843} & 276.50494 & 300.36578 & 442.53767 & 420.72607 & 418.83022 & 762.90399 & 607.44898 \\
                     & Kruskal & \textbf{28.41176} & 122.01961 & 150.5098 & 131.96078 & 211.96078 & 310.47059 & 361.98039 & 318.68627 \\
                     & WSRT &  & 1.8E-17(+) & 3.7E-18(+) & 6.2E-16(+) & 3.3E-18(+) & 3.3E-18(+) & 3.3E-18(+) & 3.3E-18(+) \\
                    \midrule
                    \multirow{4}{*}{F30} & Mean & \textbf{3545.922} & 7145878.4 & 999690.62 & 871792.39 & 1643392.5 & 185894209 & 39191527 & 11067002 \\
                     & Std & \textbf{391.276} & 1760462.2 & 178439.45 & 166130.3 & 412704.26 & 269064073 & 12085920 & 2988111.9 \\
                     & Kruskal & \textbf{26} & 236.62745 & 116.11765 & 93.411765 & 174.47059 & 383 & 331.88235 & 274.4902 \\
                     & WSRT &  & 3.3E-18(+) & 3.3E-18(+) & 3.3E-18(+) & 3.3E-18(+) & 3.3E-18(+) & 3.3E-18(+) & 3.3E-18(+) \\
                    \midrule
                    \multicolumn{2}{c}{FMR} & \textbf{1.207573} & 3.1406356 & 3.2792427 & 4.643002 & 4.4347532 & 6.8891143 & 6.2623394 & 6.1433401 \\
                    \multicolumn{2}{c}{F-Rank} & \textbf{1} & 2 & 3 & 5 & 4 & 8 & 7 & 6 \\
                    \bottomrule
                \end{tabular}
                \vspace{-4pt}
                \begin{flushleft}
                     \footnotesize
                     \rmfamily
                     FMR and F-Rank are the statistical results over all CEC2017 benchmark test functions (50D).
                \end{flushleft}
            \end{table*}

            As shown in the numerical results from Tables~\ref{tab9}, ~\ref{tab10} and~\ref{tab11}, on the 50-dimensional CEC2017 benchmark test functions, DoS achieves the best values in all three indicators, Mean, Std and Kruskal on 21 functions, except for F4, F6, F10, F11, F16, F22, F25 and F28. This demonstrates a clear advantage in terms of performance and stability. In particular, for the unimodal functions F1 and F3, DoS converges to the global optimal solution in all 51 independent runs, reflecting its strong exploitation ability. Moreover, in multimodal, hybrid and composition functions, DoS maintains a significant comparative advantage, indicating its excellent exploration ability as well.

            Although in F4, F6, F10, F11, F16, F22 and F25, DoS only outperforms competitors in terms of Mean and is slightly inferior to WAA, FFA and AOO in Std, this actually reflects that its dynamic selection mechanism achieves a better balance between exploration and exploitation. Compared with other algorithms, DoS can more thoroughly explore the feasible domain and avoid getting trapped in local optima. Although this introduces some fluctuations, the overall optimization ability is significantly improved. In the Kruskal-Wallis test, DoS ranks first on 29 test functions, showing outstanding performance.

            In terms of statistical significance, DoS performs significantly better than its competitors on 28 functions and shows comparable performance with AOO and PSA only on F28. The Friedman test results further show that DoS ranks first in average ranking, with a lead of 1.94 over the second-best algorithm, AOO, highlighting its stability and advantage in medium dimensional optimization problems.

            \begin{table*}[htbp!]
                \centering
                \caption{Experimental results of DoS and advanced competitors on 100D CEC2017 benchmark test functions (Unimodal and Multimodal).}
                \label{tab12}
                \begin{tabular}{\cm{0.02\textwidth}\cm{0.06\textwidth}\cm{0.085\textwidth}\cm{0.085\textwidth}\cm{0.085\textwidth}\cm{0.085\textwidth}\cm{0.085\textwidth}\cm{0.085\textwidth}\cm{0.085\textwidth}\cm{0.085\textwidth}\cm{0.085\textwidth}}
                    \toprule
                    No. & Metrics & DoS & AOO & PSA & FFA & AVOA & ETO & SGA & WAA \\
                    \midrule
                    \multirow{4}{*}{F1} & Mean & \textbf{100} & 11434.827 & 9153.6919 & 2544239.6 & 15030.121 & 8.79E+10 & 553528597 & 28429335 \\
                     & Std & \textbf{1.79E-09} & 12022.747 & 10985.079 & 1057815.3 & 26942.256 & 1.178E+10 & 478815373 & 2733487.8 \\
                     & Kruskal & \textbf{26} & 132.80392 & 123.88235 & 230 & 127.31373 & 383 & 332 & 281 \\
                     & WSRT &  & 3.3E-18(+) & 3.3E-18(+) & 3.3E-18(+) & 3.3E-18(+) & 3.3E-18(+) & 3.3E-18(+) & 3.3E-18(+) \\
                    \midrule
                    \multirow{4}{*}{F3} & Mean & \textbf{300.0001} & 316.54711 & 60838.978 & 489769.43 & 16618.625 & 247187.64 & 125616.3 & 15583.486 \\
                     & Std & \textbf{0.000108} & 13.01766 & 63938.281 & 35277.33 & 4668.4778 & 19100.352 & 14829.806 & 3253.5886 \\
                     & Kruskal & \textbf{26} & 77 & 215.64706 & 383 & 168.98039 & 330.82353 & 274.47059 & 160.07843 \\
                     & WSRT &  & 3.3E-18(+) & 3.3E-18(+) & 3.3E-18(+) & 3.3E-18(+) & 3.3E-18(+) & 3.3E-18(+) & 3.3E-18(+) \\
                    \midrule
                    \multirow{4}{*}{F4} & Mean & \textbf{499.4334} & 643.94986 & 636.87283 & 771.30545 & 638.51502 & 11434.472 & 1061.8479 & 667.13172 \\
                     & Std & 64.788848 & \textbf{25.95666} & 49.47919 & 46.836753 & 40.477182 & 2834.7527 & 107.44049 & 43.502178 \\
                     & Kruskal & \textbf{30.66667} & 148.86275 & 137.96078 & 276.84314 & 143.17647 & 383 & 331.94118 & 183.54902 \\
                     & WSRT &  & 2.1E-17(+) & 1.1E-15(+) & 3.3E-18(+) & 2.2E-16(+) & 3.3E-18(+) & 3.3E-18(+) & 2.0E-17(+) \\
                    \midrule
                    \multirow{4}{*}{F5} & Mean & \textbf{904.6199} & 980.36006 & 1238.994 & 1516.454 & 1314.5973 & 1660.0572 & 1393.9979 & 1408.4371 \\
                     & Std & 46.84234 & 69.67801 & 98.469226 & 132.64523 & 76.511927 & 51.396168 & 67.294038 & \textbf{33.53873} \\
                     & Kruskal & \textbf{35.90196} & 67.666667 & 152.15686 & 308.11765 & 188.92157 & 376.84314 & 245.33333 & 261.05882 \\
                     & WSRT &  & 1.0E-07(+) & 3.3E-18(+) & 3.3E-18(+) & 3.3E-18(+) & 3.3E-18(+) & 3.3E-18(+) & 3.3E-18(+) \\
                    \midrule
                    \multirow{4}{*}{F6} & Mean & \textbf{600} & 636.02547 & 601.81291 & 602.87896 & 643.78282 & 675.19617 & 672.02774 & 671.22654 \\
                     & Std & \textbf{1.79E-07} & 5.4579063 & 1.330067 & 0.9751589 & 4.5218746 & 5.4426438 & 3.5616361 & 2.0948545 \\
                     & Kruskal & \textbf{26} & 185.4902 & 88.254902 & 116.7451 & 223.5098 & 357.07843 & 326.62745 & 312.29412 \\
                     & WSRT &  & 3.2E-18(+) & 3.2E-18(+) & 3.2E-18(+) & 3.2E-18(+) & 3.2E-18(+) & 3.2E-18(+) & 3.2E-18(+) \\
                    \midrule
                    \multirow{4}{*}{F7} & Mean & \textbf{1299.162} & 1438.8492 & 2075.9228 & 1637.8928 & 2826.4009 & 2805.5654 & 3212.2258 & 3491.3025 \\
                     & Std & 70.832237 & 103.97397 & 191.46112 & 148.07439 & 159.68196 & 136.39185 & 158.30975 & \textbf{68.38764} \\
                     & Kruskal & \textbf{33} & 78.215686 & 177.45098 & 121.64706 & 259.45098 & 255.05882 & 329.76471 & 381.41176 \\
                     & WSRT &  & 2.3E-11(+) & 3.3E-18(+) & 8.4E-17(+) & 3.3E-18(+) & 3.3E-18(+) & 3.3E-18(+) & 3.3E-18(+) \\
                    \midrule
                    \multirow{4}{*}{F8} & Mean & \textbf{1196.997} & 1301.1995 & 1580.1334 & 1778.4256 & 1711.509 & 1999.2754 & 1829.3321 & 1892.6368 \\
                     & Std & 44.761791 & 85.12443 & 103.00011 & 166.94534 & 63.089571 & 72.90298 & 66.657071 & \textbf{31.49254} \\
                     & Kruskal & \textbf{32.41176} & 73.137255 & 141.84314 & 247.37255 & 191.07843 & 372.90196 & 263.56863 & 313.68627 \\
                     & WSRT &  & 5.2E-11(+) & 3.3E-18(+) & 5.3E-18(+) & 3.3E-18(+) & 3.3E-18(+) & 3.3E-18(+) & 3.3E-18(+) \\
                    \midrule
                    \multirow{4}{*}{F9} & Mean & \textbf{9842.292} & 17371.124 & 26628.345 & 25952.071 & 22605.19 & 67107.475 & 27707.789 & 43774.083 \\
                     & Std & 2121.9743 & 4064.3951 & 3857.9097 & 8646.918 & \textbf{1051.23} & 6576.3486 & 2179.1687 & 3166.527 \\
                     & Kruskal & \textbf{28.52941} & 93.568627 & 215.72549 & 192.88235 & 152.03922 & 382.90196 & 240.7451 & 329.60784 \\
                     & WSRT &  & 8.7E-16(+) & 3.3E-18(+) & 2.0E-17(+) & 3.3E-18(+) & 3.3E-18(+) & 3.3E-18(+) & 3.3E-18(+) \\
                    \midrule
                    \multirow{4}{*}{F10} & Mean & \textbf{11269.18} & 13069.852 & 14570.173 & 30959.545 & 16523.415 & 27990.615 & 20080.44 & 17274.403 \\
                     & Std & 622.0695 & 1295.7596 & 1332.2332 & \textbf{461.1482} & 1602.8631 & 1348.1399 & 2956.8032 & 972.66823 \\
                     & Kruskal & \textbf{31.45098} & 85.117647 & 128.72549 & 382.60784 & 190.86275 & 331.2549 & 266.52941 & 219.45098 \\
                     & WSRT &  & 1.7E-12(+) & 2.1E-17(+) & 3.3E-18(+) & 3.3E-18(+) & 3.3E-18(+) & 3.3E-18(+) & 3.3E-18(+) \\
                    \bottomrule
                \end{tabular}
            \end{table*}

            \begin{table*}[htbp!]
                \centering
                \caption{Experimental results of DoS and advanced competitors on 100D CEC2017 benchmark test functions (Hybrid).}
                \label{tab13}
                \begin{tabular}{\cm{0.02\textwidth}\cm{0.06\textwidth}\cm{0.085\textwidth}\cm{0.085\textwidth}\cm{0.085\textwidth}\cm{0.085\textwidth}\cm{0.085\textwidth}\cm{0.085\textwidth}\cm{0.085\textwidth}\cm{0.085\textwidth}\cm{0.085\textwidth}}
                    \toprule
                    No. & Metrics & DoS & AOO & PSA & FFA & AVOA & ETO & SGA & WAA \\
                    \midrule
                    \multirow{4}{*}{F11} & Mean & \textbf{2093.826} & 2336.136 & 2499.4601 & 45253.726 & 2364.2269 & 63454.389 & 5624.7345 & 2666.6838 \\
                     & Std & 260.32616 & 223.32864 & 365.46358 & 6151.0608 & 203.67499 & 11307.545 & 1001.9915 & \textbf{185.3469} \\
                     & Kruskal & \textbf{60.82353} & 112.07843 & 150.45098 & 335.92157 & 119.47059 & 379.07843 & 281 & 197.17647 \\
                     & WSRT &  & 1.2E-05(+) & 6.1E-08(+) & 3.3E-18(+) & 1.1E-06(+) & 3.3E-18(+) & 3.3E-18(+) & 1.6E-16(+) \\
                    \midrule
                    \multirow{4}{*}{F12} & Mean & \textbf{70696.23} & 27497571 & 1321869 & 26204306 & 10543347 & 2.452E+10 & 378055056 & 46529684 \\
                     & Std & \textbf{46332.36} & 12742127 & 689098.2 & 9472511.5 & 4533071.7 & 1.194E+10 & 159585023 & 6876212.1 \\
                     & Kruskal & \textbf{26} & 205.33333 & 77.156863 & 203.39216 & 134.80392 & 383 & 332 & 274.31373 \\
                     & WSRT &  & 3.3E-18(+) & 3.3E-18(+) & 3.3E-18(+) & 3.3E-18(+) & 3.3E-18(+) & 3.3E-18(+) & 3.3E-18(+) \\
                    \midrule
                    \multirow{4}{*}{F13} & Mean & \textbf{4454.267} & 85648.223 & 8132.8355 & 15356.532 & 47599.804 & 2.789E+09 & 104445.51 & 599208.66 \\
                     & Std & \textbf{2231.859} & 36140.043 & 7294.2077 & 51117.923 & 15756.806 & 2.655E+09 & 116417.53 & 66059.953 \\
                     & Kruskal & \textbf{70} & 247.45098 & 95.803922 & 73.745098 & 188.7451 & 383 & 246.80392 & 330.45098 \\
                     & WSRT &  & 3.3E-18(+) & 1.6E-03(+) & 4.4E-01($\approx$) & 3.3E-18(+) & 3.3E-18(+) & 3.3E-18(+) & 3.3E-18(+) \\
                    \midrule
                    \multirow{4}{*}{F14} & Mean & \textbf{2176.241} & 215634.62 & 102155.61 & 8802351.6 & 162433.31 & 6010552.4 & 1463966.9 & 299704.97 \\
                     & Std & \textbf{223.8973} & 123978.52 & 50784.534 & 3015230.9 & 61713.227 & 2961045.6 & 801206.11 & 72743.323 \\
                     & Kruskal & \textbf{26} & 162.23529 & 97.862745 & 369.82353 & 140.5098 & 342.88235 & 281.84314 & 214.84314 \\
                     & WSRT &  & 3.3E-18(+) & 3.3E-18(+) & 3.3E-18(+) & 3.3E-18(+) & 3.3E-18(+) & 3.3E-18(+) & 3.3E-18(+) \\
                    \midrule
                    \multirow{4}{*}{F15} & Mean & \textbf{2391.339} & 69900.429 & 5518.5835 & 21516.984 & 31328.556 & 361516597 & 119079.57 & 153572 \\
                     & Std & \textbf{1418.561} & 30682.356 & 4783.5966 & 58341.072 & 10585.893 & 608735919 & 302248.11 & 30003.227 \\
                     & Kruskal & \textbf{44.07843} & 245.4902 & 97.843137 & 107.84314 & 178.62745 & 383 & 255.03922 & 324.07843 \\
                     & WSRT &  & 3.3E-18(+) & 9.7E-11(+) & 1.9E-06(+) & 3.5E-18(+) & 3.3E-18(+) & 3.3E-18(+) & 3.3E-18(+) \\
                    \midrule
                    \multirow{4}{*}{F16} & Mean & \textbf{4322.565} & 4951.892 & 5795.6355 & 10163.523 & 6349.6419 & 9710.3545 & 7750.8686 & 7414.9598 \\
                     & Std & \textbf{405.9632} & 651.60008 & 609.45896 & 357.30588 & 682.62745 & 1258.1596 & 1060.2266 & 577.5631 \\
                     & Kruskal & \textbf{36.88235} & 79.490196 & 136.4902 & 364.17647 & 174.52941 & 342.92157 & 258 & 243.5098 \\
                     & WSRT &  & 1.1E-07(+) & 3.2E-17(+) & 3.3E-18(+) & 5.9E-18(+) & 3.3E-18(+) & 3.3E-18(+) & 3.3E-18(+) \\
                    \midrule
                    \multirow{4}{*}{F17} & Mean & \textbf{3932.694} & 4477.6843 & 5437.6177 & 6831.2505 & 5955.2657 & 8248.3973 & 6919.1373 & 5988.4034 \\
                     & Std & \textbf{327.2295} & 536.4395 & 641.08481 & 493.61691 & 686.81887 & 3852.3467 & 946.52654 & 690.73377 \\
                     & Kruskal & \textbf{36.82353} & 79.078431 & 160.31373 & 310.47059 & 211.88235 & 320.64706 & 301.52941 & 215.2549 \\
                     & WSRT &  & 9.6E-08(+) & 1.1E-17(+) & 3.3E-18(+) & 1.7E-17(+) & 3.3E-18(+) & 3.3E-18(+) & 3.3E-18(+) \\
                    \midrule
                    \multirow{4}{*}{F18} & Mean & \textbf{4298.386} & 416232.2 & 280470.84 & 17342725 & 284063.12 & 6321902.1 & 2416922.2 & 885691.47 \\
                     & Std & \textbf{1602.503} & 165783.79 & 163227.62 & 3925354 & 99000.537 & 2860777.4 & 882296.51 & 394334.48 \\
                     & Kruskal & \textbf{26} & 158.7451 & 110.78431 & 382.31373 & 120.09804 & 328.72549 & 281.58824 & 227.7451 \\
                     & WSRT &  & 3.3E-18(+) & 3.3E-18(+) & 3.3E-18(+) & 3.3E-18(+) & 3.3E-18(+) & 3.3E-18(+) & 3.3E-18(+) \\
                    \midrule
                    \multirow{4}{*}{F19} & Mean & \textbf{3710.711} & 290837.38 & 5722.7038 & 12303.77 & 16963.956 & 514617222 & 5425954.7 & 1585898.6 \\
                     & Std & \textbf{2211.767} & 135120.16 & 4728.225 & 24051.218 & 7892.9895 & 616405725 & 3141094.9 & 548486.25 \\
                     & Kruskal & \textbf{60.86275} & 230.09804 & 92.019608 & 90.333333 & 166.78431 & 383 & 326.76471 & 286.13725 \\
                     & WSRT &  & 3.3E-18(+) & 3.4E-04(+) & 2.9E-02(+) & 3.2E-17(+) & 3.3E-18(+) & 3.3E-18(+) & 3.3E-18(+) \\
                    \midrule
                    \multirow{4}{*}{F20} & Mean & \textbf{4022.997} & 4587.5177 & 5355.894 & 7095.0004 & 5735.6564 & 6100.8094 & 6071.921 & 6237.9918 \\
                     & Std & 269.95347 & 460.41147 & 596.22632 & \textbf{198.7064} & 633.39178 & 605.70915 & 590.19839 & 556.8732 \\
                     & Kruskal & \textbf{35.09804} & 84.686275 & 162.33333 & 371.62745 & 208.76471 & 249.96078 & 250.92157 & 272.60784 \\
                     & WSRT &  & 3.7E-10(+) & 7.3E-16(+) & 3.3E-18(+) & 4.4E-18(+) & 3.3E-18(+) & 3.3E-18(+) & 3.3E-18(+) \\
                    \bottomrule
                \end{tabular}
            \end{table*}

            \begin{table*}[htbp!]
                \centering
                \caption{Experimental results of DoS and advanced competitors on 100D CEC2017 benchmark test functions (Composition).}
                \label{tab14}
                \begin{tabular}{\cm{0.02\textwidth}\cm{0.06\textwidth}\cm{0.085\textwidth}\cm{0.085\textwidth}\cm{0.085\textwidth}\cm{0.085\textwidth}\cm{0.085\textwidth}\cm{0.085\textwidth}\cm{0.085\textwidth}\cm{0.085\textwidth}\cm{0.085\textwidth}}
                    \toprule
                    No. & Metrics & DoS & AOO & PSA & FFA & AVOA & ETO & SGA & WAA \\
                    \midrule
                    \multirow{4}{*}{F21} & Mean & \textbf{2687.31} & 2817.1405 & 3089.7996 & 3195.7371 & 3549.1034 & 3773.2339 & 3891.7374 & 4531.5168 \\
                     & Std & \textbf{42.61848} & 85.264587 & 104.49185 & 176.04385 & 192.28252 & 111.68982 & 239.68755 & 165.59949 \\
                     & Kruskal & \textbf{30.2549} & 75.627451 & 142.72549 & 166 & 240.68627 & 289.29412 & 309.23529 & 382.17647 \\
                     & WSRT &  & 2.8E-13(+) & 3.3E-18(+) & 5.3E-18(+) & 3.3E-18(+) & 3.3E-18(+) & 3.3E-18(+) & 3.3E-18(+) \\
                    \midrule
                    \multirow{4}{*}{F22} & Mean & \textbf{13378.01} & 15909.747 & 17807.914 & 33116.244 & 19590.348 & 30978.01 & 23117.352 & 20489.071 \\
                     & Std & 1682.1851 & 1481.4432 & 1560.3427 & \textbf{539.5097} & 1692.7854 & 1388.0878 & 2169.571 & 1328.2754 \\
                     & Kruskal & \textbf{30.86275} & 85.745098 & 134.98039 & 379.60784 & 187.19608 & 335.23529 & 268.58824 & 213.78431 \\
                     & WSRT &  & 1.9E-12(+) & 3.3E-18(+) & 3.3E-18(+) & 3.3E-18(+) & 3.3E-18(+) & 3.3E-18(+) & 3.3E-18(+) \\
                    \midrule
                    \multirow{4}{*}{F23} & Mean & \textbf{3032.522} & 3328.5723 & 3389.6055 & 3083.5447 & 3783.339 & 4518.4414 & 5043.2614 & 6436.9036 \\
                     & Std & \textbf{27.67774} & 73.602701 & 117.78652 & 34.219696 & 153.71478 & 124.79537 & 311.09346 & 768.66897 \\
                     & Kruskal & \textbf{32.33333} & 145.82353 & 162.09804 & 70.72549 & 229.01961 & 283.58824 & 330.82353 & 381.58824 \\
                     & WSRT &  & 3.3E-18(+) & 3.3E-18(+) & 6.2E-11(+) & 3.3E-18(+) & 3.3E-18(+) & 3.3E-18(+) & 3.3E-18(+) \\
                    \midrule
                    \multirow{4}{*}{F24} & Mean & 3674.1432 & 3820.2856 & 4128.3965 & \textbf{3598.212} & 4737.9563 & 5745.9319 & 6339.9629 & 8580.0161 \\
                     & Std & \textbf{74.67906} & 97.07425 & 152.63608 & 94.014668 & 253.97023 & 265.1987 & 514.51641 & 1548.2708 \\
                     & Kruskal & \textbf{73.11765} & 121.2549 & 177.64706 & 38.509804 & 229.82353 & 288.90196 & 329.60784 & 377.13725 \\
                     & WSRT &  & 2.3E-11(+) & 3.3E-18(+) & 8.3E-08(-) & 3.3E-18(+) & 3.3E-18(+) & 3.3E-18(+) & 3.3E-18(+) \\
                    \midrule
                    \multirow{4}{*}{F25} & Mean & \textbf{3232.027} & 3256.0455 & 3303.9447 & 3489.7101 & 3310.6253 & 8665.1744 & 3550.1983 & 3317.1892 \\
                     & Std & 67.830152 & 57.747201 & 73.760736 & \textbf{55.87312} & 64.481479 & 1336.0849 & 84.615751 & 50.813122 \\
                     & Kruskal & \textbf{76.37255} & 95.705882 & 151.54902 & 294.01961 & 156.35294 & 383 & 316.05882 & 162.94118 \\
                     & WSRT &  & 5.5E-02($\approx$) & 2.7E-06(+) & 3.3E-18(+) & 1.8E-07(+) & 3.3E-18(+) & 3.7E-18(+) & 7.1E-09(+) \\
                    \midrule
                    \multirow{4}{*}{F26} & Mean & 10554.709 & 10894.301 & 16019.937 & \textbf{9331.776} & 22504.056 & 25063.478 & 29891.053 & 26260.802 \\
                     & Std & 1059.3182 & 2333.5268 & 3108.4928 & \textbf{837.6775} & 2440.5727 & 1434.2431 & 5081.4525 & 1221.4587 \\
                     & Kruskal & 88.098039 & 107.47059 & 181.41176 & \textbf{39.58824} & 243.70588 & 290.35294 & 362.41176 & 322.96078 \\
                     & WSRT &  & 2.9E-03(+) & 9.4E-17(+) & 2.2E-10(-) & 3.3E-18(+) & 3.3E-18(+) & 4.4E-18(+) & 3.3E-18(+) \\
                    \midrule
                    \multirow{4}{*}{F27} & Mean & \textbf{3200.022} & 3473.0578 & 3694.7876 & 3550.3028 & 3957.9768 & 5138.5934 & 4704.5723 & 6511.3032 \\
                     & Std & \textbf{0.000198} & 59.393888 & 96.870455 & 80.498393 & 190.86168 & 292.03137 & 474.18335 & 1414.1513 \\
                     & Kruskal & \textbf{26} & 87.647059 & 178.31373 & 124.4902 & 226.52941 & 328.76471 & 294.05882 & 370.19608 \\
                     & WSRT &  & 3.3E-18(+) & 3.3E-18(+) & 3.3E-18(+) & 3.3E-18(+) & 3.3E-18(+) & 3.3E-18(+) & 3.3E-18(+) \\
                    \midrule
                    \multirow{4}{*}{F28} & Mean & \textbf{3300.022} & 3374.6808 & 3382.7937 & 3605.4397 & 3347.5405 & 10628.347 & 3706.4048 & 3371.1428 \\
                     & Std & \textbf{0.000224} & 33.73612 & 34.239748 & 38.34197 & 31.215903 & 1686.0586 & 73.021728 & 28.121203 \\
                     & Kruskal & \textbf{29} & 161.39216 & 177.62745 & 286.37255 & 112 & 383 & 326.62745 & 159.98039 \\
                     & WSRT &  & 3.3E-18(+) & 3.3E-18(+) & 3.3E-18(+) & 1.6E-14(+) & 3.3E-18(+) & 3.3E-18(+) & 3.3E-18(+) \\
                    \midrule
                    \multirow{4}{*}{F29} & Mean & \textbf{5214.179} & 6684.1913 & 7013.0317 & 6765.6676 & 7872.0657 & 10892.841 & 12699.242 & 9128.366 \\
                     & Std & 506.21232 & 496.21594 & \textbf{490.2713} & 1069.4739 & 667.77233 & 807.73563 & 1620.2471 & 591.11213 \\
                     & Kruskal & \textbf{31.62745} & 120.64706 & 150.88235 & 129.2549 & 213.35294 & 337.66667 & 374.29412 & 278.27451 \\
                     & WSRT &  & 8.4E-17(+) & 3.0E-17(+) & 1.0E-13(+) & 4.2E-18(+) & 3.3E-18(+) & 3.3E-18(+) & 3.3E-18(+) \\
                    \midrule
                    \multirow{4}{*}{F30} & Mean & \textbf{6430.107} & 3246799.4 & 12398.871 & 104714.3 & 207840.52 & 2.088E+09 & 82562715 & 5713969.6 \\
                     & Std & \textbf{3366.11} & 1206638 & 5176.9533 & 149532.06 & 96848.202 & 1.584E+09 & 42286195 & 1376174.1 \\
                     & Kruskal & \textbf{34.01961} & 232.60784 & 72.352941 & 133.29412 & 170.37255 & 383 & 332 & 278.35294 \\
                     & WSRT &  & 3.3E-18(+) & 1.1E-09(+) & 1.0E-17(+) & 3.3E-18(+) & 3.3E-18(+) & 3.3E-18(+) & 3.3E-18(+) \\
                    \midrule
                    \multicolumn{2}{c}{FMR} & \textbf{1.266396} & 3.0682894 & 3.2535497 & 4.8999324 & 4.0594997 & 7.2393509 & 6.3461799 & 5.8668019 \\
                    \multicolumn{2}{c}{F-Rank} & \textbf{1} & 2 & 3 & 5 & 4 & 8 & 7 & 6 \\
                    \bottomrule
                \end{tabular}
                \vspace{-4pt}
                \begin{flushleft}
                     \footnotesize
                     \rmfamily
                     FMR and F-Rank are the statistical results over all CEC2017 benchmark test functions (100D).
                \end{flushleft}
            \end{table*}

            Tables~\ref{tab12}, ~\ref{tab13} and~\ref{tab14} present the experimental results of DoS on the 100-dimensional CEC2017 benchmark test functions. Except for F24 and F26, DoS shows excellent performance in terms of Mean and Kruskal indicators on all other functions, indicating its strong adaptability in high dimensional optimization tasks.

            Due to the structural complexity and density of local optima in high dimensional functions, DoS, under the guidance of its dynamic selection mechanism, conducts more extensive global exploration, which leads to slight differences in the exploitation regions across multiple independent runs. As a result, its stability on some functions is slightly lower than that of AOO, WAA, AVOA, FFA and PSA. For example, on F4, F5, F7, F8, F9, F10, F11, F20, F22, F25 and F29 a total of 11 functions, the Std of DoS is slightly worse than those of the competitors. However, this phenomenon is mainly caused by some algorithms getting trapped in local optima early in the search process. Although these algorithms exhibit smaller fluctuations, their Mean values are generally worse than those of DoS, thus their overall performance is not superior.

            In the Wilcoxon rank-sum test, DoS performs significantly better than other competitors on 25 functions, except for F13 and F24~F26, further validating its strong robustness and performance advantages in high-dimensional optimization problems. Although DoS performs comparably to FFA on F13 and to AOO on F25 and is slightly inferior to FFA on F24 and F26, the gaps in the Mean indicator between DoS and the best results are small and its overall performance remains within an acceptable range.

            The Friedman test results also show that DoS ranks first in average ranking among all methods, with a lead of 1.8 over the second-best algorithm, AOO, fully demonstrating its competitiveness in high-dimensional tasks.

            \begin{table*}[htbp!]
                \centering
                \caption{Experimental results of DoS and advanced competitors on 10D CEC2022 benchmark test functions.}
                \label{tab15}
                \begin{tabular}{\cm{0.02\textwidth}\cm{0.06\textwidth}\cm{0.085\textwidth}\cm{0.085\textwidth}\cm{0.085\textwidth}\cm{0.085\textwidth}\cm{0.085\textwidth}\cm{0.085\textwidth}\cm{0.085\textwidth}\cm{0.085\textwidth}\cm{0.085\textwidth}}
                    \toprule
                    No. & Metrics & DoS & AOO & PSA & FFA & AVOA & ETO & SGA & WAA \\
                    \midrule
                    \multirow{4}{*}{C1} & Mean & \textbf{300} & \textbf{300} & \textbf{300} & 300.33526 & \textbf{300} & 2591.1845 & 300.02927 & 300.0381 \\
                     & Std & \textbf{0} & 5.047E-08 & 6.922E-14 & 0.5229235 & 1.301E-13 & 2181.0563 & 0.0450559 & 0.0115081 \\
                     & Kruskal & \textbf{17} & 105.5 & 55.066667 & 186.33333 & 64.433333 & 225.5 & 145.4 & 164.76667 \\
                     & WSRT &  & 1.2E-12(+) & 1.1E-11(+) & 1.2E-12(+) & 3.2E-12(+) & 1.2E-12(+) & 1.2E-12(+) & 1.2E-12(+) \\
                    \midrule
                    \multirow{4}{*}{C2} & Mean & \textbf{403.1328} & 405.99319 & 403.92253 & 407.34966 & 413.59611 & 428.71844 & 418.64804 & 409.12183 \\
                     & Std & \textbf{1.030256} & 3.5072486 & 3.7235414 & 3.2614581 & 23.134137 & 37.315088 & 28.497419 & 17.201281 \\
                     & Kruskal & \textbf{74} & 124.3 & 81.516667 & 162.86667 & 109.65 & 139.36667 & 149.6 & 122.7 \\
                     & WSRT &  & 3.4E-05(+) & 1.2E-01($\approx$) & 8.1E-05(+) & 1.3E-02(+) & 7.0E-02($\approx$) & 5.1E-06(+) & 3.8E-02(+) \\
                    \midrule
                    \multirow{4}{*}{C3} & Mean & \textbf{600} & 600.20325 & 600 & 600.00979 & 604.69499 & 613.61211 & 618.84397 & 644.14483 \\
                     & Std & \textbf{0} & 0.380026 & 9.244E-11 & 0.022072 & 5.5438782 & 8.7362625 & 11.301923 & 10.096949 \\
                     & Kruskal & \textbf{16} & 107 & 45.766667 & 78 & 138.33333 & 172.23333 & 183 & 223.66667 \\
                     & WSRT &  & 1.2E-12(+) & 2.5E-12(+) & 1.2E-12(+) & 1.2E-12(+) & 1.2E-12(+) & 1.2E-12(+) & 1.2E-12(+) \\
                    \midrule
                    \multirow{4}{*}{C4} & Mean & \textbf{804.4445} & 813.67136 & 821.25901 & 822.62518 & 830.45045 & 826.55429 & 825.40461 & 839.14213 \\
                     & Std & \textbf{1.376077} & 6.8207061 & 7.4752319 & 4.3039578 & 11.565046 & 10.617232 & 9.34342 & 7.4060587 \\
                     & Kruskal & \textbf{16.8} & 66.566667 & 109.85 & 121.4 & 162.91667 & 139.83333 & 139.36667 & 207.26667 \\
                     & WSRT &  & 8.9E-10(+) & 3.0E-11(+) & 3.0E-11(+) & 4.5E-11(+) & 3.0E-11(+) & 3.0E-11(+) & 3.0E-11(+) \\
                    \midrule
                    \multirow{4}{*}{C5} & Mean & \textbf{900} & 900.04842 & 943.63206 & 900.26687 & 1046.4564 & 1049.5779 & 1015.1158 & 1474.1793 \\
                     & Std & \textbf{0} & 0.1385791 & 76.258644 & 0.2804191 & 186.69996 & 179.76697 & 86.994952 & 144.26707 \\
                     & Kruskal & \textbf{15.5} & 48.3 & 125.96667 & 73.833333 & 158.13333 & 158.6 & 161.93333 & 221.73333 \\
                     & WSRT &  & 1.2E-12(+) & 1.2E-12(+) & 1.2E-12(+) & 1.2E-12(+) & 1.2E-12(+) & 1.2E-12(+) & 1.2E-12(+) \\
                    \midrule
                    \multirow{4}{*}{C6} & Mean & \textbf{1803.835} & 4933.8797 & 3566.9497 & 3928.8872 & 4606.0945 & 15995.753 & 5120.3224 & 2122.3238 \\
                     & Std & \textbf{5.362242} & 2222.2208 & 1611.3283 & 2317.9508 & 2216.0887 & 11247.6 & 2303.1726 & 387.10921 \\
                     & Kruskal & \textbf{15.5} & 147.1 & 113.36667 & 116.53333 & 137.06667 & 212.26667 & 154.5 & 67.666667 \\
                     & WSRT &  & 3.0E-11(+) & 3.0E-11(+) & 3.0E-11(+) & 3.0E-11(+) & 3.0E-11(+) & 3.0E-11(+) & 3.0E-11(+) \\
                    \midrule
                    \multirow{4}{*}{C7} & Mean & \textbf{2000.963} & 2022.206 & 2018.0725 & 2008.8364 & 2022.8608 & 2040.6097 & 2036.89 & 2099.5237 \\
                     & Std & \textbf{3.658985} & 8.479453 & 5.9034516 & 8.3959996 & 7.1027243 & 19.087814 & 12.816271 & 42.872582 \\
                     & Kruskal & \textbf{18.93333} & 117.76667 & 75.266667 & 60.266667 & 125.56667 & 175.3 & 172.63333 & 218.26667 \\
                     & WSRT &  & 6.7E-11(+) & 3.5E-10(+) & 1.0E-08(+) & 4.5E-11(+) & 3.3E-11(+) & 4.1E-11(+) & 3.0E-11(+) \\
                    \midrule
                    \multirow{4}{*}{C8} & Mean & \textbf{2202.504} & 2217.2387 & 2219.2726 & 2215.2868 & 2221.534 & 2226.9403 & 2228.0515 & 2302.7774 \\
                     & Std & 3.9505512 & 8.5367334 & 5.1832188 & 8.610161 & \textbf{3.73739} & 5.7251247 & 5.6842061 & 97.110312 \\
                     & Kruskal & \textbf{21.16667} & 97.166667 & 77.8 & 88.7 & 113.03333 & 178.86667 & 177.56667 & 209.7 \\
                     & WSRT &  & 2.4E-09(+) & 1.6E-08(+) & 2.7E-09(+) & 5.5E-11(+) & 4.1E-11(+) & 3.3E-11(+) & 3.0E-11(+) \\
                    \midrule
                    \multirow{4}{*}{C9} & Mean & \textbf{2486.87} & 2529.2847 & 2529.2844 & 2529.5164 & 2529.2844 & 2559.0567 & 2534.1822 & 2529.2847 \\
                     & Std & 2.0089098 & 0.0005269 & 9.956E-13 & 0.6135013 & \textbf{1.46E-13} & 31.561907 & 26.825978 & 0.0001537 \\
                     & Kruskal & \textbf{15.5} & 140.26667 & 69.9 & 192.86667 & 51.1 & 223.63333 & 116.26667 & 154.46667 \\
                     & WSRT &  & 3.0E-11(+) & 2.2E-11(+) & 3.0E-11(+) & 3.2E-12(+) & 3.0E-11(+) & 3.0E-11(+) & 3.0E-11(+) \\
                    \midrule
                    \multirow{4}{*}{C10} & Mean & \textbf{2500.415} & 2503.9091 & 2535.1462 & 2503.9086 & 2549.386 & 2525.027 & 2522.8988 & 2762.5456 \\
                     & Std & 26.952171 & 19.9886 & 53.781311 & \textbf{19.486939} & 61.10995 & 38.639138 & 50.952252 & 368.31004 \\
                     & Kruskal & \textbf{35.6} & 43.733333 & 159.5 & 75.933333 & 157.93333 & 149.13333 & 142.56667 & 199.6 \\
                     & WSRT &  & 3.8E-01($\approx$) & 2.4E-10(+) & 1.6E-07(+) & 1.8E-10(+) & 6.1E-10(+) & 8.1E-10(+) & 1.2E-10(+) \\
                    \midrule
                    \multirow{4}{*}{C11} & Mean & 2850 & \textbf{2697.484} & 2937.1769 & 2716.5658 & 2740.9847 & 2995.6384 & 2757.8508 & 2852.2844 \\
                     & Std & 113.71471 & 153.11814 & 184.48101 & \textbf{70.76167} & 137.21471 & 160.22422 & 128.9277 & 114.28792 \\
                     & Kruskal & 98.833333 & \textbf{76.13333} & 151.36667 & 80 & 82.6 & 204.36667 & 108.36667 & 162.33333 \\
                     & WSRT &  & 2.5E-01($\approx$) & 7.3E-10(+) & 1.3E-04(-) & 7.1E-01($\approx$) & 8.0E-08(+) & 2.5E-01($\approx$) & 6.3E-07(+) \\
                    \midrule
                    \multirow{4}{*}{C12} & Mean & 2889.772 & 2863.2723 & 2865.526 & \textbf{2861.705} & 2864.8995 & 2880.6942 & 2865.1139 & 2963.9553 \\
                     & Std & 20.821272 & \textbf{1.251124} & 2.182712 & 1.3734052 & 2.42454 & 19.067232 & 1.9165205 & 70.583671 \\
                     & Kruskal & 158.7 & 64.733333 & 116.53333 & \textbf{34.56667} & 100.66667 & 162.26667 & 106 & 220.53333 \\
                     & WSRT &  & 6.8E-05(-) & 6.8E-05(-) & 6.8E-05(-) & 6.8E-05(-) & 2.7E-02(-) & 6.8E-05(-) & 1.7E-07(+) \\
                    \midrule
                    \multicolumn{2}{c}{FMR} & \textbf{1.883333} & 3.6555556 & 3.7527778 & 4.0263889 & 4.3541667 & 6.4444444 & 5.3555556 & 6.5277778 \\
                    \multicolumn{2}{c}{F-Rank} & \textbf{1} & 2 & 3 & 4 & 5 & 7 & 6 & 8 \\
                    \bottomrule
                \end{tabular}
                \vspace{-4pt}
                \begin{flushleft}
                     \footnotesize
                     \rmfamily
                     FMR and F-Rank are the statistical results over all CEC2022 benchmark test functions (10D).
                \end{flushleft}
            \end{table*}

            Table~\ref{tab15} summarizes the experimental comparison results between DoS and the advanced competitors on the 10-dimensional CEC2022 benchmark test functions. The results show that DoS achieves the best values in all three indicators, Mean, Std and Kruskal on 7 functions from C1 to C7. For C8, C9 and C10, although the Std of DoS is slightly worse than those of AVOA and AOO, showing slightly higher fluctuations, its Mean is better, indicating that although the competitors are more stable, they are more prone to getting stuck in local optima and lack effective escape mechanisms. In contrast, DoS can still find better solutions through its dynamic selection mechanism and ranks first overall in the Kruskal test, demonstrating its structural advantage in balancing exploration and exploitation.

            The Wilcoxon rank-sum test results further show that DoS significantly outperforms other methods on all test functions except C11 and C12. Although it does not exhibit a statistically significant advantage on C11 and C12, its Mean values are close to those of the best performing algorithms AOO and FFA, indicating a small performance gap and demonstrating good robustness of DoS on complex low dimensional functions. The Friedman test further shows that DoS ranks first in average ranking, confirming its superior overall optimization performance over the competitors.
            
            \begin{table*}[htbp!]
                \centering
                \caption{Experimental results of DoS and advanced competitors on 20D CEC2022 benchmark test functions.}
                \label{tab16}
                \begin{tabular}        {\cm{0.02\textwidth}\cm{0.06\textwidth}\cm{0.085\textwidth}\cm{0.085\textwidth}\cm{0.085\textwidth}\cm{0.085\textwidth}\cm{0.085\textwidth}\cm{0.085\textwidth}\cm{0.085\textwidth}\cm{0.085\textwidth}\cm{0.085\textwidth}}
                    \toprule
                    No. & Metrics & DoS & AOO & PSA & FFA & AVOA & ETO & SGA & WAA \\
                    \midrule
                    \multirow{4}{*}{C1} & Mean & \textbf{300} & \textbf{300} & \textbf{300} & 306.03495 & \textbf{300} & 11803.089 & 307.16622 & 300.39313 \\
                     & Std & \textbf{3.66E-14} & 8.449E-09 & 1.84E-12 & 6.8698472 & 1.855E-13 & 4703.1851 & 8.1940942 & 0.1019005 \\
                     & Kruskal & \textbf{15.5} & 105.5 & 68.983333 & 179.23333 & 52.016667 & 225.5 & 177.26667 & 140 \\
                     & WSRT &  & 1.3E-11(+) & 1.3E-11(+) & 1.3E-11(+) & 1.2E-11(+) & 1.3E-11(+) & 1.3E-11(+) & 1.3E-11(+) \\
                    \midrule
                    \multirow{4}{*}{C2} & Mean & \textbf{409.2096} & 441.76036 & 436.91201 & 453.40179 & 431.14145 & 530.20493 & 452.20248 & 451.34328 \\
                     & Std & 9.3737047 & 16.365079 & 22.07657 & \textbf{5.766707} & 24.286936 & 60.651785 & 22.946428 & 12.37996 \\
                     & Kruskal & \textbf{29.5} & 91.633333 & 83.75 & 171.46667 & 66.016667 & 223.93333 & 151.83333 & 145.86667 \\
                     & WSRT &  & 7.9E-09(+) & 1.1E-06(+) & 2.8E-11(+) & 2.5E-05(+) & 2.8E-11(+) & 7.9E-09(+) & 1.2E-10(+) \\
                    \midrule
                    \multirow{4}{*}{C3} & Mean & \textbf{600.0003} & 601.24916 & 600.07275 & 600.09485 & 612.29284 & 626.58495 & 650.12214 & 663.4934 \\
                     & Std & \textbf{0.00065} & 1.8702363 & 0.3068289 & 0.1426903 & 7.6594782 & 8.3308069 & 12.017137 & 8.5810271 \\
                     & Kruskal & \textbf{28.96667} & 103.13333 & 35.766667 & 75.266667 & 137.3 & 164.73333 & 198.56667 & 220.26667 \\
                     & WSRT &  & 2.5E-11(+) & 3.9E-01($\approx$) & 8.3E-11(+) & 2.5E-11(+) & 2.5E-11(+) & 2.5E-11(+) & 2.5E-11(+) \\
                    \midrule
                    \multirow{4}{*}{C4} & Mean & \textbf{817.9441} & 846.99617 & 868.48742 & 890.94447 & 888.23454 & 893.1711 & 886.87962 & 902.38775 \\
                     & Std & \textbf{4.891897} & 15.185806 & 18.540806 & 7.4258835 & 21.749882 & 21.052151 & 22.494013 & 19.847342 \\
                     & Kruskal & \textbf{16.03333} & 55.966667 & 96.4 & 161.56667 & 146.66667 & 161.03333 & 142.96667 & 183.36667 \\
                     & WSRT &  & 1.5E-10(+) & 3.0E-11(+) & 3.0E-11(+) & 3.0E-11(+) & 3.0E-11(+) & 3.0E-11(+) & 3.0E-11(+) \\
                    \midrule
                    \multirow{4}{*}{C5} & Mean & \textbf{900.509} & 901.10872 & 2061.4428 & 906.11336 & 2174.7726 & 2073.8973 & 2187.695 & 2677.0431 \\
                     & Std & \textbf{0.584541} & 3.4361638 & 750.80819 & 4.8491158 & 332.38818 & 479.77182 & 331.46565 & 361.48852 \\
                     & Kruskal & 33.566667 & \textbf{30} & 144.4 & 72.933333 & 161.2 & 151.26667 & 162.66667 & 207.96667 \\
                     & WSRT &  & 2.6E-01($\approx$) & 2.9E-11(+) & 1.3E-10(+) & 2.9E-11(+) & 2.9E-11(+) & 2.9E-11(+) & 2.9E-11(+) \\
                    \midrule
                    \multirow{4}{*}{C6} & Mean & \textbf{1877.902} & 7981.3414 & 4112.4339 & 30373.841 & 6556.6909 & 956476.69 & 8621.9567 & 7154.3353 \\
                     & Std & \textbf{49.59876} & 6589.327 & 2747.5297 & 64926.322 & 5325.4433 & 1256758.3 & 6967.6583 & 3372.3559 \\
                     & Kruskal & \textbf{18.23333} & 125.43333 & 84.066667 & 138.96667 & 111.86667 & 225.26667 & 125.6 & 134.56667 \\
                     & WSRT &  & 3.0E-11(+) & 1.0E-08(+) & 3.3E-11(+) & 6.7E-11(+) & 3.0E-11(+) & 9.0E-11(+) & 3.0E-11(+) \\
                    \midrule
                    \multirow{4}{*}{C7} & Mean & \textbf{2016.568} & 2032.7279 & 2067.6759 & 2032.7674 & 2083.7473 & 2112.3879 & 2111.8595 & 2258.9069 \\
                     & Std & \textbf{6.996966} & 10.666281 & 47.067405 & 7.6841236 & 41.035303 & 42.169606 & 37.101645 & 98.870378 \\
                     & Kruskal & \textbf{17.53333} & 67.1 & 116.73333 & 68.733333 & 138.4 & 167.06667 & 168.4 & 220.03333 \\
                     & WSRT &  & 2.4E-10(+) & 3.0E-11(+) & 2.2E-10(+) & 2.2E-10(+) & 3.0E-11(+) & 3.0E-11(+) & 3.0E-11(+) \\
                    \midrule
                    \multirow{4}{*}{C8} & Mean & \textbf{2219.708} & 2224.1792 & 2246.8719 & 2229.7121 & 2228.1167 & 2246.0707 & 2275.6722 & 2419.4856 \\
                     & Std & \textbf{2.672545} & 3.8690004 & 44.58114 & 3.5858018 & 7.9241616 & 32.321295 & 54.96744 & 150.3355 \\
                     & Kruskal & \textbf{16.8} & 80.7 & 103.76667 & 126.86667 & 100.3 & 151.16667 & 180.2 & 204.2 \\
                     & WSRT &  & 5.6E-10(+) & 7.4E-11(+) & 3.0E-11(+) & 3.0E-11(+) & 3.0E-11(+) & 3.0E-11(+) & 3.0E-11(+) \\
                    \midrule
                    \multirow{4}{*}{C9} & Mean & \textbf{2466.983} & 2480.7852 & 2480.7813 & 2481.2435 & 2480.7813 & 2504.1376 & 2480.8885 & 2480.7874 \\
                     & Std & 0.47377 & 0.0031774 & \textbf{7.83E-12} & 0.4450729 & 3.055E-09 & 18.595197 & 0.0656501 & 0.0031057 \\
                     & Kruskal & \textbf{15.5} & 116.7 & 45.5 & 184.3 & 75.5 & 225.36667 & 170.06667 & 131.06667 \\
                     & WSRT &  & 3.0E-11(+) & 3.0E-11(+) & 3.0E-11(+) & 3.0E-11(+) & 3.0E-11(+) & 3.0E-11(+) & 3.0E-11(+) \\
                    \midrule
                    \multirow{4}{*}{C10} & Mean & \textbf{2432.837} & 2587.7035 & 2653.6576 & 2500.5107 & 2664.7057 & 2895.1071 & 3104.8031 & 4836.6844 \\
                     & Std & 45.606809 & 196.87289 & 202.64586 & \textbf{0.087707} & 210.60401 & 576.96241 & 1085.0082 & 966.51639 \\
                     & Kruskal & \textbf{24.93333} & 72.766667 & 142.86667 & 74.066667 & 128.8 & 160.2 & 146.53333 & 213.83333 \\
                     & WSRT &  & 8.1E-05(+) & 9.9E-11(+) & 7.1E-09(+) & 4.2E-10(+) & 3.0E-11(+) & 4.1E-11(+) & 3.0E-11(+) \\
                    \midrule
                    \multirow{4}{*}{C11} & Mean & \textbf{2900} & 2920.0002 & \textbf{2900} & 2986.0277 & 2930 & 3811.5196 & 2950.103 & 2951.9492 \\
                     & Std & \textbf{3.04E-13} & 40.68374 & 2.508E-12 & 69.291934 & 46.60916 & 522.16053 & 77.538283 & 44.121338 \\
                     & Kruskal & \textbf{15.5} & 110.5 & 48.466667 & 167.43333 & 98.033333 & 225.26667 & 143.43333 & 155.36667 \\
                     & WSRT &  & 1.4E-11(+) & 1.3E-11(+) & 1.4E-11(+) & 1.4E-11(+) & 1.4E-11(+) & 1.4E-11(+) & 1.4E-11(+) \\
                    \midrule
                    \multirow{4}{*}{C12} & Mean & \textbf{2900.004} & 2944.5165 & 2969.2382 & 2940.0136 & 2972.7629 & 3031.0166 & 3005.7321 & 3648.542 \\
                     & Std & \textbf{0.000117} & 6.0104328 & 23.658208 & 4.0042601 & 23.791441 & 53.230748 & 48.757913 & 296.09399 \\
                     & Kruskal & \textbf{15.5} & 72.966667 & 126.16667 & 55.033333 & 131.46667 & 178.86667 & 158.86667 & 225.13333 \\
                     & WSRT &  & 3.0E-11(+) & 3.0E-11(+) & 3.0E-11(+) & 3.0E-11(+) & 3.0E-11(+) & 3.0E-11(+) & 3.0E-11(+) \\
                    \midrule
                    \multicolumn{2}{c}{FMR} & \textbf{1.172222} & 3.3694444 & 3.4736111 & 4.6027778 & 4.2541667 & 6.75 & 5.8277778 & 6.55 \\
                    \multicolumn{2}{c}{F-Rank} & \textbf{1} & 2 & 3 & 5 & 4 & 8 & 6 & 7 \\
                    \bottomrule
                \end{tabular}
                \vspace{-4pt}
                \begin{flushleft}
                     \footnotesize
                     \rmfamily
                     FMR and F-Rank are the statistical results over all CEC2022 benchmark test functions (20D).
                \end{flushleft}
            \end{table*}

            Table~\ref{tab16} presents the numerical results of DoS on the 20-dimensional CEC2022 benchmark test functions. The results show that DoS achieves the best values for both Mean and Std on 9 functions (F1, F3~F8, F11 and F12), indicating strong performance and stability in solving complex optimization problems. Although DoS is slightly less stable than FFA, AOO and PSA on F2, F9 and F10, its search mechanism yields the best Mean on 12 functions and tops the Kruskal-Wallis test on all functions except F5, highlighting its overall performance advantage.

            Notably, although DoS ranks slightly lower than AOO in the Kruskal-Wallis test on F5, it still outperforms all competitors in terms of Mean and Std on this function, suggesting its performance remains unaffected. According to WSRT results, DoS is significantly better than all competitors on 10 out of 12 functions, except F3 and F5, where its performance is comparable to PSA and AOO. It’s worth noting that DoS demonstrates stronger stability on these two functions and consistently approximates the global optimal solution along with the best performing competitors.

            The Friedman test further confirms that DoS ranks first in average performance, with a significant lead over the second-best algorithm, AOO, validating its comprehensive optimization capabilities.

        \subsection{Comparison with SOTA Algorithms}

            To further highlight DoS's performance, this section compares it with the three SOTA algorithms described in Section 4.1 on the 50-dimensional CEC2017 and 20-dimensional CEC2022 benchmark test functions. The relevant results are summarized in Tables~\ref{tab17}, ~\ref{tab18}, ~\ref{tab19} and~\ref{tab20}. Beyond conventional statistical metrics like Mean, Std, Kruskal and WSRT, the tables also report the best function value (Best) achieved by each algorithm within the standard number of trials, enabling a more comprehensive evaluation of their optimization performance.

            \begin{table*}[htbp!]
                \centering
                \caption{Experimental results of DoS and SOTA algorithms on 50D CEC2017 benchmark test functions (Unimodal and Multimodal).}
                \label{tab17}
                \begin{tabular}{\cm{0.04\textwidth}\cm{0.23\textwidth}\cm{0.11\textwidth}\cm{0.11\textwidth}\cm{0.11\textwidth}\cm{0.11\textwidth}\cm{0.11\textwidth}}
                    \toprule
                    No. & Metrics & Mean & Best & Std & Kruskal & WSRT \\
                    \midrule
                    \multirow{4}{*}{F1} & DoS & \textbf{100} & \textbf{100} & 8.266E-10 & 86.676471 & \\
                     & LSHADE & \textbf{100} & \textbf{100} & 1.591E-07 & 163.98039 & 2.4E-14(+) \\
                     & LSHADE-SPACMA & \textbf{100} & \textbf{100} & \textbf{2.57E-14} & \textbf{27.84314} & 4.4E-16(-) \\
                     & AL-SHADE & \textbf{100} & \textbf{100} & 1.164E-09 & 131.5 & 5.0E-09(+) \\
                    \midrule
                    \multirow{4}{*}{F3} & DoS & \textbf{300} & \textbf{300} & \textbf{2.36E-13} & \textbf{46.16667} & \\
                     & LSHADE & 40946.98 & \textbf{300} & 79720.112 & 153.36275 & 1.8E-17(+) \\
                     & LSHADE-SPACMA & 60631.926 & \textbf{300} & 95842.571 & 95.088235 & 1.8E-04(+) \\
                     & AL-SHADE & \textbf{300} & \textbf{300} & 4.175E-12 & 115.38235 & 2.4E-12(+) \\
                    \midrule
                    \multirow{4}{*}{F4} & DoS & \textbf{422.0392} & \textbf{400} & \textbf{35.92002} & \textbf{77.55882} & \\
                     & LSHADE & 463.70769 & 400.00212 & 46.551156 & 137.7549 & 1.3E-08(+) \\
                     & LSHADE-SPACMA & 437.48371 & \textbf{400} & 46.064835 & 89.529412 & 8.4E-01($\approx$) \\
                     & AL-SHADE & 434.38821 & \textbf{400} & 42.792302 & 105.15686 & 2.4E-03(+) \\
                    \midrule
                    \multirow{4}{*}{F5} & DoS & 616.22595 & 576.65356 & 20.868763 & 178.03922 & \\
                     & LSHADE & \textbf{558.3527} & 533.82927 & 11.07337 & \textbf{62.76471} & 4.7E-18(-) \\
                     & LSHADE-SPACMA & 565.86406 & 540.79397 & \textbf{10.94323} & 92.490196 & 1.5E-17(-) \\
                     & AL-SHADE & 561.68739 & \textbf{532.8336} & 13.281825 & 76.705882 & 8.9E-18(-) \\
                    \midrule
                    \multirow{4}{*}{F6} & DoS & \textbf{600.0709} & \textbf{600} & \textbf{0.256435} & \textbf{31.80392} & \\
                     & LSHADE & 600.80875 & 600.00991 & 0.7665073 & 106.03922 & 5.2E-15(+) \\
                     & LSHADE-SPACMA & 601.9224 & 600.39045 & 1.1694028 & 161.13725 & 2.4E-17(+) \\
                     & AL-SHADE & 600.84039 & 600.09291 & 0.6344647 & 111.01961 & 3.8E-15(+) \\
                    \midrule
                    \multirow{4}{*}{F7} & DoS & 881.06362 & 826.21983 & 24.848898 & 159.01961 & \\
                     & LSHADE & 849.79627 & 811.16742 & \textbf{21.96977} & 94.588235 & 2.3E-09(-) \\
                     & LSHADE-SPACMA & \textbf{836.3118} & \textbf{783.4292} & 22.297265 & \textbf{66.84314} & 1.8E-13(-) \\
                     & AL-SHADE & 847.31205 & 798.11222 & 24.275304 & 89.54902 & 2.8E-09(-) \\
                    \midrule
                    \multirow{4}{*}{F8} & DoS & 907.91801 & 873.65473 & 19.880242 & 175.58824 & \\
                     & LSHADE & \textbf{856.9283} & \textbf{837.8107} & 13.75822 & \textbf{58.70588} & 3.6E-17(-) \\
                     & LSHADE-SPACMA & 865.23972 & 840.79385 & \textbf{12.38462} & 91.176471 & 1.4E-16(-) \\
                     & AL-SHADE & 864.5357 & 844.77313 & 13.853883 & 84.529412 & 1.6E-16(-) \\
                    \midrule
                    \multirow{4}{*}{F9} & DoS & 1257.9707 & \textbf{900} & 367.32114 & 104.52941 & \\
                     & LSHADE & 1212.3163 & 924.17463 & 273.78712 & 105.58824 & 9.7E-01($\approx$) \\
                     & LSHADE-SPACMA & 1294.1224 & 956.75563 & 325.51319 & 124.7451 & 2.3E-01($\approx$) \\
                     & AL-SHADE & \textbf{1090.38} & 918.80654 & \textbf{164.7694} & \textbf{75.13725} & 5.2E-02($\approx$) \\
                    \midrule
                    \multirow{4}{*}{F10} & DoS & 4782.9366 & 3899.8102 & 386.18247 & 137.54902 & \\
                     & LSHADE & 4625.0048 & 3830.4219 & \textbf{327.6693} & 118.11765 & 4.3E-02(-) \\
                     & LSHADE-SPACMA & \textbf{4226.487} & \textbf{3569.296} & 333.14474 & \textbf{59.15686} & 2.5E-10(-) \\
                     & AL-SHADE & 4492.4279 & 3717.5407 & 388.03587 & 95.176471 & 3.2E-04(-) \\
                    \bottomrule
                \end{tabular}
            \end{table*}

            \begin{table*}[htbp!]
                            \centering
                            \caption{Experimental results of DoS and SOTA algorithms on 50D CEC2017 benchmark test functions (Hybrid).}
                            \label{tab18}
                            \begin{tabular}{\cm{0.04\textwidth}\cm{0.23\textwidth}\cm{0.11\textwidth}\cm{0.11\textwidth}\cm{0.11\textwidth}\cm{0.11\textwidth}\cm{0.11\textwidth}}
                                \toprule
                                No. & Metrics & Mean & Best & Std & Kruskal & WSRT \\
                                \midrule
                                \multirow{4}{*}{F11} & DoS & \textbf{1279.913} & \textbf{1167.419} & \textbf{47.39033} & \textbf{77.84314} & \\
                                 & LSHADE & 1335.352 & 1237.6025 & 70.316482 & 127.86275 & 2.4E-05(+) \\
                                 & LSHADE-SPACMA & 1317.3168 & 1216.7653 & 56.918422 & 112.60784 & 2.0E-03(+) \\
                                 & AL-SHADE & 1294.5099 & 1190.8966 & 59.899022 & 91.686275 & 2.7E-01($\approx$) \\
                                \midrule
                                \multirow{4}{*}{F12} & DoS & 6940.079 & \textbf{2276.173} & 3988.2658 & 95.784314 & \\
                                 & LSHADE & 14013.363 & 4292.871 & 10769.826 & 146.66667 & 8.7E-07(+) \\
                                 & LSHADE-SPACMA & \textbf{3675.482} & 2688.0149 & \textbf{515.1498} & \textbf{33.31373} & 5.4E-11(-) \\
                                 & AL-SHADE & 10385.516 & 3735.6708 & 5425.2378 & 134.23529 & 8.5E-05(+) \\
                                \midrule
                                \multirow{4}{*}{F13} & DoS & 3664.4024 & 1430.8619 & 3178.4674 & 100 & \\
                                 & LSHADE & 3325.9359 & 1610.5161 & 3474.2883 & 93.215686 & 7.5E-01($\approx$) \\
                                 & LSHADE-SPACMA & 4252.6011 & 2380.6939 & 1034.2212 & 155.56863 & 2.0E-05(+) \\
                                 & AL-SHADE & \textbf{2126.275} & \textbf{1399.92} & \textbf{607.4895} & \textbf{61.21569} & 2.0E-03(-) \\
                                \midrule
                    \multirow{4}{*}{F14} & DoS & \textbf{1705.921} & \textbf{1480.839} & 115.13288 & \textbf{97.07843} & \\
                     & LSHADE & 1720.2395 & 1548.4749 & \textbf{70.12658} & 111.27451 & 2.3E-01($\approx$) \\
                     & LSHADE-SPACMA & 1718.6654 & 1560.3522 & 95.8092 & 103.98039 & 5.9E-01($\approx$) \\
                     & AL-SHADE & 1706.6258 & 1557.2875 & 85.717375 & 97.666667 & 9.1E-01($\approx$) \\
                    \midrule
                    \multirow{4}{*}{F15} & DoS & \textbf{1835.009} & \textbf{1529.352} & 170.31982 & \textbf{79.82353} & \\
                     & LSHADE & 1913.0865 & 1622.0463 & 162.68382 & 102.31373 & 4.3E-02(+) \\
                     & LSHADE-SPACMA & 2059.2344 & 1759.3827 & 212.34541 & 141.92157 & 3.9E-07(+) \\
                     & AL-SHADE & 1857.6854 & 1595.5396 & \textbf{156.3706} & 85.941176 & 5.3E-01($\approx$) \\
                    \midrule
                    \multirow{4}{*}{F16} & DoS & 2547.2624 & 1988.9621 & 253.96026 & 113.88235 & \\
                     & LSHADE & 2509.1228 & 1950.1857 & \textbf{215.3516} & 105.84314 & 4.1E-01($\approx$) \\
                     & LSHADE-SPACMA & 2475.4606 & \textbf{1854.779} & 287.9017 & 95.960784 & 1.3E-01($\approx$) \\
                     & AL-SHADE & \textbf{2469.641} & 1859.0595 & 305.04425 & \textbf{94.31373} & 1.3E-01($\approx$) \\
                    \midrule
                    \multirow{4}{*}{F17} & DoS & 2418.7292 & 1989.9291 & \textbf{173.8556} & 111.92157 & \\
                     & LSHADE & 2443.6957 & 2063.042 & 167.98872 & 120.47059 & 4.5E-01($\approx$) \\
                     & LSHADE-SPACMA & \textbf{2339.293} & 2032.8425 & 175.44524 & \textbf{80.70588} & 8.4E-03(-) \\
                     & AL-SHADE & 2381.2666 & \textbf{1938.228} & 187.2952 & 96.901961 & 1.9E-01($\approx$) \\
                    \midrule
                    \multirow{4}{*}{F18} & DoS & 2889.9539 & 1915.6841 & 837.54343 & 122.76471 & \\
                     & LSHADE & 3246.1577 & 1904.0504 & 1233.8132 & 134.66667 & 2.1E-01($\approx$) \\
                     & LSHADE-SPACMA & \textbf{2086.042} & \textbf{1875.893} & \textbf{132.3506} & \textbf{50.07843} & 1.9E-10(-) \\
                     & AL-SHADE & 2723.3786 & 1902.0375 & 1007.4075 & 102.4902 & 7.1E-02($\approx$) \\
                    \midrule
                    \multirow{4}{*}{F19} & DoS & \textbf{2056.599} & \textbf{1929.094} & 60.367966 & \textbf{84.66667} & \\
                     & LSHADE & 2077.7849 & 1990.4833 & \textbf{57.37541} & 94.784314 & 4.0E-01($\approx$) \\
                     & LSHADE-SPACMA & 2160.8638 & 1988.825 & 148.15809 & 145.76471 & 1.9E-07(+) \\
                     & AL-SHADE & 2063.9512 & 1993.6955 & 43.008306 & 84.784314 & 9.8E-01($\approx$) \\
                    \midrule
                    \multirow{4}{*}{F20} & DoS & 2437.4399 & 2152.9213 & \textbf{131.8813} & 103.80392 & \\
                     & LSHADE & 2504.5134 & 2172.6279 & 155.73278 & 126.47059 & 5.2E-02($\approx$) \\
                     & LSHADE-SPACMA & 2421.9738 & 2090.7928 & 149.82474 & 98.862745 & 6.8E-01($\approx$) \\
                     & AL-SHADE & \textbf{2370.908} & \textbf{2037.509} & 187.92547 & \textbf{80.86275} & 4.9E-02(-) \\
                    \bottomrule
                \end{tabular}
            \end{table*}

            \begin{table*}[htbp!]
                \centering
                \caption{Experimental results of DoS and SOTA algorithms on 50D CEC2017 benchmark test functions (Composition).}
                \label{tab19}
                \begin{tabular}{\cm{0.04\textwidth}\cm{0.23\textwidth}\cm{0.11\textwidth}\cm{0.11\textwidth}\cm{0.11\textwidth}\cm{0.11\textwidth}\cm{0.11\textwidth}}
                    \toprule
                    No. & Metrics & Mean & Best & Std & Kruskal & WSRT \\
                    \midrule
                    \multirow{4}{*}{F21} & DoS & 2404.2514 & 2372.333 & 14.806696 & 177.70588 & \\
                     & LSHADE & \textbf{2355.131} & 2334.8268 & \textbf{11.24392} & \textbf{68.76471} & 5.3E-18(-) \\
                     & LSHADE-SPACMA & 2360.6729 & 2339.5417 & 11.606271 & 88.72549 & 1.0E-17(-) \\
                     & AL-SHADE & 2356.7652 & \textbf{2331.429} & 14.092693 & 74.803922 & 3.2E-17(-) \\
                    \midrule
                    \multirow{4}{*}{F22} & DoS & 6284.9812 & \textbf{2300} & 892.27191 & 109.5098 & \\
                     & LSHADE & 6459.7616 & 2301.4107 & \textbf{685.0383} & 124 & 1.9E-01($\approx$) \\
                     & LSHADE-SPACMA & \textbf{5376.747} & \textbf{2300} & 1760.7421 & \textbf{78.86275} & 7.3E-03(-) \\
                     & AL-SHADE & 6037.6519 & \textbf{2300} & 1309.1275 & 97.627451 & 3.1E-01($\approx$) \\
                    \midrule
                    \multirow{4}{*}{F23} & DoS & \textbf{2790.402} & \textbf{2746.397} & 21.305894 & \textbf{77.66667} & \\
                     & LSHADE & 2796.1118 & 2767.7534 & \textbf{17.03535} & 92.509804 & 1.3E-01($\approx$) \\
                     & LSHADE-SPACMA & 2814.1393 & 2772.3922 & 22.115573 & 139.33333 & 1.2E-06(+) \\
                     & AL-SHADE & 2799.3236 & 2763.9507 & 20.440585 & 100.4902 & 3.6E-02(+) \\
                    \midrule
                    \multirow{4}{*}{F24} & DoS & 3009.852 & 2956.0793 & 24.03879 & 159.31373 & \\
                     & LSHADE & \textbf{2963.059} & \textbf{2934.013} & \textbf{16.60232} & \textbf{59.17647} & 5.7E-15(-) \\
                     & LSHADE-SPACMA & 2984.5581 & 2947.5657 & 24.460412 & 109.56863 & 3.3E-07(-) \\
                     & AL-SHADE & 2972.8762 & 2934.116 & 24.034438 & 81.941176 & 1.0E-10(-) \\
                    \midrule
                    \multirow{4}{*}{F25} & DoS & \textbf{2968.833} & 2931.2689 & \textbf{33.59039} & \textbf{46.70588} & \\
                     & LSHADE & 3029.4396 & 2960.1618 & 44.04549 & 115.92157 & 3.2E-09(+) \\
                     & LSHADE-SPACMA & 3034.0004 & 2958.1133 & 43.409329 & 124.37255 & 7.1E-11(+) \\
                     & AL-SHADE & 3033.648 & \textbf{2928.82} & 42.165142 & 123 & 4.3E-11(+) \\
                    \midrule
                    \multirow{4}{*}{F26} & DoS & \textbf{4028.313} & \textbf{2900} & 383.51956 & \textbf{40.17647} & \\
                     & LSHADE & 4542.3483 & 4124.9167 & \textbf{205.398} & 102.52941 & 9.5E-12(+) \\
                     & LSHADE-SPACMA & 4727.5656 & 4216.3413 & 261.95951 & 135.98039 & 3.3E-13(+) \\
                     & AL-SHADE & 4691.8314 & 4158.2567 & 226.44455 & 131.31373 & 7.6E-13(+) \\
                    \midrule
                    \multirow{4}{*}{F27} & DoS & \textbf{3200.011} & \textbf{3200.01} & \textbf{0.00014} & \textbf{26} & \\
                     & LSHADE & 3354.315 & 3236.7039 & 70.682236 & 120.58824 & 3.3E-18(+) \\
                     & LSHADE-SPACMA & 3406.4369 & 3289.8923 & 75.060222 & 150.31373 & 3.3E-18(+) \\
                     & AL-SHADE & 3340.8421 & 3221.6297 & 72.071441 & 113.09804 & 3.3E-18(+) \\
                    \midrule
                    \multirow{4}{*}{F28} & DoS & 3299.2651 & 3279.8089 & \textbf{3.595295} & 121.23529 & \\
                     & LSHADE & 3293.2885 & 3258.8487 & 26.909047 & 103.62745 & 2.5E-02(-) \\
                     & LSHADE-SPACMA & \textbf{3287.072} & \textbf{3253.352} & 30.787893 & \textbf{87.48039} & 6.9E-04(-) \\
                     & AL-SHADE & 3292.1996 & \textbf{3253.352} & 32.123185 & 97.656863 & 4.5E-01($\approx$) \\
                    \midrule
                    \multirow{4}{*}{F29} & DoS & \textbf{3442.742} & \textbf{3217.06} & \textbf{127.5843} & \textbf{83.43137} & \\
                     & LSHADE & 3568.0981 & 3245.5777 & 184.68755 & 122.62745 & 6.3E-04(+) \\
                     & LSHADE-SPACMA & 3534.3358 & 3257.1179 & 161.01547 & 114.31373 & 5.6E-03(+) \\
                     & AL-SHADE & 3484.103 & 3223.4824 & 219.0791 & 89.627451 & 7.6E-01($\approx$) \\
                    \midrule
                    \multirow{4}{*}{F30} & DoS & \textbf{3545.922} & \textbf{3270.068} & \textbf{391.276} & \textbf{26} &  \\
                     & LSHADE & 743861.5 & 582496.75 & 135523.45 & 125.90196 & 3.3E-18(+) \\
                     & LSHADE-SPACMA & 775423.83 & 582420.14 & 140828.88 & 137.84314 & 3.3E-18(+)\\
                     & AL-SHADE & 719364.27 & 582412.59 & 105318.04 & 120.2549 & 3.3E-18(+)\\
                    \bottomrule
                \end{tabular}
                \vspace{-4pt}
                \begin{flushleft}
                    \footnotesize
                    \rmfamily
                    FMR: DoS (\textbf{2.416836}), LSHADE (2.6585531), LSHADE-SPACMA (2.4989858), AL-SHADE(2.4256254). \\
                    F-Rank: DoS (\textbf{1}), LSHADE (4), LSHADE-SPACMA (3), AL-SHADE(2). \\
                    FMR and F-Rank are the statistical results over all CEC2017 benchmark test functions (50D).
                \end{flushleft}
            \end{table*}

            The results in Tables~\ref{tab17}, ~\ref{tab18} and~\ref{tab19} show that DoS achieves the best values in Mean, Best, Std and Kruskal indicators simultaneously on 7 test functions: F3, F4, F6, F11, F27, F29 and F30. This fully demonstrates its comprehensive advantage in terms of performance, stability and statistical significance. It can be seen that even when compared with representative SOTA algorithms, DoS still maintains a leading position on multiple functions in the CEC2017 benchmark tests. This is especially true for functions like F27, F29 and F30, which feature complex problem structures and irregular search spaces, further confirming DoS’s strength in solving highly complex problems.

            Although DoS only achieves the best results in Mean, Std and Kruskal on F14, F15, F19, F23 and F26 and its Std is slightly worse than some SOTA algorithms, this also reflects the effectiveness of its dynamic selection mechanism. Compared to SOTA algorithms, DoS is better able to maintain a reasonable balance between exploration and exploitation in the later stages, thus achieving more thorough searches in the feasible domain. Although this may lead to slightly increased variability in solutions, the overall performance remains excellent.

            In terms of performance significance analysis, the results of the Wilcoxon rank-sum test show that DoS exhibits significant advantages on F3, F4, F6, F11, F15, F19, F23, F25, F26, F27, F29 and F30 and performs comparably to SOTA algorithms on F1, F9, F12, F13, F14, F16, F17, F18, F20 and F22. This fully demonstrates that DoS has clear performance advantages on most test functions or at least performs on par with SOTA algorithms. Although DoS performs significantly worse than SOTA algorithms on F5, F7, F8, F10, F21, F24 and F28, its performance on these functions remains within an acceptable range in terms of Mean and Best indicators, showing no obvious failure. The FMR obtained from the Friedman test further shows that DoS ranks first among all compared algorithms, verifying its potential and adaptability in optimization problems.

            \begin{table*}[htbp!]
                \centering
                \caption{Experimental results of DoS and SOTA algorithms on 20D CEC2022 benchmark test functions.}
                \label{tab20}
                \begin{tabular}{\cm{0.04\textwidth}\cm{0.23\textwidth}\cm{0.11\textwidth}\cm{0.11\textwidth}\cm{0.11\textwidth}\cm{0.11\textwidth}\cm{0.11\textwidth}}
                    \toprule
                    No. & Metrics & Mean & Best & Std & Kruskal & WSRT\\
                    \midrule
                    \multirow{4}{*}{C1} & DoS & \textbf{300} & \textbf{300} & 3.657E-14 & 54.3 & \\
                     & LSHADE & 1321.9344 & \textbf{300} & 4227.5481 & 84.25 & 3.1E-05(+) \\
                     & LSHADE-SPACMA & 3754.3535 & \textbf{300} & 6223.297 & 54.6 & 8.4E-01($\approx$) \\
                     & AL-SHADE & \textbf{300} & \textbf{300} & \textbf{3.17E-14} & \textbf{48.85} & 4.3E-01($\approx$) \\
                    \midrule
                    \multirow{4}{*}{C2} & DoS & \textbf{409.2096} & \textbf{400} & 9.3737047 & \textbf{18.25} & \\
                     & LSHADE & 448.80521 & 444.89547 & \textbf{1.062786} & 81.216667 & 4.7E-12(+) \\
                     & LSHADE-SPACMA & 443.89676 & \textbf{400} & 14.919996 & 72.216667 & 2.1E-08(+) \\
                     & AL-SHADE & 448.10704 & 444.89547 & 1.802041 & 70.316667 & 6.9E-12(+) \\
                    \midrule
                    \multirow{4}{*}{C3} & DoS & \textbf{600.0001} & \textbf{600} & \textbf{0.000143} & \textbf{29.08333} & \\
                     & LSHADE & 600.00088 & \textbf{600} & 0.0027695 & 69.95 & 1.0E-05(+) \\
                     & LSHADE-SPACMA & 600.01526 & \textbf{600} & 0.0536951 & 86.416667 & 2.8E-08(+) \\
                     & AL-SHADE & 600.00684 & \textbf{600} & 0.0343109 & 56.55 & 2.5E-05(+) \\
                    \midrule
                    \multirow{4}{*}{C4} & DoS & 817.94414 & 810.94481 & 4.8918965 & 94.466667 & \\
                     & LSHADE & 811.67419 & 806.96472 & \textbf{2.920516} & 54.8 & 3.3E-07(-) \\
                     & LSHADE-SPACMA & 812.10533 & \textbf{804.9748} & 4.571311 & 55.383333 & 1.7E-05(-) \\
                     & AL-SHADE & \textbf{810.2149} & \textbf{804.9748} & 4.0219236 & \textbf{37.35} & 1.7E-08(-) \\
                    \midrule
                    \multirow{4}{*}{C5} & DoS & \textbf{900.509} & \textbf{900} & \textbf{0.584541} & 59.633333 & \\
                     & LSHADE & 900.89887 & \textbf{900} & 1.0320978 & 74.533333 & 8.9E-02($\approx$) \\
                     & LSHADE-SPACMA & 900.63022 & \textbf{900} & 1.126228 & 58.066667 & 8.5E-01($\approx$) \\
                     & AL-SHADE & 900.78316 & \textbf{900} & 2.0991567 & \textbf{49.76667} & 2.6E-01($\approx$) \\
                    \midrule
                    \multirow{4}{*}{C6} & DoS & 1877.9024 & \textbf{1804.273} & 49.598764 & 71.833333 & \\
                     & LSHADE & 1860.7911 & 1806.505 & 38.895073 & 59.933333 & 2.0E-01($\approx$) \\
                     & LSHADE-SPACMA & 1854.8243 & 1819.314 & \textbf{30.68686} & 56.933333 & 7.0E-02($\approx$) \\
                     & AL-SHADE & \textbf{1853.532} & 1805.4535 & 38.973975 & \textbf{53.3} & 5.7E-02($\approx$) \\
                    \midrule
                    \multirow{4}{*}{C7} & DoS & \textbf{2016.568} & 2004.9575 & 6.9969661 & 59.433333 & \\
                     & LSHADE & 2017.7427 & 2004.6305 & \textbf{6.768248} & 63.2 & 5.4E-01($\approx$) \\
                     & LSHADE-SPACMA & 2017.9714 & \textbf{2001.357} & 7.6158463 & 65.266667 & 5.0E-01($\approx$) \\
                     & AL-SHADE & 2017.4489 & 2002.3021 & 7.1361371 & \textbf{54.1} & 4.1E-01($\approx$) \\
                    \midrule
                    \multirow{4}{*}{C8}
                     & DoS & \textbf{2219.708} & \textbf{2207.679} & 2.6725453 & 62.033333 & \\
                     & LSHADE & 2220.579 & 2218.0775 & 0.773127 & 64.666667 & 9.4E-01($\approx$) \\
                     & LSHADE-SPACMA & 2220.5105 & 2216.5918 & 0.9010526 & 68.2 & 3.9E-01($\approx$) \\
                     & AL-SHADE & 2220.213 & 2217.3343 & \textbf{0.756027} & \textbf{47.1} & 1.0E-01($\approx$) \\
                    \midrule
                    \multirow{4}{*}{C9} & DoS & \textbf{2466.967} & \textbf{2465.609} & 0.4475143 & \textbf{15.5} & \\
                     & LSHADE & 2480.7813 & 2480.7813 & 4.222E-13 & 73.5 & 5.1E-12(+) \\
                     & LSHADE-SPACMA & 2480.7813 & 2480.7813 & 3.16E-13 & 87 & 1.4E-11(+) \\
                     & AL-SHADE & 2480.7813 & 2480.7813 & \textbf{0} & 66 & 1.2E-12(+) \\
                    \midrule
                    \multirow{4}{*}{C10} & DoS & \textbf{2432.837} & 2400.0312 & 45.606809 & \textbf{31.73333} & \\
                     & LSHADE & 2494.3365 & \textbf{2400} & 38.508124 & 71.683333 & 7.7E-05(+) \\
                     & LSHADE-SPACMA & 2493.7898 & 2400.0312 & \textbf{25.26385} & 64.733333 & 3.2E-05(+) \\
                     & AL-SHADE & 2512.911 & \textbf{2400} & 46.224733 & 73.85 & 3.8E-06(+) \\
                    \midrule
                    \multirow{4}{*}{C11} & DoS & \textbf{2900} & \textbf{2900} & 3.045E-13 & 57 & \\
                     & LSHADE & \textbf{2900} & \textbf{2900} & 2.801E-13 & 53 & 6.1E-01($\approx$) \\
                     & LSHADE-SPACMA & \textbf{2900} & \textbf{2900} & \textbf{2.67E-13} & \textbf{51} & 4.3E-01($\approx$) \\
                     & AL-SHADE & \textbf{2900} & \textbf{2900} & 4.222E-13 & 81 & 1.5E-03(-) \\
                    \midrule
                    \multirow{4}{*}{C12} & DoS & \textbf{2900.004} & \textbf{2900.004} & \textbf{0.00012} & \textbf{15.5} & \\
                     & LSHADE & 2938.0431 & 2931.0617 & 4.291119 & 60.283333 & 3.0E-11(+)\\
                     & LSHADE-SPACMA & 2946.4365 & 2935.419 & 9.2818255 & 90.333333 & 3.0E-11(+)\\
                     & AL-SHADE & 2941.9602 & 2932.4577 & 7.0578608 & 75.883333 & 3.0E-11(+)\\
                    \bottomrule
                \end{tabular}
                \vspace{-4pt}
                \begin{flushleft}
                    \footnotesize
                    \rmfamily
                    FMR: DoS (\textbf{2.072222}), LSHADE (2.7125), LSHADE-SPACMA (2.7597222), AL-SHADE(2.4555556). \\
                    F-Rank: DoS (\textbf{1}), LSHADE (3), LSHADE-SPACMA (4), AL-SHADE(2). \\
                    FMR and F-Rank are the statistical results over all CEC2022 benchmark test functions (20D).
                \end{flushleft}
            \end{table*}

            According to Table~\ref{tab20}’s results, DoS exhibits a clear performance advantage in terms of both the Mean and Best indicators. For example, on 8 functions except for C4, C6, C7 and C10, DoS achieves the best values in both Mean and Best, indicating that even when facing the more structurally complex CEC2022 benchmark test functions, DoS still shows strong optimization capability and search efficiency compared to SOTA methods.

            However, due to the higher complexity of the CEC2022 benchmark test functions, the stability of DoS is somewhat affected. Except for C3, C5 and C12, DoS does not outperform the three SOTA algorithms in Std on the remaining 9 functions, suggesting relatively larger result variability on some functions.

            Based on the statistical results of the Wilcoxon rank-sum test, DoS performs significantly better than the SOTA algorithms on C1, C2, C3, C9, C10 and C12. On C5, C6, C7 and C8, DoS performs comparably to the three SOTA algorithms, indicating that DoS possesses strong or equivalent optimization ability on most problems in the CEC2022 benchmark test functions. Although DoS shows slightly inferior performance on C4 and C11 compared to SOTA algorithms, its Mean and Best values are close to the best and the overall impact on performance is minimal, remaining within an acceptable range. Finally, the average ranking obtained from the Friedman test shows that DoS once again ranks first among all algorithms, with a greater statistically significant gap compared to the second ranked AL-SHADE, further verifying its strong competitiveness and robustness in solving complex optimization problems.

        \subsection{Exploration and Exploitation Analysis}

\begin{figure*}[htbp!]
                \centering
                \begin{tabular}{\cm{0.31\textwidth}\cm{0.31\textwidth}\cm{0.31\textwidth}}
                    \multicolumn{3}{c}{\includegraphics[width=0.4\linewidth]{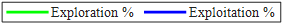}} \\
                    \includegraphics[width=0.95\linewidth]{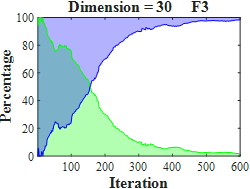} & \includegraphics[width=0.95\linewidth]{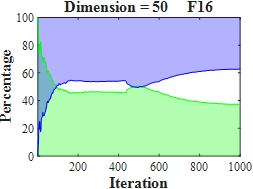} & \includegraphics[width=0.95\linewidth]{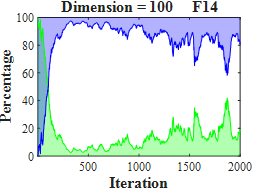} \\
                    \includegraphics[width=0.95\linewidth]{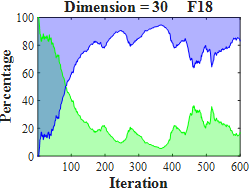} & \includegraphics[width=0.95\linewidth]{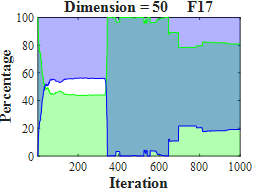} & \includegraphics[width=0.95\linewidth]{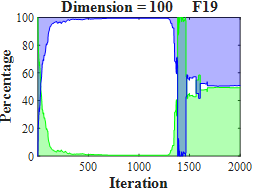} \\
                    \includegraphics[width=0.95\linewidth]{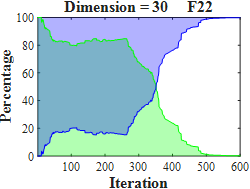} & \includegraphics[width=0.95\linewidth]{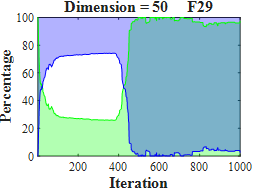} & \includegraphics[width=0.95\linewidth]{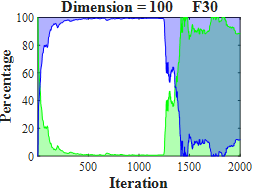} \\
                    \includegraphics[width=0.95\linewidth]{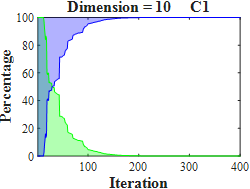} & \includegraphics[width=0.95\linewidth]{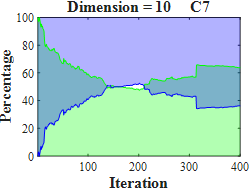} & \includegraphics[width=0.95\linewidth]{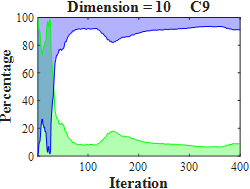} \\
                    \includegraphics[width=0.95\linewidth]{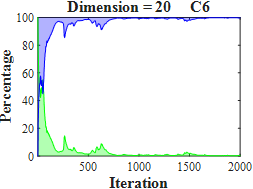} & \includegraphics[width=0.95\linewidth]{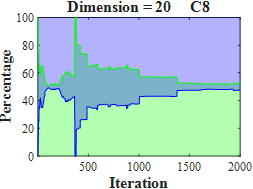} & \includegraphics[width=0.95\linewidth]{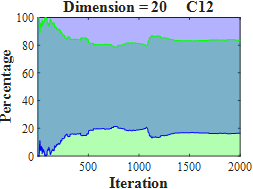}
                \end{tabular}
                \caption{Convergence curves of objective function values.}
                \label{fig7}
            \end{figure*}

            To qualitatively analyze the dynamic characteristics of exploration and exploitation during the search process of DoS, this section adopts the dimension-wise diversity measurement method \citep{hussain2019exploration} to analyze the exploration rate and exploitation rate of DoS. The number of evaluations is set to one-tenth of the standard evaluation count for the experiment and the resulting variation curves of the exploration rate and exploitation rate are shown in Fig.~\ref{fig7}.

            As shown in Fig.~\ref{fig7}, for F18 in 30-dimensions and F14 in 100-dimensions, the exploration rate and exploitation rate curves of DoS exhibit a trend of focusing on exploration in the early stages and on exploitation in the later stages, with frequent fluctuations. This indicates that even in the later stages of iteration, DoS can still maintain global exploration in order to pursue the globally optimal solution.

            According to existing research, many metaheuristic algorithms increase the execution probability of exploitation strategies in the later stages in order to ensure good convergence. Therefore, the exploration and exploitation rate curves of these algorithms usually show that the exploration rate is higher than the exploitation rate in the early stage, while the exploration rate approaches zero in the later stage. However, as shown in Fig.~\ref{fig7}, DoS exhibits a different trend in F17 and F29 (50-dimensions), F30 (100-dimensions), C7 (10-dimensions) and C8 and C12 (20-dimensions), where the exploration rate remains higher than the exploitation rate even in the later stages of iteration and the dynamic changes in exploration and exploitation rates are very significant. Although the search behavior on these functions differs greatly from that of most metaheuristic algorithms, the experimental results in Tables~\ref{tab6}, ~\ref{tab7}, ~\ref{tab8}, ~\ref{tab9}, ~\ref{tab10}, ~\ref{tab11}, ~\ref{tab12}, ~\ref{tab13}, ~\ref{tab14}, ~\ref{tab15} and~\ref{tab16} show that DoS performs significantly better than advanced competitors on these functions, further confirming the performance advantage brought by the unique structural design of DoS.

            In summary, DoS exhibits different optimization behaviors on different problems. Some of these behaviors diverge differ significantly from traditional patterns, further highlighting DoS’s adaptability to diverse scenarios.
            
        \subsection{Convergence Analysis}

\begin{figure*}[htbp!]
                \centering
                \begin{tabular}{\cm{0.3\textwidth}\cm{0.3\textwidth}\cm{0.3\textwidth}}
                    \multicolumn{3}{c}{\includegraphics[width=0.8\linewidth]{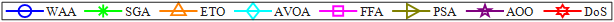}} \\
                    \includegraphics[width=0.95\linewidth]{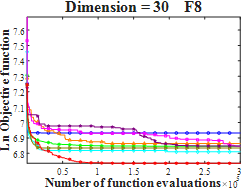} & \includegraphics[width=0.95\linewidth]{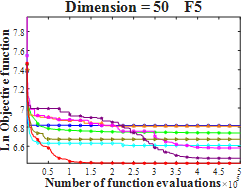} & \includegraphics[width=0.95\linewidth]{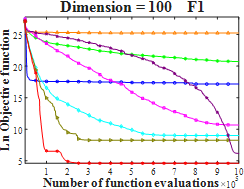} \\
                    \includegraphics[width=0.95\linewidth]{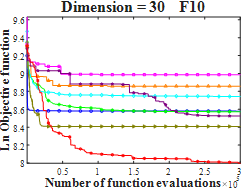} & \includegraphics[width=0.95\linewidth]{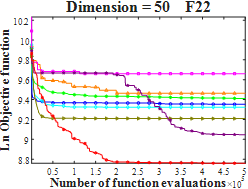} & \includegraphics[width=0.95\linewidth]{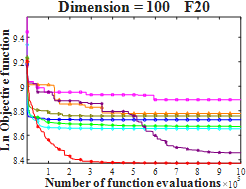} \\
                    \includegraphics[width=0.95\linewidth]{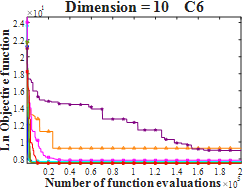} & \includegraphics[width=0.95\linewidth]{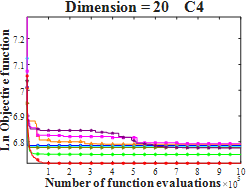} & \includegraphics[width=0.95\linewidth]{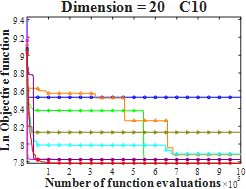}
                \end{tabular}
                \caption{Convergence curves of objective function values.}
                \label{fig8}
            \end{figure*}

            Convergence is also one of the key indicators for evaluating the performance of metaheuristic algorithms. To comprehensively assess the convergence ability of DoS, this paper conducts a visual analysis of the function value trends of DoS and 7 advanced competitors on CEC2017 and CEC2022 benchmark test functions with different dimensions. The results are shown in Fig.~\ref{fig8}.

            As shown in Fig.~\ref{fig8}, DoS is able to identify high quality exploitation regions earlier than its competitors on multiple test functions, achieving more efficient convergence. Although AOO reaches final function values close to those of DoS on certain test functions, its search speed is significantly slower and often relies on occasional late-stage exploration strategies to escape local optima. Overall, its convergence performance is still inferior to DoS.

            On functions such as F10 (30-dimensions), F22 (50-dimensions), F20 (100-dimensions) and C4 (20-dimensions), DoS avoids the problem of falling into local optima in the early stages compared to competitors such as WAA, SGA, ETO, AVOA, FFA and PSA. This demonstrates DoS's unique ability to comprehensively explore the feasible region, resulting in superior optimization performance. Although AOO also shows a second round of convergence on these functions, its final function values remain significantly worse than those of DoS.

            In summary, when solving simple optimization problems, DoS can quickly locate the optimal solution domain through a well balanced exploration and exploitation mechanism. When facing complex optimization problems, it can effectively avoid falling into local optima. This feature not only enables the algorithm to perform broad exploration but also allows it to continue exploitation operations. In terms of both convergence speed and convergence quality, DoS demonstrates significant advantages.

        \subsection{Scalability Analysis}

\begin{figure*}[htbp!]
                \centering
                \begin{tabular}{\cm{0.3\textwidth}\cm{0.3\textwidth}\cm{0.3\textwidth}}
                    \multicolumn{3}{c}{\includegraphics[width=0.8\linewidth]{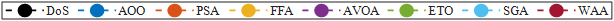}} \\
                    \includegraphics[width=0.95\linewidth]{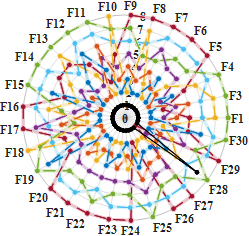} & \includegraphics[width=0.95\linewidth]{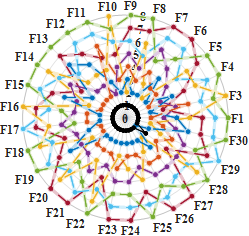} & \includegraphics[width=0.95\linewidth]{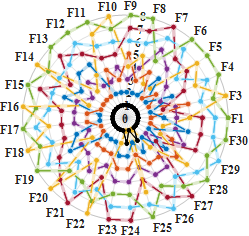}
                \end{tabular}
                \caption{Radar chart of Mean rankings on CEC2017 benchmark test functions.}
                \label{fig9}
            \end{figure*}

\begin{figure*}[htbp!]
                \centering
                \includegraphics[width=0.9\linewidth]{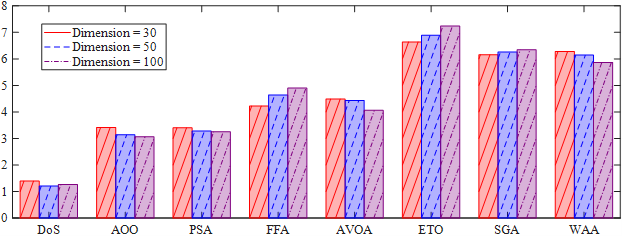}
                \caption{Bar chart of Friedman rankings on CEC2017 benchmark test functions.}
                \label{fig10}
            \end{figure*}

\begin{figure*}[htbp!]
                \centering
                \begin{tabular}{\cm{0.35\textwidth}\cm{0.35\textwidth}}
                    \multicolumn{2}{c}{\includegraphics[width=0.8\linewidth]{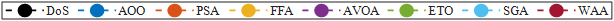}} \\
                    \includegraphics[width=0.95\linewidth]{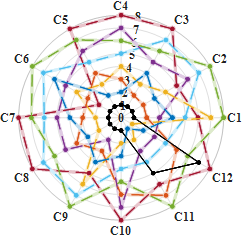} & \includegraphics[width=0.95\linewidth]{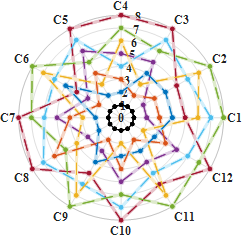}
                \end{tabular}
                \caption{Radar chart of Mean rankings on CEC2022 benchmark test functions.}
                \label{fig11}
            \end{figure*}

\begin{figure*}[htbp!]
                \centering
                \includegraphics[width=0.9\linewidth]{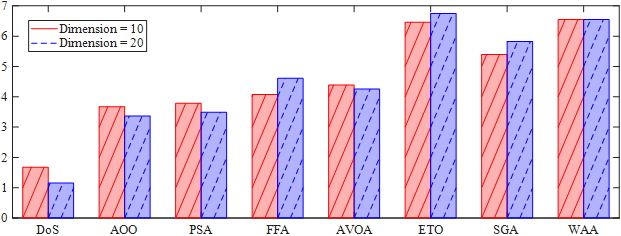}
                \caption{Bar chart of Friedman rankings on CEC2022 benchmark test functions.}
                \label{fig12}
            \end{figure*}

            Dimensionality is one of the key factors affecting the complexity of optimization problems and has a direct impact on algorithm performance. Therefore, this section analyzes the scalability of DoS based on the Mean rankings and Friedman rankings of CEC2017 and CEC2022 benchmark test functions under different dimensions, as presented in Section 4.3. The visualizations of the Mean rankings and Friedman rankings of DoS compared with other competitors are shown in Fig.~\ref{fig9}, ~\ref{fig10}, ~\ref{fig11} and~\ref{fig12}.

            Fig.~\ref{fig9} shows the Mean ranking of DoS and its competitors on the 30D, 50D and 100D CEC2017 benchmark test functions. According to the results, except for slightly inferior performance on F28 in 30D, DoS performs excellently on all other functions. As the dimensionality increases, its performance on F28 gradually improves ranking second in 50D (slightly lower than AOO) and rising to first place in 100D. This indicates that the extra evaluations from higher dimensionality give DoS more exploration time, thus improving its performance on this function.

            For functions F24 and F26, DoS ranks first in terms of the Mean metric in both 30D and 50D, but is surpassed by FFA in 100D, dropping to second place. Nevertheless, in terms of numerical values, the Mean difference between DoS and FFA is small and both are significantly better than the other competitors.

            Fig.~\ref{fig10} presents the Friedman average rankings of DoS and its competitors on the CEC2017 test functions. It can be observed that the Friedman rankings of DoS remain relatively stable across the three dimensions, maintaining first place with a significant lead, which demonstrates good scalability. Meanwhile, AOO, PSA, AVOA and WAA show slight improvements in their Friedman rankings as the dimensionality increases, also indicating relatively strong scalability. In contrast, the average rankings of FFA, ETO and SGA decline noticeably, especially for FFA and ETO, whose performance drops significantly, suggesting weaker adaptability to higher dimensions and poorer scalability.

            Fig.~\ref{fig11} illustrates the radar chart of the Mean rankings of DoS and advanced competitors on the 10D and 20D CEC2022 benchmark test functions. The results show that DoS consistently ranks first on ten out of the twelve functions, except for C11 and C12. In 10D, DoS performs relatively poorly on C11 and C12, but in 20D, its performance improves significantly, surpassing all competitors. This discrepancy is mainly due to the fewer evaluation times in the 10D tests, which limit the full effectiveness of DoS’s search mechanisms on these two functions.

            Fig.~\ref{fig12} shows the bar chart of the Friedman average rankings of each algorithm on the CEC2022 benchmark test functions. Combining Fig.~\ref{fig11} and~\ref{fig12}, it can be seen that DoS maintains strong overall performance in both 10D and 20D settings and its Friedman ranking continuously improves as the dimensionality increases, reflecting excellent scalability. Similarly, AOO, PSA and AVOA also exhibit a certain degree of scalability, as their rankings improve with increasing dimensions. In contrast, WAA performs relatively poorly in both dimensions, with little change in ranking, indicating limited scalability. FFA, ETO and SGA show a declining trend in their Friedman rankings across both dimensions, suggesting poor scalability on this benchmark.

            In summary, DoS maintains stable and improving performance on the CEC2022 benchmark test functions as dimensionality increases. The continuous improvement in its Friedman ranking further confirms its strong scalability. Although AOO, PSA and AVOA do not outperform DoS in overall performance, their improvements with increasing dimensionality indicate they are also suitable for higher-dimensional optimization tasks. By comparison, FFA, ETO and SGA show decreasing performance as dimensionality increases, highlighting their insufficient scalability.

        \subsection{Behavior Analysis}

\begin{figure*}[htbp!]
                \centering
                \begin{tabular}{\cm{0.23\textwidth}\cm{0.23\textwidth}\cm{0.23\textwidth}\cm{0.23\textwidth}}
                    \includegraphics[width=0.95\linewidth]{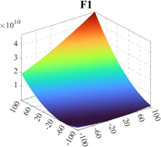} & \includegraphics[width=0.95\linewidth]{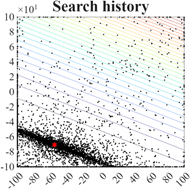} & \includegraphics[width=0.95\linewidth]{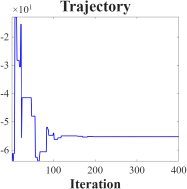} & \includegraphics[width=0.95\linewidth]{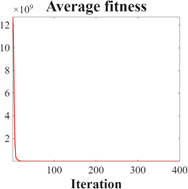} \\
                    \includegraphics[width=0.95\linewidth]{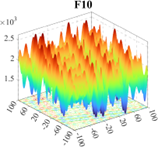} & \includegraphics[width=0.95\linewidth]{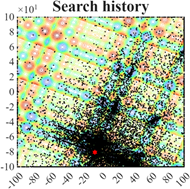} & \includegraphics[width=0.95\linewidth]{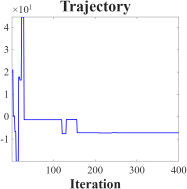} & \includegraphics[width=0.95\linewidth]{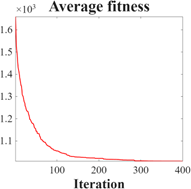} \\
                    \includegraphics[width=0.95\linewidth]{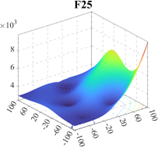} & \includegraphics[width=0.95\linewidth]{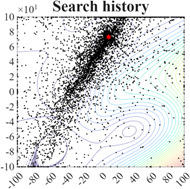} & \includegraphics[width=0.95\linewidth]{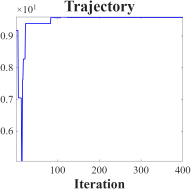} & \includegraphics[width=0.95\linewidth]{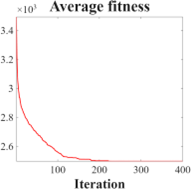} \\
                    \includegraphics[width=0.95\linewidth]{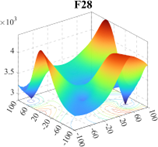} & \includegraphics[width=0.95\linewidth]{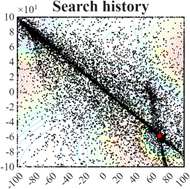} & \includegraphics[width=0.95\linewidth]{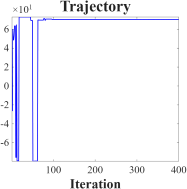} & \includegraphics[width=0.95\linewidth]{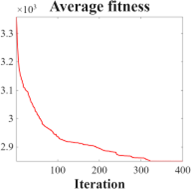} \\
                    \includegraphics[width=0.95\linewidth]{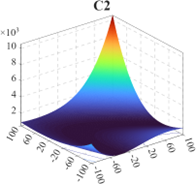} & \includegraphics[width=0.95\linewidth]{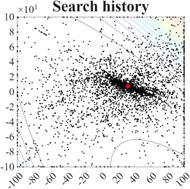} & \includegraphics[width=0.95\linewidth]{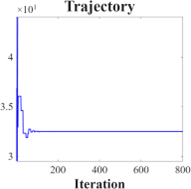} & \includegraphics[width=0.95\linewidth]{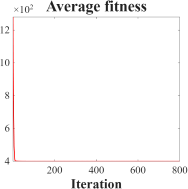} \\
                    \includegraphics[width=0.95\linewidth]{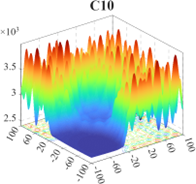} & \includegraphics[width=0.95\linewidth]{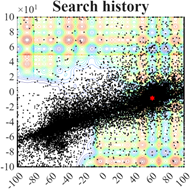} & \includegraphics[width=0.95\linewidth]{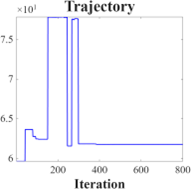} & \includegraphics[width=0.95\linewidth]{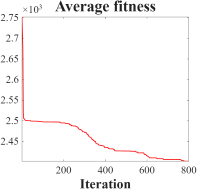}
                \end{tabular}
                \caption{Search History, Trajectory and Average Fitness of DoS on Some CEC2017 Benchmark Test Functions (2D).}
                \label{fig13}
            \end{figure*}

            Behavior analysis is an important method for gaining an in depth understanding of the search process of metaheuristic algorithms, as it intuitively reveals the evolutionary characteristics of algorithm strategies during iterations. To further analyze the search mechanism of DoS, this paper conducts a visual analysis of selected 2-dimensional CEC2017 and CEC2022 benchmark test functions in Fig.~\ref{fig13}. The analysis includes: historical search trajectories, the variation curves of the centroids of the formations on the first dimension coordinate and the evolutionary trends of the expected function values.

            As shown in the search trajectories in Fig. 13, combined with the 3D structural plots of each function, DoS successfully converges to the global optimal solution (indicated by the red hexagram) on all 6 test functions. It forms dense exploitation paths near the best solutions, particularly evident in functions like F10 and C10, where local optima are densely distributed. For functions with relatively easy to locate global optimal solution, such as F1 and C2, DoS exhibits a faster convergence rate. From the variation trends of the centroid on the first dimension, it can be observed that DoS typically locks onto the global optimal solution within the first 1/16 of the total iterations and then rapidly transitions into a deep exploitation phase under the guidance of the dynamic selection mechanism. Even in functions such as F25 and F28, which are intentionally designed with many local optima traps, DoS is able to accurately locate the global optimal solution through effective early-stage exploration and simultaneously achieve efficient exploitation.

            Analysis of the search trajectories for F1, F25 and F28 reveals that the DoS search paths exhibit distinct linear characteristics on these functions. The center-of-mass curve for the first-dimensional coordinate fluctuates significantly during iterations, reflecting how DoS enhances the accuracy of the global optimal solution by dynamically switching between different guide solutions using offensive strategies during multiple approaches to the global optimal solution. Particularly in F28, DoS formed two primary search trajectories, ultimately converging at the intersection point to develop the solution domain for the global optimal solution. This phenomenon demonstrates that under balanced control of exploration and exploitation strategies, DoS can switch between multiple high-potential regions, thereby breaking through local optima constraints.

            Furthermore, the jamming avoidance strategy proposed in Section 3.2.5 is especially evident in function C2. Since the global optimal solution of this function lies near the boundary of the feasible region and is connected to the upper and lower bound vertices, DoS utilizes the jamming avoidance strategy to effectively search the boundary region. As shown in the figure, DoS adaptively guides the search from the boundary toward the interior and finally converges to the global optimal solution.

            In conclusion, the behavioral analysis on CEC2017 and CEC2022 benchmark test functions demonstrates that DoS exhibits strong exploratory and exploitative capabilities. With the exploration and exploitation balance achieved by the dynamic selection mechanism, DoS not only escapes local optima effectively but also locates the global optimal solution in a short time and efficiently exploits the surrounding region, thereby showcasing excellent optimization performance and search efficiency.

        \subsection{Computational Time Cost Analysis}

            In the evaluation of metaheuristic algorithms, performance and computational time cost are two critical metrics. Although most SOTA algorithms demonstrate excellent optimization performance, they are often accompanied by high computational overhead due to the increased complexity of their internal structural designs. Therefore, assessing the computational efficiency of DoS is of significant importance for evaluating its practical applicability.

            This section evaluates the computational time cost of DoS based on the 30-dimensional CEC2017 benchmark test functions. To ensure fairness, the population size of all algorithms is uniformly set to 50. Four influencing factors are introduced in the experiment:

            \textbf{$T_{0}$}: the time required to execute Equation (30) one million times under the initial condition $x$ = 0.55;
            
            \textbf{$T_{1}$}: the time consumed to perform 200,000 function evaluations of F18 with a fixed random solution;
            
            \textbf{$T_{2}$}: the total time required to run the complete algorithm for 200,000 evaluations of F18 with a fixed initial solution;
            
            \textbf{$T_{2}^{mean}$}: the average value of T2 obtained from five independent experiments.
            \begin{equation}
                \begin{aligned}
                    &x = x + x; \quad x = \frac{x}{2}; \quad x = x \times x; \quad x = \sqrt{x}; \\
                    &x = \ln(x); \quad x = e^{x}; \quad x = \frac{x}{x + 2}
                \end{aligned}
                \label{eq29}
            \end{equation}

            \begin{table*}[htbp!]
                \centering
                \caption{Results of computational time cost.}
                \label{tab21}
                \begin{tabular}{\cm{0.12\textwidth}\cm{0.07\textwidth}\cm{0.07\textwidth}\cm{0.07\textwidth}\cm{0.07\textwidth}\cm{0.07\textwidth}\cm{0.07\textwidth}\cm{0.07\textwidth}\cm{0.07\textwidth}\cm{0.07\textwidth}}
                    \toprule
                     & DoS & AOO & PSA & FFA & AVOA & ETO & SGA & WAA \\
                    \midrule
                    $T_{0}$ & 0.0073590 & 0.0075530 & 0.0076597 & 0.0075053 & 0.0068882 & 0.0074479 & 0.0071382 & 0.0070808 \\
                    $T_{1}$ & 0.3870446 & 0.4021236 & 0.3944610 & 0.4043148 & 0.3995194 & 0.387512 & 0.3841997 & 0.3897004 \\
                    $T_{2}^{mean}$ & 1.1057091 & 1.3215604 & 0.8729215 & 1.5493374 & 0.9475669 & 1.9344475 & 0.6012483 & 4.7229055 \\
                    $(T_{2}^{mean}-T_{1})/T_{0}$ & 97.657899 & 121.73134 & 62.464655 & 152.56187 & 79.563241 & 207.70089 & 30.406635 & 611.96547 \\
                    \bottomrule
                \end{tabular}
            \end{table*}

            Table~\ref{tab21} summarizes the computational time costs of DoS and the seven advanced competing algorithms. The metric $(T_{2}^{mean}-T_{1})/T_{0}$ eflects the computational time cost variable of the algorithm, the smaller the value, the lower the computational time cost. It can be observed that the computational cost of DoS is lower than that of AOO, FFA, ETO and WAA, indicating its certain advantage in balancing performance and efficiency. Combined with the structural analysis in Section 3, DoS introduces multiple search strategies and a dynamic selection mechanism, making its overall design slightly more complex compared to PSA (which adopts a single strategy), as well as the structurally simpler AVOA and SGA. As a result, its computational cost is slightly higher than these three algorithms. However, as demonstrated in Sections 4.2 and 4.3, this added complexity brings significant improvements in optimization performance, leading DoS to outperform PSA, AVOA and SGA across benchmark test functions.

            In conclusion, DoS maintains high performance while keeping computational time costs within a reasonable range, achieving a sound balance between efficiency and optimization capability and demonstrating strong practical value.

    \section{Engineering Problems}

        The original intention behind the design of metaheuristic algorithms is to address the efficiency bottlenecks encountered by traditional methods when solving highly nonlinear, non-differentiable problems with multiple local optima. Compared with traditional optimization methods, metaheuristic algorithms demonstrate stronger adaptability and problem-solving capabilities for such complex problems. However, real-world engineering optimization problems are not only influenced by dimensionality ($D$), but also involve complex inequality constraints ($g$) and equality constraints ($h$), further increasing the difficulty of solution and posing challenges to algorithm generalization.

        To verify the performance of DoS on constrained optimization problems, this paper selects 10 real-world engineering optimization problems from CEC2020, covering four major categories: Industrial Chemical Processes, Process Synthesis and Design, Mechanical Engineering and Power Systems. Detailed information on the specific problems is provided in Table~\ref{tab22}.

        \begin{table*}[htbp!]
            \centering
            \caption{Basic information of constrained optimization problems.}
            \label{tab22}
            \begin{tabular}{\cm{0.05\textwidth}\cm{0.25\textwidth}\cm{0.38\textwidth}\cm{0.05\textwidth}\cm{0.05\textwidth}\cm{0.05\textwidth}}
                \toprule
                No. & Type & Name & $D$ & $g$ & $h$ \\
                \midrule
                R1 & Industrial Chemical Processes & Optimal operation of alkylation unit & 7 & 14 & 0 \\
                R2 & Process Synthesis and Design & Process flow sheeting & 3 & 3 & 0 \\
                R3 & Mechanical Engineering & Welded beam design & 4 & 7 & 0 \\
                R4 & Mechanical Engineering & Pressure vessel design & 4 & 4 & 0 \\
                R5 & Mechanical Engineering & Side impact design of automobiles & 11 & 10 & 0 \\
                R6 & Mechanical Engineering & Optimal design of industrial refrigeration system & 14 & 15 & 0 \\
                R7 & Mechanical Engineering & Step-cone pulley & 5 & 8 & 3 \\
                R8 & Mechanical Engineering & Hydro-static thrust bearing design & 4 & 7 & 0 \\
                R9 & Mechanical Engineering & Four-stage gear box & 22 & 86 & 0 \\
                R10 & Power System & Wind farm layout & 30 & 91 & 0 \\
                \bottomrule
            \end{tabular}
        \end{table*}

        For the above 10 problems, this paper adopts the standard experimental settings for CEC2020 engineering problems proposed by Kumar et al. \citep{kumar2020test} and performs 25 independent runs. To enhance the distinguishability of performance differences among algorithms, This paper halves the standard evaluation count specified in the original text, with the specific setting for the maximum evaluation count shown in Equation~\ref{eq30}.
        \begin{equation}
            E =
            \begin{cases}
                5 \times 10^{4}, & D \leq 10 \\
                1 \times 10^{5}, & 10 < D \leq 30 \\
                2 \times 10^{5}, & 30 < D \leq 50 \\
                4 \times 10^{5}, & 50 < D \leq 150 \\
                5 \times 10^{5}, & D > 150
            \end{cases}
            \label{eq30}
        \end{equation}

        The statistical content of the experimental results includes: the best value (Best), mean value (Mean), standard deviation (Std), feasible solution rate (Success), Wilcoxon rank-sum test results (WSRT) at a 5\% significance level, Friedman ranking (FMR) and overall ranking (F-Rank). All results in Tables~\ref{tab23}, ~\ref{tab24}, ~\ref{tab25}, ~\ref{tab26}, ~\ref{tab27}, ~\ref{tab28}, ~\ref{tab29}, ~\ref{tab30}, ~\ref{tab31}, ~\ref{tab32}, ~\ref{tab33}, ~\ref{tab34}, ~\ref{tab35}, ~\ref{tab36}, ~\ref{tab37}, ~\ref{tab38}, ~\ref{tab39}, ~\ref{tab40}, ~\ref{tab41} and~\ref{tab42} represent valid results satisfying the constraints; solutions failing to meet constraints are denoted as NaN. The best values for each indicator are shown in bold.

        \subsection{Optimal Operation of Alkylation Unit}

            The Optimal Operation of Alkylation Unit problem aims to maximize the economic benefit or product yield of the alkylation reaction system while satisfying 14 inequality constraints. Its decision variables involve several key process parameters, such as feed flow rate, reaction temperature and reflux ratio of the distillation column, denoted as $\bm{\mathrm{x}} = (x_{1}, x_{2}, \cdots , x_{7})$. The mathematical model of this problem is formulated as follows:

            \textbf{Minimize:}
            \begin{equation}
                \begin{aligned}
                    f(\bm{\mathrm{x}}) = \ &0.035 x_{1} \cdot x_{6} + 1.715 x_{1} + 10.0 x_{2} + \\
                    &4.0565 x_{3} - 0.063 x_{3} \cdot x_{5}
                \end{aligned}
                \label{eq31}
            \end{equation}

            \textbf{Subject to:}
            \begin{equation*}
                \begin{aligned}
                    g_{1}(\bm{\mathrm{x}}) = \ &0.0059553571 x_{6}^{2} \cdot x_{1} + 0.88392857 x_{3} - \\
                    &0.1175625 x_{6} \cdot x_{1} - x_{1} \leq 0
                \end{aligned}
            \end{equation*}
            \begin{equation*}
                \begin{aligned}
                    g_{2}(\bm{\mathrm{x}}) = \ &1.1088 x_{1} + 0.1303533 x_{1} \cdot x_{6} - \\
                    &0.0066033 x_{1} \cdot x_{6}^{2} - x_{3} \leq 0
                \end{aligned}
            \end{equation*}
            \begin{equation*}
                \begin{aligned}
                    g_{3}(\bm{\mathrm{x}}) = \ &6.66173269 x_{6}^{2} - 56.596669 x_{4} + \\
                    &172.39878 x_{5} - 191.20592 x_{6} - \\
                    &10000 \leq 0
                \end{aligned}
            \end{equation*}
            \begin{equation*}
                \begin{aligned}
                    g_{4}(\bm{\mathrm{x}}) = \ &1.08702 x_{6} - 0.03762 x_{6}^{2} + \\
                    &0.32175 x_{4} + 56.85075 - x_{5} \leq 0
                \end{aligned}
            \end{equation*}
            \begin{equation*}
                \begin{aligned}
                    g_{5}(\bm{\mathrm{x}}) = \ &0.006198 x_{7} \cdot x_{4} \cdot x_{3} + 2462.3121 x_{2} - \\
                    &25.125634 x_{2} \cdot x_{4} - x_{3} \cdot x_{4} \leq 0
                \end{aligned}
            \end{equation*}
            \begin{equation*}
                \begin{aligned}
                    g_{6}(\bm{\mathrm{x}}) = \ &161.18996 x_{3} \cdot x_{4} + 5000 x_{2} \cdot x_{4} - \\
                    &489510 x_{2} - x_{3} \cdot x_{4} \cdot x_{7} \leq 0
                \end{aligned}
            \end{equation*}
            \begin{equation*}
                g_{7}(\bm{\mathrm{x}}) = 0.33 x_{7} + 44.333333 - x_{5} \leq 0
            \end{equation*}
            \begin{equation*}
                g_{8}(\bm{\mathrm{x}}) = 0.22556 x_{5} - 1 - 0.007595 x_{7} \leq 0
            \end{equation*}
            \begin{equation*}
                g_{9}(\bm{\mathrm{x}}) = 0.00061 x_{3} - 1 - 0.0005 x_{1} \leq 0
            \end{equation*}
            \begin{equation*}
                g_{10}(\bm{\mathrm{x}}) = 0.819672 x_{1} - x_{3} + 0.819672 \leq 0
            \end{equation*}
            \begin{equation*}
                g_{11}(\bm{\mathrm{x}}) = 24500 x_{2} - 250 x_{2} \cdot x_{4} - x_{3} \cdot x_{4} \leq 0
            \end{equation*}
            \begin{equation*}
                \begin{aligned}
                    g_{12}(\bm{\mathrm{x}}) = \ &1020.1082 x_{4} \cdot x_{2} + 1.2244898 x_{3} \cdot x_{4} - \\
                    &100000 x_{2} \leq 0
                \end{aligned}
            \end{equation*}
            \begin{equation*}
                \begin{aligned}
                    g_{13}(\bm{\mathrm{x}}) = \ &6.25 x_{1} \cdot x_{6} + 6.25 x_{1} - 7.625 x_{3} - \\
                    &100000 \leq 0
                \end{aligned}
            \end{equation*}
            \begin{equation*}
                g_{14}(\bm{\mathrm{x}}) = 1.22 x_{3} - x_{6} \cdot x_{1} - x_{1} + 1 \leq 0
            \end{equation*}

            \textbf{With bounds:}
            \begin{equation*}
                \begin{aligned}
                    &1000 \leq x_{1} \leq 2000, \quad 0 \leq x_{2}, x_{4}, x_{5} \leq 100, \\
                    &2000 \leq x_{3} \leq 4000, \quad 0 \leq x_{6} \leq 20, \\
                    &0 \leq x_{7} \leq 200
                \end{aligned}
            \end{equation*}

            \begin{table*}[htbp!]
                \centering
                \caption{The best results for the Optimal operation of alkylation unit.}
                \label{tab23}
                \begin{tabular}{\cm{0.08\textwidth}\cm{0.085\textwidth}\cm{0.085\textwidth}\cm{0.085\textwidth}\cm{0.085\textwidth}\cm{0.085\textwidth}\cm{0.085\textwidth}\cm{0.085\textwidth}\cm{0.085\textwidth}}
                    \toprule
                     & DoS & AOO & PSA & FFA & AVOA & ETO & SGA & WAA \\
                    \midrule
                    $x_{1}$ & 2000 & 1999.96884 & 1292.95679 & 1999.99371 & 2000 & 1994.89058 & NaN & 1195.04863 \\
                    $x_{2}$ & 0 & 0 & 59.1542907 & 0 & 0 & 0 & NaN & 43.9829715 \\
                    $x_{3}$ & 2576.38006 & 2498.27354 & 2158.35587 & 3162.21346 & 2403.38284 & 3223.05907 & NaN & 2116.12799 \\
                    $x_{4}$ & 0 & 0 & 85.5187486 & 0 & 0 & 0 & NaN & 82.1842824 \\
                    $x_{5}$ & 58.1606111 & 57.8291168 & 92.1783142 & 60.8575728 & 57.4311232 & 61.8920633 & NaN & 91.0010561 \\
                    $x_{6}$ & 1.25994095 & 0.92996449 & 13.4113046 & 4.33703641 & 0.5441601 & 5.18089297 & NaN & 11.3993706 \\
                    $x_{7}$ & 41.5998808 & 40.6421005 & 143.515131 & 49.794398 & 39.4993316 & 52.4282639 & NaN & 141.400686 \\
                    Best & \textbf{-4529.12} & -4527.4937 & 362.83445 & -4437.0876 & -4521.5882 & -4289.9606 & NaN & 581.693178 \\
                    \bottomrule
                \end{tabular}
            \end{table*}

            \begin{table*}[htbp!]
                \centering
                \caption{Statistical results for the Optimal operation of alkylation unit.}
                \label{tab24}
                \begin{tabular}{\cm{0.08\textwidth}\cm{0.085\textwidth}\cm{0.085\textwidth}\cm{0.085\textwidth}\cm{0.085\textwidth}\cm{0.085\textwidth}\cm{0.085\textwidth}\cm{0.085\textwidth}\cm{0.085\textwidth}}
                    \toprule
                     & DoS & AOO & PSA & FFA & AVOA & ETO & SGA & WAA \\
                    \midrule
                    Mean & \textbf{-4481.477} & -3649.8628 & 620.373012 & -2583.1734 & -4200.1079 & -4034.2287 & NaN & 726.378768 \\
                    Std & 238.213415 & 634.096046 & 165.068727 & 2075.36504 & 373.127814 & 332.001258 & NaN & \textbf{0} \\
                    Success & \textbf{1} & \textbf{1} & 0.48 & 0.6 & \textbf{1} & 0.28 & 0 & 0.04 \\
                    WSRT &  & 8.3E-09(+) & 1.0E-09(+) & 3.2E-09(+) & 2.6E-08(+) & 1.2E-08(+) & NaN & 9.7E-11(+) \\
                    FMR & \textbf{1.12} & 3.64 & 6 & 4.56 & 2.64 & 3.2 & NaN & 6.84 \\
                    F-Rank & \textbf{1} & 4 & 6 & 5 & 2 & 3 & 8 & 7 \\
                    \bottomrule
                \end{tabular}
            \end{table*}

            Tables~\ref{tab23} and~\ref{tab24} respectively present the best results and statistical results of DoS and the competing algorithms on this problem. The results indicate that this problem imposes high requirements on constraint handling capabilities, as many algorithms performed poorly in terms of feasibility. SGA failed to obtain any feasible solution in all 25 independent runs, while PSA, FFA, ETO and WAA also exhibited low success rates, revealing significant limitations in their search strategies for this problem.

            In contrast, DoS, AOO and AVOA successfully converged to feasible solutions in all runs, demonstrating stronger adaptability and constraint-handling stability. Among them, DoS not only achieved a 100\% success rate but also obtained the best values in both Best and Mean performance metrics, significantly outperforming the second ranked AVOA, clearly indicating its advantages in global search capability and solution efficiency.

            Although DoS ranked second in the Std metric, it is worth noting that WAA only obtained one feasible solution, making its stability results less meaningful for reference. Therefore, DoS exhibits not only high optimization performance but also good stability.

            In terms of statistical significance, the WSRT results show that DoS holds a significant advantage over all other algorithms. Finally, DoS ranks first in the Friedman ranking, further validating its strong adaptability and comprehensive optimization capability when addressing such complex industrial problems.

        \subsection{Process Flow Sheeting}

            Process flow sheeting is a classical optimization problem in process systems engineering, aiming to minimize the operating cost or maximize the product profit of a process system, under the constraints of mass balance, energy balance and operational limits. This problem involves three decision variables, representing the operating conditions of key equipment and the distribution ratio of material flows, denoted as $\bm{\mathrm{x}} = (x_{1}, x_{2}, x_{3})$. The specific mathematical formulation is presented in Equation~\ref{eq32}.

            \textbf{Minimize:}
            \begin{equation}
                f(\bm{\mathrm{x}}) = - 0.7 x_{3} + 0.8 + 5 (0.5 - x_{1})^{2}
                \label{eq32}
            \end{equation}

            \textbf{Subject to:}
            \begin{equation*}
                g_{1}(\bm{\mathrm{x}}) = -\exp(x_{1} - 0.2) - x_{2} \leq 0
            \end{equation*}
            \begin{equation*}
                g_{2}(\bm{\mathrm{x}}) = x_{2} + 1.1 x_{3} + 1 \leq 0
            \end{equation*}
            \begin{equation*}
                g_{3}(\bm{\mathrm{x}}) = x_{1} - x_{3} - 0.2 \leq 0
            \end{equation*}

            \textbf{With bounds:}
            \begin{equation*}
                0.2 \leq x_{1} \leq 1, \quad -2.22554 \leq x_{2} \leq -1, \quad x_{3} \in \{0,1\}
            \end{equation*}

            \begin{table*}[htbp!]
                \centering
                \caption{The best results for the Process flow sheeting.}
                \label{tab25}
                \begin{tabular}{\cm{0.08\textwidth}\cm{0.085\textwidth}\cm{0.085\textwidth}\cm{0.085\textwidth}\cm{0.085\textwidth}\cm{0.085\textwidth}\cm{0.085\textwidth}\cm{0.085\textwidth}\cm{0.085\textwidth}}
                    \toprule
                     & DoS & AOO & PSA & FFA & AVOA & ETO & SGA & WAA \\
                    \midrule
                    $x_{1}$ & 0.94193734 & 0.94193735 & 0.94193734 & 0.94193759 & 0.94193734 & 0.94212928 & 0.94193738 & 0.94201652 \\
                    $x_{2}$ & -2.1 & -2.1 & -2.1 & -2.1000003 & -2.1 & -2.1002767 & -2.1000001 & -2.1001605 \\
                    $x_{3}$ & 0.75209273 & 1.31123275 & 0.95099277 & 1.1686118 & 1.48992977 & 1.00614078 & 1.26073299 & 1.1339543 \\
                    Best & \textbf{1.0765431} & 1.07654309 & 1.07654308 & 1.07654415 & 1.07654308 & 1.07739151 & 1.07654326 & 1.07689302 \\
                    \bottomrule
                \end{tabular}
            \end{table*}

            \begin{table*}[htbp!]
                \centering
                \caption{Statistical results for the Process flow sheeting.}
                \label{tab26}
                \begin{tabular}{\cm{0.08\textwidth}\cm{0.085\textwidth}\cm{0.088\textwidth}\cm{0.085\textwidth}\cm{0.085\textwidth}\cm{0.085\textwidth}\cm{0.085\textwidth}\cm{0.085\textwidth}\cm{0.085\textwidth}}
                    \toprule
                     & DoS & AOO & PSA & FFA & AVOA & ETO & SGA & WAA \\
                    \midrule
                    Mean & \textbf{1.0765431} & 1.07654364 & 1.1320493 & 1.08980276 & 1.18061723 & 1.0804378 & 1.19451316 & 1.12838518 \\
                    Std & \textbf{4.532E-16} & 1.3065E-06 & 0.08258201 & 0.02321725 & 0.08672846 & 0.00176624 & 0.08255323 & 0.07745561 \\
                    Success & \textbf{1} & \textbf{1} & \textbf{1} & \textbf{1} & \textbf{1} & 0.6 & \textbf{1} & \textbf{1} \\
                    WSRT &  & 9.7E-11(+) & 2.4E-03(+) & 9.7E-11(+) & 4.9E-11(+) & 8.0E-11(+) & 3.6E-11(+) & 9.1E-11(+) \\
                    FMR & \textbf{1.34} & 3.2 & 3.4 & 5.04 & 5.44 & 5.44 & 6.26 & 5.88 \\
                    F-Rank & \textbf{1} & 2 & 3 & 4 & 5 & 6 & 8 & 7 \\
                    \bottomrule
                \end{tabular}
            \end{table*}

            Tables~\ref{tab25} and~\ref{tab26} present the best results and statistical indicators of each algorithm on this problem. The results show that DoS achieves the best performance in terms of Best, Mean and Std, fully demonstrating its excellent convergence capability and stability. Although most algorithms achieve values close to DoS in the Best metric, DoS shows significantly better performance in Mean and Std, with a standard deviation as low as 4.532E-16, revealing strong robustness and consistency.

            In terms of statistical significance, the WSRT show that DoS exhibits significant superiority over all competitors. The Friedman ranking further confirms this conclusion, with DoS ranking first on this problem, demonstrating its outstanding adaptability and overall performance in Process Synthesis and Design engineering optimization problems.

        \subsection{Welded Beam Design}

            The welded beam design problem seeks to minimize manufacturing cost and structural weight while satisfying strength, geometric, and manufacturability constraints. This problem involves four decision variables, representing the weld seam height, top flange width, beam height and beam length, denoted as: $\bm{\mathrm{x}} = (x_{1}, x_{2}, x_{3}, x_{4})$. The constraints include seven inequality constraints, covering aspects such as maximum stress, shear stress, deflection and geometric boundaries. The complete mathematical model is provided in Equation~\ref{eq33}.

            \textbf{Minimize:}
            \begin{equation}
                f(\bm{\mathrm{x}}) = 1.1047 x_{1}^{2} \cdot x_{2} + 0.04811 x_{3} \cdot x_{4} (14 + x_{2})
                \label{eq33}
            \end{equation}

            \textbf{Subject to:}
            \begin{equation*}
                g_{1}(\bm{\mathrm{x}}) = \tau(\bm{\mathrm{x}}) - 13600 \leq 0
            \end{equation*}
            \begin{equation*}
                g_{2}(\bm{\mathrm{x}}) = \sigma(\bm{\mathrm{x}}) - 30000 \leq 0
            \end{equation*}
            \begin{equation*}
                g_{3}(\bm{\mathrm{x}}) = \delta(\bm{\mathrm{x}}) - 0.25 \leq 0
            \end{equation*}
            \begin{equation*}
                g_{4}(\bm{\mathrm{x}}) = x_{1} - x_{4} \leq 0
            \end{equation*}
            \begin{equation*}
                g_{5}(\bm{\mathrm{x}}) = 6000 - P_{c}(\bm{\mathrm{x}}) \leq 0
            \end{equation*}
            \begin{equation*}
                g_{6}(\bm{\mathrm{x}}) = 0.125 - x_{1} \leq 0
            \end{equation*}
            \begin{equation*}
                g_{7}(\bm{\mathrm{x}}) = 1.1047 x_{1} + 0.04811 x_{3} \cdot x_{4} (14 + x_{2}) - 5 \leq 0
            \end{equation*}

            \textbf{Where:}
            \begin{equation*}
                \tau(\bm{\mathrm{x}}) = \sqrt{ (\tau')^{2} + (\tau'')^{2} + \frac{x_{2} \cdot \tau' \cdot \tau''} {\sqrt{0.25 \left[x_{2}^{2} + (x_{1} + x_{3})^{2}\right]}} }
            \end{equation*}
            \begin{equation*}
                \tau' = \frac{6000}{\sqrt{2} x_{1} \cdot x_{2}}
            \end{equation*}
            \begin{equation*}
                \tau'' = \frac{6000 (14 + 0.5 x_{2}) \sqrt{ 0.25 \left[x_{2}^{2} + (x_{1} + x_{3})^{2}\right] } } { 2 \times 0.707 x_{1} \cdot x_{2} \left[ \frac{x_{2}^{2}}{12} + 0.25 (x_{1} + x_{3})^{2} \right] }
            \end{equation*}
            \begin{equation*}
                \sigma(\bm{\mathrm{x}}) = \frac{504000}{x_{3}^{2} x_{4}}
            \end{equation*}
            \begin{equation*}
                \delta(\bm{\mathrm{x}}) = \frac{65856000}{30 \times 10^{6} x_{4} \cdot x_{3}^{3}}
            \end{equation*}
            \begin{equation*}
                P_{c}(\bm{\mathrm{x}}) = 64746.022 \left(1 - 0.0282346 x_{3}\right) x_{3} \cdot x_{4}^{3}
            \end{equation*}

            \textbf{With bounds:}
            \begin{equation*}
                0.1 \leq x_{1}, x_{4} \leq 2, \quad 0.1 \leq x_{2}, x_{3} \leq 10
            \end{equation*}

            \begin{table*}[htbp!]
                \centering
                \caption{The best results for the Welded beam design.}
                \label{tab27}
                \begin{tabular}{\cm{0.08\textwidth}\cm{0.085\textwidth}\cm{0.085\textwidth}\cm{0.085\textwidth}\cm{0.085\textwidth}\cm{0.085\textwidth}\cm{0.085\textwidth}\cm{0.085\textwidth}\cm{0.085\textwidth}}
                    \toprule
                     & DoS & AOO & PSA & FFA & AVOA & ETO & SGA & WAA \\
                    \midrule
                    $x_{1}$ & 0.20572964 & 0.20460826 & 0.20572964 & 0.20572963 & 0.20565992 & 0.20379636 & 0.20385754 & 0.20151701 \\
                    $x_{2}$ & 3.25312004 & 3.2733965 & 3.25312004 & 3.25312028 & 3.25436302 & 3.28201771 & 3.28630361 & 3.34964561 \\
                    $x_{3}$ & 9.03662391 & 9.03605126 & 9.0366239 & 9.03662391 & 9.03662131 & 9.07355424 & 9.0385155 & 9.08696903 \\
                    $x_{4}$ & 0.20572964 & 0.20575609 & 0.20572964 & 0.20572964 & 0.20572976 & 0.20554814 & 0.20572023 & 0.20557583 \\
                    Best & \textbf{1.6952472} & 1.69644556 & 1.69524717 & 1.6952472 & 1.69531381 & 1.70126017 & 1.69723581 & 1.70952413 \\
                    \bottomrule
                \end{tabular}
            \end{table*}

            \begin{table*}[htbp!]
                \centering
                \caption{Statistical results for the Welded beam design.}
                \label{tab28}
                \begin{tabular}{\cm{0.08\textwidth}\cm{0.085\textwidth}\cm{0.085\textwidth}\cm{0.085\textwidth}\cm{0.085\textwidth}\cm{0.085\textwidth}\cm{0.085\textwidth}\cm{0.085\textwidth}\cm{0.085\textwidth}}
                    \toprule
                     & DoS & AOO & PSA & FFA & AVOA & ETO & SGA & WAA \\
                    \midrule
                    Mean & \textbf{1.6952472} & 1.7033707 & 1.76975031 & 1.73309548 & 1.71426008 & 1.70811517 & 1.71747601 & 1.82058308 \\
                    Std & \textbf{2.266E-16} & 0.01235605 & 0.1188277 & 0.04718389 & 0.01906936 & 0.00371913 & 0.024741 & 0.16032823 \\
                    Success & \textbf{1} & \textbf{1} & \textbf{1} & 0.08 & \textbf{1} & \textbf{1} & \textbf{1} & \textbf{1} \\
                    WSRT &  & 9.7E-11(+) & 9.7E-11(+) & 6.7E-12(+) & 9.7E-11(+) & 9.7E-11(+) & 9.7E-11(+) & 9.7E-11(+) \\
                    FMR & \textbf{1} & 3.44 & 4.72 & 6.44 & 4.4 & 4.32 & 4.68 & 7 \\
                    F-Rank & \textbf{1} & 2 & 6 & 7 & 4 & 3 & 5 & 8 \\
                    \bottomrule
                \end{tabular}
            \end{table*}

            Tables~\ref{tab27} and~\ref{tab28} present the comparative experimental results of DoS against several advanced competing algorithms on this problem. DoS achieved the best performance across all key metrics, including Best, Mean, Std and Success, fully demonstrating its superior optimization ability and convergence stability. Although algorithms such as AOO, PSA and FFA performed closely in terms of the Best value, the Mean and Std results show that DoS exhibited higher consistency and robustness over 25 independent runs.

            In addition, DoS achieved a 100\% success rate, indicating its strong adaptability in handling complex constrained problems. In contrast, although FFA obtained a Best value close to that of DoS, its success rate was only 0.08, suggesting significant shortcomings in its constraint handling strategy. This limits its ability to effectively explore feasible regions in engineering problems.

            From the perspective of statistical significance, the Wilcoxon rank-sum test indicates that DoS significantly outperformed all competing algorithms. In the Friedman test, DoS further widened the gap with the second-best AOO by a margin of 2.44, reaffirming its overall performance advantage in this constrained optimization problem.

        \subsection{Pressure Vessel Design}

            Pressure vessel design aims to minimize the manufacturing cost of a pressure vessel while satisfying constraints related to structural strength and operational safety. This problem involves four decision variables, representing the inner radius of the cylindrical shell, head thickness, shell thickness and length of the vessel, denoted as: $\bm{\mathrm{x}} = (x_{1}, x_{2}, x_{3}, x_{4})$. The problem includes four inequality constraints. The detailed mathematical formulation is given in Equation~\ref{eq34}.

            \textbf{Minimize:}
            \begin{equation}
                \begin{aligned}
                    f(\bm{\mathrm{x}}) = \ &1.7781 z_{2} x_{3}^{2} + 0.6224 z_{1} x_{3} x_{4} + \\
                    &3.1661 z_{1}^{2} x_{4} + 19.84 z_{1}^{2} x_{3}
                \end{aligned}
                \label{eq34}
            \end{equation}

            \textbf{Subject to:}
            \begin{equation*}
                g_{1}(\bm{\mathrm{x}}) = 0.00954 x_{3} - z_{2} \leq 0
            \end{equation*}
            \begin{equation*}
                g_{2}(\bm{\mathrm{x}}) = 0.0193 x_{3} - z_{1} \leq 0
            \end{equation*}
            \begin{equation*}
                g_{3}(\bm{\mathrm{x}}) = x_{4} - 240 \leq 0
            \end{equation*}
            \begin{equation*}
                g_{4}(\bm{\mathrm{x}}) = -\pi x_{3}^{2} \cdot x_{4} - \frac{4}{3} \pi x_{3}^{3} + 1296000 \leq 0
            \end{equation*}

            \textbf{Where:}
            \begin{equation*}
                z_{1} = 0.0625 x_{1}
            \end{equation*}
            \begin{equation*}
                z_{2} = 0.0625 x_{2}
            \end{equation*}

            \textbf{With bounds:}
            \begin{equation*}
                x_{1}, x_{2} \in \{1,2,\ldots,99\}, \quad 10 \leq x_{3}, x_{4} \leq 200
            \end{equation*}

            \begin{table*}[htbp!]
                \centering
                \caption{The best results for the Pressure vessel design.}
                \label{tab29}
                \begin{tabular}{\cm{0.08\textwidth}\cm{0.085\textwidth}\cm{0.085\textwidth}\cm{0.085\textwidth}\cm{0.085\textwidth}\cm{0.085\textwidth}\cm{0.085\textwidth}\cm{0.085\textwidth}\cm{0.085\textwidth}}
                    \toprule
                     & DoS & AOO & PSA & FFA & AVOA & ETO & SGA & WAA \\
                    \midrule
                    $x_{1}$ & 12.7927137 & 12.8923688 & 13.018652 & 13.4194694 & 13.0437498 & 12.7584516 & 13.3917073 & 15.7473509 \\
                    $x_{2}$ & 7.31679176 & 7.22512106 & 6.88413741 & 7.17723922 & 7.10333375 & 6.78178608 & 6.71662934 & 8.40498716 \\
                    $x_{3}$ & 42.0984456 & 42.0982098 & 42.0984456 & 42.0721363 & 42.0984456 & 42.0512357 & 42.0984135 & 51.8132818 \\
                    $x_{4}$ & 176.636596 & 176.639524 & 176.636596 & 176.963169 & 176.636596 & 177.425062 & 176.636993 & 84.5827817 \\
                    Best & \textbf{6059.7143} & 6059.74319 & 6059.71434 & 6062.92769 & 6059.71434 & 6070.20351 & 6059.71824 & 6410.20622 \\
                    \bottomrule
                \end{tabular}
            \end{table*}

            \begin{table*}[htbp!]
                \centering
                \caption{Statistical results for the Pressure vessel design.}
                \label{tab30}
                \begin{tabular}{\cm{0.08\textwidth}\cm{0.085\textwidth}\cm{0.085\textwidth}\cm{0.085\textwidth}\cm{0.085\textwidth}\cm{0.085\textwidth}\cm{0.085\textwidth}\cm{0.085\textwidth}\cm{0.085\textwidth}}
                    \toprule
                     & DoS & AOO & PSA & FFA & AVOA & ETO & SGA & WAA \\
                    \midrule
                    Mean & \textbf{6067.5069} & 6349.98002 & 6399.20768 & 6539.31568 & 6744.7382 & 6471.54327 & 6495.56359 & 6808.79589 \\
                    Std & \textbf{14.273598} & 498.958892 & 221.108314 & 264.377381 & 532.303828 & 424.078767 & 385.542707 & 161.908079 \\
                    Success & \textbf{1} & \textbf{1} & \textbf{1} & \textbf{1} & \textbf{1} & 0.96 & \textbf{1} & \textbf{1} \\
                    WSRT &  & 1.3E-05(+) & 4.5E-09(+) & 7.1E-10(+) & 9.2E-09(+) & 5.7E-09(+) & 1.8E-08(+) & 5.5E-10(+) \\
                    FMR & \textbf{1.32} & 3.52 & 4.16 & 5.08 & 5.44 & 5.16 & 4.48 & 6.84 \\
                    F-Rank & \textbf{1} & 2 & 3 & 5 & 7 & 6 & 4 & 8 \\
                    \bottomrule
                \end{tabular}
            \end{table*}

            Tables~\ref{tab29} and~\ref{tab30} present the experimental results of DoS and competitors on this problem. The statistical results show that DoS achieved the best performance in three key indicators: Best, Mean and Std and reached a 100\% success rate across 25 independent runs, demonstrating its excellent optimization performance and convergence stability.

            In terms of statistical significance analysis, the Wilcoxon rank-sum test (WSRT) indicates that DoS shows a statistically significant advantage over all competitors. In the Friedman test, DoS ranks first among all algorithms, outperforming the second-best algorithm AOO by 1.2, further validating its overall performance advantage and robustness in this type of engineering optimization problem.

        \subsection{Side Impact Design of Automobiles}

            The Side impact design of automobiles aims to minimize the overall vehicle weight by optimizing the structural parameters of the car body, under the premise of satisfying safety and structural constraints. This problem involves 11 decision variables, each representing the structural thickness of key body components, denoted as: $\bm{\mathrm{x}} = (x_{1}, x_{2}, \cdots , x_{11})$. The constraints cover occupant injury indices, structural intrusion and overall vehicle mass, making this a typical constrained single objective optimization problem. The formulation is given in Equation~\ref{eq35}.

            \textbf{Minimize:}
            \begin{equation}
                \begin{aligned}
                    f(\bm{\mathrm{x}}) = \ &1.98 + 4.9 x_{1} + 6.67 x_{2} + 6.98 x_{3} + \\
                    &4.01 x_{4} + 1.78 x_{5} + 2.73 x_{7}
                \end{aligned}
                \label{eq35}
            \end{equation}

            \textbf{Subject to:}
            \begin{equation*}
                \begin{aligned}
                    g_{1}(\bm{\mathrm{x}}) = \ &1.16 - 0.3717 x_{2} \cdot x_{4} - 0.00931 x_{2} \cdot x_{10} - \\
                    &0.484 x_{3} \cdot x_{9} + 0.01343 x_{6} \cdot x_{10} - 1 \leq 0
                \end{aligned}
            \end{equation*}
            \begin{equation*}
                \begin{aligned}
                    g_{2}(\bm{\mathrm{x}}) = \ &46.36 - 9.9 x_{2} - 12.9 x_{1} \cdot x_{2} + \\
                    &0.1107 x_{3} \cdot x_{10} - 32 \leq 0
                \end{aligned}
            \end{equation*}
            \begin{equation*}
                \begin{aligned}
                    g_{3}(\bm{\mathrm{x}}) = \ &33.86 + 2.95 x_{3} + 0.1792 x_{3} - \\
                    &5.057 x_{1} \cdot x_{2} - 11 x_{2} \cdot x_{8} - 0.0215 x_{5} \cdot x_{10} - \\
                    &9.98 x_{7} \cdot x_{8} + 22 x_{8} \cdot x_{9} - 32 \leq 0
                \end{aligned}
            \end{equation*}
            \begin{equation*}
                \begin{aligned}
                    g_{4}(\bm{\mathrm{x}}) = \ &28.98 + 3.818 x_{3} - 4.2 x_{1} \cdot x_{2} + \\
                    &0.0207 x_{5} \cdot x_{10} + 6.63 x_{6} \cdot x_{9} - \\
                    &7.7 x_{7} \cdot x_{8} + 0.32 x_{9} \cdot x_{10} - 32 \leq 0
                \end{aligned}
            \end{equation*}
            \begin{equation*}
                \begin{aligned}
                    g_{5}(\bm{\mathrm{x}}) = \ &0.261 - 0.0159 x_{1} \cdot x_{2} - 0.188 x_{1} \cdot x_{8} - \\
                    &0.019 x_{2} \cdot x_{7} + 0.0144 x_{3} \cdot x_{5} + \\
                    &0.0008757 x_{5} \cdot x_{10} + 0.08045 x_{6} \cdot x_{9} + \\
                    &0.00139 x_{8} \cdot x_{11} + 0.00001575 x_{10} \cdot x_{11} - \\
                    &0.32 \leq 0
                \end{aligned}
            \end{equation*}
            \begin{equation*}
                \begin{aligned}
                    g_{6}(\bm{\mathrm{x}}) = \ &0.214 + 0.00817 x_{5} - 0.131 x_{1} \cdot x_{8} - \\
                    &0.0704 x_{1} \cdot x_{9} + 0.03099 x_{2} \cdot x_{6} - \\
                    &0.018 x_{2} \cdot x_{7} + 0.0208 x_{3} \cdot x_{8} + 0.121 x_{3} \cdot x_{9} - \\
                    &0.00364 x_{5} \cdot x_{6} + 0.0007715 x_{5} \cdot x_{10} - \\
                    &0.0005354 x_{6} \cdot x_{10} + 0.00121 x_{8} \cdot x_{11} + \\
                    &0.00184 x_{9} \cdot x_{10} - 0.02 x_{2}^{2} - 0.32 \leq 0
                \end{aligned}
            \end{equation*}
            \begin{equation*}
                \begin{aligned}
                    g_{7}(\bm{\mathrm{x}}) = \ &0.74 - 0.61 x_{2} - 0.163 x_{3} \cdot x_{8} + \\
                    &0.001232 x_{3} \cdot x_{10} - 0.166 x_{7} \cdot x_{9} + \\
                    &0.227 x_{2}^{2} - 0.32 \leq 0
                \end{aligned}
            \end{equation*}
            \begin{equation*}
                \begin{aligned}
                    g_{8}(\bm{\mathrm{x}}) = \ &4.72 - 0.5 x_{4} - 0.19 x_{2} \cdot x_{3} - \\
                    &0.0122 x_{4} \cdot x_{10} + 0.009325 x_{6} \cdot x_{10} + \\
                    &0.000191 x_{11}^{2} - 4 \leq 0
                \end{aligned}
            \end{equation*}
            \begin{equation*}
                \begin{aligned}
                    g_{9}(\bm{\mathrm{x}}) = \ &10.58 - 0.674 x_{1} \cdot x_{2} - 1.95 x_{2} \cdot x_{8} + \\
                    &0.02054 x_{3} \cdot x_{10} - 0.0198 x_{4} \cdot x_{10} + \\
                    &0.028 x_{6} \cdot x_{10} - 9.9 \leq 0
                \end{aligned}
            \end{equation*}
            \begin{equation*}
                \begin{aligned}
                    g_{10}(\bm{\mathrm{x}}) = \ &16.45 - 0.489 x_{3} x_{7} - 0.843 x_{5} x_{6} + \\
                    &0.0432 x_{9} x_{10} - 0.0556 x_{9} x_{11} - \\
                    &0.000786 x_{11}^{2} - 15.7 \leq 0
                \end{aligned}
            \end{equation*}

            \textbf{With bounds:}
            \begin{equation*}
                0.5 \leq x_{1}, x_{2}, x_{3}, x_{4}, x_{5}, x_{6}, x_{7} \leq 1.5,
            \end{equation*}
            \begin{equation*}
                x_{8}, x_{9} \in \{0.192, 0.345\}, \quad -30 \leq x_{10}, x_{11} \leq 30
            \end{equation*}

            \begin{table*}[htbp!]
                \centering
                \caption{The best results for the Side impact design of automobiles.}
                \label{tab31}
                \begin{tabular}{\cm{0.08\textwidth}\cm{0.088\textwidth}\cm{0.085\textwidth}\cm{0.085\textwidth}\cm{0.085\textwidth}\cm{0.085\textwidth}\cm{0.085\textwidth}\cm{0.085\textwidth}\cm{0.085\textwidth}}
                    \toprule
                     & DoS & AOO & PSA & FFA & AVOA & ETO & SGA & WAA \\
                    \midrule
                    $x_{1}$ & 0.5 & 0.50263548 & 0.62281348 & NaN & 0.5 & 0.5 & 0.51948093 & 0.5 \\
                    $x_{2}$ & 0.97984709 & 0.97778074 & 0.88730099 & NaN & 0.96383745 & 0.92325516 & 0.96325679 & 0.94018371 \\
                    $x_{3}$ & 0.5 & 0.50000005 & 0.5 & NaN & 0.5 & 0.5 & 0.5 & 0.5 \\
                    $x_{4}$ & 1.00244987 & 1.0056303 & 1.02435596 & NaN & 1.04103104 & 1.14925255 & 1.00876643 & 1.12618121 \\
                    $x_{5}$ & 0.5 & 0.50004965 & 0.5 & NaN & 0.50000002 & 0.5 & 0.5 & 0.5 \\
                    $x_{6}$ & 0.5 & 0.50854536 & 0.5 & NaN & 0.5000684 & 0.51201813 & 0.50000657 & 0.52275068 \\
                    $x_{7}$ & 0.5 & 0.50012063 & 0.5 & NaN & 0.5 & 0.5 & 0.5 & 0.5 \\
                    $x_{8}$ & 0.99702629 & 0.91019198 & 1 & NaN & 0.78736634 & 0.53490886 & 0.72674984 & 0.58931079 \\
                    $x_{9}$ & 5.9061E-15 & 0.0014404 & 0.56869178 & NaN & 0.17163354 & 0.63277634 & 0.08640629 & 0.3080347 \\
                    $x_{10}$ & 30 & 29.9901901 & 28.0599228 & NaN & 25.270859 & 12.5572851 & 29.4727881 & 18.283222 \\
                    $x_{11}$ & 23.0333313 & -23.043959 & 22.4554294 & NaN & 22.3247219 & 17.3388072 & 23.0176274 & 22.1871779 \\
                    Best & \textbf{20.730404} & 20.7427069 & 20.8027511 & NaN & 20.7783303 & 20.9416146 & 20.7405328 & 20.962012 \\
                    \bottomrule
                \end{tabular}
            \end{table*}

            \begin{table*}[htbp!]
                \centering
                \caption{Statistical results for the Side impact design of automobiles.}
                \label{tab32}
                \begin{tabular}{\cm{0.08\textwidth}\cm{0.085\textwidth}\cm{0.085\textwidth}\cm{0.085\textwidth}\cm{0.085\textwidth}\cm{0.085\textwidth}\cm{0.085\textwidth}\cm{0.085\textwidth}\cm{0.085\textwidth}}
                    \toprule
                     & DoS & AOO & PSA & FFA & AVOA & ETO & SGA & WAA \\
                    \midrule
                    Mean & \textbf{20.850749} & 21.0935379 & 21.093146 & NaN & 21.0122346 & 21.1392367 & 21.1213781 & 21.2455305 \\
                    Std & 0.1790251 & \textbf{0.1333431} & 0.17630027 & NaN & 0.36347931 & 0.16308521 & 0.39526243 & 0.20830952 \\
                    Success & \textbf{1} & \textbf{1} & \textbf{1} & 0 & \textbf{1} & \textbf{1} & \textbf{1} & \textbf{1} \\
                    WSRT &  & 4.0E-08(+) & 5.3E-06(+) & NaN & 4.1E-04(+) & 2.2E-07(+) & 4.5E-05(+) & 4.0E-08(+) \\
                    FMR & \textbf{1.76} & 3.92 & 3.72 & NaN & 3.92 & 5 & 3.72 & 5.96 \\
                    F-Rank & \textbf{1} & 4 & 2 & 8 & 5 & 6 & 3 & 7 \\
                    \bottomrule
                \end{tabular}
            \end{table*}

            Tables~\ref{tab31} and~\ref{tab32} present the best solutions and statistical indicators of each algorithm for the side impact design problem. The results show that all algorithms except FFA successfully converged to feasible solutions in all 25 independent runs, achieving a success rate of 1. In contrast, FFA failed to obtain any feasible solution, indicating that its search strategy lacks adaptability to this problem.

            In terms of optimization performance, DoS achieved the best results in both Best and Mean indicators, demonstrating excellent optimization capability. Although DoS ranked slightly lower than AOO in the Std indicator, indicating slightly lower stability, the significant advantage in Mean shows that DoS is more consistently able to approach the global optimal solution across most runs. In contrast, AOO, although occasionally producing better results in individual runs, is more prone to getting trapped in local optima, reflecting weaker global search capability.

            The Wilcoxon rank-sum test further confirms that DoS shows statistically significant performance advantages over all competitors. In the Friedman test, DoS consistently ranks first, fully demonstrating its superior performance, robustness and practical value in this complex constrained optimization problem.

        \subsection{Optimal Design of Industrial Refrigeration System}

            Optimal design of industrial refrigeration system aims to minimize the total energy consumption and investment cost of the equipment while ensuring thermodynamic performance, energy efficiency ratio and system stability. This problem involves 14 decision variables, covering key design parameters such as compressor power, evaporation temperature and condensation pressure, denoted as: $\bm{\mathrm{x}} = (x_{1}, x_{2}, \cdots , x_{14})$. The objective function and 15 inequality constraints form the complete mathematical model, as shown in Equation~\ref{eq36}.

            \textbf{Minimize:}
            \begin{equation}
                \begin{aligned}
                    f(\bm{\mathrm{x}}) = \ &63098.88 x_{2} \cdot x_{4} \cdot x_{12} + 5441.5 x_{2}^{2} \cdot x_{12} + \\
                    &115055.5 x_{2}^{1.664} \cdot x_{6} + 6172.27 x_{2}^{2} \cdot x_{6} + \\
                    &63098.88 x_{1} \cdot x_{3} \cdot x_{11} + 5441.5 x_{1}^{2} \cdot x_{11} + \\
                    &115055.5 x_{1}^{1.664} \cdot x_{5} + 6172.27 x_{1}^{2} \cdot x_{5} + \\
                    &140.53 x_{1} \cdot x_{11} + 281.29 x_{3} \cdot x_{11} + \\
                    &70.26 x_{1}^{2} + 281.29 x_{1} \cdot x_{3} + 281.29 x_{3}^{2} + \\
                    &14437 x_{8}^{1.8812} \cdot x_{12}^{0.3424} \cdot x_{10}^{-1} \cdot x_{1}^{2} \cdot x_{7} \cdot x_{9}^{-1} + \\
                    &20470.2 x_{7}^{2.893} \cdot x_{11}^{0.316} \cdot x_{1}^{2}
                \end{aligned}
                \label{eq36}
            \end{equation}

            \textbf{Subject to:}
            \begin{equation*}
                g_{1}(\bm{\mathrm{x}}) = 1.524 x_{7}^{-1} \leq 1
            \end{equation*}
            \begin{equation*}
                g_{2}(\bm{\mathrm{x}}) = 1.524 x_{8}^{-1} \leq 1
            \end{equation*}
            \begin{equation*}
                g_{3}(\bm{\mathrm{x}}) = 0.07789 x_{1} - 2 x_{7}^{-1} \cdot x_{9} - 1 \leq 0
            \end{equation*}
            \begin{equation*}
                \begin{aligned}
                    g_{4}(\bm{\mathrm{x}}) = \ &7.05305 x_{9}^{-1} \cdot x_{1}^{2} \cdot x_{10} \cdot x_{8}^{-1} \cdot x_{2}^{-1} \cdot x_{14}^{-1} - \\
                    &1 \leq 0
                \end{aligned}
            \end{equation*}
            \begin{equation*}
                g_{5}(\bm{\mathrm{x}}) = 0.0833 x_{13}^{-1} \cdot x_{14} - 1 \leq 0
            \end{equation*}
            \begin{equation*}
                \begin{aligned}
                    g_{6}(\bm{\mathrm{x}}) = \ &47.136 x_{2}^{0.333} \cdot x_{10}^{-1} \cdot x_{12} - 1.333 x_{8} \cdot x_{13}^{2.1195} + \\
                    &62.08 x_{13}^{2.1195} \cdot x_{12}^{-1} \cdot x_{8}^{0.2} \cdot x_{10}^{-1} - 1 \leq 0
                \end{aligned}
            \end{equation*}
            \begin{equation*}
                g_{7}(\bm{\mathrm{x}}) = 0.04771 x_{10} \cdot x_{8}^{1.8812} \cdot x_{12}^{0.3424} - 1 \leq 0
            \end{equation*}
            \begin{equation*}
                g_{8}(\bm{\mathrm{x}}) = 0.0488 x_{9} \cdot x_{7}^{1.893} \cdot x_{11}^{0.316} - 1 \leq 0
            \end{equation*}
            \begin{equation*}
                g_{9}(\bm{\mathrm{x}}) = 0.0099 x_{1} \cdot x_{3}^{-1} - 1 \leq 0
            \end{equation*}
            \begin{equation*}
                g_{10}(\bm{\mathrm{x}}) = 0.0193 x_{2} \cdot x_{4}^{-1} - 1 \leq 0
            \end{equation*}
            \begin{equation*}
                g_{11}(\bm{\mathrm{x}}) = 0.0298 x_{1} \cdot x_{5}^{-1} - 1 \leq 0
            \end{equation*}
            \begin{equation*}
                g_{12}(\bm{\mathrm{x}}) = 0.056 x_{2} \cdot x_{6}^{-1} - 1 \leq 0
            \end{equation*}
            \begin{equation*}
                g_{13}(\bm{\mathrm{x}}) = 2 x_{9}^{-1} - 1 \leq 0
            \end{equation*}
            \begin{equation*}
                g_{14}(\bm{\mathrm{x}}) = 2 x_{10}^{-1} - 1 \leq 0
            \end{equation*}
            \begin{equation*}
                g_{15}(\bm{\mathrm{x}}) = x_{12} \cdot x_{11}^{-1} - 1 \leq 0
            \end{equation*}

            \textbf{With bounds:}
            \begin{equation*}
                0.001 \leq x_{i} \leq 5, \quad i = 1, 2, \ldots , 14
            \end{equation*}

            \begin{table*}[htbp!]
                \centering
                \caption{The best results for the Optimal design of industrial refrigeration system.}
                \label{tab33}
                \begin{tabular}{\cm{0.08\textwidth}\cm{0.085\textwidth}\cm{0.085\textwidth}\cm{0.085\textwidth}\cm{0.085\textwidth}\cm{0.085\textwidth}\cm{0.085\textwidth}\cm{0.085\textwidth}\cm{0.085\textwidth}}
                    \toprule
                     & DoS & AOO & PSA & FFA & AVOA & ETO & SGA & WAA \\
                    \midrule
                    $x_{1}$ & 0.001 & 0.001 & 0.001 & 0.001 & 0.001 & 0.00100193 & 0.001 & 0.001 \\
                    $x_{2}$ & 0.001 & 0.00100134 & 0.001 & 0.001 & 0.001 & 0.0012729 & 0.001 & 0.001 \\
                    $x_{3}$ & 0.001 & 0.00100008 & 0.001 & 0.001 & 0.001 & 0.00130519 & 0.001 & 0.0010957 \\
                    $x_{4}$ & 0.001 & 0.01079786 & 0.001 & 0.001 & 0.00100005 & 0.00153374 & 0.001571 & 0.01001128 \\
                    $x_{5}$ & 0.001 & 0.00100067 & 0.001 & 0.001 & 0.0010001 & 0.001 & 0.001 & 0.00659281 \\
                    $x_{6}$ & 0.001 & 0.00100002 & 0.001 & 0.001 & 0.001 & 0.001 & 0.00105786 & 0.00191787 \\
                    $x_{7}$ & 1.524 & 1.52435993 & 1.524 & 1.5240264 & 1.524 & 1.56881568 & 1.52403173 & 1.65669373 \\
                    $x_{8}$ & 1.524 & 1.53740159 & 1.524 & 1.52400052 & 1.524 & 1.55532936 & 1.52404964 & 1.5253391 \\
                    $x_{9}$ & 5 & 4.98194606 & 5 & 4.99999918 & 4.99999988 & 4.92988068 & 4.99998572 & 2.91251926 \\
                    $x_{10}$ & 2 & 2.03389883 & 2.00059666 & 2.00604538 & 2.00000143 & 2.09761448 & 2.12228525 & 3.77780929 \\
                    $x_{11}$ & 0.001 & 0.00100011 & 0.00100028 & 0.00100441 & 0.00100012 & 0.00161822 & 0.00100005 & 0.17997694 \\
                    $x_{12}$ & 0.001 & 0.001 & 0.00100028 & 0.0010044 & 0.001 & 0.0015924 & 0.001 & 0.08448447 \\
                    $x_{13}$ & 0.0072934 & 0.00731364 & 0.00729538 & 0.00731893 & 0.00729342 & 0.00897963 & 0.0075009 & 0.03468082 \\
                    $x_{14}$ & 0.08755583 & 0.08772952 & 0.08757953 & 0.08786226 & 0.08755602 & 0.10752741 & 0.0900468 & 0.39091399 \\
                    Best & \textbf{0.032213} & 0.03358755 & 0.03221638 & 0.03224857 & 0.03221349 & 0.03770509 & 0.03298267 & 0.28539794 \\
                    \bottomrule
                \end{tabular}
            \end{table*}

            \begin{table*}[htbp!]
                \centering
                \caption{Statistical results for the Optimal design of industrial refrigeration system.}
                \label{tab34}
                \begin{tabular}{\cm{0.08\textwidth}\cm{0.085\textwidth}\cm{0.085\textwidth}\cm{0.085\textwidth}\cm{0.085\textwidth}\cm{0.085\textwidth}\cm{0.085\textwidth}\cm{0.085\textwidth}\cm{0.085\textwidth}}
                    \toprule
                     & DoS & AOO & PSA & FFA & AVOA & ETO & SGA & WAA \\
                    \midrule
                    Mean & \textbf{0.032213} & 0.04270679 & 0.0651192 & 0.11484867 & 0.0403492 & 0.05902045 & 0.05023075 & 1.34361959 \\
                    Std & \textbf{1.018E-07} & 0.01463627 & 0.02767783 & 0.05306657 & 0.01459214 & 0.02002695 & 0.01265639 & 1.2266915 \\
                    Success & 0.64 & \textbf{1} & 0.72 & \textbf{1} & 0.68 & 0.88 & \textbf{1} & 0.84 \\
                    WSRT &  & 1.3E-09(+) & 1.3E-09(+) & 1.3E-09(+) & 1.2E-09(+) & 1.3E-09(+) & 1.3E-09(+) & 1.2E-09(+) \\
                    FMR & \textbf{1} & 3.12 & 5.2 & 6.52 & 3.08 & 4.92 & 4.16 & 8 \\
                    F-Rank & \textbf{1} & 3 & 6 & 7 & 2 & 5 & 4 & 8 \\
                    \bottomrule
                \end{tabular}
            \end{table*}

            Tables~\ref{tab33} and~\ref{tab34} summarize the experimental results of DoS and the competing algorithms on this problem. The results show that DoS achieved the best performance across all three key metrics Best, Mean and Std demonstrating excellent optimization accuracy and convergence stability. Specifically, the Mean value obtained by DoS is significantly better than that of AOO and AVOA and the Std is as low as 1.018E-07, indicating extremely low variance and highly stable convergence on feasible solutions.

            Although DoS has a slightly lower success rate than some competitors on this problem, in successful experiments, DoS has higher optimization accuracy than its competitors in other metrics, indicating that DoS still has good applicability to this problem.

            From a statistical significance perspective, the Wilcoxon rank-sum test indicates that DoS outperforms all competitors significantly. The Friedman test further confirms that DoS ranks first overall, with a margin of 2.08 over the second ranked AVOA, validating its global search capability and overall optimization effectiveness in solving this complex engineering problem.

            In summary, despite a slightly lower feasible solution rate, DoS exhibits outstanding performance in terms of optimization precision, stability and statistical superiority, clearly demonstrating its adaptability and engineering applicability in high-dimensional, constrained engineering optimization problems.

        \subsection{Step-Cone Pulley}

            Step-cone pulley is a typical mechanical structural optimization problem, aiming to minimize the total volume or weight of the pulley while satisfying multiple constraints such as transmission ratio, profile dimensions and material strength. This problem involves five continuous decision variables, denoted as $\bm{\mathrm{x}} = (d_{1}, d_{2}, d_{3}, d_{4}, \omega)$ and includes 8 inequality constraints and three equality constraints, which significantly increase the problem's complexity and pose higher demands on the algorithm's constraint handling capability and convergence precision. The complete mathematical formulation is provided in Equation~\ref{eq37}.

            Considering that equality constraints typically impose stricter requirements on both convergence speed and solution accuracy, the evaluation budget specified in the original Equation~\ref{eq30} is insufficient to ensure effective optimization. Therefore, a uniform evaluation budget of 100,000 is used in this experiment.

            \textbf{Minimize:}
            \begin{equation}
                \begin{aligned}
                    f(\bm{\mathrm{x}}) = \ &\rho \cdot \omega \Bigg\{ d_{1}^{2} \left[ 11 + \left(\frac{N_{1}}{N}\right)^{2} \right] + \\
                    &d_{2}^{2} \left[ 1 + \left(\frac{N_{2}}{N}\right)^{2} \right] + d_{3}^{2} \left[ 1 + \left(\frac{N_{3}}{N}\right)^{2} \right] + \\
                    &d_{4}^{2} \left[ 1 + \left(\frac{N_{4}}{N}\right)^{2} \right] \Bigg\}
                \end{aligned}
                \label{eq37}
            \end{equation}

            \textbf{Subject to:}
            \begin{equation*}
                h_{1}(\bm{\mathrm{x}}) = C_{1} - C_{2} = 0
            \end{equation*}
            \begin{equation*}
                h_{2}(\bm{\mathrm{x}}) = C_{1} - C_{3} = 0
            \end{equation*}
            \begin{equation*}
                h_{3}(\bm{\mathrm{x}}) = C_{1} - C_{4} = 0
            \end{equation*}
            \begin{equation*}
                g_{i}(\bm{\mathrm{x}}) = -R_{i} - 2 \leq 0, \quad i = 1,2,3,4
            \end{equation*}
            \begin{equation*}
                g_{i}(\bm{\mathrm{x}}) = 0.75 \times 745.6998 - P_{i} \leq 0, \quad i = 5,6,7,8
            \end{equation*}

            \textbf{Where:}
            \begin{equation*}
                \begin{aligned}
                    C_{i} = \ &\frac{\pi \cdot d_{i}}{2} \left( 1 + \frac{N_{i}}{N} \right) + \frac{ \left [ (N_{i}/{N}) - 1 \right ]^{2} }{4a} + \\
                    &2a, \quad i = 1,2,3,4
                \end{aligned}
            \end{equation*}
            \begin{equation*}
                \begin{aligned}
                    R_{i} = \ &\exp \Bigg\{ - 2 \mu \cdot \sin^{-1} \left[ \left( \frac{N_{i}}{N} - 1 \right) \frac{d_{i}}{2a} \right] + \\
                    &\pi \cdot \mu\Bigg\}, \quad i = 1,2,3,4
                \end{aligned}
            \end{equation*}
            \begin{equation*}
                P_{i} = s \cdot t \cdot \omega (1 - R_{i}) \frac{ \pi \cdot d_{i} \cdot N_{i} }{60}, \quad i = 1,2,3,4
            \end{equation*}
            \begin{equation*}
                t = 8 \ \mathrm{mm}, \quad s = 1.75 \ \mathrm{MPa}, \quad \mu = 0.35
            \end{equation*}
            \begin{equation*}
                \rho = 7200 \ \mathrm{kg/m}^{3}, \quad a = 3 \ \mathrm{mm}
            \end{equation*}

            \textbf{With bounds:}
            \begin{equation*}
                0 \leq d_{1}, d_{2} \leq 60, \quad 0 \leq d_{3}, d_{4}, \omega \leq 90
            \end{equation*}

            \begin{table*}[htbp!]
                \centering
                \caption{The best results for the Step-cone pulley.}
                \label{tab35}
                \begin{tabular}{\cm{0.08\textwidth}\cm{0.085\textwidth}\cm{0.085\textwidth}\cm{0.085\textwidth}\cm{0.085\textwidth}\cm{0.085\textwidth}\cm{0.085\textwidth}\cm{0.085\textwidth}\cm{0.085\textwidth}}
                    \toprule
                     & DoS & AOO & PSA & FFA & AVOA & ETO & SGA & WAA \\
                    \midrule
                    $d_{1}$ & 38.4139618 & NaN & 38.4415703 & 38.4160269 & 39.6685254 & NaN & NaN & NaN \\
                    $d_{2}$ & 52.8586378 & NaN & 52.8966562 & 52.8614815 & 54.5862806 & NaN & NaN & NaN \\
                    $d_{3}$ & 70.4726957 & NaN & 70.523379 & 70.4764868 & 72.775855 & NaN & NaN & NaN \\
                    $d_{4}$ & 84.4957161 & NaN & 84.5564332 & 84.5002576 & 87.2547612 & NaN & NaN & NaN \\
                    $\omega$ & 90 & NaN & 89.9351938 & 89.9951493 & 87.2375442 & NaN & NaN & NaN \\
                    Best & \textbf{16.090273} & NaN & 16.1018141 & 16.0911367 & 16.6320727 & NaN & NaN & NaN \\
                    \bottomrule
                \end{tabular}
            \end{table*}

            \begin{table*}[htbp!]
                \centering
                \caption{Statistical results for the Step-cone pulley.}
                \label{tab36}
                \begin{tabular}{\cm{0.08\textwidth}\cm{0.085\textwidth}\cm{0.085\textwidth}\cm{0.085\textwidth}\cm{0.085\textwidth}\cm{0.085\textwidth}\cm{0.085\textwidth}\cm{0.085\textwidth}\cm{0.085\textwidth}}
                    \toprule
                     & DoS & AOO & PSA & FFA & AVOA & ETO & SGA & WAA \\
                    \midrule
                    Mean & \textbf{16.090273} & NaN & 16.6411059 & 16.4156844 & 17.2273553 & NaN & NaN & NaN \\
                    Std & \textbf{6.405E-15} & NaN & 0.35976121 & 0.3277681 & 0.47618814 & NaN & NaN & NaN \\
                    Success & \textbf{1} & 0 & \textbf{1} & 0.52 & 0.84 & 0 & 0 & 0 \\
                    WSRT &  & NaN & 4.7E-10(+) & 3.5E-10(+) & 4.7E-10(+) & NaN & NaN & NaN \\
                    FMR & \textbf{1} & NaN & 2.84 & 2.36 & 3.8 & NaN & NaN & NaN \\
                    F-Rank & \textbf{1} & 5 & 3 & 2 & 4 & 6 & 7 & 8 \\
                    \bottomrule
                \end{tabular}
            \end{table*}

            Tables~\ref{tab35} and~\ref{tab36} present the optimization results of DoS and other competing algorithms on this problem. The results show that AOO, ETO, SGA and WAA failed to find any feasible solution across all 25 independent runs, indicating their inadequate adaptability in handling equality constrained problems. The success rates of FFA and AVOA were 0.52 and 0.84, respectively, also reflecting partial failure in achieving constraint satisfaction.

            Only DoS and PSA successfully converged to feasible solutions in all 25 runs, demonstrating robust performance. However, in terms of the three key metrics Best, Mean and Std DoS outperformed PSA across the board. Specifically, DoS achieved the best Best value, a lower Mean and an almost zero Std, indicating not only high stability across multiple runs, but also greater search accuracy. Although both algorithms converged to a similar solution domain, DoS yielded higher-quality feasible solutions, showcasing superior practical value.

            In the Wilcoxon rank-sum test, the WSRT indicate that DoS exhibited statistically significant performance advantages over all other algorithms. Additionally, the Friedman ranking places DoS at the top position, outperforming the second-best algorithm FFA by a margin of 1.36, further validating its superior performance and stability in tackling complex equality-constrained optimization problems.

        \subsection{Hydro-Static Thrust Bearing Design}

            Hydro-static Thrust Bearing Design is a classic problem in mechanical design optimization. The objective is to minimize the oil supply flow or total energy consumption of the bearing, while satisfying constraints related to operational performance, structural dimensions and energy consumption. This problem involves 4 continuous decision variables, which respectively represent the diameter of the supply orifice, the throttling parameter, the working oil film thickness and the bearing surface area, denoted as $\bm{\mathrm{x}} = (x_{1}, x_{2}, x_{3}, x_{4})$. The full model is shown in Equation~\ref{eq38}.

            \textbf{Minimize:}
            \begin{equation}
                f(\bm{\mathrm{x}}) = \frac{x_{4} \cdot P_{0}}{0.7} + E_{f}
                \label{eq38}
            \end{equation}

            \textbf{Subject to:}
            \begin{equation*}
                g_{1}(\bm{\mathrm{x}}) = 1000 - P_{0} \leq 0
            \end{equation*}
            \begin{equation*}
                g_{2}(\bm{\mathrm{x}}) = W - 101000 \leq 0
            \end{equation*}
            \begin{equation*}
                g_{3}(\bm{\mathrm{x}}) = 5000 - \frac{W}{\pi \left(x_{1}^{2} - x_{2}^{2}\right)} \leq 0
            \end{equation*}
            \begin{equation*}
                g_{4}(\bm{\mathrm{x}}) = 50 - P_{0} \leq 0
            \end{equation*}
            \begin{equation*}
                g_{5}(\bm{\mathrm{x}}) = 0.001 - \frac{0.0307}{386.4 P_{0}} \left( \frac{x_{4}}{2 \pi \cdot x_{1} \cdot h} \right) \leq 0
            \end{equation*}
            \begin{equation*}
                g_{6}(\bm{\mathrm{x}}) = x_{1} - x_{2} \leq 0
            \end{equation*}
            \begin{equation*}
                g_{7}(\bm{\mathrm{x}}) = h - 0.001 \leq 0
            \end{equation*}

            \textbf{Where:}
            \begin{equation*}
                W = \frac{\pi \cdot P_{0}}{2} \cdot \frac{x_{1}^{2} - x_{2}^{2}}{\ln\left(\frac{x_{1}}{x_{2}}\right)}\end{equation*}
            \begin{equation*}
                P_{0} = \frac{6 x_{3} \cdot x_{4}}{\pi \cdot h^{3}} \cdot \ln\left(\frac{x_{1}}{x_{2}}\right)
            \end{equation*}
            \begin{equation*}
                E_{f} = 9336 x_{4} \times 0.0307 \times 0.5 \Delta T
            \end{equation*}
            \begin{equation*}
                \Delta T = 2 \left(10^{P} - 559.7\right)
            \end{equation*}
            \begin{equation*}
                P = \frac{\mathrm{lg} \left[ \mathrm{lg} \left(8.122 \times 10^{6} x_{3} + 0.8\right) \right] + 3.55}{10.04}
            \end{equation*}
            \begin{equation*}
                h = \left( \frac{2 \pi \times 750}{60} \right)^{2} \cdot \frac{2 \pi \cdot x_{3}}{E_{f}} \cdot \left( \frac{x_{1}^{4}}{4} - \frac{x_{2}^{4}}{4} \right)
            \end{equation*}

            \textbf{With bounds:}
            \begin{equation*}
                1 \leq x_{1}, x_{2}, x_{4} \leq 16, \quad 10^{-6} \leq x_{3} \leq 1.6 \times 10^{7}
            \end{equation*}

            \begin{table*}[htbp!]
                \centering
                \caption{The best results for the Hydro-static thrust bearing design.}
                \label{tab37}
                \begin{tabular}{\cm{0.08\textwidth}\cm{0.088\textwidth}\cm{0.088\textwidth}\cm{0.088\textwidth}\cm{0.088\textwidth}\cm{0.088\textwidth}\cm{0.088\textwidth}\cm{0.088\textwidth}\cm{0.088\textwidth}}
                    \toprule
                     & DoS & AOO & PSA & FFA & AVOA & ETO & SGA & WAA \\
                    \midrule
                    $x_{1}$ & 5.95551185 & 5.98000801 & 6.4667959 & 6.26446032 & 6.0918833 & 6.16587283 & 7.20625039 & 6.37381799 \\
                    $x_{2}$ & 5.38871638 & 5.41558346 & 5.94891526 & 5.70944689 & 5.53900748 & 5.55478777 & 5.68618996 & 5.76494585 \\
                    $x_{3}$ & 5.3587E-06 & 6.4328E-06 & 8.2294E-06 & 5.9139E-06 & 6.6888E-06 & 5.5526E-06 & 7.5162E-06 & 7.9497E-06 \\
                    $x_{4}$ & 2.25664104 & 3.26750791 & 7.65466293 & 3.01829092 & 3.68679675 & 2.69046358 & 10.3889963 & 7.00843564 \\
                    Best & \textbf{1616.1204} & 1744.98267 & 2193.90348 & 1836.03564 & 1813.66823 & 1819.17843 & 3761.60047 & 2283.20929 \\
                    \bottomrule
                \end{tabular}
            \end{table*}

            \begin{table*}[htbp!]
                \centering
                \caption{Statistical results for the Hydro-static thrust bearing design.}
                \label{tab38}
                \begin{tabular}{\cm{0.08\textwidth}\cm{0.085\textwidth}\cm{0.085\textwidth}\cm{0.085\textwidth}\cm{0.085\textwidth}\cm{0.085\textwidth}\cm{0.085\textwidth}\cm{0.085\textwidth}\cm{0.085\textwidth}}
                    \toprule
                     & DoS & AOO & PSA & FFA & AVOA & ETO & SGA & WAA \\
                    \midrule
                    Mean & \textbf{1650.2441} & 2284.54068 & 2728.81527 & 2206.51479 & 2170.76997 & 2156.02939 & 5601.09312 & 5118.3029 \\
                    Std & \textbf{57.051917} & 369.796558 & 478.199976 & 290.012353 & 276.19162 & 314.054991 & 2556.66797 & 2376.17662 \\
                    Success & \textbf{1} & 0.96 & \textbf{1} & \textbf{1} & 0.96 & \textbf{1} & 0.2 & 0.44 \\
                    WSRT &  & 7.4E-09(+) & 1.4E-09(+) & 1.6E-09(+) & 2.3E-09(+) & 2.3E-09(+) & 3.9E-10(+) & 9.3E-10(+) \\
                    FMR & \textbf{1} & 4.12 & 5.36 & 3.72 & 3.52 & 3.36 & 7.72 & 7.2 \\
                    F-Rank & \textbf{1} & 5 & 6 & 4 & 3 & 2 & 8 & 7 \\
                    \bottomrule
                \end{tabular}
            \end{table*}

            Tables~\ref{tab37} and~\ref{tab38} summarize the experimental results of DoS and the competing algorithms on this problem. As shown, DoS outperforms all competitors across all performance indicators, demonstrating exceptional optimization capability and robustness. In terms of Best, DoS achieved the best solution of 1616.1204 for the entire game, with a significant gap of 203 in objective value compared to the second-best result by ETO. Regarding the Mean indicator, DoS also leads all competitors with an absolute advantage of 506. As for the Std, DoS achieves a remarkably low standard deviation of 57.051917, much lower than that of other competitors, indicating not only its ability to locate superior solution regions but also its stable convergence performance across multiple independent runs.

            Moreover, DoS achieves a Success rate of 1, meaning that all 25 independent experiments successfully converged to feasible solutions, further verifying its strong adaptability to the complex constraints of this problem. Although algorithms such as AOO, PSA, FFA and AVOA show a certain level of feasibility, their Mean values are generally higher and Std values fluctuate significantly, suggesting that these algorithms are still inferior to DoS in terms of stability and convergence quality.

            Wilcoxon rank-sum test results show that DoS's optimization performance on this problem is significantly better than its competitors. In the Friedman test, DoS once again ranks first and leads the second place algorithm ETO by a margin of 2.36, further demonstrating its stability and practical advantage in this type of mechanical design constrained optimization problem.

        \subsection{Four-Stage Gear Box}

            Four-stage gear box is a high-dimensional mechanical system optimization problem. The objective is to minimize the overall weight or cost of the gearbox under constraints such as transmission ratio, material strength and spatial dimensions. This problem involves 22 decision variables, covering parameters such as the radius and module of each gear stage. It is a typical high-dimensional constrained optimization problem and its mathematical model is detailed in Equation~\ref{eq39}.

            \textbf{Minimize:}
            \begin{equation}
                f(\bm{\mathrm{x}}) = \frac{\pi}{1000} \left[\sum_{i=1}^{4} \frac{b_{i} \cdot c_{i}^{2} \left( N_{pi}^{2} + N_{gi}^{2} \right)}{\left( N_{pi} + N_{gi} \right)^{2}} \right]
                \label{eq39}
            \end{equation}

            \textbf{Subject to:}
            \begin{equation*}
                \begin{aligned}
                    g_{1}(\bm{\mathrm{x}}) = \ &\frac{366000}{\pi \cdot \omega_{1}} \left[ \frac{ \left( N_{p1} + N_{g1} \right)^{2}}{4 b_{1} \cdot c_{1}^{2} \cdot N_{p1}} \right]  + \\
                    &\frac{2 c_{1} \cdot N_{p1}}{N_{p1} + N_{g1}} \left[ \frac{ \left( N_{p1} + N_{g1} \right)^{2}}{4 b_{1} \cdot c_{1}^{2} \cdot N_{p1}} \right] - \\
                    &\frac{ \sigma_{N} \cdot J_{R}}{0.0167 W \cdot K_{o} \cdot K_{m}} \leq 0
                \end{aligned}
            \end{equation*}
            \begin{equation*}
                \begin{aligned}
                    g_{2}(\bm{\mathrm{x}}) = \ &\frac{366000 N_{g1}}{\pi \cdot \omega_{1} \cdot N_{p1}}\left[ \frac{ \left( N_{p2} + N_{g2} \right)^{2}}{4 b_{2} \cdot c_{2}^{2} \cdot N_{p2}} \right] + \\
                    &\frac{2 c_{2} \cdot N_{p2}}{N_{p2} + N_{g2}} \left[ \frac{ \left( N_{p2} + N_{g2} \right)^{2}}{4 b_{2} \cdot c_{2}^{2} \cdot N_{p2}} \right] - \\
                    &\frac{\sigma_{N} \cdot J_{R}}{0.0167 W \cdot K_{o} \cdot K_{m}} \leq 0
                \end{aligned}
            \end{equation*}
            \begin{equation*}
                \begin{aligned}
                    g_{3}(\bm{\mathrm{x}}) = \ &\frac{366000 N_{g1} \cdot N_{g2}}{\pi \cdot \omega_{1} \cdot N_{p1} \cdot N_{p2}} \left[ \frac{ \left( N_{p3} + N_{g3} \right)^{2}}{4 b_{3} \cdot c_{3}^{2} \cdot N_{p3}} \right] + \\
                    &\frac{2 c_{3} \cdot N_{p3}}{N_{p3} + N_{g3}} \left[ \frac{ \left( N_{p3} + N_{g3} \right)^{2}}{4 b_{3} \cdot c_{3}^{2} \cdot N_{p3}} \right] - \\
                    &\frac{\sigma_{N} \cdot J_{R}}{0.0167 W \cdot K_{o} \cdot K_{m}} \leq 0
                \end{aligned}
            \end{equation*}
            \begin{equation*}
                \begin{aligned}
                    g_{4}(\bm{\mathrm{x}}) = \ &\frac{366000 N_{g1} \cdot N_{g2} \cdot N_{g3}}{\pi \cdot \omega_{1} \cdot N_{p1} \cdot N_{p2} \cdot N_{p3}} \left[ \frac{ \left( N_{p4} + N_{g4} \right)^{2}}{4 b_{4} \cdot c_{4}^{2} \cdot N_{p4}} \right] + \\
                    &\frac{2 c_{4} \cdot N_{p4}}{N_{p4} + N_{g4}} \left[ \frac{ \left( N_{p4} + N_{g4} \right)^{2}}{4 b_{4} \cdot c_{4}^{2} \cdot N_{p4}} \right] - \\
                    &\frac{\sigma_{N} \cdot J_{R}}{0.0167 W \cdot K_{o} \cdot K_{m}} \leq 0
                \end{aligned}
            \end{equation*}
            \begin{equation*}
                \begin{aligned}
                    g_{5}(\bm{\mathrm{x}}) = \ &\frac{366000}{\pi \cdot \omega_{1}} \left[ \frac{ \left( N_{p1} + N_{g1} \right)^{3}}{4 b_{1} \cdot c_{1}^{2} \cdot N_{g1} \cdot N_{p1}^{2}} \right] + \\
                    &\frac{2 c_{1} \cdot N_{p1}}{N_{p1} + N_{g1}} \left[ \frac{ \left( N_{p1} + N_{g1} \right)^{3}}{4 b_{1} \cdot c_{1}^{2} \cdot N_{g1} \cdot N_{p1}^{2}} \right] - \\
                    &\left( \frac{\sigma_{H}}{C_{p}} \right)^{2} \left[ \frac{\sin(\phi) \cdot \cos(\phi)}{0.0334 W \cdot K_{o} \cdot K_{m}} \right] \leq 0
                \end{aligned}
            \end{equation*}
            \begin{equation*}
                \begin{aligned}
                    g_{6}(\bm{\mathrm{x}}) = \ &\frac{366000 N_{g1}}{\pi \cdot \omega_{1} \cdot N_{p1}} \left[ \frac{ \left( N_{p2} + N_{g2} \right)^{3}}{4 b_{2} \cdot c_{2}^{2} \cdot N_{g2} \cdot N_{p2}^{2}} \right] + \\
                    &\frac{2 c_{2} \cdot N_{p2}}{N_{p2} + N_{g2}} \left[ \frac{ \left( N_{p2} + N_{g2} \right)^{3}}{4 b_{2} \cdot c_{2}^{2} \cdot N_{g2} \cdot N_{p2}^{2}} \right] - \\
                    &\left( \frac{\sigma_{H}}{C_{p}} \right)^{2} \left[ \frac{\sin(\phi) \cdot \cos(\phi)} {0.0334 W \cdot K_{o} \cdot K_{m}} \right] \leq 0
                \end{aligned}
            \end{equation*}
            \begin{equation*}
                \begin{aligned}
                    g_{7}(\bm{\mathrm{x}}) = \ &\frac{366000 N_{g1} \cdot N_{g2}}{\pi \cdot \omega_{1} \cdot N_{p1} \cdot N_{p2}} \left[ \frac{ \left( N_{p3} + N_{g3} \right)^{3}}{ 4 b_{3} \cdot c_{3}^{2} \cdot N_{g3} \cdot N_{p3}^{2}} \right] + \\
                    &\frac{2 c_{3} \cdot N_{p3}}{N_{p3} + N_{g3}} \left[ \frac{ \left( N_{p3} + N_{g3} \right)^{3}}{ 4 b_{3} \cdot c_{3}^{2} \cdot N_{g3} \cdot N_{p3}^{2}} \right] -\\
                    &\left( \frac{\sigma_{H}}{C_{p}} \right)^{2} \left[ \frac{\sin(\phi) \cdot \cos(\phi)}{0.0334 W \cdot K_{o} \cdot K_{m}} \right] \leq 0
                \end{aligned}
            \end{equation*}
            \begin{equation*}
                \begin{aligned}
                    g_{8}(\bm{\mathrm{x}}) = \ &\left(\frac{366000 N_{g1} \cdot N_{g2} \cdot N_{g3}}{\pi \cdot \omega_{1} \cdot N_{p1} \cdot N_{p2} \cdot N_{p3}} + \frac{2 c_{4} \cdot N_{p4}}{N_{p4} + N_{g4}}\right) \cdot \\
                    &\left[ \frac{ \left( N_{p4} + N_{g4} \right)^{3}}{4 b_{4} \cdot c_{4}^{2} \cdot N_{g4} \cdot N_{p4}^{2}} \right] - \\
                    &\left( \frac{\sigma_{H}}{C_{p}} \right)^{2} \left[ \frac{\sin(\phi) \cdot \cos(\phi)} {0.0334 W \cdot K_{o} \cdot K_{m}} \right] \leq 0
                \end{aligned}
            \end{equation*}
            \begin{equation*}
                \begin{aligned}
                    g_{9-12}(\bm{\mathrm{x}}) = \ &N_{gi} \sqrt{\frac{\sin^{2}(\phi)}{4} + \frac{1}{N_{gi}}  \cdot \left( \frac{1}{N_{gi}} \right)^{2}} - \\
                    &N_{pi} \sqrt{\frac{\sin^{2}(\phi)}{4} - \frac{1}{N_{pi}} + \left( \frac{1}{N_{pi}} \right)^{2}} + \\
                    &\frac{ \left( N_{pi} + N_{gi} \right) \sin(\phi)}{2} + \\
                    &CR_{min} \cdot \pi \cdot \cos(\phi) \leq 0, \quad i = 1, 2, 3, 4
                \end{aligned}
            \end{equation*}
            \begin{equation*}
                g_{13-16}(\bm{\mathrm{x}}) = d_{min} - \frac{2 c_{i} \cdot N_{pi}}{N_{pi} + N_{gi}} \leq 0, \quad i = 1, 2, 3, 4
            \end{equation*}
            \begin{equation*}
                g_{17-20}(\bm{\mathrm{x}}) = d_{min} - \frac{2 c_{i} \cdot N_{gi}}{N_{pi} + N_{gi}} \leq 0, \quad i = 1, 2, 3, 4
            \end{equation*}
            \begin{equation*}
                g_{21}(\bm{\mathrm{x}}) = x_{p1} + \frac{\left( N_{p1} + 2 \right) c_{1}}{N_{p1} + N_{g1}} - L_{max} \leq 0
            \end{equation*}
            \begin{equation*}
                \begin{aligned}
                    g_{22-24}(\bm{\mathrm{x}}) = \ &\frac{\left( N_{pi} + 2 \right) c_{i}}{N_{gi} + N_{pi}} - L_{max} + \\
                    &x_{g(i-1)} \leq 0, \quad i = 2, 3, 4
                \end{aligned}
            \end{equation*}
            \begin{equation*}
                g_{25}(\bm{\mathrm{x}}) = \frac{\left( N_{p1} + 2 \right) c_{1}}{N_{p1} + N_{g1}} - x_{p1} \leq 0
            \end{equation*}
            \begin{equation*}
                g_{26-28}(\bm{\mathrm{x}}) = \frac{(N_{pi} + 2) c_{i}}{N_{pi} + N_{gi}} - x_{g(i-1)} \leq 0, \quad i = 2, 3, 4
            \end{equation*}
            \begin{equation*}
                g_{29}(\bm{\mathrm{x}}) = y_{p1} + \frac{(N_{p1} + 2) c_{1}}{N_{p1} + N_{g1}} - L_{max} \leq 0
            \end{equation*}
            \begin{equation*}
                \begin{aligned}
                    g_{30-32}(\bm{\mathrm{x}}) = \ &\frac{(2 + N_{pi}) c_{i}}{N_{pi} + N_{gi}} - L_{max} + \\
                    &y_{g(i-1)} \leq 0, \quad i = 2, 3, 4
                \end{aligned}
            \end{equation*}
            \begin{equation*}
                g_{33}(\bm{\mathrm{x}}) = \frac{(2 + N_{p1}) c_{1}}{N_{p1} + N_{g1}} - y_{p1} \leq 0
            \end{equation*}
            \begin{equation*}
                g_{34-36}(\bm{\mathrm{x}}) = \frac{(2 + N_{pi}) c_{i}}{N_{pi} + N_{gi}} - y_{g(i-1)} \leq 0, \quad i = 2, 3, 4
            \end{equation*}
            \begin{equation*}
                \begin{aligned}
                    g_{37-40}(\bm{\mathrm{x}}) = \ &\frac{(2 + N_{gi}) c_{i}}{N_{pi} + N_{gi}} - L_{max} + \\
                    &x_{gi} \leq 0, \quad i = 1, 2, 3, 4
                \end{aligned}
            \end{equation*}
            \begin{equation*}
                g_{41-44}(\bm{\mathrm{x}}) = \frac{(N_{gi} + 2) c_{i}}{N_{pi} + N_{gi}} - x_{gi} \leq 0, \quad i = 1, 2, 3, 4
            \end{equation*}
            \begin{equation*}
                \begin{aligned}
                    g_{45-48}(\bm{\mathrm{x}}) = \ &\frac{(N_{gi} + 2) c_{i}}{N_{pi} + N_{gi}} - L_{max} + \\
                    &y_{gi} \leq 0, \quad i = 1, 2, 3, 4
                \end{aligned}
            \end{equation*}
            \begin{equation*}
                g_{49-52}(\bm{\mathrm{x}}) = \frac{(N_{gi} + 2) c_{i}}{N_{pi} + N_{gi}} - y_{gi} \leq 0, \quad i = 1, 2, 3, 4
            \end{equation*}
            \begin{equation*}
                \begin{aligned}
                    g_{53-56}(\bm{\mathrm{x}}) = \ &(N_{pi} + N_{gi}) (b_{i} - 8.255) \cdot \\
                    &(b_{i} - 5.715) (b_{i} - 12.7) - 0.945 c_{i} \cdot \\
                    &(b_{i} - 8.255) (b_{i} - 5.715) \cdot \\
                    &(b_{i} - 12.7) \leq 0, \quad i = 1, 2, 3, 4
                \end{aligned}
            \end{equation*}
            \begin{equation*}
                \begin{aligned}
                    g_{57-60}(\bm{\mathrm{x}}) = \ &(-N_{pi} - N_{gi}) (b_{i} - 8.255) \cdot \\
                    &(b_{i} - 3.175) (b_{i} - 12.7) + 0.646 c_{i} \cdot \\
                    &(b_{i} - 8.255) (b_{i} - 3.175) \cdot \\
                    &(b_{i} - 12.7) \leq 0, \quad i = 1, 2, 3, 4
                \end{aligned}
            \end{equation*}
            \begin{equation*}
                \begin{aligned}
                    g_{61-64}(\bm{\mathrm{x}}) = \ &(-N_{pi} - N_{gi}) (b_{i} - 5.715) \cdot \\
                    &(b_{i} - 3.175) (b_{i} - 12.7) + 0.504 c_{i} \cdot \\
                    &(b_{i} - 5.715) (b_{i} - 3.175) \cdot \\
                    &(b_{i} - 12.7) \leq 0, \quad i = 1, 2, 3, 4
                \end{aligned}
            \end{equation*}
            \begin{equation*}
                \begin{aligned}
                    g_{65-68}(\bm{\mathrm{x}}) = \ &(-N_{gi} - N_{pi}) (b_{i} - 5.715) \cdot \\
                    &(b_{i} - 3.175) (b_{i} - 8.255) \leq \\
                    &0, \quad i = 1, 2, 3, 4
                \end{aligned}
            \end{equation*}
            \begin{equation*}
                \begin{aligned}
                    g_{69-72}(\bm{\mathrm{x}}) = \ &(-N_{gi} - N_{pi}) (b_{i} - 8.255) \cdot \\
                    &(b_{i} - 5.715) (b_{i} - 12.7) + 1.812 c_{i} \cdot \\
                    &(b_{i} - 8.255) (b_{i} - 5.715) \cdot \\
                    &(b_{i} - 12.7) \leq 0, \quad i = 1, 2, 3, 4
                \end{aligned}
            \end{equation*}
            \begin{equation*}
                \begin{aligned}
                    g_{73-76}(\bm{\mathrm{x}}) = \ &(N_{pi} + N_{gi}) (b_{i} - 8.255) \cdot \\
                    &(b_{i} - 3.175) (b_{i} - 12.7) - 0.945 c_{i} \cdot \\
                    &(b_{i} - 8.255) (b_{i} - 3.175) \cdot \\
                    &(b_{i} - 12.7) \leq 0, \quad i = 1, 2, 3, 4
                \end{aligned}
            \end{equation*}
            \begin{equation*}
                \begin{aligned}
                    g_{77-80}(\bm{\mathrm{x}}) = \ &(-N_{pi} - N_{gi}) (b_{i} - 5.715) \cdot \\
                    &(b_{i} - 3.175) (b_{i} - 12.7) + 0.646 c_{i} \cdot \\
                    &(b_{i} - 5.715) (b_{i} - 3.175) \cdot \\
                    &(b_{i} - 12.7) \leq 0, \quad i = 1, 2, 3, 4
                \end{aligned}
            \end{equation*}
            \begin{equation*}
                \begin{aligned}
                    g_{81-84}(\bm{\mathrm{x}}) = \ &(N_{pi} + N_{gi}) (b_{i} - 5.715) \cdot \\
                    &(b_{i} - 3.175) (b_{i} - 8.255) - 0.504 c_{i} \cdot \\
                    &(b_{i} - 5.715) (b_{i} - 3.175) \cdot \\
                    &(b_{i} - 8.255) \leq 0, \quad i = 1, 2, 3, 4
                \end{aligned}
            \end{equation*}
            \begin{equation*}
                g_{85}(\bm{\mathrm{x}}) = \omega_{min} - \frac{\omega_{1} \cdot N_{p1} \cdot N_{p2} \cdot N_{p3} \cdot N_{p4}}{N_{g1} \cdot N_{g2} \cdot N_{g3} \cdot N_{g4}} \leq 0
            \end{equation*}
            \begin{equation*}
                g_{86}(\bm{\mathrm{x}}) = \frac{\omega_{1} \cdot N_{p1} \cdot N_{p2} \cdot N_{p3} \cdot N_{p4}}{N_{g1} \cdot N_{g2} \cdot N_{g3} \cdot N_{g4}} - \omega_{max} \leq 0
            \end{equation*}

            \textbf{Where:}
            \begin{equation*}
                c_{i} = \sqrt{(y_{gi} - y_{pi})^{2} + (x_{gi} - x_{pi})^{2}}
            \end{equation*}
            \begin{equation*}
                K_{0} = 1.5, \quad d_{min} = 25, \quad J_{R} = 0.2, \quad \phi = 120^{\circ}
            \end{equation*}
            \begin{equation*}
                W = 55.9, \quad K_{M} = 1.6, \quad CR_{min} = 1.4
            \end{equation*}
            \begin{equation*}
                L_{max} = 127C_{p} = 464, \quad \sigma_{H} = 3290, \quad \omega_{max} = 255
            \end{equation*}
            \begin{equation*}
                \omega_{1} = 5000, \quad \sigma_{N} = 2090, \quad \omega_{min} = 245
            \end{equation*}

            \textbf{With bounds:}
            \begin{equation*}
                b_{i} \in \{3.175, 12.7, 8.255, 5.715\},
            \end{equation*}
            \begin{equation*}
                N_{gi}, N_{pi} \in \{7, 8, \ldots, 76\},
            \end{equation*}
            \begin{equation*}
                \begin{aligned}
                    x_{p1}, y_{p1}, x_{gi}, y_{gi} \in \{&12.7, 38.1, 25.4, 50.8, 76.2, \\
                &63.5, 88.9, 114.3, 101.6\}
                \end{aligned}
            \end{equation*}
            
            \begin{table*}[htbp!]
                \centering
                \caption{The best results for the Four-stage gear box.}
                \label{tab39}
                \begin{tabular}{\cm{0.08\textwidth}\cm{0.085\textwidth}\cm{0.085\textwidth}\cm{0.085\textwidth}\cm{0.085\textwidth}\cm{0.085\textwidth}\cm{0.085\textwidth}\cm{0.085\textwidth}\cm{0.085\textwidth}}
                    \toprule
                     & DoS & AOO & PSA & FFA & AVOA & ETO & SGA & WAA \\
                    \midrule
                    $N_{p1}$ & 21.5320385 & 14.4365735 & 16.3522535 & 27.6467997 & 14.2885826 & 28.4713456 & 14.541271 & 24.1665764 \\
                    $N_{g1}$ & 46.6205336 & 31.0391991 & 38.1855094 & 40.6005345 & 28.4852633 & 40.2217087 & 45.1645854 & 63.3848144 \\
                    $N_{p2}$ & 21.0587073 & 18.453822 & 21.4922182 & 23.1810638 & 17.9460929 & 14.009929 & 15.2618438 & 14.2695344 \\
                    $N_{g2}$ & 42.4589623 & 40.109072 & 38.4061982 & 49.1868321 & 41.1803625 & 35.8717594 & 19.7370829 & 31.0863413 \\
                    $N_{p3}$ & 21.4872325 & 21.2135998 & 10.7031089 & 18.2407005 & 18.6306136 & 21.4574928 & 15.2525391 & 16.5003483 \\
                    $N_{g3}$ & 45.3812778 & 42.3807148 & 23.4193648 & 46.304301 & 40.6289769 & 40.5799586 & 30.8480731 & 41.069902 \\
                    $N_{p4}$ & 20.9287857 & 13.5623083 & 21.2845599 & 18.1278907 & 19.0777106 & 16.1181924 & 17.013716 & 22.2135821 \\
                    $N_{g4}$ & 45.4514195 & 28.1343203 & 46.3987004 & 45.1736973 & 38.2164589 & 44.1485892 & 40.6262691 & 31.1860917 \\
                    $b_{1}$ & 0.69418949 & 1.38169063 & 0.9020677 & 0.69946145 & 0.91685304 & 0.94970777 & 1.25089483 & 0.71725398 \\
                    $b_{2}$ & 0.87511381 & 1.24663426 & 1.08084847 & 1.07914246 & 0.6267313 & 1.07177167 & 2.32819288 & 0.64327457 \\
                    $b_{3}$ & 0.75252488 & 1.44831389 & 1.70107568 & 0.61402979 & 0.93292115 & 1.36342825 & 1.10315959 & 0.51 \\
                    $b_{4}$ & 1.40641289 & 0.74910218 & 0.5587868 & 1.22326827 & 0.77502445 & 1.04994095 & 1.06786147 & 0.76960197 \\
                    $x_{p1}$ & 1.71829571 & 3.05314121 & 2.59790079 & 2.81997842 & 3.57116076 & 2.38254436 & 6.2343942 & 4.44054981 \\
                    $x_{g1}$ & 4.81044719 & 4.8875494 & 6.48071995 & 5.82510208 & 3.59341973 & 1.91232332 & 5.51931233 & 3.94956775 \\
                    $x_{g2}$ & 4.89132384 & 4.53238538 & 5.15785114 & 6.45530085 & 5.5288688 & 3.53161928 & 5.51279197 & 5.71913391 \\
                    $x_{g3}$ & 3.17502923 & 6.42118539 & 4.33422852 & 4.60217008 & 5.31358539 & 4.77717887 & 4.99913312 & 4.15664445 \\
                    $x_{g4}$ & 4.93962556 & 5.57983126 & 2.52752637 & 5.6855937 & 3.62376827 & 5.56627464 & 3.79873123 & 5.92031571 \\
                    $y_{p1}$ & 5.96903726 & 2.62620248 & 2.87021188 & 2.81976882 & 7.02282306 & 5.87311718 & 2.39255506 & 8.46009973 \\
                    $y_{g1}$ & 5.47950026 & 5.98449881 & 4.95026566 & 4.5288527 & 4.45671595 & 2.853717 & 5.62743707 & 4.13487523 \\
                    $y_{g2}$ & 5.63879698 & 6.22093695 & 5.18254811 & 3.55437704 & 4.36812185 & 2.72434446 & 5.31190738 & 5.19498598 \\
                    $y_{g3}$ & 3.27357283 & 2.84150747 & 6.27275755 & 5.60450977 & 4.34024296 & 3.79176373 & 4.86560138 & 3.87938439 \\
                    $y_{g4}$ & 5.48673809 & 2.84181107 & 6.80551646 & 4.55627422 & 4.01089936 & 5.029828 & 5.40524818 & 6.37188674 \\
                    Best & \textbf{35.359232} & 39.9992606 & 50.1074146 & 44.7579233 & 37.2597421 & 48.144882 & 50.6500247 & 49.1011141 \\
                    \bottomrule
                \end{tabular}
            \end{table*}

            \begin{table*}[htbp!]
                \centering
                \caption{Statistical results for the Four-stage gear box.}
                \label{tab40}
                \begin{tabular}{\cm{0.08\textwidth}\cm{0.085\textwidth}\cm{0.085\textwidth}\cm{0.085\textwidth}\cm{0.085\textwidth}\cm{0.085\textwidth}\cm{0.085\textwidth}\cm{0.085\textwidth}\cm{0.085\textwidth}}
                    \toprule
                     & DoS & AOO & PSA & FFA & AVOA & ETO & SGA & WAA \\
                    \midrule
                    Mean & \textbf{42.973019} & 62.8111705 & 45.9896535 & 60.0108344 & 47.3720259 & 75.0605268 & 93.1821241 & 51.349004 \\
                    Std & 7.65659425 & 20.7589046 & \textbf{0} & 13.570429 & 5.49153005 & \textbf{0} & \textbf{0} & 8.21682197 \\
                    Success & 0.84 & 0.4 & 0.04 & \textbf{0.92} & 0.2 & 0.04 & 0.04 & 0.24 \\
                    WSRT &  & 2.6E-08(+) & 1.1E-05(+) & 4.5E-07(+) & 6.0E-05(+) & 9.6E-11(+) & 9.6E-11(+) & 5.3E-06(+) \\
                    FMR & \textbf{1.56} & 5.36 & 2.08 & 5.04 & 3.16 & 6.76 & 7.92 & 4.12 \\
                    F-Rank & \textbf{1} & 6 & 2 & 5 & 3 & 7 & 8 & 4 \\
                    \bottomrule
                \end{tabular}
            \end{table*}

            Tables~\ref{tab39} and~\ref{tab40} summarize the experimental results of DoS and the competing algorithms on this problem. The problem includes 86 inequality constraints, imposing high demands on the algorithm’s feasibility maintaining ability and convergence accuracy. As shown by the Success indicator, none of the algorithms achieved a 100\% success rate. Among them, PSA, ETO and SGA each obtained only one feasible solution out of 25 runs, indicating poor stability and adaptability. Although FFA achieved a 92\% success rate, slightly higher than the 84\% of DoS, DoS still outperformed all competitors in both Best and Mean metrics, demonstrating that it not only finds more feasible solutions but also produces higher quality solutions.

            In terms of result dispersion, the Std of DoS was slightly higher than that of AVOA, but significantly lower than the other algorithms, indicating that DoS possesses strong local exploitation capability around feasible regions. The Wilcoxon rank-sum test further confirms that the performance advantages of DoS on this problem are statistically significant. Finally, DoS ranked first in the Friedman test, highlighting its optimization performance and practical value in high dimensional, single objective constrained optimization problems.

        \subsection{Wind Farm Layout}

            Wind farm layout is a renewable energy optimization problem characterized by spatial coupling. The objective is to maximize the total power output of the wind farm or minimize the levelized cost of energy, under environmental constraints such as geographical layout and wind speed distribution. The problem involves 30 decision variables, representing the positions and spacing design of wind turbines, denoted as $\bm{\mathrm{x}} = (x_{1}, \cdots , x_{15}, y_{1}, \cdots , y_{15})$. It includes 91 inequality constraints, making it a typical high dimensional, complex constrained optimization problem. The mathematical model is defined in Equation~\ref{eq40}.

            \textbf{Minimize:}
            \begin{equation}
                f(\bm{\mathrm{x}}) = \sum_{i=1}^{N} E(P_i)
                \label{eq40}
            \end{equation}

            \textbf{Subject to:}
            \begin{equation*}
                \begin{aligned}
                    g_{\frac{(28-i)(i-1)}{2} + (j-i)}(\bm{\mathrm{x}}) &= \sqrt{(x_i - x_j)^2 + (y_i - y_j)^2} \geq 5R, \\
                    i &= 1,2,\ldots,15, \quad i < j < 14
                \end{aligned}
            \end{equation*}

            \textbf{Where:}
            \begin{equation*}
                \begin{aligned}
                    E(P_i) = \ &\sum_{n=1}^{h} \xi_n \cdot P_r \Bigg( e^{- \left \{ \frac{v_r}{c_i' \left[ (\theta_{n-1} + \theta_n)/2 \right]} \right \} ^{k_i(\theta_{n-1} + \theta_n)/2}} - \\
                    &e^{-\left \{ \frac{ v_{co}}{c_i' \left[ (\theta_{n-1} + \theta_n)/2 \right]} \right \} ^{k_i(\theta_{n-1} + \theta_n)/2}} \Bigg) + \sum_{n=1}^{h} \xi_{n} \cdot \\
                    &\Bigg\{ \sum_{j=1}^{s} \Bigg[ \Bigg( e^{ - \left \{\frac{v_{j-1}}{c_i' \left[(\theta_{n-1} + \theta_n)/2 \right]} \right \} ^{k_i(\theta_{n-1} + \theta_n)/2}} - \\
                    &e^{-\left \{\frac{v_j}{c_i' \left[ (\theta_{n-1} + \theta_n)/2 \right]} \right \} ^{k_i(\theta_{n-1} + \theta_n)/2}} \Bigg) \cdot \\
                    &\frac{e^{(v_{j-1} + v_j)/2}}{\alpha + \beta \cdot e^{(v_{j-1} + v_j)/2}} \Bigg] \Bigg\}
                \end{aligned}
            \end{equation*}

            \textbf{With bounds:}
            \begin{equation*}
                40 \leq x_i \leq 1960, \quad 40 \leq y_i \leq 1960, \quad i,j = 1, 2, \ldots,15
            \end{equation*}

            \begin{table*}[htbp!]
                \centering
                \caption{The best results for the Wind farm layout.}
                \label{tab41}
                \begin{tabular}{\cm{0.08\textwidth}\cm{0.085\textwidth}\cm{0.085\textwidth}\cm{0.085\textwidth}\cm{0.085\textwidth}\cm{0.085\textwidth}\cm{0.085\textwidth}\cm{0.085\textwidth}\cm{0.085\textwidth}}
                    \toprule
                     & DoS & AOO & PSA & FFA & AVOA & ETO & SGA & WAA \\
                    \midrule
                    $x_{1}$ & 1462.63623 & 686.727895 & 40.0266304 & 40 & 1144.00883 & NaN & 1077.02246 & 569.828992 \\
                    $x_{2}$ & 788.942125 & 522.561888 & 1959.9987 & 40 & 1392.90003 & NaN & 1704.95946 & 1377.66413 \\
                    $x_{3}$ & 1547.23036 & 945.95155 & 1841.77117 & 1959.68289 & 147.650041 & NaN & 1555.44313 & 1859.44004 \\
                    $x_{4}$ & 1960 & 1329.53655 & 40.000101 & 46.0147221 & 230.504712 & NaN & 86.8739308 & 1954.81841 \\
                    $x_{5}$ & 272.001925 & 1922.48364 & 40.0000206 & 300.774847 & 40.0000368 & NaN & 316.67443 & 1018.64235 \\
                    $x_{6}$ & 842.392038 & 1324.08245 & 40.0576999 & 490.453593 & 41.1868254 & NaN & 634.290145 & 1727.97915 \\
                    $x_{7}$ & 1390.37725 & 1908.0378 & 884.311937 & 1340.8296 & 863.90748 & NaN & 1802.51133 & 161.93923 \\
                    $x_{8}$ & 1206.13361 & 776.749987 & 1959.99999 & 1050.91353 & 1049.11352 & NaN & 549.996432 & 1833.87805 \\
                    $x_{9}$ & 204.925464 & 40.0010407 & 1854.60593 & 1482.3778 & 263.304927 & NaN & 40.0002408 & 700.456074 \\
                    $x_{10}$ & 1166.54878 & 47.0173447 & 940.869654 & 1667.20654 & 432.259573 & NaN & 40.0000978 & 277.850598 \\
                    $x_{11}$ & 1880.12588 & 1959.94023 & 61.1963451 & 1032.63369 & 871.317476 & NaN & 40 & 1172.18597 \\
                    $x_{12}$ & 40 & 1773.72515 & 312.665753 & 41.2434693 & 40.0002229 & NaN & 1028.90997 & 1953.05298 \\
                    $x_{13}$ & 1450.03348 & 1018.67944 & 80.1754666 & 548.781031 & 40 & NaN & 1935.82257 & 1959.5862 \\
                    $x_{14}$ & 1568.89917 & 51.7275145 & 1530.50398 & 833.668985 & 835.664244 & NaN & 1493.48728 & 766.354613 \\
                    $x_{15}$ & 767.013635 & 77.4062641 & 1477.98049 & 174.166882 & 1960 & NaN & 1960 & 79.9992837 \\
                    $x_{16}$ & 1960 & 1352.85886 & 1619.74265 & 1959.51005 & 1960 & NaN & 40.227523 & 890.87749 \\
                    $x_{17}$ & 40 & 1282.64443 & 1938.97 & 48.2132852 & 40.3312575 & NaN & 1075.44001 & 1271.45673 \\
                    $x_{18}$ & 40 & 1956.8046 & 1417.09561 & 1332.7715 & 1051.02581 & NaN & 1960 & 1177.37466 \\
                    $x_{19}$ & 75.1699377 & 1908.16395 & 1004.07086 & 1261.23706 & 1054.3472 & NaN & 1959.9974 & 144.458628 \\
                    $x_{20}$ & 1960 & 991.344764 & 40.0013151 & 1318.91042 & 1959.38814 & NaN & 1698.13098 & 1348.60015 \\
                    $x_{21}$ & 1831.25155 & 1812.41881 & 1030.56244 & 49.9481137 & 1960 & NaN & 42.6953317 & 441.650516 \\
                    $x_{22}$ & 386.167552 & 516.356072 & 921.338735 & 1005.27441 & 1404.10714 & NaN & 1960 & 395.806647 \\
                    $x_{23}$ & 88.6507558 & 101.379005 & 1876.83318 & 1101.3501 & 40.0000698 & NaN & 1960 & 670.118121 \\
                    $x_{24}$ & 1554.62265 & 1903.4665 & 414.691574 & 1958.34857 & 1395.69065 & NaN & 1960 & 833.40477 \\
                    $x_{25}$ & 710.239709 & 1750.42917 & 98.5915384 & 257.201323 & 1959.99999 & NaN & 40.3455457 & 624.67494 \\
                    $x_{26}$ & 40 & 57.5901493 & 888.69544 & 1623.99716 & 52.7654481 & NaN & 417.095965 & 1872.00174 \\
                    $x_{27}$ & 123.208964 & 130.895825 & 1960 & 1931.81424 & 210.534434 & NaN & 1960 & 920.743972 \\
                    $x_{28}$ & 377.04397 & 876.670144 & 1959.99963 & 1960 & 1960 & NaN & 393.128088 & 91.6274283 \\
                    $x_{29}$ & 40 & 1956.62571 & 61.786189 & 1346.83705 & 1960 & NaN & 1960 & 368.043789 \\
                    $x_{30}$ & 40 & 1955.55095 & 528.352934 & 773.437113 & 1960 & NaN & 1960 & 661.56747 \\
                    Best & \textbf{-6262.972} & -5986.9772 & -6084.7437 & -5815.6126 & -6043.438 & NaN & -5903.1683 & -5247.7343 \\
                    \bottomrule
                \end{tabular}
            \end{table*}

            \begin{table*}[htbp!]
                \centering
                \caption{Statistical results for the Wind farm layout.}
                \label{tab42}
                \begin{tabular}{\cm{0.08\textwidth}\cm{0.085\textwidth}\cm{0.085\textwidth}\cm{0.085\textwidth}\cm{0.085\textwidth}\cm{0.085\textwidth}\cm{0.085\textwidth}\cm{0.085\textwidth}\cm{0.085\textwidth}}
                    \toprule
                     & DoS & AOO & PSA & FFA & AVOA & ETO & SGA & WAA \\
                    \midrule
                    Mean & \textbf{-6205.214} & -5806.3767 & -5913.9266 & -5496.6945 & -5747.3537 & NaN & -5546.75 & -4913.6592 \\
                    Std & \textbf{30.581606} & 125.853091 & 86.1246327 & 76.656197 & 135.130191 & NaN & 193.119997 & 239.132835 \\
                    Success & \textbf{1} & \textbf{1} & \textbf{1} & \textbf{1} & \textbf{1} & 0 & 0.88 & \textbf{1} \\
                    WSRT &  & 1.4E-09(+) & 1.4E-09(+) & 1.4E-09(+) & 1.4E-09(+) & NaN & 1.4E-09(+) & 1.4E-09(+) \\
                    FMR & \textbf{1} & 3.16 & 2.4 & 5.4 & 3.92 & NaN & 5.16 & 6.96 \\
                    F-Rank & \textbf{1} & 3 & 2 & 6 & 4 & 8 & 5 & 7 \\
                    \bottomrule
                \end{tabular}
            \end{table*}

            Tables~\ref{tab41} and~\ref{tab42} present the experimental results of DoS and its competitors on this problem. Except for ETO, which failed to converge to any feasible solution in all 25 independent runs, the success rates of the other algorithms are relatively high. However, in terms of the Best and Mean indicators, DoS achieves the best performance, indicating stronger search capabilities and a higher likelihood of approaching the global optimal solution. DoS also significantly outperforms competitors in terms of the Std indicator, demonstrating high convergence accuracy and good stability.

            The results of the Wilcoxon rank-sum test show that DoS is statistically significantly better than all other competitors. In the Friedman test, DoS also ranks first. In summary, DoS demonstrates outstanding performance and practical value in high dimensional constrained optimization problems and Power System related tasks.
 
    \section{Application to Mountainous Terrain Path Planning}

        In unmanned aerial vehicle (UAV) power inspection tasks, the planning of flight paths is directly related to the operational safety of transmission lines. Therefore, developing a scientifically sound and reasonable flight path for UAVs \citep{hrabar20083d, ergezer20143d, yang2014literature} is key to ensuring task safety and operational efficiency. This problem involves multiple complex discontinuous constraints [73], making it highly significant in engineering applications. To verify the adaptability of DoS in real-world engineering problems, this study uses this problem as a representative case for evaluation.

        Section 6.1 presents the mathematical model of the mountainous terrain path planning problem. In Section 6.2, DoS and advanced competitors are evaluated through 25 independent experiments under typical mountainous terrain. The performance of DoS on this problem is comprehensively analyzed using statistical indicators, including the best value (Best), average value (Mean), standard deviation (Std), success rate (Success), Wilcoxon rank-sum test (WSRT) at a 5\% significance level, Friedman mean rank (FMR) and overall rank (F-Rank).

        \subsection{Mathematical Model}

            This problem aims to plan a safe and efficient flight path for UAVs based on mountainous terrain, starting point and destination. In the mathematical model, the flight path is discretized into 7 nodes and the variations in the X-axis, Y-axis and relative terrain height between adjacent nodes are defined as the optimization decision variables. Accordingly, the coordinates of the i-th node are expressed as:
            \begin{equation}
                \begin{aligned}
                    x_i &= \mathrm{round} \left( 0.5 + x_s + \sum_{j=1}^{i} \Delta x_j \right), \\
                    y_i &= \mathrm{round} \left( 0.5 + y_s + \sum_{j=1}^{i} \Delta y_j \right), \\
                    z_i &= H_i + \Delta z_i
                \end{aligned}
                \label{eq41}
            \end{equation}

            \noindent
            where $\Delta x_j$ and $\Delta y_j$ denote the variations in the X-axis and Y-axis directions of the j-th node relative to its previous node and $\Delta z_i$ is the relative change in terrain height. Hi represents the terrain elevation at coordinates $\left(x_{i}, y_{i}\right)$. Thus, the decision variables can be represented as: $\bm{\mathrm{x}} = (\Delta x_{1}, \Delta y_{1}, \Delta x_{2}, \ldots, \Delta x_{7}, \Delta y_{7}, \Delta z_{7})$.

            Considering that using only seven nodes is insufficient to generate a smooth trajectory in complex terrain, a cubic spline interpolation is applied to smooth the flight path after determining all node coordinates. Let the number of interpolated trajectory points be m, with coordinates denoted as $X_{1}, X_{2}, \cdots , X_{m}$, where each $X_{i} = \left( \tilde{x}_i, \tilde{y}_i, \tilde{z}_i \right)$.

            The optimization objectives for the UAV flight path primarily include path length minimization and altitude smoothness, which are formulated as follows:
            \begin{equation}
                F_{1}(\bm{\mathrm{x}}) = \sum_{i=1}^{m-1} \left\| X_{i+1} - X_{i} \right\|
                \label{eq42}
            \end{equation}
            \begin{equation}
                F_{2}(\bm{\mathrm{x}}) = \sqrt{\frac{1}{m} \sum_{i=1}^{m} \left( \tilde{z}_{i} - z_{mean} \right)^{2}}
                \label{eq43}
            \end{equation}

            \noindent
            where $z_{mean}$ represents the average flight altitude of all trajectory points. In addition, the mountainous terrain path planning problem involves four types of constraints:

            Turning angle constraint: Limits the angle between adjacent trajectory segments from being smaller than a critical value $\psi$, in order to avoid sharp turns:
            \begin{equation}
                \begin{aligned}
                    G_{1}(\bm{\mathrm{x}}) = \ &\arccos \left( \frac{A_i \cdot B_i}{\left\|A_i\right\| \left\|B_i\right\|} \right) - \\
                    &\psi \leq 0, \quad i = 2, 3, \ldots ,m-1
                \end{aligned}
                \label{eq44}
            \end{equation}

            \noindent
            where $A_{i}$ and $B_{i}$ represent the direction vectors formed between the $i$-th trajectory point and the $(i-1)$-th and $(i+1)$-th trajectory points, respectively. The flight turning angle can be calculated by combining the cosine law with the inverse trigonometric function.
            \begin{equation*}
                A_{i} = X_{i} - X_{i - 1}, \quad B_{i} = X_{i + 1} - X_{i}
            \end{equation*}

            Collision avoidance constraint: Trajectory points must maintain sufficient clearance above the terrain:
            \begin{equation}
                \begin{aligned}
                    G_{2}(\bm{\mathrm{x}}) &= \ \tilde{z}_{(\tilde{x}_i,\tilde{y}_i)} + \Delta h_{\min} -
                    \tilde{z}_i \leq 0, \\ i &= 1, 2, \ldots ,m
                \end{aligned}
                \label{eq45}
            \end{equation}

            \noindent
            where $\tilde{z}_{(\tilde{x}_i,\tilde{y}_i)}$ denotes the interpolated terrain height at point $(\tilde{x}_i,\tilde{y}_i)$ based on the terrain results and $\Delta h_{\min}$ represents the minimum safety distance required to avoid terrain collision.

            Since the trajectory points are generated via interpolation from the discrete nodes, there is a possibility that some points may fall outside the boundary. To address this, the following boundary constraint is established:
            \begin{equation}
                G_{3}(\bm{\mathrm{x}}) =
                \begin{cases}
                    M_{L} - \tilde{x}_i \leq 0 \\
                    \tilde{x}_i - M_{U} \leq 0 \\
                    M_{L} - \tilde{y}_i \leq 0 \\
                    \tilde{y}_i - M_{U} \leq 0
                \end{cases}, \quad i = 1, 2, \ldots,m
                \label{eq46}
            \end{equation}

            \noindent
            where $M_{L}$ and $M_{U}$ represent the lower and upper bounds of the ground area respectively.

            Since the UAV’s flight altitude is also a critical factor affecting safety, the following altitude constraint is introduced in the mathematical model:
            \begin{equation}
                G_{4}(\bm{\mathrm{x}}) = \tilde{z}_i - h_{\max} \leq 0, \quad i = 1, 2, \ldots ,m
                \label{eq47}
            \end{equation}

            \noindent
            where $h_{max}$ denotes the maximum allowable flight altitude for the UAV.

            Since optimizing a single objective before verifying constraints significantly reduces the extraction rate of feasible solutions, we also use the penalty function method to handle constraints in this problem. That is, for solutions that violate constraints, their objective function values are adjusted by adding penalty terms. The corresponding auxiliary function is defined as follows:
            \begin{equation}
                Q\!\left(G_s(\bm{\mathrm{x}})\right) =
                \begin{cases}
                    0, & G_s(\bm{\mathrm{x}}) \le 0 \\
                    10000, & G_s(\bm{\mathrm{x}}) > 0
                \end{cases}, \quad s = 1,2,3,4
                \label{eq48}
            \end{equation}

            Finally, the optimization objective of the mountainous terrain path planning problem is formulated as:
            \begin{equation}
                \begin{aligned}
                    F = \ &F_{1} + F_{2} + Q\left[G_{1}(\bm{\mathrm{x}})\right] + Q\left[G_{2}(\bm{\mathrm{x}})\right] + \\
                    &Q\left[G_{3}(\bm{\mathrm{x}})\right] + Q\left[G_{4}(\bm{\mathrm{x}})\right]
                \end{aligned}
                \label{eq49}
            \end{equation}

        \subsection{Performance Analysis}

            Based on multiple preliminary experiments, the evaluation budget for each algorithm in this section is set to 10,000 function evaluations and 25 independent runs are conducted to assess algorithm performance. The best solution results of each algorithm are shown in Table~\ref{tab43} and the statistical indicators are presented in Table~\ref{tab44}.

            \begin{table*}[htbp!]
                \centering
                \caption{The best results for mountainous terrain path planning.}
                \label{tab43}
                \begin{tabular}{\cm{0.08\textwidth}\cm{0.085\textwidth}\cm{0.085\textwidth}\cm{0.085\textwidth}\cm{0.085\textwidth}\cm{0.085\textwidth}\cm{0.085\textwidth}\cm{0.085\textwidth}\cm{0.085\textwidth}}
                    \toprule
                     & DoS & AOO & PSA & FFA & AVOA & ETO & SGA & WAA \\
                    \midrule
                    $x_{1}$ & 12.9888909 & 13.3089578 & 14.3835712 & NaN & 10.3624115 & 6.10146067 & 2.2170757 & 9.90246732 \\
                    $x_{2}$ & 13.8690372 & 13.7325103 & 14.8694544 & NaN & 8.77075093 & 4.52803833 & 2.05372255 & 10.4415022 \\
                    $x_{3}$ & 2.50699876 & 2.40820912 & 2.59246594 & NaN & 2.77900294 & 2.72990543 & 2.21872515 & 3.07949599 \\
                    $x_{4}$ & 6.48131699 & 9.37718037 & 10.5927012 & NaN & 7.25823212 & 0.72235314 & 4.2335812 & 5.81442362 \\
                    $x_{5}$ & 6.33422166 & 11.6377417 & 12.3866256 & NaN & 9.5216832 & 2.72726476 & 1.46826182 & 5.2906585 \\
                    $x_{6}$ & 3.95329554 & 3.40000558 & 2.39321317 & NaN & 3.78592371 & 2.54084754 & 3.9254009 & 4.17795437 \\
                    $x_{7}$ & 8.18867199 & 2.12047252 & 2.21619303 & NaN & 9.30716946 & 1.46474564 & 6.19247403 & 9.41297312 \\
                    $x_{8}$ & 9.49747645 & 1.45326478 & 2.03145342 & NaN & 10.8418392 & 1.99592892 & 8.26377253 & 10.4078648 \\
                    $x_{9}$ & 1.56757028 & 2.99775567 & 1.76327833 & NaN & 1.77738873 & 1.50679381 & 3.35051702 & 2.87116885 \\
                    $x_{10}$ & 10.8948181 & 6.96976651 & 10.7605182 & NaN & 10.6614498 & 1.02559479 & 9.15197199 & 10.2419927 \\
                    $x_{11}$ & 11.3355857 & 7.10015945 & 10.4132149 & NaN & 10.9663482 & -0.0027454 & 12.6743471 & 9.52118694 \\
                    $x_{12}$ & 3.19192349 & 1.81821068 & 3.07970717 & NaN & 2.65054604 & 1.91553843 & 4.36618523 & 2.32941245 \\
                    $x_{13}$ & 9.11686592 & 11.9292805 & 9.67542029 & NaN & 11.8307123 & 4.89502925 & 16.8599053 & 8.73907293 \\
                    $x_{14}$ & 9.51696602 & 11.6578647 & 12.7824881 & NaN & 12.9130449 & 4.04181626 & 15.5477379 & 10.5080317 \\
                    $x_{15}$ & 1.71999706 & 2.13675858 & 2.13118587 & NaN & 2.81417943 & 3.02497411 & 4.61624824 & 2.50188693 \\
                    $x_{16}$ & 5.63803607 & 7.83496928 & 6.6648707 & NaN & 10.5297882 & 2.33637529 & 13.7479204 & 6.94191714 \\
                    $x_{17}$ & 5.96924972 & 8.92472529 & 6.92384978 & NaN & 11.531419 & 3.27194708 & 13.2380064 & 9.03256409 \\
                    $x_{18}$ & 2.48864298 & 2.90505994 & 1.53419283 & NaN & 2.39877312 & 3.88518424 & 3.15550806 & 2.83347527 \\
                    $x_{19}$ & 7.97981168 & 10.7482806 & 7.28957934 & NaN & 10.2280648 & 17.9821989 & 10.243686 & 7.45934563 \\
                    $x_{20}$ & 8.1123124 & 10.6989156 & 6.77729394 & NaN & 8.7034067 & 23.4440988 & 11.5825813 & 7.35780639 \\
                    $x_{21}$ & 2.30288825 & 2.65683761 & 2.63959671 & NaN & 2.72170739 & 2.06283091 & 2.85644707 & 1.5002853 \\
                    Best & \textbf{115.457868} & 115.613739 & 115.525492 & NaN & 115.786401 & 117.535791 & 117.470719 & 115.66231 \\
                    \bottomrule
                \end{tabular}
            \end{table*}

            \begin{table*}[htbp!]
                \centering
                \caption{Statistical results for mountainous terrain path planning.}
                \label{tab44}
                \begin{tabular}{\cm{0.08\textwidth}\cm{0.085\textwidth}\cm{0.085\textwidth}\cm{0.085\textwidth}\cm{0.085\textwidth}\cm{0.085\textwidth}\cm{0.085\textwidth}\cm{0.085\textwidth}\cm{0.085\textwidth}}
                    \toprule
                     & DoS & AOO & PSA & FFA & AVOA & ETO & SGA & WAA \\
                    \midrule
                    Mean & \textbf{115.70175} & 116.664178 & 117.451852 & NaN & 116.386235 & 126.408888 & 124.175462 & 115.897928 \\
                    Std & 0.15175712 & 1.12785545 & 2.36536695 & NaN & 1.37315754 & 4.90373033 & 8.51917331 & \textbf{0.1045847} \\
                    Success & \textbf{1} & 0.96 & 0.92 & 0 & \textbf{1} & \textbf{1} & \textbf{1} & \textbf{1} \\
                    WSRT &  & 2.4E-06(+) & 2.2E-07(+) & NaN & 5.5E-03(+) & 1.4E-09(+) & 1.4E-09(+) & 3.3E-07(+) \\
                    FMR & \textbf{1.52} & 3.84 & 4.12 & NaN & 2.96 & 6.52 & 6.2 & 2.84 \\
                    F-Rank & \textbf{1} & 4 & 5 & 8 & 3 & 7 & 6 & 2 \\
                    \bottomrule
                \end{tabular}
            \end{table*}

            As shown in Table~\ref{tab43}, DoS achieved the best Best value in 25 independent runs, indicating its strong local exploitation ability on this problem. Further, Table~\ref{tab44} reveals that DoS outperforms all competitors in terms of Mean, demonstrating its more stable and reliable overall performance. Although its Std is slightly higher than that of WAA, it remains close to zero, reflecting good stability.

            In the Wilcoxon rank-sum test, DoS exhibits comparable performance to WAA and AVOA, while significantly outperforming other competitors. In the Friedman test, DoS ranks first, ahead of WAA, further validating its advantage in this problem. In summary, DoS demonstrates high convergence accuracy and stability in mountainous terrain path planning tasks, showing strong engineering adaptability.

\begin{figure*}[htbp!]
                \centering
                \begin{tabular}{\cm{0.45\textwidth}\cm{0.45\textwidth}}
                    \multicolumn{2}{c}{\includegraphics[width=0.75\linewidth]{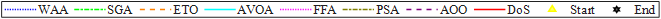}} \\
                    \includegraphics[width=0.75\linewidth]{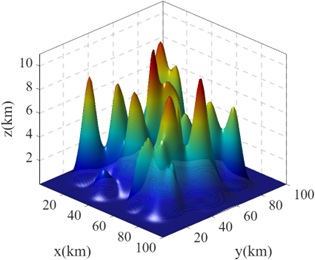} & \includegraphics[width=0.75\linewidth]{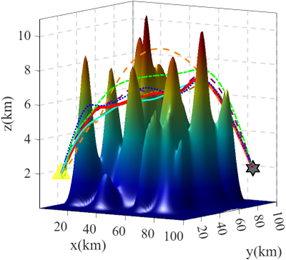} \\
                    (a) Mountainous terrain map without no fly zones & (b) Right view \\
                    \includegraphics[width=0.75\linewidth]{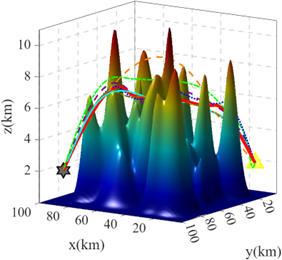} & \includegraphics[width=0.75\linewidth]{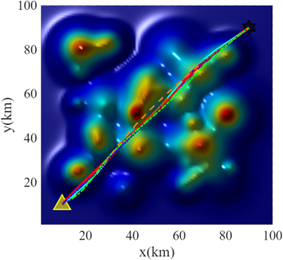} \\
                    (c) Left view & (d) Top view
                \end{tabular}
                \caption{Visualization of mountainous terrain path planning.}
                \label{fig14}
            \end{figure*}

            To help readers intuitively understand the performance differences among algorithms, Fig.~\ref{fig14} provides a visualization of the mountainous terrain and the corresponding flight paths obtained during the experiments.

        \subsection{Mountainous Terrain Path Planning with No Fly Zones}

            As verified in Section 6.2, the performance of DoS in mountainous terrain path planning under standard terrain conditions has been demonstrated. To further evaluate its effectiveness in discontinuous terrains, this section introduces five nofly zone cylinders on the original mountainous terrain. The UAV is required to minimize the path length without intersecting any of these no-fly zones. The center coordinates $(x_{c}, y_{c})$, radius $(r_{c})$ and height $(h_{c})$ of each no-fly zone are detailed in Table~\ref{tab45}.

            \begin{table}[htbp!]
                \centering
                \caption{Information of no fly zones.}
                \label{tab45}
                \begin{tabular}{\cm{0.02\textwidth}\cm{0.06\textwidth}\cm{0.06\textwidth}\cm{0.06\textwidth}\cm{0.06\textwidth}\cm{0.06\textwidth}}
                    \toprule
                     & No fly zone 1 & No fly zone 2 & No fly zone 3 & No fly zone 4 & No fly zone 5 \\
                    \midrule
                    $x_{c}$ & 56.7157 & 32.6590 & 13.1987 & 65.9033 & 64.0290 \\
                    $y_{c}$ & 18.5965 & 39.5000 & 45.5621 & 70.4151 & 44.1858 \\
                    $r_{c}$ & 5.0676 & 4.7530 & 10.5505 & 4.9743 & 8.7564 \\
                    $h_{c}$ & 6.4409 & 8.2620 & 3.8032 & 8.5927 & 9.5035 \\
                    \bottomrule
                \end{tabular}
            \end{table}

            To account for the no fly zone restrictions, a fifth constraint is added to the mathematical model from Section 6.1, as shown in Equation~\ref{eq50}.
            \begin{equation}
                \begin{aligned}
                    G_{5}(\bm{\mathrm{x}}) = \ &\sqrt{\left( \tilde{x}_i - x_c^{j} \right)^2 + \left( \tilde{y}_i - y_c^{j} \right)^2} > r_c^{j} \ \ \mathrm{and} \\
                    &\tilde{z}_i > h_c^{j}, \quad i = 1, \ldots ,m, \quad j = 1, \ldots ,5
                \end{aligned}
                \label{eq50}
            \end{equation}

            \noindent
            where $x_{c}^{j}$, $y_{c}^{j}$, $h_{c}^{j}$ and $r_{c}^{j}$ denote the center coordinates, height and radius of the $j$-th no fly zone, respectively. Accordingly, the original objective function (Equation~\ref{eq50}) is updated to Equation~\ref{eq51}.
            \begin{equation}
                \begin{aligned}
                    F = \ &F_{1} + F_{2} + Q\left[G_{1}(\bm{\mathrm{x}})\right] + Q\left[G_{2}(\bm{\mathrm{x}})\right] + \\
                    &Q\left[G_{3}(\bm{\mathrm{x}})\right] + Q\left[G_{4}(\bm{\mathrm{x}})\right] + Q\left[G_{5}(\bm{\mathrm{x}})\right]
                \end{aligned}
                \label{eq51}
            \end{equation}

            Based on Equation~\ref{eq51}, 25 independent experiments were conducted for DoS and the advanced competitors under 10,000 evaluations. The best solutions are listed in Table~\ref{tab46} and the statistical results are shown in Table~\ref{tab47}.

            \begin{table*}[htbp!]
                \centering
                \caption{The best results of mountainous terrain path planning with no fly zones.}
                \label{tab46}
                \begin{tabular}{\cm{0.08\textwidth}\cm{0.085\textwidth}\cm{0.085\textwidth}\cm{0.085\textwidth}\cm{0.085\textwidth}\cm{0.085\textwidth}\cm{0.085\textwidth}\cm{0.085\textwidth}\cm{0.085\textwidth}}
                    \toprule
                     & DoS & AOO & PSA & FFA & AVOA & ETO & SGA & WAA \\
                    \midrule
                    $x_{1}$ & 24.6151727 & 6.19575074 & 5.83758569 & NaN & 9.653058 & 8.35983887 & 1.35877043 & 9.10803916 \\
                    $x_{2}$ & 24.3137897 & 7.18586063 & 6.44106836 & NaN & 9.27005726 & 7.07298245 & -0.1705658 & 8.19400313 \\
                    $x_{3}$ & 3.69506866 & 3.39562003 & 3.25904729 & NaN & 2.69756453 & 1.91506739 & 2.38150614 & 3.55724855 \\
                    $x_{4}$ & 10.4177494 & 5.70624126 & 11.9844314 & NaN & 9.03563824 & 0.33891949 & 2.50262731 & 9.27551588 \\
                    $x_{5}$ & 9.43944082 & 6.02840097 & 9.26556372 & NaN & 6.61331424 & 0.62194985 & 1.69667884 & 8.90176369 \\
                    $x_{6}$ & 3.02437287 & 1.75030092 & 4.31175872 & NaN & 4.22084634 & 1.96825997 & 3.71290435 & 4.76670039 \\
                    $x_{7}$ & 12.1206834 & 13.2568275 & 19.3167965 & NaN & 7.87895285 & 2.11268223 & 6.58050107 & 8.50796062 \\
                    $x_{8}$ & 11.1233053 & 10.5697383 & 21.9059203 & NaN & 10.2328007 & 2.07845911 & 6.36037763 & 8.8398287 \\
                    $x_{9}$ & 3.64572418 & 3.45413548 & 3.73734884 & NaN & 2.41774344 & 1.36523701 & 4.10445474 & 3.06226768 \\
                    $x_{10}$ & 8.72585555 & 4.00668196 & 10.7922801 & NaN & 7.75352486 & 7.34281971 & 11.1271871 & 9.43526576 \\
                    $x_{11}$ & 8.2382542 & 4.29188002 & 20.5549274 & NaN & 8.75100226 & 3.84754704 & 12.7144057 & 9.46919635 \\
                    $x_{12}$ & 3.22731975 & 1.56173136 & 3.8820003 & NaN & 1.93019374 & 2.07339549 & 4.42039303 & 3.05984284 \\
                    $x_{13}$ & 2.24884308 & 5.31807437 & 1.73387989 & NaN & 4.1888267 & 10.5138342 & 13.9973874 & 9.46240553 \\
                    $x_{14}$ & 1.82812146 & 5.02473278 & 3.63358891 & NaN & 6.31302907 & 2.91577219 & 15.7358933 & 8.4837311 \\
                    $x_{15}$ & 2.89644401 & 2.07465456 & 2.44401085 & NaN & 3.53826108 & 3 & 3.83985846 & 4.31680178 \\
                    $x_{16}$ & 8.68999588 & 12.9995088 & 4.7233468 & NaN & 19.1727522 & 16.6093988 & 13.3722614 & 8.56006039 \\
                    $x_{17}$ & 9.5890141 & 13.2086192 & 4.53099461 & NaN & 13.7208065 & 5.67094357 & 9.69266408 & 8.43153517 \\
                    $x_{18}$ & 3.16433811 & 2.76574832 & 1.91770311 & NaN & 2.11226421 & 4.45068555 & 3.39367589 & 3.70199248 \\
                    $x_{19}$ & 6.30480295 & 16.9155338 & 10.9790725 & NaN & 4.93688188 & 14.4558711 & 15.8873627 & 9.00155613 \\
                    $x_{20}$ & 7.52608428 & 12.4222365 & 5.64191406 & NaN & 8.56534178 & 4.5718354 & 14.7960556 & 8.65555913 \\
                    $x_{21}$ & 3.96077692 & 3.14030899 & 2.32702171 & NaN & 2.91844852 & 2.2050623 & 4.59898876 & 3.66490752 \\
                    Best & \textbf{116.451419} & 116.884378 & 118.018618 & NaN & 117.230664 & 129.290161 & 118.430385 & 116.845406 \\
                    \bottomrule
                \end{tabular}
            \end{table*}

            \begin{table*}[htbp!]
                \centering
                \caption{Statistical results of mountainous terrain path planning with no fly zones.}
                \label{tab47}
                \begin{tabular}{\cm{0.08\textwidth}\cm{0.085\textwidth}\cm{0.085\textwidth}\cm{0.085\textwidth}\cm{0.085\textwidth}\cm{0.085\textwidth}\cm{0.085\textwidth}\cm{0.085\textwidth}\cm{0.085\textwidth}}
                    \toprule
                     & DoS & AOO & PSA & FFA & AVOA & ETO & SGA & WAA \\
                    \midrule
                    Mean & \textbf{118.36611} & 122.097895 & 129.442151 & NaN & 120.574604 & 138.719413 & 133.276447 & 119.678824 \\
                    Std & \textbf{1.2245701} & 8.08930431 & 8.54127858 & NaN & 7.3629909 & 7.81870547 & 12.0252177 & 7.52920892 \\
                    Success & \textbf{1} & 0.44 & 0.6 & 0 & 0.96 & 0.88 & 0.96 & 0.56 \\
                    WSRT &  & 3.8E-05(+) & 1.8E-07(+) & NaN & 9.4E-01($\approx$) & 1.4E-09(+) & 1.2E-08(+) & 6.8E-01($\approx$) \\
                    FMR & \textbf{2.2} & 3.88 & 5.08 & NaN & 2.6 & 6.48 & 5.32 & 2.44 \\
                    F-Rank & \textbf{1} & 4 & 5 & 8 & 3 & 7 & 6 & 2 \\
                    \bottomrule
                \end{tabular}
            \end{table*}

            As shown in Tables~\ref{tab46} and~\ref{tab47}, DoS ranks first in five key performance metrics demonstrating the best overall performance. With the addition of no fly zones, only DoS maintains a 100\% success rate, further validating its adaptability to no-fly zone constraints.

            In terms of optimization accuracy, DoS outperforms other algorithms in both Best and Mean values and has the smallest Std, indicating stronger exploitation capability and better stability in this problem. The WSRT results show that, except for AVOA and WAA, all other algorithms are statistically significantly inferior to DoS. In the Friedman test, DoS again ranks first, further verifying its practicality in complex mountainous terrain path planning problems.

\begin{figure*}[htbp!]
                \centering
                \begin{tabular}{\cm{0.45\textwidth}\cm{0.45\textwidth}}
                    \multicolumn{2}{c}{\includegraphics[width=0.75\linewidth]{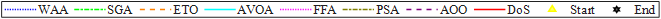}} \\
                    \includegraphics[width=0.75\linewidth]{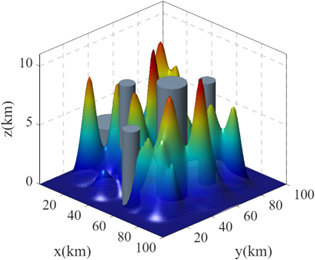} & \includegraphics[width=0.75\linewidth]{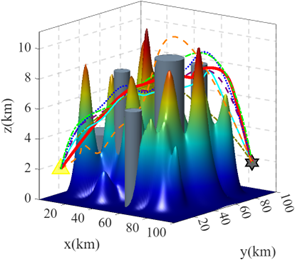} \\
                    (a) Mountainous terrain map with no fly zones & (b) Right view \\
                    \includegraphics[width=0.75\linewidth]{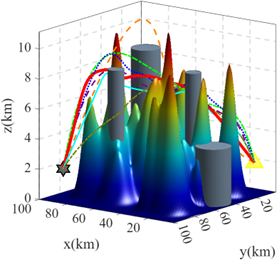} & \includegraphics[width=0.75\linewidth]{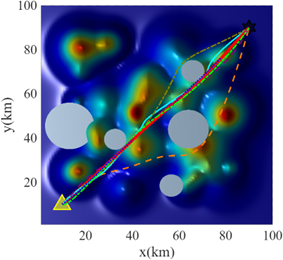} \\
                    (c) Left view & (d) Top view
                \end{tabular}
                \caption{Visualization of mountainous terrain path planning with no fly zones.}
                \label{fig15}
            \end{figure*}

            Fig.~\ref{fig15} visualizes the flight paths obtained by each algorithm over terrain with no fly zones, enabling a more intuitive comparison of their performances in this scenario.

    \section{Discussion}

        Based on the experimental results of DoS in Sections 4 to 6, covering CEC2017 and CEC2022 benchmark test functions, 10 engineering optimization problems, and the mountainous terrain path planning application, this subsection will discuss the key distinctions between DoS and traditional metaheuristic algorithms, and analyze the fundamental reasons behind its performance advantages.

        \subsection{Advantages of DoS}

            As shown in Fig.~\ref{fig7}, the exploration and exploitation rate curves of DoS vary significantly across different functions, indicating that DoS exhibits strong adaptability when solving various types of optimization problems. Furthermore, the information in Fig.~\ref{fig8} demonstrates that DoS achieves faster convergence speed and higher solution accuracy compared to advanced competitors across diverse functions. This fully demonstrates that under the same computational cost, DoS can achieve better optimization results.

            The results in Fig.~\ref{fig9} and 11 show that for 92.7\% of the CEC2017 and CEC2022 benchmark test functions, the performance of DoS is not significantly affected by increasing dimensionality. Moreover, on F28, C11 and C12, the performance of DoS improves with increasing dimension, further verifying that DoS possesses good scalability across various types of optimization problems. In addition, the results in Fig.~\ref{fig10} and~\ref{fig12} indicate that DoS consistently ranks first in Friedman ranking across five different dimensions of benchmark test functions, significantly outperforming advanced competitors. Furthermore, in comparison with SOTA algorithms, DoS demonstrates superior performance, validating its outstanding optimization capability and great potential across various problem types.

            In the engineering problem experiments of Section 5, DoS was evaluated on low- dimensional and high-dimensional problems (up to 30-dimensions), with both few and many constraints (up to 91 inequality constraints). The experimental results consistently show that DoS achieves higher success rates and better optimization results than advanced competitors, demonstrating its strong competitiveness in solving constrained optimization problems.

            In the mountainous terrain path planning application experiments, DoS also demonstrates strong competitiveness. According to the results in Tables 42 to 45, DoS outperforms advanced competitors in terms of higher success rates and superior optimization outcomes, fully highlighting its advantage in mountainous terrain path planning tasks.

        \subsection{Disadvantages of DoS}

            Although DoS demonstrates strong optimization capabilities, its performance remains suboptimal on F28 (30-dimensions) and on C11 and C12 (10-dimensions). Moreover, the proposed DoS is currently limited to single objective optimization problems and its extension to multi objective versions has not yet been explored. Compared with metaheuristic algorithms such as PSA and SGA, which feature simpler designs, DoS does not hold an advantage in computational cost.

    \section{Conclusions}

        Inspired by the dogfighting behavior of fighter jets, this paper proposes a metaphor-free metaheuristic algorithm named DoS, aiming to introduce more effective solution methods for optimization problems. Its performance is first evaluated on CEC2017 (30D, 50D, 100D) and CEC2022 (10D, 20D) benchmark test functions. We analyze DoS's design advantages and characteristics from the perspectives of exploration and exploitation rates, convergence, scalability, behavioral characteristics, and computational cost. Subsequently, we validated DoS's performance on 10 constrained engineering optimization problems and analyzed its applicability. Finally, we applied DoS to mountainous terrain path planning task, further demonstrating its advantages in this domain.

        Based on comparisons with advanced competitors and SOTA algorithms, the main conclusions of this study can be summarized as follows:

        \begin{enumerate}[(1)]
            \item The design of DoS alleviates the dependency of swarm-based algorithms on the current best solution by integrating multiple search strategies and a dynamic selection mechanism, thus achieving a better balance between exploration and exploitation;
            \item In experiments on the CEC2017 and CEC2022 benchmark test functions, DoS significantly outperforms advanced competitors and shows strong competitiveness compared to SOTA algorithms;
            \item On 10 constrained engineering problems, DoS achieves higher success rates and better optimization results than its competitors, demonstrating clear performance advantages;
            \item In the practical application of mountainous terrain path planning, DoS consistently outperforms its competitors in both the absence and presence of no fly zones, with higher success rates and superior optimization results.
        \end{enumerate}

        Given DoS’s promising design, performance, and applicability, future work may consider the following directions:

        \begin{enumerate}[(1)]
            \item Developing a multi objective version of DoS;
            \item Extending the dynamic selection mechanism of DoS to other metaheuristic algorithms to improve the balance between exploitation and exploitation;
            \item Applying DoS to more challenging and high complexity engineering problems to further validate its practical value;
            \item Expanding the application scope of DoS, such as solving numerical differential equations.
        \end{enumerate}

    \section{Declaration of Generative AI in Scientific Writing}
        
        After drafting the manuscript, the authors used ChatGPT for language editing. The content was subsequently reviewed and revised as necessary by the authors, who take full responsibility for the final publication.

    \printcredits
    
	\bibliographystyle{apalike}
	\bibliography{references}

@article{hao2025composite,
  title={A composite particle swarm optimization algorithm with future information inspired by non-equidistant grey predictive evolution for global optimization problems and engineering problems},
  author={Hao, Rui and Hu, Zhongbo and Xiong, WenTao and Jiang, Shaojie},
  journal={Advances in Engineering Software},
  volume={202},
  pages={103868},
  year={2025},
  publisher={Elsevier}
}

@article{osaba2019discrete,
  title={A discrete and improved bat algorithm for solving a medical goods distribution problem with pharmacological waste collection},
  author={Osaba, Eneko and Yang, Xin-She and Fister Jr, Iztok and Del Ser, Javier and Lopez-Garcia, Pedro and Vazquez-Pardavila, Alejo J},
  journal={Swarm and evolutionary computation},
  volume={44},
  pages={273--286},
  year={2019},
  publisher={Elsevier}
}

@article{chu2021physarum,
  title={A Physarum-inspired algorithm for logistics optimization: From the perspective of effective distance},
  author={Chu, Dong and Ma, Wenjian and Yang, Zhuocheng and Li, Jingyu and Deng, Yong and Cheong, Kang Hao},
  journal={Swarm and Evolutionary Computation},
  volume={64},
  pages={100890},
  year={2021},
  publisher={Elsevier}
}

@article{dye2012finite,
  title={A finite horizon deteriorating inventory model with two-phase pricing and time-varying demand and cost under trade credit financing using particle swarm optimization},
  author={Dye, Chung-Yuan},
  journal={Swarm and Evolutionary Computation},
  volume={5},
  pages={37--53},
  year={2012},
  publisher={Elsevier}
}

@article{su2011two,
  title={A two-stage algorithm integrating genetic algorithm and modified Newton method for neural network training in engineering systems},
  author={Su, Ching-Long and Yang, Shih-Ming and Huang, WL},
  journal={Expert Systems with Applications},
  volume={38},
  number={10},
  pages={12189--12194},
  year={2011},
  publisher={Elsevier}
}

@article{hosseinali2024accelerated,
  title={Accelerated gradient descent using improved Selective Backpropagation},
  author={Hosseinali, Farzad},
  journal={Expert Systems with Applications},
  volume={255},
  pages={124426},
  year={2024},
  publisher={Elsevier}
}

@article{keivanian2022novel,
  title={A novel hybrid fuzzy--metaheuristic approach for multimodal single and multi-objective optimization problems},
  author={Keivanian, Farshid and Chiong, Raymond},
  journal={Expert Systems with Applications},
  volume={195},
  pages={116199},
  year={2022},
  publisher={Elsevier}
}

@article{deb2002fast,
  title={A fast and elitist multiobjective genetic algorithm: NSGA-II},
  author={Deb, Kalyanmoy and Pratap, Amrit and Agarwal, Sameer and Meyarivan, TAMT},
  journal={IEEE transactions on evolutionary computation},
  volume={6},
  number={2},
  pages={182--197},
  year={2002},
  publisher={Ieee}
}

@article{myers2016application,
  title={Application of a multi-objective evolutionary algorithm to the spacecraft stationkeeping problem},
  author={Myers, Philip L and Spencer, David B},
  journal={Acta Astronautica},
  volume={127},
  pages={76--86},
  year={2016},
  publisher={Elsevier}
}

@article{gong2010bbo,
  title={DE/BBO: a hybrid differential evolution with biogeography-based optimization for global numerical optimization},
  author={Gong, Wenyin and Cai, Zhihua and Ling, Charles X},
  journal={Soft Computing},
  volume={15},
  number={4},
  pages={645--665},
  year={2010},
  publisher={Springer}
}

@article{dhiman2019seagull,
  title={Seagull optimization algorithm: Theory and its applications for large-scale industrial engineering problems},
  author={Dhiman, Gaurav and Kumar, Vijay},
  journal={Knowledge-based systems},
  volume={165},
  pages={169--196},
  year={2019},
  publisher={Elsevier}
}

@article{gholizadeh2020new,
  title={A new Newton metaheuristic algorithm for discrete performance-based design optimization of steel moment frames},
  author={Gholizadeh, Saeed and Danesh, Masood and Gheyratmand, Changiz},
  journal={Computers \& Structures},
  volume={234},
  pages={106250},
  year={2020},
  publisher={Elsevier}
}

@article{houssein2025recent,
  title={Recent metaheuristic algorithms for solving some civil engineering optimization problems},
  author={Houssein, Essam H and Hossam Abdel Gafar, Mohamed and Fawzy, Naglaa and Sayed, Ahmed Y},
  journal={Scientific Reports},
  volume={15},
  number={1},
  pages={7929},
  year={2025},
  publisher={Nature Publishing Group UK London}
}

@article{lu2025research,
  title={Research on optimization and regulation of air volume in mine ventilation network based on Multi-Strategy beetle swarm optimization},
  author={Lu, Feng and Wang, Kai and Wang, Jishuo and Wang, Zhijing and Ma, Yanfei},
  journal={Advanced Engineering Informatics},
  volume={68},
  pages={103654},
  year={2025},
  publisher={Elsevier}
}

@article{wang2025towards,
  title={Towards large-scale cotton blending optimization: dual-pheromone crossover ant colony algorithm with expert heuristic cognition},
  author={Wang, Menglei and Wang, Jingan and Gao, Weidong},
  journal={Advanced Engineering Informatics},
  volume={68},
  pages={103657},
  year={2025},
  publisher={Elsevier}
}

@article{wu2025improved,
  title={Improved discrete particle swarm optimization algorithm for solving fuzzy flexible job shop machines and automated guided vehicles fusion scheduling problem},
  author={Wu, Rui and Tian, Zheng and Li, Xixing and Wu, Chenchen and Tang, Hongtao and Li, Yibing},
  journal={Engineering Applications of Artificial Intelligence},
  volume={160},
  pages={111951},
  year={2025},
  publisher={Elsevier}
}

@article{kuo2025artificial,
  title={Artificial rabbits optimization--based motion balance system for the impact recovery of a bipedal robot},
  author={Kuo, Ping-Huan and Yang, Wei-Cyuan and Lin, Yu-Sian and Peng, Chao-Chung},
  journal={Advanced Engineering Informatics},
  volume={63},
  pages={102965},
  year={2025},
  publisher={Elsevier}
}

@article{wangying2025scheduling,
  title={Scheduling and route planning for forests rescue: Applications with a novel ant colony optimization algorithm},
  author={Wangying, Xu and Naiming, Xie},
  journal={Engineering Applications of Artificial Intelligence},
  volume={155},
  pages={111042},
  year={2025},
  publisher={Elsevier}
}

@article{wolpert2002no,
  title={No free lunch theorems for optimization},
  author={Wolpert, David H and Macready, William G},
  journal={IEEE transactions on evolutionary computation},
  volume={1},
  number={1},
  pages={67--82},
  year={2002},
  publisher={IEEE}
}

@article{arani2013improved,
  title={An improved PSO algorithm with a territorial diversity-preserving scheme and enhanced exploration--exploitation balance},
  author={Arani, Behrooz Ostadmohammadi and Mirzabeygi, Pooya and Panahi, Masoud Shariat},
  journal={Swarm and Evolutionary Computation},
  volume={11},
  pages={1--15},
  year={2013},
  publisher={Elsevier}
}

@inproceedings{kennedy1995particle,
  title={Particle swarm optimization},
  author={Kennedy, James and Eberhart, Russell},
  booktitle={Proceedings of ICNN'95-international conference on neural networks},
  volume={4},
  pages={1942--1948},
  year={1995},
  organization={ieee}
}

@article{mirjalili2014grey,
  title={Grey wolf optimizer},
  author={Mirjalili, Seyedali and Mirjalili, Seyed Mohammad and Lewis, Andrew},
  journal={Advances in engineering software},
  volume={69},
  pages={46--61},
  year={2014},
  publisher={Elsevier}
}

@article{wang2024black,
  title={Black-winged kite algorithm: a nature-inspired meta-heuristic for solving benchmark functions and engineering problems},
  author={Wang, Jun and Wang, Wen-chuan and Hu, Xiao-xue and Qiu, Lin and Zang, Hong-fei},
  journal={Artificial Intelligence Review},
  volume={57},
  number={4},
  pages={98},
  year={2024},
  publisher={Springer}
}

@article{gao2025freedom,
  title={Freedom from inspiration! Achieving efficient metaheuristic optimization with delta plus},
  author={Gao, Yuansheng and Wang, Jinpeng and Qin, Lang and Zhang, Jiahui and Wang, Yulin},
  journal={Cluster Computing},
  volume={28},
  number={7},
  pages={424},
  year={2025},
  publisher={Springer}
}

@article{gao2023pid,
  title={PID-based search algorithm: A novel metaheuristic algorithm based on PID algorithm},
  author={Gao, Yuansheng},
  journal={Expert Systems with Applications},
  volume={232},
  pages={120886},
  year={2023},
  publisher={Elsevier}
}

@article{zhang2009jade,
  title={JADE: adaptive differential evolution with optional external archive},
  author={Zhang, Jingqiao and Sanderson, Arthur C},
  journal={IEEE Transactions on evolutionary computation},
  volume={13},
  number={5},
  pages={945--958},
  year={2009},
  publisher={IEEE}
}

@article{sorensen2015metaheuristics,
  title={Metaheuristics—the metaphor exposed},
  author={S{\"o}rensen, Kenneth},
  journal={International Transactions in Operational Research},
  volume={22},
  number={1},
  pages={3--18},
  year={2015},
  publisher={Wiley Online Library}
}

@article{camacho2023exposing,
  title={Exposing the grey wolf, moth-flame, whale, firefly, bat, and antlion algorithms: six misleading optimization techniques inspired by bestial metaphors},
  author={Camacho-Villal{\'o}n, Christian L and Dorigo, Marco and St{\"u}tzle, Thomas},
  journal={International Transactions in Operational Research},
  volume={30},
  number={6},
  pages={2945--2971},
  year={2023},
  publisher={Wiley Online Library}
}

@article{wang2025logistic,
  title={Logistic-Gauss Circle optimizer: Theory and applications},
  author={Wang, Jinpeng and Gao, Yuansheng and Qin, Lang and Li, Yike},
  journal={Applied Mathematical Modelling},
  volume={143},
  pages={116052},
  year={2025},
  publisher={Elsevier}
}

@article{holland1992genetic,
  title={Genetic algorithms},
  author={Holland, John H},
  journal={Scientific american},
  volume={267},
  number={1},
  pages={66--73},
  year={1992},
  publisher={JSTOR}
}

@article{zohar2011mobile,
  title={Mobile robot characterized by dynamic and kinematic equations and actuator dynamics: Trajectory tracking and related application},
  author={Zohar, Ilan and Ailon, Amit and Rabinovici, Raul},
  journal={Robotics and autonomous systems},
  volume={59},
  number={6},
  pages={343--353},
  year={2011},
  publisher={Elsevier}
}

@inproceedings{tanabe2014improving,
  title={Improving the search performance of SHADE using linear population size reduction},
  author={Tanabe, Ryoji and Fukunaga, Alex S},
  booktitle={2014 IEEE congress on evolutionary computation (CEC)},
  pages={1658--1665},
  year={2014},
  organization={IEEE}
}

@inproceedings{mohamed2017lshade,
  title={LSHADE with semi-parameter adaptation hybrid with CMA-ES for solving CEC 2017 benchmark problems},
  author={Mohamed, Ali W and Hadi, Anas A and Fattouh, Anas M and Jambi, Kamal M},
  booktitle={2017 IEEE Congress on evolutionary computation (CEC)},
  pages={145--152},
  year={2017},
  organization={IEEE}
}

@article{li2022novel,
  title={A novel adaptive L-SHADE algorithm and its application in UAV swarm resource configuration problem},
  author={Li, Yintong and Han, Tong and Zhou, Huan and Tang, Shangqin and Zhao, Hui},
  journal={Information Sciences},
  volume={606},
  pages={350--367},
  year={2022},
  publisher={Elsevier}
}

@article{kirkpatrick1983optimization,
  title={Optimization by simulated annealing},
  author={Kirkpatrick, Scott and Gelatt Jr, C Daniel and Vecchi, Mario P},
  journal={science},
  volume={220},
  number={4598},
  pages={671--680},
  year={1983},
  publisher={American association for the advancement of science}
}

@inproceedings{prajapati2020tabu,
  title={Tabu search algorithm (TSA): A comprehensive survey},
  author={Prajapati, Vishnu Kumar and Jain, Mayank and Chouhan, Lokesh},
  booktitle={2020 3rd International Conference on Emerging Technologies in Computer Engineering: Machine Learning and Internet of Things (ICETCE)},
  pages={1--8},
  year={2020},
  organization={IEEE}
}

@article{zabinsky2009random,
  title={Random search algorithms},
  author={Zabinsky, Zelda B and others},
  journal={Department of Industrial and Systems Engineering, University of Washington, USA},
  pages={1--16},
  year={2009}
}

@article{abualigah2024non,
  title={The non-monopolize search (NO): A novel single-based local search optimization algorithm},
  author={Abualigah, Laith and Al-qaness, Mohammed AA and Abd Elaziz, Mohamed and Ewees, Ahmed A and Oliva, Diego and Cuong-Le, Thanh},
  journal={Neural Computing and Applications},
  volume={36},
  number={10},
  pages={5305--5332},
  year={2024},
  publisher={Springer}
}

@article{ezugwu2021metaheuristics,
  title={Metaheuristics: a comprehensive overview and classification along with bibliometric analysis},
  author={Ezugwu, Absalom E and Shukla, Amit K and Nath, Rahul and Akinyelu, Andronicus A and Agushaka, Jeffery O and Chiroma, Haruna and Muhuri, Pranab K},
  journal={Artificial Intelligence Review},
  volume={54},
  number={6},
  pages={4237--4316},
  year={2021},
  publisher={Springer}
}

@article{storn1997differential,
  title={Differential evolution--a simple and efficient heuristic for global optimization over continuous spaces},
  author={Storn, Rainer and Price, Kenneth},
  journal={Journal of global optimization},
  volume={11},
  number={4},
  pages={341--359},
  year={1997},
  publisher={Springer}
}

@article{koza1994genetic,
  title={Genetic programming as a means for programming computers by natural selection},
  author={Koza, John R},
  journal={Statistics and computing},
  volume={4},
  number={2},
  pages={87--112},
  year={1994},
  publisher={Springer}
}

@article{gao2024love,
  title={Love evolution algorithm: A stimulus--value--role theory-inspired evolutionary algorithm for global optimization},
  author={Gao, Yuansheng and Zhang, Jiahui and Wang, Yulin and Wang, Jinpeng and Qin, Lang},
  journal={The Journal of Supercomputing},
  volume={80},
  number={9},
  pages={12346--12407},
  year={2024},
  publisher={Springer}
}

@article{dorigo1991ant,
  title={Ant Colony Optimization—new optimization techniques in engineering},
  author={Dorigo, M},
  journal={by Onwubolu, GC, and BV Babu, Springer-Verlag Berlin Heidelberg},
  pages={101--117},
  year={1991}
}

@article{rashedi2009gsa,
  title={GSA: a gravitational search algorithm},
  author={Rashedi, Esmat and Nezamabadi-Pour, Hossein and Saryazdi, Saeid},
  journal={Information sciences},
  volume={179},
  number={13},
  pages={2232--2248},
  year={2009},
  publisher={Elsevier}
}

@article{mirjalili2016sca,
  title={SCA: a sine cosine algorithm for solving optimization problems},
  author={Mirjalili, Seyedali},
  journal={Knowledge-based systems},
  volume={96},
  pages={120--133},
  year={2016},
  publisher={Elsevier}
}

@article{ouyang2025beaver,
  title={Beaver behavior optimizer: A novel metaheuristic algorithm for solar PV parameter identification and engineering problems},
  author={Ouyang, Kaichen and Wei, Dedai and Sha, Xinye and Yu, Juntao and Zhao, Yuanli and Qiu, Minyu and Fu, Shengwei and Heidari, Ali Asghar and Chen, Huiling},
  journal={Journal of Advanced Research},
  year={2025},
  publisher={Elsevier}
}

@article{gao2025escape,
  title={Escape after love: Philoponella prominens optimizer and its application to 3D path planning},
  author={Gao, Yuansheng and Wang, Jinpeng and Li, Changlin},
  journal={Cluster Computing},
  volume={28},
  number={2},
  pages={81},
  year={2025},
  publisher={Springer}
}

@article{ouyang2024escape,
  title={Escape: an optimization method based on crowd evacuation behaviors},
  author={Ouyang, Kaichen and Fu, Shengwei and Chen, Yi and Cai, Qifeng and Heidari, Ali Asghar and Chen, Huiling},
  journal={Artificial Intelligence Review},
  volume={58},
  number={1},
  pages={19},
  year={2024},
  publisher={Springer}
}

@article{wang2025traffic,
  title={Traffic jam optimizer: A novel swarm-based metaheuristic algorithm for solving global optimization problems},
  author={Wang, Jinpeng and Shang, Ziyang},
  journal={Applied Mathematical Modelling},
  pages={116410},
  year={2025},
  publisher={Elsevier}
}

@article{mirjalili2016multi,
  title={Multi-verse optimizer: a nature-inspired algorithm for global optimization},
  author={Mirjalili, Seyedali and Mirjalili, Seyed Mohammad and Hatamlou, Abdolreza},
  journal={Neural computing and applications},
  volume={27},
  number={2},
  pages={495--513},
  year={2016},
  publisher={Springer}
}

@article{zhao2019atom,
  title={Atom search optimization and its application to solve a hydrogeologic parameter estimation problem},
  author={Zhao, Weiguo and Wang, Liying and Zhang, Zhenxing},
  journal={Knowledge-Based Systems},
  volume={163},
  pages={283--304},
  year={2019},
  publisher={Elsevier}
}

@article{cheng2024weighted,
  title={Weighted average algorithm: A novel meta-heuristic optimization algorithm based on the weighted average position concept},
  author={Cheng, Jun and De Waele, Wim},
  journal={Knowledge-Based Systems},
  volume={305},
  pages={112564},
  year={2024},
  publisher={Elsevier}
}

@inproceedings{ashraf2021dogfight,
  title={Dogfight: Detecting drones from drones videos},
  author={Ashraf, Muhammad Waseem and Sultani, Waqas and Shah, Mubarak},
  booktitle={Proceedings of the IEEE/CVF Conference on Computer Vision and Pattern Recognition},
  pages={7067--7076},
  year={2021}
}

@article{greenwood1992differential,
  title={A differential game in three dimensions: The aerial dogfight scenario},
  author={Greenwood, Nigel},
  journal={Dynamics and Control},
  volume={2},
  number={2},
  pages={161--200},
  year={1992},
  publisher={Springer}
}

@article{olsder1974role,
  title={Role determination in an aerial dogfight},
  author={Olsder, GJ and Breakwell, JV},
  journal={International Journal of Game Theory},
  volume={3},
  number={1},
  pages={47--66},
  year={1974},
  publisher={Springer}
}

@article{winder1966partitions,
  title={Partitions of N-space by hyperplanes},
  author={Winder, Robert O},
  journal={SIAM Journal on Applied Mathematics},
  volume={14},
  number={4},
  pages={811--818},
  year={1966},
  publisher={SIAM}
}

@article{sudholt2019benefits,
  title={The benefits of population diversity in evolutionary algorithms: a survey of rigorous runtime analyses},
  author={Sudholt, Dirk},
  journal={Theory of evolutionary computation: Recent developments in discrete optimization},
  pages={359--404},
  year={2019},
  publisher={Springer}
}

@misc{thomas2009introduction,
  title={Introduction to algorithms third edition},
  author={Thomas H, Cormen and Charles, E and Ronald L, Rivest and Clifford, Stein and others},
  year={2009},
  publisher={Mit Press}
}

@article{wu2017problem,
  title={Problem definitions and evaluation criteria for the CEC 2017 competition on constrained real-parameter optimization},
  author={Wu, Guohua and Mallipeddi, Rammohan and Suganthan, Ponnuthurai Nagaratnam},
  journal={National University of Defense Technology, Changsha, Hunan, PR China and Kyungpook National University, Daegu, South Korea and Nanyang Technological University, Singapore, Technical Report},
  volume={9},
  pages={2017},
  year={2017},
  publisher={National University of Defense Technology Changsha, China}
}

@inproceedings{ahrari2022problem,
  title={Problem definition and evaluation criteria for the CEC’2022 competition on dynamic multimodal optimization},
  author={Ahrari, Ali and Elsayed, Saber and Sarker, Ruhul and Essam, Daryl and Coello, Carlos A Coello},
  booktitle={Proceedings of the IEEE World Congress on Computational Intelligence (IEEE WCCI 2022), Padua, Italy},
  pages={18--23},
  year={2022}
}

@article{wang2025animated,
  title={The Animated Oat Optimization Algorithm: A nature-inspired metaheuristic for engineering optimization and a case study on Wireless Sensor Networks},
  author={Wang, Ruo-Bin and Hu, Rui-Bin and Geng, Fang-Dong and Xu, Lin and Chu, Shu-Chuan and Pan, Jeng-Shyang and Meng, Zhen-Yu and Mirjalili, Seyedali},
  journal={Knowledge-Based Systems},
  pages={113589},
  year={2025},
  publisher={Elsevier}
}

@article{shayanfar2018farmland,
  title={Farmland fertility: A new metaheuristic algorithm for solving continuous optimization problems},
  author={Shayanfar, Human and Gharehchopogh, Farhad Soleimanian},
  journal={Applied Soft Computing},
  volume={71},
  pages={728--746},
  year={2018},
  publisher={Elsevier}
}

@article{abdollahzadeh2021african,
  title={African vultures optimization algorithm: A new nature-inspired metaheuristic algorithm for global optimization problems},
  author={Abdollahzadeh, Benyamin and Gharehchopogh, Farhad Soleimanian and Mirjalili, Seyedali},
  journal={Computers \& Industrial Engineering},
  volume={158},
  pages={107408},
  year={2021},
  publisher={Elsevier}
}

@article{luan2024exponential,
  title={Exponential-trigonometric optimization algorithm for solving complicated engineering problems},
  author={Luan, Tran Minh and Khatir, Samir and Tran, Minh Thi and De Baets, Bernard and Cuong-Le, Thanh},
  journal={Computer Methods in Applied Mechanics and Engineering},
  volume={432},
  pages={117411},
  year={2024},
  publisher={Elsevier}
}

@article{tian2024snow,
  title={Snow Geese Algorithm: A novel migration-inspired meta-heuristic algorithm for constrained engineering optimization problems},
  author={Tian, Ai-Qing and Liu, Fei-Fei and Lv, Hong-Xia},
  journal={Applied Mathematical Modelling},
  volume={126},
  pages={327--347},
  year={2024},
  publisher={Elsevier}
}

@article{mckight2010kruskal,
  title={Kruskal-wallis test},
  author={McKight, Patrick E and Najab, Julius},
  journal={The corsini encyclopedia of psychology},
  pages={1--1},
  year={2010},
  publisher={Wiley Online Library}
}

@incollection{wilcoxon1992individual,
  title={Individual comparisons by ranking methods},
  author={Wilcoxon, Frank},
  booktitle={Breakthroughs in statistics: Methodology and distribution},
  pages={196--202},
  year={1992},
  publisher={Springer}
}

@article{hussain2019exploration,
  title={On the exploration and exploitation in popular swarm-based metaheuristic algorithms},
  author={Hussain, Kashif and Salleh, Mohd Najib Mohd and Cheng, Shi and Shi, Yuhui},
  journal={Neural Computing and Applications},
  volume={31},
  number={11},
  pages={7665--7683},
  year={2019},
  publisher={Springer}
}

@article{kumar2020test,
  title={A test-suite of non-convex constrained optimization problems from the real-world and some baseline results},
  author={Kumar, Abhishek and Wu, Guohua and Ali, Mostafa Z and Mallipeddi, Rammohan and Suganthan, Ponnuthurai Nagaratnam and Das, Swagatam},
  journal={Swarm and evolutionary computation},
  volume={56},
  pages={100693},
  year={2020},
  publisher={Elsevier}
}

@inproceedings{hrabar20083d,
  title={3D path planning and stereo-based obstacle avoidance for rotorcraft UAVs},
  author={Hrabar, Stefan},
  booktitle={2008 IEEE/RSJ International Conference on Intelligent Robots and Systems},
  pages={807--814},
  year={2008},
  organization={IEEE}
}

@article{ergezer20143d,
  title={3D path planning for multiple UAVs for maximum information collection},
  author={Ergezer, Halit and Leblebicio{\u{g}}lu, Kemal},
  journal={Journal of Intelligent \& Robotic Systems},
  volume={73},
  number={1},
  pages={737--762},
  year={2014},
  publisher={Springer}
}

@inproceedings{yang2014literature,
  title={A literature review of UAV 3D path planning},
  author={Yang, Liang and Qi, Juntong and Xiao, Jizhong and Yong, Xia},
  booktitle={Proceeding of the 11th world congress on intelligent control and automation},
  pages={2376--2381},
  year={2014},
  organization={IEEE}
}
    
\end{document}